\documentclass{article}

\usepackage{arxiv}

\usepackage[utf8]{inputenc} %
\usepackage[T1]{fontenc}    %
\usepackage{url}            %
\usepackage{graphicx}
\usepackage{booktabs}       %
\usepackage{amsfonts}       %
\usepackage{nicefrac}       %
\usepackage{microtype}      %
\usepackage{xcolor}     %
\usepackage{natbib}

\usepackage{amsmath}
\usepackage{amssymb}
\usepackage{amsthm}
\usepackage{mathtools}
\usepackage{thm-restate}

\makeatletter \@ifundefined{newcounteralias}{}{%
  \renewcommand\thmt@autorefsetup{%
    \@xa\def\csname\thmt@envname autorefname\@xa\endcsname\@xa{\thmt@thmname}%
  }%
} \makeatother

\usepackage{bm}
\usepackage{algorithm}
\usepackage{algpseudocode}
\usepackage{enumitem}
\usepackage{etoc}
\usepackage{hyperref}
\hypersetup{ colorlinks=true, linkcolor=blue, citecolor=blue, urlcolor=blue }
\usepackage[capitalize]{cleveref}
\usepackage{macros}
\usepackage{nccmath}

\makeatletter \providecommand*{\theHALG@line}{}
\renewcommand*{\theHALG@line}{\thealgorithm.\arabic{ALG@line}}
\makeatother

\setlist[enumerate]{itemsep=0pt, topsep=0pt, leftmargin=2.5em} \allowdisplaybreaks[1]
\expandafter\gdef\csname r@HI-def:kappa_ast\endcsname{{3.1}{12}{Assumptions and some further notation}{equation.3.1}{https://arxiv.org/pdf/2505.04898v1}}
\expandafter\gdef\csname r@HI-def:rho_se_matrix\endcsname{{3.4}{14}{}{equation.3.4}{https://arxiv.org/pdf/2505.04898v1}}
\expandafter\gdef\csname r@HI-def:GFOM_se_asym\endcsname{{B.5}{76}{}{problem.B.5}{https://arxiv.org/pdf/2505.04898v1}}
\expandafter\gdef\csname r@HI-def:gd_NN_se\endcsname{{3.1}{13}{}{problem.3.1}{https://arxiv.org/pdf/2505.04898v1}}
\expandafter\gdef\csname r@HI-lem:U_apr_err_est\endcsname{{11.7}{58}{}{problem.11.7}{https://arxiv.org/pdf/2505.04898v1}}
\expandafter\gdef\csname r@HI-prop:GD_l2_est\endcsname{{9.1}{32}{}{problem.9.1}{https://arxiv.org/pdf/2505.04898v1}}
\expandafter\gdef\csname r@HI-prop:GFOM_deloc_asym\endcsname{{B.8}{77}{}{problem.B.8}{https://arxiv.org/pdf/2505.04898v1}}
\expandafter\gdef\csname r@HI-prop:apr_est_derivative\endcsname{{8.3}{31}{}{problem.8.3}{https://arxiv.org/pdf/2505.04898v1}}
\expandafter\gdef\csname r@HI-prop:apr_est_det_fcn\endcsname{{8.1}{29}{}{problem.8.1}{https://arxiv.org/pdf/2505.04898v1}}
\expandafter\gdef\csname r@HI-prop:apr_est_se\endcsname{{8.4}{32}{}{problem.8.4}{https://arxiv.org/pdf/2505.04898v1}}
\expandafter\gdef\csname r@HI-prop:err_gd_GFOM\endcsname{{7.2}{28}{}{problem.7.2}{https://arxiv.org/pdf/2505.04898v1}}
\expandafter\gdef\csname r@HI-prop:se_aux_GFOM\endcsname{{7.1}{27}{}{problem.7.1}{https://arxiv.org/pdf/2505.04898v1}}
\expandafter\gdef\csname r@HI-prop:tau_rho_est_err\endcsname{{11.4}{55}{}{problem.11.4}{https://arxiv.org/pdf/2505.04898v1}}
\expandafter\gdef\csname r@HI-thm:GFOM_se_asym_avg\endcsname{{B.7}{77}{}{problem.B.7}{https://arxiv.org/pdf/2505.04898v1}}
\expandafter\gdef\csname r@HI-thm:se_gd_NN\endcsname{{3.2}{15}{}{problem.3.2}{https://arxiv.org/pdf/2505.04898v1}}

\newtheorem{theorem}{Theorem}[section] \newtheorem{proposition}[theorem]{Proposition} \newtheorem{corollary}[theorem]{Corollary} \newtheorem{lemma}[theorem]{Lemma}  \newtheorem{suppproposition}{Proposition}

\newtheorem{supplemma}{Lemma}[suppproposition]

\crefname{suppproposition}{Proposition}{Propositions} \Crefname{suppproposition}{Proposition}{Propositions} \crefname{supplemma}{Lemma}{Lemmas} \Crefname{supplemma}{Lemma}{Lemmas} \crefname{suppclaim}{Claim}{Claims} \Crefname{suppclaim}{Claim}{Claims} \theoremstyle{definition} \newtheorem{assumption}[theorem]{Assumption} \newtheorem{definition}[theorem]{Definition} \theoremstyle{remark} \newtheorem{remark}[theorem]{Remark}

\crefname{assumption}{Assumption}{Assumptions} \Crefname{assumption}{Assumption}{Assumptions} \crefname{claim}{Claim}{Claims} \Crefname{claim}{Claim}{Claims}

\title{High-Dimensional Off-Policy Evaluation without Sample Splitting using an Onsager Correction} \hypersetup{pdftitle={High-Dimensional Off-Policy Evaluation without Sample Splitting using an Onsager Correction}}

\author{ Saksham Jain \\
  University of Washington \And Guillaume Braun \\
  RIKEN AIP \And Masaaki Imaizumi \\
  University of Tokyo \\ 
  RIKEN AIP \\ Kyoto University } \date{}

\begin{document}
\maketitle

\begin{abstract}
Reward models used for off-policy evaluation (OPE) are commonly selected using held-out data. When sample size and context dimension grow proportionally, the difference in test risk between models fitted on the full sample and on a fixed fraction of it need not vanish. Held-out evaluation can therefore consistently estimate the test risk of the model trained on the smaller sample, but because the relative ranking of candidate reward models can change after full-data refitting, held-out selection can be suboptimal. We develop an Onsager-corrected, target-policy-weighted criterion that estimates the full-data reward model's out-of-sample risk under the target policy using only its training observations. This enables early stopping along the full-data gradient-descent trajectory without sample splitting. For neural networks with fixed depth and hidden width under Gaussian contexts, single-index rewards, context-dependent multi-index target policies, and known context-independent logging, we prove uniform consistency of our criterion over a fixed finite set of gradient-descent iterates and a near-oracle guarantee for the selected reward model. The same correction yields a doubly robust estimator that consistently estimates the true policy value without sample splitting, even when reward-model prediction error does not vanish. Experiments show decreasing risk-estimation error as dimension grows at a fixed sample-to-dimension ratio, and improved reward-model selection relative to held-out baselines. Empirically, our correction also reduces the root mean squared error of doubly robust value estimates computed from the same data used to fit and select the reward models.
\end{abstract}

\section{Introduction}

Off-policy evaluation (OPE) estimates the value of a target policy from data collected by a different logging policy. In contextual bandits, only the rewards of the logged actions are observed, so evaluating a new policy requires estimating its expected reward from partial feedback. Many OPE estimators therefore rely on learned reward models to predict unobserved rewards~\citep{beygelzimer2009offset,dudik2014,wang2017optimal,farajtabar2018more}. The out-of-sample performance of these models matters for value estimation~\citep{su2020doubly,metelli2021subgaussian} and for related decision-making tasks such as policy optimization~\citep{dudik2014,athey2021policy} and staged online bandit learning~\citep{kang2026single,ma2025nonparametric,arya2026kernel}. Selecting a reward model with good prediction performance under the target policy is therefore an important part of OPE.

A natural approach is to select a reward model using held-out prediction error or cross-validation, with importance weighting when appropriate, and then refit it on the full logged sample \citep{arlot2010survey,sugiyama2007covariate}. Although the held-out criterion estimates the test risk of the model trained on the remaining observations, it does not directly estimate the test risk of the model obtained after refitting \citep{bates2024crossvalidation}. This distinction is particularly important in proportional high-dimensional regimes, where sample size and context dimension grow proportionally and prediction risk can depend nontrivially on the sample-to-dimension ratio \citep{dobriban2018highdimensional, han2024gradient, han2025}. Holding out a fixed fraction changes this ratio, so the difference between the full-data and reduced-sample models' test risks need not vanish.

Estimating full-data test risk from training observations requires accounting for each observation's influence on its own fitted prediction~\citep{bellec2026uncertainty,han2025}. In the proportional regime, this influence can persist, so the importance-weighted training error need not consistently estimate test risk. The same dependence can make a doubly robust (DR) policy-value estimator inconsistent when evaluated on its reward model's training
data. For instance, Proposition~\ref{prop:one-step-naive-bias} shows that this can occur after a single gradient-descent step. We therefore ask:
\begin{center}
  \emph{Can one train and select a reward model on the full logged sample, \\ and consistently estimate policy value from the same sample?}
\end{center}

We answer this affirmatively for neural reward models trained by full-batch gradient descent. Building on state-evolution theory for first-order methods~\citep{han2025entrywise,bellec2026uncertainty} and the neural Onsager correction of \citet{han2025}, we construct an in-sample criterion that estimates target-policy-weighted out-of-sample error. The primary technical challenge of this paper is importance weighting: the target-to-logging ratio depends on the same covariates as the learned representation, so its dependence on the prediction error must be retained. We augment the state evolution to jointly track the linear projections that determine the target-policy probabilities and the learned hidden states. The corrected predictions are used both to select the stopping gradient descent iterate and to form `adjusted' direct and doubly robust value estimates.

We study single-index rewards, where each arm's mean reward is an unknown function of a one-dimensional projection of the context. Our analysis in the proportional asymptotics regime assumes Gaussian contexts, known context-independent logging, and neural networks of fixed depth and width. Our contributions are as follows:
\begin{enumerate}
    \item We construct an Onsager-corrected criterion and prove that it consistently estimates the target-policy-weighted conditional test risk of neural reward models trained via gradient descent on the full logged sample.
    \item We show that using this criterion for reward model selection yields near-optimal early stopping without sample splitting for hold-out or cross validation.
    \item We establish guarantees for value-estimation on the full logged sample by using the Onsager-corrected criterion to construct adjusted value estimators that are consistent for their true population targets, without sample splitting, cross-fitting, or requiring prediction error relative to the true reward model to vanish.
    \item We demonstrate the benefits of our methods empirically: at a fixed sample-to-dimension ratio, our criterion's risk-estimation error and selection regret decrease as dimension grows. When sample size outgrows dimension, our adjusted doubly robust estimator attains a mean squared error close to the classical semiparametric efficiency benchmark at the largest sample size studied.
\end{enumerate}
A detailed discussion of related work is provided in Appendix~\ref{app:related-work}.

\section{Statistical Framework}
\label{sec:stat-framework}

\subsection{Problem Formulation} %
\label{subsec:problem-formulation}

\paragraph{OPE in contextual bandit settings.}

We observe an i.i.d.\ sample $\mathcal{D}_m \coloneqq \{Q_i\}_{i=1}^m$ collected under a context-independent logging policy $\mu$. Each observation $Q_i = (X_i, A_i, R_i) \in \mathbb{R}^d \times [K] \times \mathbb{R}$ with $K \geq 2$ comprises the context, the logged action, and the resulting reward. We assume the standard positivity condition~\citep{dudik2014}: $\mu(a) \geq c > 0$ for every $a \in [K]$. We assume rewards are generated as $R_i \coloneqq g^\ast(X_i, A_i) + \varepsilon_i$ with some $g^\ast: \R^d \times [K] \to \R$, where, conditional on $A_i=a$, the noise $\varepsilon_i \in \mathbb{R}$ is independent of $X_i$, mean-zero, and has finite variance. We wish to evaluate a fixed target policy $\nu$, specified independently of the logged sample and allowed to depend on the contexts. Let $\nu(a \mid x)$ denote the probability of choosing action $a$ given context $x$. The value of the target policy $\nu$ is defined as:
\begin{equation}\label{eq:induced_value}
    \Vs(\nu) \coloneqq
    \E_{P_\nu}[R] =
    \E_{P_X} \E_{\nu_{A\mymid X}} \E_{P_{R\mymid A,X}}[R] =
    \E \left[ \sum_{a=1}^K \nu(a\mymid X)g^\ast(X,a)\right].
\end{equation}
Here the final expectation is over the marginal distribution of $X$. To make the dependence on the true reward function explicit, we also write this value as $\Vs_{g^\ast}(\nu)$. For any fixed measurable reward model $g:\R^d\times[K]\to\R$ satisfying $\E[\sum_a\nu(a\mid X)|g(X,a)|]<\infty$, define $\Vs_g(\nu)$ by replacing $g^\ast$ with $g$ in the final expression above. This is the population quantity estimated by the direct method (DM)~\citep{beygelzimer2009offset,dudik2014}.

\paragraph{Importance-weighted risk as selection criterion.}
Let $w_a(x) := \nu(a \mymid x)/\mu(a)$ be the target-to-logging importance ratio~\citep{sugiyama2007covariate, farajtabar2018more} for arm $a$, and observe that since $\nu(\cdot\mymid x)$ is a distribution, $w_a(x)\ge0$ and $\sum_{a}\mu(a)w_a(x)=1$ for all $x\in\R^d$.  We distinguish the reward model's squared error relative to the true reward function, $\Rs_\nu^\circ(g)$, from its squared error in predicting a new reward, $\Rs_\nu(g)$. Averaging under the target policy, we define these quantities as
\begin{align} \label{eq:target-prediction-error}
    \Rs_\nu^\circ(g)
    \coloneqq
    \E\sum_{a=1}^K \nu(a\mymid X) \p{g(X,a)-g^\ast(X,a)}^2,\quad
    \Rs_\nu(g)
    \coloneqq
    \E\br{w_A(X)\p{R-g(X,A)}^2}.
\end{align}
For a single-arm predictor $g_a$, let $\Rs_{\nu,a}^\circ(g_a) \coloneqq \E[\nu(a\mymid X)\p{g_a(X)-g^\ast(X,a)}^2]$. Two facts make $\mathcal{R}_\nu^\circ(g)$ useful for OPE. First, Cauchy--Schwarz gives $\abs{\Vs_g(\nu) - \Vs(\nu)}^2\le \Rs_\nu^\circ(g)$, so prediction error bounds the squared error in the model-induced policy value. Second, \cref{prop:adr-proxy-main} uses this prediction error to bound the additional variance of a DR score evaluated on an observation independent of model fitting, relative to the score using $g^\ast$. These bounds motivate prediction risk as a selection objective; minimizing it need not minimize the value estimator's mean squared error.

Although $g^\ast$ is unknown, the zero-mean independent noise assumption gives $\Rs_\nu(g) = \E[w_A(X)\varepsilon^2] + \Rs_\nu^\circ(g)$. The noise term does not depend on $g$, so minimizing test risk also minimizes prediction error over the same candidate models.

Our goal is to construct a model selector based on $\Rs_\nu(g)$ and to use the selected reward model for policy value estimation, without sample splitting. Note that for \textit{learned} predictors, all risk and value functionals are understood to be for an independent test draw, conditional on the training data and network initializations; these are defined formally in Appendix~\ref{sec:formal-regime}.

\paragraph{Doubly robust value estimation.}
For a reward model $g$, the doubly robust score is defined as
\begin{equation}\label{eq:ortho-score}
\psi_g(X,A,R)
\coloneqq \sum_{a=1}^K\nu(a\mid X)g(X,a)
+w_A(X)\{R-g(X,A)\}.
\end{equation}
Let \(P_d\) denote expectation over a new observation \(Q=(X,A,R)\) from the logging distribution, holding the reward model fixed; the subscript allows the data law to depend on dimension \(d\). For any fixed \(g\) satisfying \(\mathbb E[\sum_a\nu(a\mid X)|g(X,a)|]<\infty\), and assuming \(\mathbb E[w_A(X)|R|]<\infty\), conditioning on \(X\) yields
\begin{align*}
\E[\psi_g(X,A,R)\mid X]
&=\sum_{a=1}^K\nu(a\mid X)g(X,a)
  +\sum_{a=1}^K\mu(a)w_a(X)\{g^\ast(X,a)-g(X,a)\} =\sum_{a=1}^K\nu(a\mid X)g^\ast(X,a).
\end{align*}
Because the logging probabilities are known, \(P_d\psi_g=\mathcal V(\nu)\) for every fixed reward model, even when \(g\ne g^\ast\) (setting \(g=0\) gives the inverse propensity scoring, or IPS, score). The identity also holds for a fitted model, provided the evaluation data is independent of the data used for fitting and selection.

Cross-fitting avoids the need for an additional evaluation sample by applying this identity on disjoint partitions of the data: each observation is evaluated using a model fitted and selected without its fold, so its score has a conditional mean of $\Vs(\nu)$ given its training complement~\citep{schick1986asymptotically,chernozhukov2018double}. Thus, with a fixed number of folds, and under suitable moment bounds~\citep{chernozhukov2018double}, value estimation is consistent even when prediction error persists. 

However, each reward model is consequently trained on a reduced sample, and reward model selection still matters for the precision of value estimation. When an inner validation split selects the model, it assesses models trained on fewer observations than the subsequent refit on the complete training complement. This training-size mismatch can make stopping-time selection suboptimal after refitting (see Appendix~\ref{sec:appendix-reduction}). In Section~\ref{sec:experiments}, we compare this procedure with our corrected criterion, which selects directly on that complement.

An alternative is to reuse the full sample for training, selection, and evaluation and show that the resulting dependence has a negligible effect. Let \(P_{m,d}\) be the empirical measure of the logged sample and let \(\mathbb G_{m,d}=\sqrt m(P_{m,d}-P_d)\). If \(\widehat g\) is fitted and selected on this sample, the same-sample DR estimator \(P_{m,d}\psi_{\widehat g}\) satisfies the exact decomposition

\begin{equation}\label{eq:main-empirical-process}
\sqrt m\{P_{m,d}\psi_{\widehat g}-\Vs(\nu)\}
=\sqrt m\{P_{m,d}\psi_{\widehat g}-P_d\psi_{g^\ast}\}
=\mathbb G_{m,d}\psi_{g^\ast}
+\mathbb G_{m,d}(\psi_{\widehat g}-\psi_{g^\ast}).
\end{equation}
The first term is the fluctuation of the score using the true reward function. The second is the empirical-process term that arises due to replacing $g^\ast$ with the fitted model $\widehat g$. Note that there is no population remainder because we assume known logging probabilities. If the second term is $o_p(1)$, fitting and selecting the model on the same observations as the evaluation data does not change the first-order behavior of estimator $P_{m,d}\psi_{\widehat g}$ relative to $P_{m,d}\psi_{g^\ast}$ at the $\sqrt{m}$ scale.

For a fixed distribution \(P\), the second term is \(o_p(1)\) if \(\psi_{\widehat g}\) belongs with probability tending to one to a fixed \(P\)-Donsker class and \(\|\psi_{\widehat g}-\psi_{g^\ast}\|_{L^2(P)}\to_p0\) \citep{kennedy2024semiparametric}. Suitable algorithmic stability, together with nuisance convergence and regularity conditions, also permits asymptotically normal estimation without sample splitting \citep{chen2022debiased}.

These sufficient conditions are not automatic in our proportional regime. Indeed, prediction error need not vanish, and as we show in Appendix~\ref{app:same-sample-empirical-process}, even a class of linear reward models with uniformly vanishing \(L^2(P_d)\) score errors can have a nonvanishing empirical-process supremum. 

Rather than assuming the effect of sample reuse is negligible at first-order, we analyze the fitted reward models at candidate (for selection) gradient-descent iterates and apply an Onsager correction to their same-sample evaluation. Section~\ref{subsec:theory-value} establishes consistency of our adjusted doubly robust (ADR) estimator (developed in Section~\ref{subsec:adjusted}) on the full training sample without requiring prediction error to vanish. It also shows why the correction matters: uncorrected same-sample DR estimation can be inconsistent after a single gradient-descent step.    

\subsection{Neural Reward Learning}
\label{sec:setting}

We fit a separate reward model for each arm. For $a \in [K]$, let $\mathcal{I}_a \coloneqq \{i \in [m] : A_i = a\}$ be the indices of observations assigned to arm $a$, with count $N_a \coloneqq |\mathcal{I}_a|$. Because logging is context-independent, conditioning on the action labels preserves the common context distribution within each arm. Each model is a feed-forward neural network of depth $L \ge 2$, hidden width $q \ge 1$, and scalar output. The learnable weight parameter is $W_a = (W_{a,1}, W_{a,[2:L]})$, comprising the first layer $W_{a,1} \in \mathbb{R}^{d \times q}$ and the deeper layers $W_{a,[2:L]}$. The learned predictor is defined by:
\begin{equation*}
    f_{W_a}(x) \coloneqq G_{W_{a,[2:L]}}\bigl(x^\top W_{a,1}\bigr),
\end{equation*}
where $G_{W_{a,[2:L]}} : \R^q \to \R$ is the scalar feed-forward map defined by the deeper layers acting on the first-layer hidden state $u = x^\top W_{a,1} \in \R^q$. We identify hidden-state rows in $\R^{1\times q}$ with elements of $\R^q$ when evaluating the network, and write gradients with respect to these coordinates as column vectors. We use smooth activations throughout. The collection of arm-wise initializations $(W_a^{(0)})_{a\in[K]}$ is jointly independent of $\Ds_m$, with first-layer weights scaled by $d^{-1/2}$. The prescribed deterministic layer-wise learning-rate schedule is denoted by $\{\eta_{a,\alpha}^{(s)}:\alpha\in[1:L],s\ge0\}$, where $\alpha$ indexes network layers. Appendix~\ref{sec:formal-regime} states the architecture, initialization, and regularity conditions, including the representation of the scalar output used to apply the state-evolution results.

Simultaneous full-batch gradient descent (GD) on every layer of the arm-wise squared loss $$\mathcal L_a(W_a) \coloneqq  (2N_a)^{-1} \sum_{i \in \Is_a} \bigl(R_i - f_{W_{a}}(X_i)\bigr)^2$$ produces iterates $\{W_a^{(t)}\}_{t \in \Ts}$ and fitted arm-wise models $g_{a,t}(x) \coloneqq  f_{W_a^{(t)}}(x)$. For an iterate vector $\tb = (t_1, \ldots, t_K) \in \Ts^K$, we write $g_\tb(x, a) \coloneqq  g_{a, t_a}(x)$.

We analyze this arm-wise reward learning in the high-dimensional proportional regime, where the aspect ratio $m/d \to \phi \in (0, \infty)$ with the number of arms $K$, network depth $L$, width $q$, and nonempty GD iterate grid $\Ts \subset \mathbb Z_{\ge 0}$ all finite and fixed. Unlike neural tangent kernel-type regimes, where representations remain close to initialization during training~\citep{jacot2018neural, bartlett2021deep}, our finite-width setting allows non-trivial feature learning across multiple layers.

\subsection{Assumptions}

We make the following assumptions for our analysis. Most are standard regularity conditions used in finite-width state-evolution analyses, as in \cite{han2025}; we emphasize the assumptions that concern the OPE problem structure and target-policy weighting. The formal exposition of all conditions in this section, including on the true reward function and neural network architecture, is provided in Appendix~\ref{sec:formal-regime}.

\begin{assumption}[Known context-independent logging under positivity]
\label{ass:no-empty-arms}
The logging policy is known and context-independent, with $\mu(a)\ge c>0$ for every $a\in[K]$, where $c$ is independent of $d$.
\end{assumption}

\begin{assumption}[Single-index rewards under Gaussian design]
\label{ass:single-index-logging}
The contexts satisfy $X_i \overset{iid}{\sim} \Ns(0, I_d)$. For each arm $a$, assume $g^\ast(x, a) = \varphi_a^\ast(\beta_a^\top x)$ with $\norm{\beta_a}_2 = 1$, $\sqrt d\norm{\beta_a}_\infty=d^{o(1)}$, and $\varphi_a^\ast$ smooth and of polynomial growth. We also let the arm-wise noise $\varepsilon_i \mymid A_i = a$ be sub-Gaussian with a dimension-independent parameter.

\end{assumption}

\begin{assumption}[Multi-index importance weights]
\label{ass:multi-index-weights}
    We let $w_a$ be a multi-index function, i.e., $w_a(x) = h_a(\Upsilon^\top x)$ with $\Upsilon=(\gamma_1,\ldots,\gamma_\ell) \in \R^{d \times \ell}$, $\Upsilon^\top \Upsilon = I_\ell$, $\sqrt d\,\norm{\gamma_j}_\infty \le C_\Upsilon$ for each $j$, and $h_a$ is nonnegative, bounded, and Lipschitz, with $\sum_a \mu(a)h_a(u) = 1$ for every $u \in \R^\ell$, for some fixed $\ell \ge 1$. The coordinate bound and the common boundedness and Lipschitz constants are independent of $d$.
\end{assumption}

Assumption~\ref{ass:no-empty-arms} accommodates practically relevant logging policies such as pure exploration, randomized controlled trials and A/B tests. Our guarantees assume known logging probabilities. Under the proportional regime, this assumption ensures that the arm-wise aspect ratios $\phi_a=N_a/d$ belong to a fixed interval $[c_-,c_+]\subset(0,\infty)$ with probability tending to one. Proposition~\ref{prop:supp:arm-wise_conditioning} yields this count bound and the resulting reduction to arm-wise regression. Assumption~\ref{ass:single-index-logging} permits unknown nonlinear, arm-specific link functions---a setting of active interest in recent bandit literature~\citep{kang2026single, ma2025nonparametric, arya2026kernel}. The isotropic Gaussian design is a standard, if restrictive, assumption in state evolution theory~\citep{bayati2011dynamics, berthier2020state, bellec2026uncertainty}; extensions of the weighted comparison beyond Gaussian contexts are discussed in Section~\ref{sec:discussion}. Finally, sub-Gaussian noise is assumed merely for convenience; our proofs hold under milder moment and envelope conditions detailed in Appendix~\ref{sec:formal-regime}. Assumption~\ref{ass:multi-index-weights} includes e.g. logistic, softmax, and probit policies whose scores depend on a fixed number of linear projections satisfying the stated bounds. Different arms may use different projections, provided their joint span admits an orthonormal basis $\Upsilon$ satisfying the coordinate bounds above. For isotropic Gaussian contexts, Corollary~\ref{cor:initialization-rotation} permits arbitrary fixed reward and policy-index directions, including axis-aligned directions, under the stated data-independent initialization and size conditions.

\section{Full-Sample OPE via Onsager Correction}
\label{sec:full-sample-via-Onsager}

This section introduces our full-sample OPE procedure based on the general first-order method (GFOM) Onsager correction~\citep{bayati2011dynamics,celentano2020estimation,han2024gradient,han2025}. We first correct each training observation's hidden state to account for its contribution to the fitted weights, removing the leading sample-reuse bias induced by gradient-based training. We then use the resulting predictions for model selection and for direct and doubly robust value estimation, without sample splitting.

\subsection{GFOM-Corrected Hidden States}
\label{subsec:corrected-hidden-state}

Our goal is to estimate the weighted test risk $\mathcal{R}_\nu(g)$~\eqref{eq:target-prediction-error} without a holdout set. We apply the GFOM correction separately to each arm's fitted network. Fix an arm $a$, and let $X_a\in\R^{N_a\times d}$ be its design matrix. At iterate $t$, the first-layer pre-activation on training observation $i$ is $\widetilde U_{i,a}^{(t)}=X_i^\top W_{a,1}^{(t)}$ for $i\in\Is_a$. To estimate test error from the training residuals, we account for observation $i$'s influence on the gradient updates that produced $W_{a,1}^{(t)}$. The neural Onsager correction of \citet{han2025} removes the leading self-interaction by adding a term to the hidden state. To define it, for each step $r$, let
\[
    \Psi_{i,a}^{(r)}
    \coloneqq
    \left(
        G_{W_{a,[2:L]}^{(r)}}(\widetilde U_{i,a}^{(r)})-R_i
    \right)
    \nabla_u
    G_{W_{a,[2:L]}^{(r)}}(\widetilde U_{i,a}^{(r)})^\top
    \in\R^{1\times q},
\]
and let $\widetilde U_{a}^{(r)}$ and $\Psi_a^{(r)}\in\R^{N_a\times q}$ collect the corresponding rows.

Thus, each row's $\Psi_{i,a}^{(r)}$ is the gradient of (one half of) the squared residual with respect to the first-layer pre-activation, written as a row vector, so the first-layer loss gradient is $N_a^{-1}X_a^\top\Psi_a^{(r)}$. From this training trajectory, the recursion detailed in Appendix~\ref{sec:onsager} computes the Onsager matrices $\widehat\rho_{a,t,s}\in\R^{q\times q}$, $s=1,\ldots,t$.

The GFOM-corrected hidden state is then defined as follows. We set $\widehat U_{i,a}^{(0)}=\widetilde U_{i,a}^{(0)}$ at initialization. At each subsequent iterate, the corrected hidden state for $i\in\Is_a$ is
\begin{equation}
    \widehat U_{i,a}^{(t)}
    =
    X_i^\top W_{a,1}^{(t)}
    +
    \phi_a^{-1}
    \sum_{s=1}^t
    \eta_{a,1}^{(s-1)}
    \Psi_{i,a}^{(s-1)}
    \widehat\rho_{a,t,s}^{\top}
    \in\R^{1\times q}.
    \label{eq:onsager-correction-row}
\end{equation}
Let $\widehat U_{a}^{(t)}\in\R^{N_a\times q}$ collect these rows. It is computed from the same arm-wise training trajectory and does not require sample splitting, leave-one-out retraining, or knowledge of the true reward function.

\subsection{Model Selection and Value Estimation without Sample Splitting}
\label{subsec:adjusted}

The corrected hidden states~\eqref{eq:onsager-correction-row} are used to select the gradient-descent stopping iterate and estimate the target-policy value from the training sample only.

\paragraph{Model selection.}
For each arm $a$ and each iterate $t \in \mathcal{T}$, we define the corrected weighted residual criterion
\begin{equation}
    \widehat{\mathcal{E}}_a^{(t)}
    \coloneqq
    \frac{1}{N_a} \sum_{i \in \Is_a}
    w_a(X_i)
    \left(
        R_i -
        G_{W_{a,[2:L]}^{(t)}}(\widehat{U}_{i,a}^{(t)})
    \right)^2.
    \label{eq:weighted-criterion}
\end{equation}
The criterion replaces each fitted prediction in the importance-weighted average of squared training residuals with a prediction based on the corrected hidden state. Theorem~\ref{thm:uniform-risk} proves that it consistently estimates the fitted network's weighted test risk, including the reward-noise contribution. We select the stopping iterate by minimizing this criterion:
\begin{align} \label{def:that}
\hat{t}_a \in \arg\min_{t \in \mathcal{T}} \widehat{\mathcal{E}}_a^{(t)},
\end{align}
breaking ties using a fixed ordering of $\mathcal{T}$. The selected iterate vector for the arms is therefore $\hat{\mathbf{t}} = (\hat{t}_1, \ldots, \hat{t}_K)$.

\paragraph{Value estimation.}
Fix an iterate vector $\mathbf{t} \in \mathcal{T}^K$. For observation $i$, the model for action $A_i$ was trained using that observation, so we evaluate it with the corrected hidden state. The models for the other actions were trained on disjoint observations, so we use their fitted predictions without correction. Define
\begin{equation}\label{eq:adm-model}
    \widehat{g}_{\mathbf{t}}(X_i, a)
    \coloneqq
    \begin{cases}
        G_{W_{a,[2:L]}^{(t_a)}}(\widehat{U}_{i,a}^{(t_a)}),
        & a = A_i, \\
        G_{W_{a,[2:L]}^{(t_a)}}(X_i^\top W_{a,1}^{(t_a)}),
        & a \neq A_i.
    \end{cases}
\end{equation}
The adjusted direct method (ADM) value estimator is
\begin{equation}
    \widehat{\mathcal{V}}_{\mathrm{ADM}}(\mathbf{t})
    \coloneqq
    \frac{1}{m} \sum_{i=1}^m \sum_{a=1}^K
    \nu(a \mymid X_i)\,
    \widehat{g}_{\mathbf{t}}(X_i, a),
    \label{eq:adm}
\end{equation}
and the adjusted doubly robust (ADR) value estimator is
\begin{equation}
    \widehat{\mathcal{V}}_{\mathrm{ADR}}(\mathbf{t})
    \coloneqq
    \widehat{\mathcal{V}}_{\mathrm{ADM}}(\mathbf{t})
    +
    \frac{1}{m} \sum_{i=1}^m
    w_{A_i}(X_i)
    \left(
        R_i -
        \widehat{g}_{\mathbf{t}}(X_i, A_i)
    \right).
    \label{eq:adr}
\end{equation}

The two procedures use the same corrected hidden states but target different quantities. The selection criterion estimates target-policy-weighted test risk. ADM estimates the fitted-model value, while ADR estimates the true policy value. Section~\ref{sec:guarantees} establishes consistency uniformly over the candidate iterations, without sample splitting. For a new context $x$, we use the selected network $g_{\widehat{\mathbf t}}(x,a)$; the adjustment in \eqref{eq:adm-model} is used only to evaluate observations included in that arm's training sample.

\begin{algorithm}[t]
\caption{Full-sample model selection and adjusted policy evaluation}
\label{alg:full-sample-ope}
\begin{algorithmic}[1]
\Require Logged sample $\Ds_m$, known $\mu$, fixed target policy $\nu$, initializations, learning-rate schedules, and fixed iterate grid $\Ts$.
\For{each arm $a\in[K]$}
\State Train all layers by full-batch GD on $\mathcal L_a$ through $T=\max\Ts$.
\State Compute $\widehat\rho_{a,t,s}$ by Appendix~\ref{sec:onsager} and $\widehat U_{i,a}^{(t)}$ by \eqref{eq:onsager-correction-row}.
\State Evaluate $\widehat{\mathcal E}_a^{(t)}$ for $t\in\Ts$ and select $\hat t_a$ by \eqref{def:that}.
\EndFor
\State Form the adjusted training predictions \eqref{eq:adm-model} at $\widehat{\mathbf t}$.
\State Compute ADM and ADR by \eqref{eq:adm} and \eqref{eq:adr}.
\State \Return $g_{\widehat{\mathbf t}}$ for new-context prediction, and the two adjusted value estimates.
\end{algorithmic}
\end{algorithm}

\section{Theoretical Guarantees}
\label{sec:guarantees}

We first show that our in-sample selection criterion~\eqref{eq:weighted-criterion} uniformly estimates conditional arm-wise weighted test risk, enabling near-oracle early stopping (Section~\ref{subsec:theory-calibration}). We then establish consistency of ADM for the model-induced value and of ADR for the true policy value, contrast these guarantees with a one-step inconsistency example for uncorrected DR, and bound the conditional variance of a DR score evaluated on an independent observation (Section~\ref{subsec:theory-value}). Finally, we outline the proof strategy (Section~\ref{subsec:theory-proofs}). All results assume the statistical framework of \Cref{sec:stat-framework} and the formal regularity conditions in Appendix~\ref{sec:formal-regime}.

\subsection{Consistent Risk Estimation and Near-Oracle Model Selection}
\label{subsec:theory-calibration}
For each arm $a$, let $Z_a \sim \mathcal{N}(0, I_d)$ and let $\xi_a^{\mathrm{new}}$ have the law of $\varepsilon_i\mid A_i=a$. These draws are independent of each other and of the training sample and network initializations. The weighted test risk at iterate $t$, conditional on the training data and initializations, is
\begin{equation}
    \label{eq:weighted-armwise-ind-risk}
    \mathcal{E}_{a, \mathrm{ind}}^{(t)}
    \coloneqq  \mathbb{E}\!\left[
        w_a(Z_a)
        \left(\varphi_a^\ast(\beta_a^\top Z_a) + \xi_a^{\mathrm{new}}
              - f_{W_a^{(t)}}(Z_a)\right)^2
        \mymid \mathcal{F}_{\mathrm{tr}}
        \right],
\end{equation}
where $\mathcal{F}_{\mathrm{tr}}$ is generated by the training sample and all network initializations. The total test risk is $\mathcal{R}_\nu(g_{\mathbf{t}}) = \sum_a \mu(a)\mathcal{E}_{a, \mathrm{ind}}^{(t_a)}$ almost surely. The following theorem shows that the criterion consistently estimates each arm's conditional test risk, uniformly over $t\in\Ts$.

\begin{restatable}[Uniform consistency]{theorem}{thmreduction}
    \label{thm:uniform-risk}
    Assume the statistical framework of Section~\ref{sec:stat-framework} and the formal conditions in Appendix~\ref{sec:formal-regime}. For every arm $a\in[K]$, as $N_a,d\to\infty$ with $\phi_a$ bounded away from zero and infinity, the criterion in \eqref{eq:weighted-criterion} and the conditional risk in \eqref{eq:weighted-armwise-ind-risk} satisfy
    \begin{equation*}
        \sup_{t \in \mathcal{T}}
        \abs{\widehat{\mathcal{E}}_a^{(t)} - \mathcal{E}_{a, \mathrm{ind}}^{(t)}}
        \xrightarrow{p} 0.
    \end{equation*}
\end{restatable}
This uniform consistency yields the following guarantee for the stopping rule $\hat t_a$ in \eqref{def:that}. The proof is given in Appendix~\ref{app:thm-selection-main-proof}.

\begin{restatable}[Arm-wise iterate selection and oracle inequality]{theorem}{thmselectionmain}
    \label{thm:selection-main}
    Under the assumptions and asymptotic regime of Theorem~\ref{thm:uniform-risk}, the selector in \eqref{def:that} satisfies, for every $a\in[K]$,
    \begin{equation*}
      \mathcal E_{a,\mathrm{ind}}^{(\hat t_a)}
      \le \min_{t\in\Ts}\mathcal E_{a,\mathrm{ind}}^{(t)}+o_p(1).
    \end{equation*}
    Moreover,
    \begin{align*}
    \mathcal R_\nu(g_{\hat{\mathbf t}})
    &\le \min_{\mathbf t\in\mathcal T^K}\mathcal R_\nu(g_{\mathbf t})+o_p(1),\\
    \left|\mathcal V_{g_{\hat{\mathbf t}}}(\nu)-\mathcal V(\nu)\right|^2
    &\le \min_{\mathbf t\in\mathcal T^K}\mathcal R_\nu^\circ(g_{\mathbf t})+o_p(1).
    \end{align*}
\end{restatable}

The weighted test risk $\mathcal{R}_\nu$ is the sum of the weighted prediction error $\mathcal{R}_\nu^\circ$ and a noise term that does not depend on the fitted model. The bound on total test risk therefore also holds for $\mathcal{R}_\nu^\circ$. Since $\mathcal{R}_\nu^\circ(g)$ bounds $|\mathcal{V}_g(\nu)-\mathcal{V}(\nu)|^2$, as shown in Section~\ref{subsec:problem-formulation}, this gives the stated bound on the fitted model's policy value error. Thus, without sample splitting, the selected models attain the smallest prediction risk among the candidate iterates up to $o_p(1)$.

\subsection{Consistency of Value Estimation}
\label{subsec:theory-value}

ADM and ADR consistently estimate their respective values using the full training sample. A single gradient-descent step example first shows why the population identity $P_d\psi_g=\Vs(\nu)$ alone does not ensure consistency of uncorrected same-sample DR.

\paragraph{Inconsistency of DR evaluation on the training sample.}
Consider the DR estimator formed by evaluating the fitted models directly on their training observations, $\widehat{\Vs}_{\DR}^{\mathrm{nv}}(\mathbf t)=m^{-1}\sum_i\psi_{g_{\mathbf t}}(Q_i)$. Take $\nu=\mu$ and a constant reward $R=c$ almost surely. With $\Lambda$ as in Assumption~\ref{ass:formal-nn-regularity}, fix $0<|c|\le\Lambda$ and $0<\eta\le\Lambda$. For each arm, use a network with depth $L=2$, width $q=1$, and identity activation $\sigma_1=\mathrm{id}$. Initialize $W_{a,1}^{(0)}=0$ and $W_{a,2}^{(0)}=1$, and set the first-layer learning rate to $\eta_{a,1}^{(0)}=\eta$.

For each observed arm, write $\overline X_a=N_a^{-1}\sum_{i\in\Is_a}X_i$. The loss gradients at initialization are
\begin{equation*}
  \nabla_{W_{a,1}}\mathcal L_a(W_a^{(0)})=-c\,\overline X_a,
  \qquad
  \nabla_{W_{a,2}}\mathcal L_a(W_a^{(0)})=0.
\end{equation*}
Thus, one simultaneous full-batch GD step produces
\begin{equation*}
  g_{a,1}(x)=\eta c\,x^\top\overline X_a.
\end{equation*}
For $i\in\Is_a$, the prediction $g_{a,1}(X_i)$ contains the term $\eta c\|X_i\|_2^2/N_a$, which is the contribution of observation $i$ to its own fitted prediction. This contribution does not vanish when $N_a$ and $d$ grow proportionally. The following proposition yields the resulting policy-value error.

\begin{restatable}[Inconsistency after one gradient-descent step]{proposition}{proponestepnaivebias}
\label{prop:one-step-naive-bias}
For the constant-reward example in Section~\ref{subsec:theory-value}, assume the conditions of Theorem~\ref{thm:uniform-risk}, $1\in\Ts$, and $m/d\to\phi\in(0,\infty)$. Then
\[
  \widehat{\Vs}_{\DR}^{\mathrm{nv}}(1,\ldots,1)
  -\Vs(\nu)
  \xrightarrow{p}
  -\frac{\eta c(K-1)}{\phi}.
\]
\end{restatable}

The limit is nonzero because $c\ne0$, $\eta>0$, and $K\ge2$. Thus, the DR estimator is inconsistent at the fixed first GD iterate. Appendix~\ref{app:same-sample-empirical-process} gives the proof and also characterizes the error of $\widehat{\Vs}_{\DR}^{\mathrm{nv}}$ uniformly over the candidate iterations, including choices made from the training data.

\paragraph{Consistency of ADM and ADR.}
Correcting the training predictions yields the following guarantees for ADM~\eqref{eq:adm} and ADR~\eqref{eq:adr}. ADM targets the population value induced by the learned reward model, while ADR targets the true policy value $\mathcal{V}(\nu)$~\eqref{eq:induced_value}. The proof is in Appendix~\ref{app:proof-value-estimation}.
\begin{theorem}[Uniform consistency of value estimation]
    \label{thm:value_estimation}
Under the assumptions and asymptotic regime of Theorem~\ref{thm:uniform-risk}, the estimators in \eqref{eq:adm} and \eqref{eq:adr} satisfy
    \begin{equation*}
        \sup_{\mathbf{t} \in \mathcal{T}^K}
        \abs{\widehat{\mathcal{V}}_{\ADM}(\mathbf{t}) -
            \mathcal{V}_{g_{\mathbf{t}}}(\nu)}
        \xrightarrow{p} 0,
        \qquad
        \sup_{\mathbf{t} \in \mathcal{T}^K}
        \abs{\widehat{\mathcal{V}}_{\ADR}(\mathbf{t}) -
            \mathcal{V}(\nu)}
        \xrightarrow{p} 0.
    \end{equation*}

\end{theorem}

Both conclusions hold for any choice $\widehat{\mathbf t}\in\Ts^K$ based on the training data and initializations, including the rule in \eqref{def:that}. ADR remains consistent when the selected model's prediction error does not converge to zero. ADM estimates the value induced by that model, so ADM minus ADR consistently estimates $\mathcal V_{g_{\widehat{\mathbf t}}}(\nu)-\mathcal V(\nu)$, as formalized in \Cref{cor:adr-selected}.

\paragraph{Variance of a DR score on an independent observation.}
We now return to the role of prediction risk in controlling the variance of the score $\psi_g(X,A,R)$ in \eqref{eq:ortho-score}, evaluated on an observation independent of the training sample and initializations. Conditional on $\mathcal F_{\mathrm{tr}}$, the fitted model is fixed. Define the weighted prediction error
\begin{equation}
\mathcal J_{\mathrm{DR}}(g)
\coloneqq
\E\left[
  \sum_{a=1}^K
  \nu(a\mid X)\frac{1-\mu(a)}{\mu(a)}
  \{g(X,a)-g^\ast(X,a)\}^2
  \mymid
  \mathcal F_{\mathrm{tr}}
\right].
\label{eq:dr-weighted-prediction-error}
\end{equation}
By the variance decomposition in \Cref{prop:adr-proxy-main}, $\mathcal J_{\mathrm{DR}}(g)$ bounds the increase in this conditional variance relative to the score using the true reward function $g^\ast$. To state the consequence for model selection, define
\begin{equation}
\begin{aligned}
  c_-(\mu)&\coloneqq\min_{a\in[K]}\frac{1-\mu(a)}{\mu(a)},
  &c_+(\mu)&\coloneqq\max_{a\in[K]}\frac{1-\mu(a)}{\mu(a)}, &\sigma_{\mathrm{eff}}^2
  &\coloneqq\operatorname{Var}\{\psi_{g^\ast}(X,A,R)\}.
\end{aligned}
\label{eq:dr-variance-constants}
\end{equation}
Positivity and $K\ge2$ imply $0<c_-(\mu)\le c_+(\mu)<\infty$. By definition,
\begin{equation*}
  c_-(\mu)\mathcal R_\nu^\circ(g)
  \le\mathcal J_{\mathrm{DR}}(g)
  \le c_+(\mu)\mathcal R_\nu^\circ(g).
\end{equation*}
Applying Theorem~\ref{thm:selection-main} separately to each arm and weighting its regret by $(1-\mu(a))/\mu(a)$ gives the following bound.

\begin{restatable}[Conditional variance on an independent observation]{corollary}{corindscorevariance}
\label{cor:independent-score-variance}
Under the assumptions and asymptotic regime of Theorem~\ref{thm:selection-main}, let $\hat{\mathbf t}$ be selected by \eqref{def:that}. For an observation $(X,A,R)$ independent of $\mathcal F_{\mathrm{tr}}$, and with $\mathcal J_{\mathrm{DR}}$ and the constants defined in \eqref{eq:dr-weighted-prediction-error} and \eqref{eq:dr-variance-constants},
\[
\operatorname{Var}\{\psi_{g_{\hat{\mathbf t}}}(X,A,R)
\mymid\mathcal F_{\mathrm{tr}}\}
\le
\sigma_{\mathrm{eff}}^2
+
\min_{\mathbf t\in\mathcal T^K}
\mathcal J_{\mathrm{DR}}(g_{\mathbf t})
+o_p(1).
\]
In particular, if $\min_{\mathbf t\in\mathcal T^K}\mathcal R_\nu^\circ(g_{\mathbf t}) =o_p(1)$, then
\[
\operatorname{Var}\{\psi_{g_{\hat{\mathbf t}}}(X,A,R)
\mymid\mathcal F_{\mathrm{tr}}\}
=
\sigma_{\mathrm{eff}}^2+o_p(1).
\]
\end{restatable}

The proof is given in Appendix~\ref{app:DR-var-control}. The corollary holds the selected model $g_{\hat{\mathbf t}}$ fixed and bounds the variance of its DR score \eqref{eq:ortho-score} on a fresh observation. ADR in \eqref{eq:adr} averages corrected scores on the training sample. Its sampling variance also depends on covariances between these scores, which share fitted models and correction terms. Characterizing these joint fluctuations is left to future work.

We also note here that the cross-fitted value estimator requires a separate variance analysis in the proportional regime. Its standard $\sqrt{m}$-convergence to $P_{m,d}\psi_{g^\ast}$ argument uses vanishing fitted-score error, which is a much stronger condition under proportionally growing $d$. In a related setting with linear outcomes and estimated logistic propensities, \citet{jiang2025new} characterize variance inflation and cross-fold covariance. This motivates our comparison of ADR both with cross-fitted DR and with the oracle DR benchmark empirically in Section~\ref{sec:experiments}.

\subsection{Proof Strategy and Key Novelty}
\label{subsec:theory-proofs}

We outline the proof that the corrected criterion consistently estimates conditional test risk. The Onsager correction replaces $\widetilde U_{i,a}^{(t)}$ with $\widehat U_{i,a}^{(t)}$~\eqref{eq:onsager-correction-row} to account for observation $i$'s contribution to its own fitted prediction; see Section~\ref{subsec:corrected-hidden-state}.

State evolution approximates empirical averages along the training trajectory by expectations of functions of a jointly Gaussian sequence. Let $\mathfrak U_a^{(0)}\in\R^q$ represent the signal coordinate $(\beta_a^\top X_i,0,\ldots,0)$, with zeros added to match the hidden-state dimension, and let $\mathfrak U_a^{(t+1)}\in\R^q$ be the Gaussian vector corresponding to the corrected hidden state at iterate $t$. The unweighted theory of \citet{han2025} compares averages of corrected fitted values and squared residuals with expectations under this state evolution.

A key technical challenge is the weighted risk in \eqref{eq:weighted-criterion}, which multiplies each squared residual by $w_a(X_i)=h_a(\Upsilon^\top X_i)$. Consequently, the analysis must track the policy index and learned hidden state jointly, since both depend on $X_i$.

We account for this dependence by appending a Gaussian vector $\mathfrak U_{\Upsilon,a}\sim\mathcal N(0,I_\ell)$ to the state evolution. The augmented recursion explicitly tracks its covariance with the learned hidden states:
\begin{equation*}
    \operatorname{Cov}(\mathfrak U_a^{(s)},\mathfrak U_{\Upsilon,a})
    =d^{-1/2}D_{a,s-1}^\top\Upsilon,
\end{equation*}
where $D_{a,s-1}\in\R^{d\times q}$ is the shift matrix defined by the arm-wise state-evolution recursion. The Gaussian law and covariance above are conditional on $\mathcal G_a$, which fixes the action labels, initialization, and realized noise. The recursion and conditioning are specified in Appendix~\ref{sec:formal-regime} and Proposition~\ref{prop:supp:auxiliary_gfom_policy_index}. The comparison is first established for the auxiliary GFOM and then transferred to the actual corrected states. For example, let $V_a^{(t)}$ denote the deeper-layer parameters in the state evolution. The weighted average of corrected predictions satisfies
\begin{align*}
    &\frac1{N_a}\sum_{i\in\Is_a} h_a(\Upsilon^\top X_i)
    G_{W_{a,[2:L]}^{(t)}}(\widehat U_{i,a}^{(t)})
    =
    \E\!\left[h_a(\mathfrak U_{\Upsilon,a})
    G_{V_a^{(t)}}(\mathfrak U_a^{(t+1)})\mid\mathcal G_a\right]
    +o_p(1).
\end{align*}
An analogous comparison applies to the weighted squared residual, with the state-evolution expectation averaged over the realized noise values in arm $a$. These comparisons hold uniformly over the fixed set of candidate iterations. They imply that $\widehat{\mathcal E}_a^{(t)}$ consistently estimates an intermediate test risk $\mathcal E_{a,\pi}^{(t)}$ that uses a new context and a noise value sampled uniformly from the realized training noises (\Cref{prop:arm-wise-calibration}). \Cref{prop:independent-noise-comparison} then shows that replacing this empirical noise distribution with the true arm-specific noise law changes the risk by $o_p(1)$, uniformly over the candidate iterations. This yields consistency for the conditional test risk in \eqref{eq:weighted-armwise-ind-risk}.

The Onsager matrices are those of \citet{han2025} since the target-policy weights enter evaluation, while training still minimizes the unweighted arm-wise squared loss.

\section{Experiments}
\label{sec:experiments}

In this section, we examine whether the proposed criterion estimates the test risk of the full-data reward model, how its stopping decisions depend on the target policy, and how the correction affects value estimation. We also compare the value estimators' mean squared errors at both fixed and increasing sample-to-dimension ratios. Experimental details are provided in Appendix~\ref{app:exp-details}.

\paragraph{Risk estimation and early stopping.}
We use a two-arm Gaussian bandit with uniform logging, nonlinear single-index rewards, and a context-dependent target policy. The reward models are sigmoid multilayer perceptrons trained by full-batch gradient descent. We increase \(d\in\{128,256,512,1024,2048\}\) with \(m=2d\), using 200 independent datasets per dimension. Appendix~\ref{app:exp-details} specifies the data-generating model and training settings.

\begin{figure}[ht]
\centering
\includegraphics[width=0.85\textwidth]{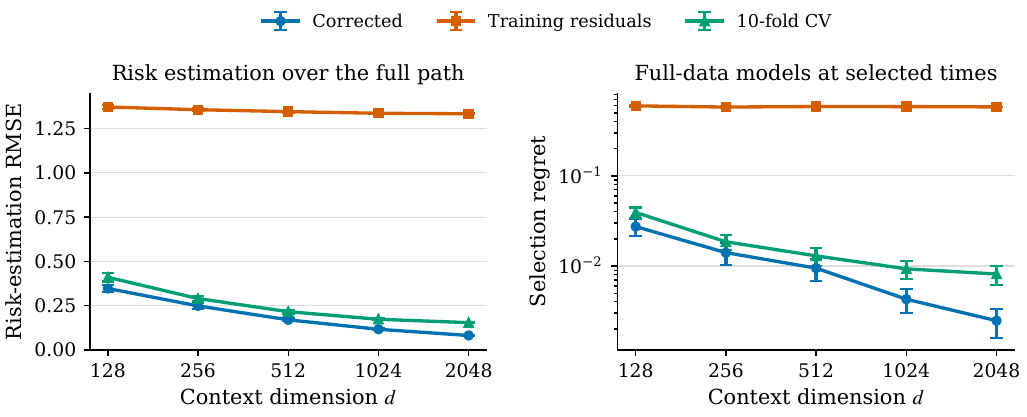}
\caption{Risk estimation and stopping-time selection with \(m=2d\) and 200 datasets per dimension. Left: RMSE in estimating the full-data models’ conditional arm-wise test risks, with squared errors averaged over arms, candidate iterations, and datasets. Right: mean total target-policy prediction-risk regret of the selected full-data model, on a logarithmic scale. Bars give pointwise \(95\%\) Monte Carlo intervals.}
\label{fig:calibration}
\end{figure}

Figure~\ref{fig:calibration} (left) compares risk estimates against the conditional test risk of the full-data model. Our corrected criterion’s root mean squared error (RMSE) decreases from \(0.347\) to \(0.081\), compared with \(0.410\) to \(0.154\) for ten-fold cross-validation (CV). At the largest dimension, correction nearly halves CV’s RMSE. Because CV evaluates models trained on fold complements, its discrepancy from full-data risk includes any difference between the test risks of the fold-complement and full-data models. Meanwhile, the importance-weighted mean squared training residuals remain inaccurate across all dimensions, with RMSE between \(1.33\) and \(1.37\).

Our corrected criterion also improves stopping decisions (Figure~\ref{fig:calibration}, right). We apply each rule’s selected iteration to the same full-data training trajectory. Selection regret is the excess total target-policy prediction risk \(\mathcal R_\nu^\circ\) over the best arm-wise stopping times on that trajectory. At \(d=2048\), the corrected rule’s mean regret is \(0.0025\), less than one-third of CV’s \(0.0082\), while training-residual stopping has regret near \(0.59\). Appendix~\ref{app:exp-risk-paths} provides the risk curves and comparisons with networks retrained after omitting the evaluated observation.

The selection rule based on the corrected criterion and CV also achieve lower risk-estimation error and selection regret than single holdouts of sizes \(\lceil\sqrt m\rceil\) and \(\lceil m^{2/3}\rceil\) at every tested dimension (Appendix~\ref{app:exp-vanishing-holdouts}). These holdouts are faster; Appendix~\ref{app:exp-runtime} reports the computational comparison. Additional experiments with nonuniform logging or non-Gaussian contexts show decreasing corrected risk-estimation error and selection regret as dimension grows. ADR has lower absolute empirical bias than uncorrected DR in these settings (Appendices~\ref{app:exp-nonuniform} and~\ref{app:exp-nongaussian}).

\paragraph{The role of target-policy weighting.}
The stopping criterion should measure prediction error in the contexts where the target policy uses each arm. To isolate this dependence, we vary the target policy while holding the training data, network trajectories, and Onsager correction fixed. We use three arms and four target policies, with \(m=3d\) at \(d\in\{128,256,512\}\). Appendix~\ref{app:exp-policy-weighting} specifies the reward links and policies.

\begin{figure}[ht]
\centering
\includegraphics[width=0.85\textwidth]{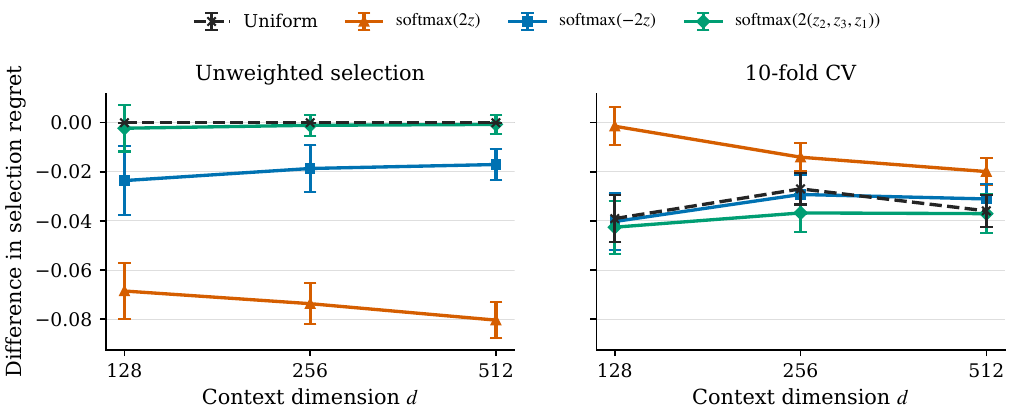}
\caption{Effect of target-policy weighting on stopping decisions for identical full-data trajectories. Points show the mean paired difference across datasets in selection regret between the corrected weighted criterion and the corrected unweighted criterion (left) or target-weighted ten-fold CV (right). Negative values favor the corrected weighted criterion. Bars give pointwise \(95\%\) Monte Carlo intervals over 200 datasets per dimension. All four target policies are shown.}
\label{fig:policy-weighting}
\end{figure}

Figure~\ref{fig:policy-weighting} compares the corrected weighted criterion with unweighted selection and target-weighted ten-fold CV. Averaging over the three nonuniform policies at each dimension, weighting reduces the selected model’s prediction risk by \(0.031\)–\(0.033\) relative to unweighted selection. For the uniform policy, the weighted and unweighted rules coincide.

Relative to target-weighted CV, the corrected weighted rule reduces prediction risk by \(0.027\)–\(0.029\), again averaging over the three nonuniform policies at each dimension. Its mean selection regret is lower than CV’s for every policy and dimension. Appendix~\ref{app:exp-policy-weighting} reports the paired comparisons.

\begin{figure}[ht]
\centering
\includegraphics[width=0.85\textwidth]{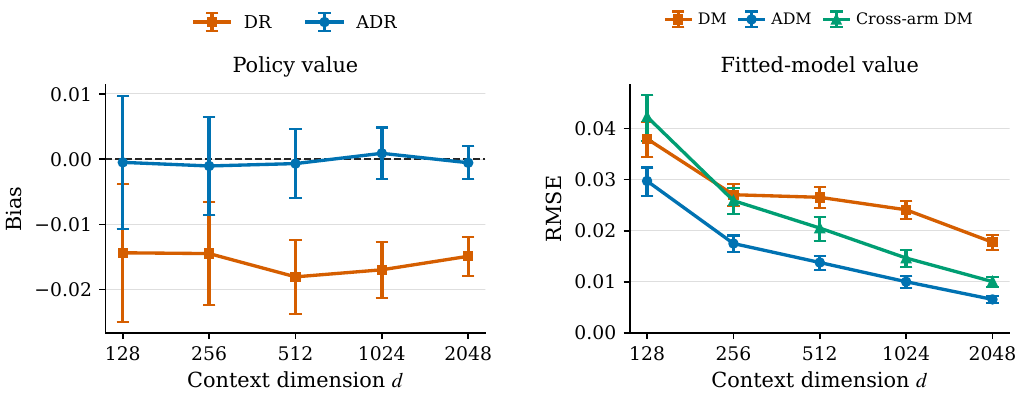}
\caption{Value estimation for identical reward models selected by the corrected criterion, with \(m=2d\). Left: empirical bias of DR and ADR relative to the true target-policy value. Right: RMSE of DM, ADM, and cross-arm DM relative to the population value induced by the fitted model in each dataset. Bars give pointwise \(95\%\) Monte Carlo intervals over 200 datasets per dimension.}
\label{fig:value-correction}
\end{figure}

\paragraph{Value estimation at the same fitted model.}
We next isolate the effect of the value correction. Returning to the two-arm experiment with \(m=2d\), we select the reward model using the corrected criterion and compute DR and ADR from the same fitted networks, stopping times, and logged observations. Figure~\ref{fig:value-correction} (left) shows that correction reduces the persistent bias of same-sample DR. At \(d=2048\), the empirical bias magnitude falls by over \(95\%\), from \(0.0149\) for DR to \(0.0006\) for ADR.

To compare precision for the fitted-model value, we also evaluate DM and ADM against the population value induced by their shared fitted model, \(\mathcal V_{g_{\widehat{\mathbf t}}}(\nu)\), computed separately for each dataset. This is the target of the ADM guarantee in Theorem~\ref{thm:value_estimation}. Cross-arm DM provides an independent-evaluation comparison: it evaluates each selected arm model on contexts logged under the other arm, avoiding reuse of its training observations. ADM reduces MSE relative to cross-arm DM by \(50.7\%\)–\(57.4\%\) across the five dimensions, with every paired \(95\%\) interval favoring ADM (Figure~\ref{fig:value-correction}, right; Appendix~\ref{app:exp-stopping}).

\paragraph{Comparison with cross-fitted DR.}
The corrected criterion also improves model selection within ten-fold cross-fitted DR (CF10). Our inner-validation CF10 baseline selects the stopping time on an inner split before refitting on the complete training complement. Selecting directly on that complement with the corrected criterion avoids this additional training-size mismatch while keeping the evaluation fold excluded from fitting and selection. Corrected stopping reduces the complement models’ prediction error by \(19.3\%\)–\(42.3\%\) relative to inner validation (Appendix~\ref{app:exp-fixed-ratio}). We compare both CF10 procedures with ADR, which trains and selects on all available observations for each arm.

\begin{figure}[ht]
\centering
\includegraphics[width=.7\textwidth]{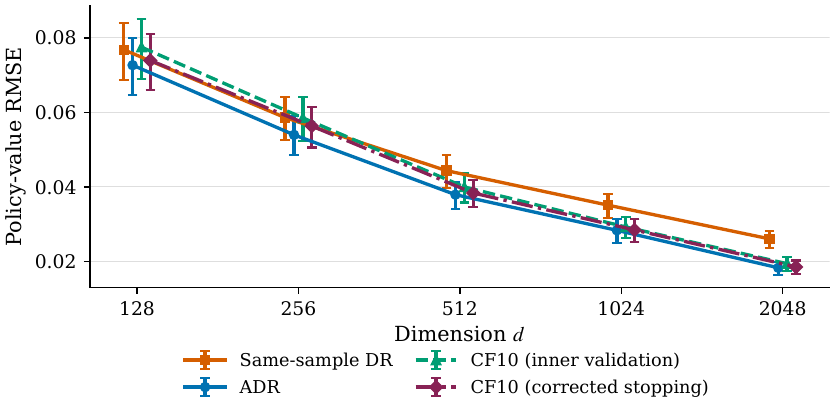}
\caption{Policy-value RMSE with \(m=2d\). ADR and uncorrected same-sample DR use the same full-data fits selected by the corrected risk criterion. CF10 uses either corrected stopping on each complete training complement or inner validation followed by refitting. Bars give pointwise \(95\%\) Monte Carlo intervals over the same 200 datasets per dimension.}
\label{fig:fixed-ratio-value}
\end{figure}

Figure~\ref{fig:fixed-ratio-value} shows that ADR has similar observed MSE to corrected-stopping CF10, with substantially lower runtime (Appendix~\ref{app:exp-runtime}). At \(d=2048\), their RMSEs are \(0.0183\) and \(0.0185\), respectively. ADR has lower MSE point estimates at every dimension, with most paired \(95\%\) intervals including zero. At \(d=2048\), ADR also has lower MSE than IPS and self-normalized IPS (SNIPS), with paired \(95\%\) intervals favoring ADR. Appendix~\ref{app:exp-fixed-ratio} reports the paired comparisons and bias–variance decomposition.

\begin{figure}[ht]
\centering
\includegraphics[width=0.85\textwidth]{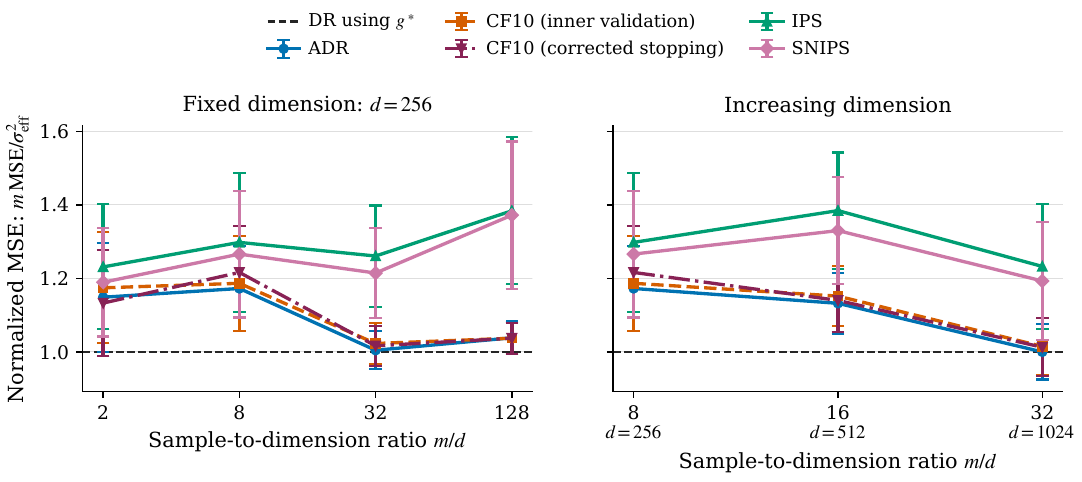}
\caption{Policy-value MSE normalized by \(\sigma_{\mathrm{eff}}^2/m\). Left: fixed dimension \(d=256\) with increasing sample size. Right: increasing dimension and sample-to-dimension ratio. The dashed line marks the oracle DR benchmark. Bars give pointwise \(95\%\) Monte Carlo intervals.}
\label{fig:ratio-precision}
\end{figure}

\paragraph{Value estimation under increasing sample-to-dimension ratio.}
We next examine how the value estimation precision comparison changes as sample size outgrows dimension, using the oracle DR benchmark introduced in Section~\ref{subsec:theory-value}. Finally, we examine value estimation as \(m/d\) increases, both at fixed \(d=256\) and while dimension grows. Figure~\ref{fig:ratio-precision} reports MSE normalized by \(\sigma_{\mathrm{eff}}^2/m\), the MSE of DR using the true reward function. At the largest sample sizes, ADR and both CF10 procedures have similar MSEs near this oracle benchmark. At \((d,m)=(1024,32768)\), the estimated normalized MSEs of ADR and corrected-stopping CF10 are \(1.000\) and \(1.013\), respectively. ADR also has lower MSE than IPS and SNIPS at both endpoints with \(m=32768\). Appendix~\ref{app:exp-increasing-ratio} gives the implementation details and paired comparisons.

Across the tested regimes, ADR achieves precision comparable to corrected-stopping CF10, with substantially lower runtime in the matched benchmarks (Appendix~\ref{app:exp-runtime}).

We report the wall-clock runtime for complete ADR, cross-validation, and cross-fitting procedures, including model selection and refitting where required in Appendix~\ref{app:exp-runtime}. 

\section{Discussion}
\label{sec:discussion}

We study how to select a reward model trained on the full logged sample and estimate policy value using the same observations. In the proportional regime, our Onsager-corrected, target-policy-weighted criterion consistently estimates conditional test risk and yields near-oracle stopping among the candidate gradient descent iterates. The correction also enables consistent estimation of the fitted-model value through ADM~\eqref{eq:adm} and the true policy value through ADR~\eqref{eq:adr}, without requiring reward prediction error to vanish.

The experiments in Section~\ref{sec:experiments} demonstrate benefits for both full-sample learning and cross-fitting. The corrected criterion improves full-data stopping selection over cross validation and two holdout sample schedules. For estimating the policy value induced by the fitted model, ADM achieves lower RMSE than both (uncorrected) DM on the full sample and cross-arm evaluation. For doubly robust value estimation with 10-fold cross-fitting (CF10), our Onsager-corrected criterion selects directly on each complete training complement and yields reward models with lower mean target-policy-weighted prediction risk than standard inner validation followed by refitting. Full-data ADR achieves comparable observed RMSE to corrected-stopping CF10 at substantially lower runtime.

Note that our experiments (Figure~\ref{fig:ratio-precision}) show that ADR and CF10 attain similar MSEs near the oracle DR benchmark as sample size outgrows dimension. These findings motivate for future work a fluctuation theory for ADR that accounts for dependence between its Onsager-corrected scores, as discussed in Section~\ref{subsec:theory-value}. Two goals remain: to derive a central limit theorem (CLT) for ADR in the proportional regime and to identify conditions for recovering the same limit as the oracle DR estimator, when \(d\to\infty\) with \(d/m\to0\). For example, showing that ADR differs from \(P_{m,d}\psi_{g^\ast}\) by \(o_p(m^{-1/2})\) would give ADR the same limiting distribution whenever the oracle estimator satisfies a CLT.

Finally, our analysis assumes Gaussian contexts and known context-independent logging. Promising results on experiments with non-Gaussian designs (see Appendix~\ref{app:exp-nongaussian}) motivate extending our weighted state-evolution analysis beyond Gaussian designs. An important next step is also to extend our analysis to context-dependent logging: since it changes the distribution of contexts within each arm’s training sample, it requires a new characterization of the resulting training dynamics and correction. 

\bibliography{refs}
\bibliographystyle{plainnat}

\appendix
\renewcommand{\thepart}{}
\renewcommand{\partname}{}
\part{Appendix} \etocsettocstyle{}{} \localtableofcontents

After related work, Appendices~\ref{sec:onsager} and~\ref{sec:formal-regime} give the computable Onsager recursion and formal assumptions. The proof has five steps. First, we condition on the logged actions to obtain an arm-wise regression problem (Proposition~\ref{prop:supp:arm-wise_conditioning}). Second, we append the target-policy index to the state evolution (Proposition~\ref{prop:supp:auxiliary_gfom_policy_index}). Third, we compare corrected training averages with independent-context expectations (Proposition~\ref{prop:supp:weighted-observable-transfer}). Fourth, we replace resampled training noise with independent test noise (Proposition~\ref{prop:independent-noise-comparison}). Finally, a finite union over the iterate grid yields the selection and ADM/ADR consistency results (Appendices~\ref{sec:appendix-reduction} and~\ref{app:proof-value-estimation}). Appendix~\ref{sec:appendix-quantitative-refinements} provides rates under a stronger envelope-moment condition. Appendix~\ref{app:same-sample-empirical-process} characterizes uncorrected same-sample DR error and its connection to classical empirical-process conditions for first-order inference.

\section{Related Work}
\label{app:related-work}

\paragraph{Reward models in off-policy evaluation.}
Classic OPE baselines include the direct method (DM)~\citep{beygelzimer2009offset}, inverse propensity scoring (IPS)~\citep{precup2000eligibility,strehl2010learning}, and doubly robust (DR) estimators~\citep{dudik2011,dudik2014}. Subsequent work improves the value estimator's bias--variance tradeoff through sample-wise interpolation between DR and DM (SWITCH~\citep{wang2017optimal}), variance-targeted reward modeling (MRDR~\citep{farajtabar2018more}), adaptive blending (CAB~\citep{su2019cab}), and smooth importance-weight corrections (DRos~\citep{su2020doubly} and IPS-$\lambda$/DR-$\lambda$~\citep{metelli2021subgaussian}). Our criterion estimates the target-policy-weighted test risk of a neural reward model fitted on the full sample. The corresponding prediction error bounds the squared difference between the model-induced and true policy values, and the additional variance of a DR score evaluated on independent observations.

\paragraph{Semiparametric estimation and sample reuse.}
The same-sample empirical-process term in \eqref{eq:main-empirical-process} is standard in semiparametric analysis~\citep{kennedy2024semiparametric}. When the data distribution is fixed, stochastic equicontinuity of a Donsker empirical process and $L^2$ convergence of the fitted score to the score using the true reward function make this term $o_p(1)$. Cross-fitting uses independent evaluation observations for each nuisance fit~\citep{schick1986asymptotically,chernozhukov2018double}; own-observation effects and cross-fitting also appear in series-based semiparametric estimation~\citep{newey2018cross}. Stability conditions permit inference without splitting for suitable learners~\citep{chen2022debiased}. Our correction establishes risk and value consistency along a neural training trajectory in proportional dimension. In this regime, \citet{jiang2025new} find variance inflation and non-negligible cross-fold covariance for cross-fitted DR with linear outcomes and logistic propensities. These findings motivate future work in studying ADR’s limiting distribution and asymptotic variance, the analysis of which must account for dependence among its corrected scores.

\paragraph{Value estimator selection and reward-model selection.}
Several OPE methods select the value estimator or its hyperparameters. SLOPE~\citep{su2020adaptive,tucker2021improved} tunes parameters that control the value estimator's bias--variance tradeoff, while PAS-IF~\citep{udagawa2023policy} selects value estimators using pseudo-evaluation policies. Cross-validated OPE also selects value estimators and their hyperparameters~\citep{cief2025cross}. \citet{mcgrath2025optimal} study ridge-based estimation of expected conditional covariance under proportional asymptotics. Prediction-optimal nuisance tuning need not minimize the estimator's asymptotic variance. Our selector targets weighted test risk; the corresponding prediction risk bounds the additional variance under independent DR evaluation. A held-out importance-weighted residual criterion estimates weighted test risk for a model trained on the complementary sample~\citep{sugiyama2007covariate}. Cross-validation followed by full-data refitting is therefore a natural comparison for our stopping rule. 

Cross-fitting likewise uses models trained on fewer observations than the full-data fit. Within each fold's training sample, one may use a further validation split to select a stopping iterate and then refit on that training sample. The validation criterion then estimates the risk of the models before refitting, while DR evaluation on the excluded fold uses the refitted model. Our Onsager-corrected criterion can select directly on the complete training complement, while the excluded fold remains independent for DR evaluation; Appendix~\ref{app:exp-fixed-ratio} examines this use.

\paragraph{Reward modeling and neural state evolution.}
Single-index reward models have been studied for online regret minimization through kernel regression, cross-fitting, and staged exploration~\citep{kang2026single,ma2025nonparametric,arya2026kernel}. Our setting concerns offline evaluation and selection of neural reward models in proportional dimension. Related work constructs risk estimates along gradient-descent trajectories for high-dimensional linear and robust regression~\citep{bellec2026uncertainty,tan2024estimating} and develops inference for empirical risk minimization~\citep{han2024gradient}. We build directly on \citet{han2025}, who establish state evolution for finite-width neural networks, a computable Onsager correction, consistent estimation of unweighted test error, and early stopping based on that estimate. We retain their correction for the unweighted arm-wise training loss and extend the joint state evolution to include the linear projections that determine the target-policy probabilities. Tracking the covariance between these projections and the learned features gives the weighted risk-estimation and OPE results proved here.

\section{Details of Onsager Correction} \label{sec:onsager}

The Onsager correction is a linear combination of the earlier hidden-state loss gradients $\Psi_a^{(0)},\ldots,\Psi_a^{(t-1)}$ defined in Section~\ref{subsec:corrected-hidden-state}. Its coefficients are $q\times q$ matrices $\widehat\rho_{a,t,s}$, $s=1,\ldots,t$, computed from the same training trajectory by the recursion of \cite{han2025}. We give that recursion below. For observation \(i\in\mathcal I_a\) and step \(r\), write its loss as a function of the hidden state:
\[
    \ell_{i,a}^{(r)}(u)
    \coloneqq
    \frac12
    \left(
        R_i - G_{W_{a,[2:L]}^{(r)}}(u)
    \right)^2,
    \qquad u\in\mathbb R^q,
\]
where \(G_{W_{a,[2:L]}^{(r)}}\) denotes the scalar network output obtained by applying the deeper layers at step \(r\) to the hidden state \(u\). Then, for each row $i\in\Is_a$ and each step $r$, define the local loss Hessian
\[
    J_{i,a}^{(r)}
    \coloneqq
    \nabla_u^2 \ell_{i,a}^{(r)}(u)
    \big|_{u=\widetilde U_{i,a}^{(r)}}
    \in\R^{q\times q}.
\]
Equivalently, $J_{i,a}^{(r)}$ is the Jacobian of the loss gradient with respect to the hidden state. Automatic differentiation or the explicit forward--backward derivative recursion in \cite{han2025} computes these Hessians. Fix an ordering $\Is_a=\{i_{a,1},\ldots,i_{a,N_a}\}$. For $t\ge1$, define the block-diagonal matrix
\[
    \widehat{\bm L}_{a,k}^{[t-1]}
    \coloneqq
    \operatorname{diag}
    \left(
        \eta_{a,1}^{(0)}J_{i_{a,k},a}^{(0)},
        \eta_{a,1}^{(1)}J_{i_{a,k},a}^{(1)},
        \ldots,
        \eta_{a,1}^{(t-1)}J_{i_{a,k},a}^{(t-1)}
    \right),
    \qquad k=1,\ldots,N_a.
\]
This is a $t\times t$ block matrix with $q\times q$ blocks. Let \(I_{t,q}\coloneqq I_t\otimes I_q\in\mathbb R^{tq\times tq}\), viewed as a \(t\times t\) block identity with \(q\times q\) diagonal blocks. For a $(t-1)\times(t-1)$ block matrix $B$, define the one-step block-shift operator
\[
    \mathfrak O_t(B)
    \coloneqq
    \begin{pmatrix}
        0 & 0 \\
        B & 0
    \end{pmatrix},
\]
with the convention $\mathfrak O_1(\emptyset)=0$. Starting from $\widehat{\bm\rho}_a^{[0]}=\emptyset$, compute
\[
    \widehat{\bm\tau}_a^{[t]}
    \coloneqq
    -
    \frac1{N_a}
    \sum_{k=1}^{N_a}
    \left[
        I_{t,q}
        +
        \phi_a^{-1}
        \widehat{\bm L}_{a,k}^{[t-1]}
        \mathfrak O_t(\widehat{\bm\rho}_a^{[t-1]})
    \right]^{-1}
    \widehat{\bm L}_{a,k}^{[t-1]},
\]
and
\[
    \widehat{\bm\rho}_a^{[t]}
    \coloneqq
    I_{t,q}
    +
    \left(
        \widehat{\bm\tau}_a^{[t]}
        +
        I_{t,q}
    \right)
    \mathfrak O_t(\widehat{\bm\rho}_a^{[t-1]}).
\]
We write $\widehat\rho_{a,t,s}\in\R^{q\times q}$ for the $(t,s)$ block of $\widehat{\bm\rho}_a^{[t]}$. To see that the recursion is well defined, set
\begin{equation*}
  O_t\coloneqq\mathfrak O_t(\widehat{\bm\rho}_a^{[t-1]}),
  \qquad
  B_{a,k}^{[t]}\coloneqq I_{t,q}
    +\phi_a^{-1}\widehat{\bm L}_{a,k}^{[t-1]}O_t.
\end{equation*}
We show by induction that $\widehat{\bm\rho}_a^{[t]}$ is block lower triangular with identity diagonal blocks. At $t=1$, $O_1=0$ and $B_{a,k}^{[1]}=\widehat{\bm\rho}_a^{[1]}=I_q$. Suppose the assertion holds at $t-1$. Then $O_t$ is strictly block lower triangular, as is $\widehat{\bm L}_{a,k}^{[t-1]}O_t$ because $\widehat{\bm L}_{a,k}^{[t-1]}$ is block diagonal. Thus, $B_{a,k}^{[t]}$ is block lower triangular with identity diagonal blocks. In particular, $\det B_{a,k}^{[t]}=1$, and its inverse has the same triangular form. Consequently,
\begin{equation*}
  Q_{a,k}^{[t]}\coloneqq(B_{a,k}^{[t]})^{-1}\widehat{\bm L}_{a,k}^{[t-1]},
  \qquad
  \widehat{\bm\tau}_a^{[t]}=-\frac1{N_a}\sum_{k=1}^{N_a}Q_{a,k}^{[t]}
\end{equation*}
are block lower triangular. The product $(\widehat{\bm\tau}_a^{[t]}+I_{t,q})O_t$ is strictly block lower triangular, so adding $I_{t,q}$ proves the induction for $\widehat{\bm\rho}_a^{[t]}$.

The same recursion also permits us to reuse the matrices computed at the preceding iteration. For $t\ge2$, suppose inductively that the leading $(t-2)\times(t-2)$ block of $\widehat{\bm\rho}_a^{[t-1]}$ equals $\widehat{\bm\rho}_a^{[t-2]}$; for $t=2$ this is the empty-block convention. The leading $(t-1)\times(t-1)$ block of $O_t$ is then $O_{t-1}$. Let $b_{a,k}^{[t]}\in\R^{q\times(t-1)q}$ denote the last block row of $B_{a,k}^{[t]}$, excluding its diagonal block. The block forms and a direct multiplication give
\begin{align*}
  B_{a,k}^{[t]}
  &=\begin{pmatrix}
      B_{a,k}^{[t-1]}&0\\
      b_{a,k}^{[t]}&I_q
    \end{pmatrix},\\
  (B_{a,k}^{[t]})^{-1}
  &=\begin{pmatrix}
      (B_{a,k}^{[t-1]})^{-1}&0\\
      -b_{a,k}^{[t]}(B_{a,k}^{[t-1]})^{-1}&I_q
    \end{pmatrix}.
\end{align*}
Multiplying by the block-diagonal loss Hessian matrix therefore yields
\begin{align*}
  Q_{a,k}^{[t]}
  &=(B_{a,k}^{[t]})^{-1}
    \begin{pmatrix}
      \widehat{\bm L}_{a,k}^{[t-2]}&0\\
      0&\eta_{a,1}^{(t-1)}J_{i_{a,k},a}^{(t-1)}
    \end{pmatrix}\\
  &=\begin{pmatrix}
      Q_{a,k}^{[t-1]}&0\\
      -b_{a,k}^{[t]}Q_{a,k}^{[t-1]}
      &\eta_{a,1}^{(t-1)}J_{i_{a,k},a}^{(t-1)}
    \end{pmatrix}.
\end{align*}
In particular, the leading blocks of $Q_{a,k}^{[t]}$ and $\widehat{\bm\tau}_a^{[t]}$ agree with their preceding values. The leading block of $\widehat{\bm\rho}_a^{[t]}$ is consequently
\begin{equation*}
  I_{t-1,q}
  +(\widehat{\bm\tau}_a^{[t-1]}+I_{t-1,q})
    \mathfrak O_{t-1}(\widehat{\bm\rho}_a^{[t-2]})
  =\widehat{\bm\rho}_a^{[t-1]},
\end{equation*}
which completes the induction. Thus, the system $B_{a,k}^{[t]}Q_{a,k}^{[t]}=\widehat{\bm L}_{a,k}^{[t-1]}$ can be solved by updating only the new block row. This is block forward substitution using the cached solution, and requires no dense inverse.

Algorithm~\ref{alg:onsager-postprocessing} summarizes the resulting post-processing computation.
\begin{algorithm}[t]
\caption{Arm-wise Onsager post-processing}
\label{alg:onsager-postprocessing}
\begin{algorithmic}[1]
\Require Arm-wise rewards $\{R_i:i\in\mathcal I_a\}$ and trajectory
$\{W_a^{(r)},\widetilde U_a^{(r)},\Psi_a^{(r)}\}_{r=0}^{T}$,
first-layer step sizes $\{\eta_{a,1}^{(r)}\}_{r=0}^{T-1}$, and $\phi_a$
\State $\widehat{\bm\rho}_a^{[0]}\gets\emptyset$ and
$\widehat U_a^{(0)}\gets\widetilde U_a^{(0)}$; set
$\widehat{\bm L}_{a,k}^{[-1]},Q_{a,k}^{[0]}\gets\emptyset$ for every $k$
\For{$t=1,\ldots,T$}
  \For{$k=1,\ldots,N_a$}
    \State Compute $J_{i_{a,k},a}^{(t-1)}$ and append
    $\eta_{a,1}^{(t-1)}J_{i_{a,k},a}^{(t-1)}$ to the cached block diagonal
    $\widehat{\bm L}_{a,k}^{[t-2]}$
    \State Set $B_{a,k}^{[t]}\gets I_{t,q}
    +\phi_a^{-1}\widehat{\bm L}_{a,k}^{[t-1]}
    \mathfrak O_t(\widehat{\bm\rho}_a^{[t-1]})$
    \State Extend the cached solution $Q_{a,k}^{[t-1]}$ by computing only
    the new block row of
    $B_{a,k}^{[t]}Q_{a,k}^{[t]}=\widehat{\bm L}_{a,k}^{[t-1]}$
    by block forward substitution
  \EndFor
  \State $\widehat{\bm\tau}_a^{[t]}
  \gets-N_a^{-1}\sum_{k=1}^{N_a}Q_{a,k}^{[t]}$
  \State Extend the cached $\widehat{\bm\rho}_a^{[t-1]}$ by computing
  the new block row of $\widehat{\bm\rho}_a^{[t]}$ from the recursion above
  \State Form $\widehat U_a^{(t)}$ from
  \eqref{eq:onsager-correction-row}
\EndFor
\State Use $\{\widehat U_a^{(t)}\}_{t=0}^T$ in
\eqref{eq:weighted-criterion}; after arm-wise selection, compute
\eqref{eq:adm}--\eqref{eq:adr}
\end{algorithmic}
\end{algorithm}

\paragraph{Implementation and complexity.}
The algorithm reuses the leading solution and Onsager blocks from the preceding iteration. Computing the new block rows at iteration $t$ costs $O(N_aq^3t^2)$ operations for one arm, and all $T$ iterations cost
\begin{equation*}
  \sum_{t=1}^T O(N_aq^3t^2)
  =O(N_aq^3T^3).
\end{equation*}
At the final iteration, each row cache has $T^2$ blocks of size $q\times q$, so the block cache stores $O(N_aT^2q^2)$ numbers. Computing the local derivatives adds $O(N_aLq^3T)$ operations. For fixed $L$, this cost remains within the correction bound. These costs are additional to full-batch training, whose first-layer operations cost $O(N_adqT)$. End-to-end storage includes $O(N_ad)$ for the design and $O(dqT)$ for the first-layer parameter trajectory, in addition to the correction cache and lower-order fixed-width arrays. Thus, when $d\asymp N_a$ and $K,L,q,T$ are fixed, the correction preserves the leading asymptotic dependence on sample size. Its finite-sample overhead can be substantial.

\section{Notation and Formal Assumptions for the Working Regime}
\label{sec:formal-regime}

The working regime uses the i.i.d. observations of Section~\ref{sec:stat-framework}, with Gaussian contexts, single-index conditional reward means, known context-independent logging under positivity, and a target policy specified independently of the logged sample with the stated multi-index importance weights. We take $m/d\to\phi\in(0,\infty)$, with $K,L,q,\ell$ fixed and $\Ts\subset\mathbb Z_{\ge0}$ finite, nonempty, and fixed. The noise and initialization conditions used in the proofs are specified below.

Throughout this appendix, HI25 denotes \citet{han2025}. Unless stated otherwise, $C_{t,r},c_{t,r}\in(0,\infty)$ are uniform in $d$, $N_a$, and the arm $a$. They may depend on the fixed iteration $t$, moment order $r$, and fixed model and regularity parameters, and their values may change from line to line. When $r$ is fixed, we suppress it and write $C_t,c_t$. We set maxima over empty index sets equal to zero. Unsubscripted vector norms are Euclidean and unsubscripted matrix norms are Frobenius; $\|\cdot\|_\infty$ denotes the largest absolute entry and $\|\cdot\|_{\op}$ the spectral norm. For a nonnegative random quantity $Y_d$, the notation $Y_d=d^{o_p(1)}$ denotes a subpolynomial upper bound: $d^{-\epsilon}Y_d\xrightarrow{p}0$ for every fixed $\epsilon>0$. For integers $r\le s$, write $[r:s]=\{r,r+1,\ldots,s\}$ and $[n]=[1:n]$. We write $z^{([r:s])}=(z^{(r)},\ldots,z^{(s)})$. Thus, $\R^{q[r:s]}$ denotes the product of $s-r+1$ copies of $\R^q$. We write $M_{k\cdot}$ and $M_{\cdot j}$ for a matrix's $k$th row and $j$th column. When forming augmented matrices, $(U,V)$ denotes column concatenation of matrices with the same number of rows. For a real-valued function $f$, its Lipschitz seminorm is $\|f\|_{\mathrm{Lip}}=\sup_{x\ne y}|f(x)-f(y)|/\|x-y\|$. A function $f:\R^n\to\R$ is $\Lambda_f$-pseudo-Lipschitz of order $p\ge1$ if
\begin{equation*}
  |f(x)-f(y)|
  \le \Lambda_f(1+\|x\|+\|y\|)^{p-1}\|x-y\|
  \qquad\text{for all }x,y\in\R^n.
\end{equation*}
When this condition is imposed on a family of functions, the order and constant are uniform over the family.

Since the logging policy is context-independent, the sample decomposes arm-wise. For each arm $a$, let $\Is_a\coloneqq\cl{i\in[m]:A_i=a}$ and $N_a\coloneqq|\Is_a|$. Because $\mu(a)\ge c>0$ and $K$ is fixed, $\Prob\p{\min_{a\in[K]}N_a\ge1}\to1$. Any arm-wise quantity whose formula requires $N_a\ge1$ is defined by that formula on $\{N_a\ge1\}$. On $\{N_a=0\}$, use any measurable definition. Because this event has vanishing probability, every such convention gives the same asymptotic statements.

Define the full training sigma-field by
\begin{equation*}
    \mathcal W^{(0)}
    \coloneqq
    \bigl(W_1^{(0)},\ldots,W_K^{(0)}\bigr),
    \qquad
    \mathcal F_{\mathrm{tr}}
    \coloneqq
    \sigma\p{\Ds_m,\mathcal W^{(0)}}.
\end{equation*}
If $\tilde g$ is $\mathcal F_{\mathrm{tr}}$-measurable and $(X^{\mathrm{new}},A^{\mathrm{new}},R^{\mathrm{new}})$ is a draw from the data law independent of $\mathcal F_{\mathrm{tr}}$, then
\begin{align*}
    \Rs_\nu(\tilde g)
    &\coloneqq
    \mathbb{E}\br{
        w_{A^{\mathrm{new}}}(X^{\mathrm{new}})
        \p{R^{\mathrm{new}}-\tilde g(X^{\mathrm{new}},A^{\mathrm{new}})}^2
        \mymid
        \mathcal F_{\mathrm{tr}}
    },\\
    \Vs_{\tilde g}(\nu)
    &\coloneqq
    \mathbb{E}\br{
        \sum_{a=1}^K \nu(a \mymid X^{\mathrm{new}})\,\tilde g(X^{\mathrm{new}},a)
        \mymid
        \mathcal F_{\mathrm{tr}}
    },\\
    \Rs_\nu^\circ(\tilde g)
    &\coloneqq
    \mathbb{E}\br{
        \sum_{a=1}^K
        \nu(a \mymid X^{\mathrm{new}})
        \p{\tilde g(X^{\mathrm{new}},a)-g^\ast(X^{\mathrm{new}},a)}^2
        \mymid
        \mathcal F_{\mathrm{tr}}
    }.
\end{align*}
For an arm-wise $\mathcal F_{\mathrm{tr}}$-measurable predictor $\tilde g_a:\mathbb R^d\to\mathbb R$, we similarly write
\begin{equation*}
    \Rs_{\nu,a}^\circ(\tilde g_a)
    \coloneqq
    \mathbb E\br{
        \nu(a\mymid X^{\mathrm{new}})
        \p{\tilde g_a(X^{\mathrm{new}})-g^\ast(X^{\mathrm{new}},a)}^2
        \mymid
        \mathcal F_{\mathrm{tr}}
    }.
\end{equation*}
The signed value functional $\Vs_{\tilde g}(\nu)$ is defined when
\begin{equation*}
  \int\sum_{a=1}^K\nu(a\mymid x)|\tilde g(x,a)|\,dP_X(x)<\infty
  \qquad\text{almost surely},
\end{equation*}
and its conditional expectation is the corresponding integral with the training sample and network initializations held fixed. Every fitted predictor considered below satisfies this condition by \Cref{prop:prediction-moment-bound}. When $\tilde g$ is deterministic, these reduce to unconditional definitions in the main text.

\paragraph{Neural-network architecture.}
We embed the scalar-output network in the $q$-dimensional output architecture of \citet{han2025} by setting its remaining output coordinates to zero. The layer index is denoted by $\alpha\in[1:L]$. For each arm $a$, the first layer has $W_{a,1}\in\R^{d\times q}$ and each deeper layer has $W_{a,\alpha}\in\R^{q\times q}$ for $\alpha\in[2:L]$. Let $\sigma_\alpha:\R\to\R$ be the activation map at layer $\alpha$, applied coordinatewise to vectors and matrices, with $\sigma_0=\sigma_L=\mathrm{id}$. For a single input $x\in\R^d$, define the vector-valued forward pass
\[
    h_{a,W;0}(x)=x,
    \qquad
    h_{a,W;\alpha}(x)
    =
    W_{a,\alpha}^\top
    \sigma_{\alpha-1}\bigl(h_{a,W;\alpha-1}(x)\bigr),
    \qquad
    \alpha\in[1:L].
\]
For a generic arm-wise parameter $W$ of this shape, the padded network output is $F_W(x)\coloneqq h_{a,W;L}(x)\in\R^q$, and the scalar predictor used in the main text is its first coordinate, $f_W(x)=e_1^\top F_W(x)$. Equivalently, for a deeper-layer tuple $B=(B_2,\ldots,B_L)\in(\R^{q\times q})^{[2:L]}$ and a first-layer hidden state $u\in\R^q$, let
\[
    z_{B,1}(u)=u,
    \qquad
    z_{B,\alpha}(u)
    =
    B_\alpha^\top
    \sigma_{\alpha-1}\bigl(z_{B,\alpha-1}(u)\bigr),
    \qquad
    \alpha\in[2:L].
\]
We write $\Gsc_B(u)\coloneqq z_{B,L}(u)\in\R^q$ for the padded deeper-layer output and $G_B(u)\coloneqq e_1^\top\Gsc_B(u)$ for the scalar output used in the main text. Thus, $f_W(x)=G_{W_{[2:L]}}(x^\top W_1)$. We measure the size of a deeper-layer tuple by
\begin{equation*}
  \kappa_B\coloneqq\prod_{\alpha=2}^L(1\vee\|B_\alpha\|_{\op}).
\end{equation*}

For $i\in\Is_a$, write the padded response as $R_{i,[q]}\coloneqq(R_i,0,\ldots,0)^\top\in\R^q$. The arm-wise padded squared loss is
\[
    \mathsf L_a(W)
    \coloneqq
    \frac{1}{2N_a}
    \sum_{i\in\Is_a}
    \bigl\|R_{i,[q]}-F_W(X_i)\bigr\|^2.
\]
To represent the scalar-output network, initialize $(W_{a,L}^{(0)})_{\cdot j}=0$ for $j=2,\ldots,q$ and draw the entire initialization collection $\mathcal W^{(0)}$ independently of the logged sample. The first column produces the scalar prediction, while the remaining output coordinates and their corresponding loss gradients are zero. These coordinates remain zero at every gradient step, so the padded loss and all parameter updates agree exactly with those of the scalar-output network in Section~\ref{sec:setting}. The same zero-column invariance holds for the auxiliary and state-evolution deeper-layer trajectories, whose update maps use the same padded gradient. The prescribed learning rates are positive, as in HI25. Every layer is updated simultaneously:
\[
    W_{a,\alpha}^{(s+1)}
    =
    W_{a,\alpha}^{(s)}
    -
    \eta_{a,\alpha}^{(s)}
    \nabla_{W_{a,\alpha}}\mathsf L_a\bigl(W_a^{(s)}\bigr),
    \qquad
    \alpha\in[1:L],\quad s\ge0.
\]

\paragraph{Initialization conventions.}
We use HI25's initialization convention
\begin{equation}
    V_{a,[2:L]}^{(0)}
    \coloneqq
    W_{a,[2:L]}^{(0)}.
    \label{eq:state-evolution-initialization}
\end{equation}
For the corresponding scaled first-layer auxiliary parameter, set $v_{1,a}^{(0)}\coloneqq\sqrt d\,W_{a,1}^{(0)}$. For the theoretical corrected state, which HI25 defines for positive iterations, we additionally set
\begin{equation*}
    U_a^{(0)}
    \coloneqq
    X_{\Is_a}W_{a,1}^{(0)}
    =
    \widehat U_a^{(0)}.
\end{equation*}
Thus, every comparison between $W_{a,[2:L]}^{(t)}$ and $V_{a,[2:L]}^{(t)}$ is exact at $t=0$. In particular, statements indexed by the grid $\mathcal T\subset\mathbb Z_{\ge0}$ include the initialization even when the imported HI25 estimate is stated only for positive iterations.

\begin{assumption}[Arm-wise neural-network regularity]
\label{ass:formal-nn-regularity}
There exist constants $\Lambda\ge2$ and $r_0\ge0$, independent of $d$ and $a$, such that the following hold.
\begin{enumerate}
    \item The learning rates are uniformly bounded:
    \[
        \max_{a\in[K]}
        \max_{\alpha\in[1:L]}
        \max_{s\ge0}
        \eta_{a,\alpha}^{(s)}
        \le \Lambda .
    \]

    \item The activation maps satisfy $\sigma_\alpha\in C^4(\R)$ and
    \[
        \max_{p\in[4]}
        \max_{\alpha\in[1:L]}
        \bigl(
            \|\sigma_\alpha^{(p)}\|_\infty
            +
            |\sigma_\alpha(0)|
        \bigr)
        \le \Lambda .
    \]

    \item The reward links satisfy $\varphi_a^\ast\in C^3(\R)$ and
    \[
        \max_{a\in[K]}
        \max_{p\in[0:3]}
        \sup_{u\in\R}
        \frac{|(\varphi_a^\ast)^{(p)}(u)|}{(1+|u|)^{r_0}}
        \le \Lambda .
    \]
\end{enumerate}
\end{assumption}
The three parts are the arm-wise versions of HI25 Assumptions~(A3)--(A5), respectively. The convention $\sigma_L=\mathrm{id}$ in the architecture above makes the final layer linear.

For the comparison with independent test noise, we assume that, for each dimension $d$ and arm $a$, the noise is independent of $X_i$ conditional on $A_i=a$, with law and moments
\begin{equation}
    \varepsilon_i \mymid (A_i=a,X_i)
    \sim
    P_{\xi,a,d},
    \qquad
    \mathbb E[\xi_{a,d}]=0,
    \qquad
    \mathbb E[\xi_{a,d}^2]=\sigma_{\xi,a,d}^2<\infty.
    \label{eq:noise-setup}
\end{equation}
Below, we suppress the dependence of $\xi_{a,d}$ and $\sigma_{\xi,a,d}^2$ on $d$. Under this triangular-array convention, the independent-noise comparison uses the following conditions on the empirical first and second moments of the noise:
\begin{equation}
    \max_{a\in[K]}
    \abs{\frac{1}{N_a}\sum_{i\in\Is_a}\varepsilon_i}
    =
    O_p\p{d^{-1/2}},
    \qquad
    \max_{a\in[K]}
    \abs{
        \frac{1}{N_a}\sum_{i\in\Is_a}\varepsilon_i^2
        -
        \sigma_{\xi,a,d}^2
    }
    \xrightarrow{p}
    0.
    \label{eq:noise-rowwise-lln}
\end{equation}
The sub-Gaussian noise condition in Assumption~\ref{ass:single-index-logging} is a sufficient condition for \eqref{eq:noise-rowwise-lln}.

Following \HI, define the arm-wise envelope
\begin{equation}
    \varkappa_{\ast,a}
    \coloneqq
    1 + \|\sqrt d\,\beta_a\|_\infty
    + \|\xi_{\Is_a}\|_\infty
    + \|\sqrt d\,W_{a,1}^{(0)}\|_\infty
    + \max_{\alpha \in [2:L]}\|W_{a,\alpha}^{(0)}\|_{\op},
    \qquad
    a \in [K],
    \label{eq:arm-wise_kappa}
\end{equation}
where $\xi_{\Is_a} \coloneqq (\varepsilon_i)_{i \in \Is_a}$, and assume that
\begin{equation}
    \max_{a \in [K]}\varkappa_{\ast,a} = d^{o_p(1)}.
    \label{eq:arm-wise_admissibility}
\end{equation}
Because $K$ is fixed, this implies, for every fixed arm $a$ and fixed $C,c>0$,
\begin{equation}
  \varkappa_{\ast,a}=d^{o_p(1)},
  \qquad
  (\varkappa_{\ast,a}\log d)^C d^{-c}=o_p(1).
  \label{eq:envelope-polynomial-vanishing}
\end{equation}
For example, consider Gaussian initialization with the scaling used in HI25: $\sqrt d\,W_{a,1}^{(0)}\in\R^{d\times q}$ and the hidden-layer matrices $q^{1/2}W_{a,\alpha}^{(0)}$ have independent centered Gaussian entries with variances bounded uniformly in $d$ for $2\le\alpha\le L-1$. The same condition holds for the first column $q^{1/2}(W_{a,L}^{(0)})_{\cdot 1}$ of the final layer. If the noise is sub-Gaussian with a dimension-free parameter and $\|\sqrt d\,\beta_a\|_\infty$ is uniformly bounded, then the noise maximum and initialization terms are $O_p(\sqrt{\log d})$. Hence, $\varkappa_{\ast,a}=O_p(\sqrt{\log d})$, which satisfies \eqref{eq:arm-wise_admissibility}. The formal results use the weaker envelope condition.

Separately, we impose delocalization of the fixed index directions used by the target-policy weights:
\begin{equation*}
    \max_{j \in [\ell]}\sqrt d\,\|\gamma_j\|_\infty \le C_\Upsilon.
\end{equation*}
For constants $B_h,L_h<\infty$ independent of $d$, we also assume
\begin{equation*}
  h_a\ge0,
  \qquad
  \max_{a\in[K]}\|h_a\|_\infty\le B_h,
  \qquad
  \max_{a\in[K]}\|h_a\|_{\mathrm{Lip}}\le L_h,
  \qquad
  \sum_{a=1}^K\mu(a)h_a(u)=1
\end{equation*}
for every $u\in\R^\ell$. These policy-index conditions are separate from the arm-wise envelope in \eqref{eq:arm-wise_kappa}.

For each arm $a$ and $w\in\R^q$, define the padded reward link
\begin{equation}
  \varphi_{a,[q]}^\ast(w)
  \coloneqq
  \bigl(\varphi_a^\ast(w_1),0,\ldots,0\bigr)^\top.
  \label{eq:arm-component-link}
\end{equation}

To justify using this padded link in the HI25 comparisons, we first allow a different reward link in each output coordinate. Fix the output width $q$. For $\mu_\ast\in\R^d$ and $Y,\xi\in\R^m$, write
\begin{equation*}
  \mu_{\ast,[q]}
  \coloneqq(\mu_\ast\,|\,0_{d\times(q-1)}),
  \qquad
  Y_{[q]}\coloneqq(Y\,|\,0_{m\times(q-1)}),
  \qquad
  \xi_{[q]}\coloneqq(\xi\,|\,0_{m\times(q-1)}).
\end{equation*}
For functions $\varphi_1,\ldots,\varphi_q\in C^3(\R)$, define
\begin{equation*}
  \boldsymbol\varphi(w)
  \coloneqq
  \bigl(\varphi_1(w_1),\ldots,\varphi_q(w_q)\bigr)^\top,
  \qquad w\in\R^q.
\end{equation*}
For a matrix argument, apply $\boldsymbol\varphi$ rowwise. The HI25 pre-gradient maps give the derivatives of the per-observation loss with respect to layer pre-activations. For $k\in[m]$, the first-layer loss gradient, written in column-vector notation, is
\begin{equation*}
 \mathfrak S_k(u,w;V)
 =D\Gsc_V(u)^\top
       \{\Gsc_V(u)-\boldsymbol\varphi(w)-\xi_k e_1\},
 \qquad u,w\in\R^q,
\end{equation*}
where $\Gsc_V$ is the deeper-layer network map defined above. We take $D^j\mathfrak S_k$ with respect to $(u,w)$ and use operator norms for its multilinear derivatives, with $D^0\mathfrak S_k=\mathfrak S_k$ measured in the Euclidean norm.

\begin{lemma}[HI25 comparison results with coordinate-specific reward links]
\label{lem:hi25-component-link}
Suppose that, for fixed $\Lambda\ge2$ and $r_0\ge0$,
\begin{equation*}
  \max_{j\in[q]}
  \max_{p\in[0:3]}
  \sup_{x\in\R}
  \frac{|\varphi_j^{(p)}(x)|}{(1+|x|)^{r_0}}
  \le \Lambda,
  \qquad
  Y_{[q]}=\boldsymbol\varphi(X\mu_{\ast,[q]})+\xi_{[q]}.
\end{equation*}
Assume the working conditions of \Cref{sec:formal-regime}. In each HI25 result below, replace the scalar entrywise reward link by $\boldsymbol\varphi$ and the data identity by the one above, retaining all other hypotheses. Then the following holds.
\begin{enumerate}
\item The value and difference bounds in \HIprop{prop:apr_est_det_fcn} and the forward- and back-propagation derivative bounds in \HIprop{prop:apr_est_derivative} hold. There is a constant $C=C(L,q,\Lambda,r_0)<\infty$ such that, for every $k\in[m]$, $u,w,u',w'\in\R^q$, and deeper-layer tuples $V,V'$,
\begin{equation*}
 \|D^j\mathfrak S_k(u,w;V)\|
 \le C(1+\|V\|+|\xi_k|+\|u\|+\|w\|)^C,
 \qquad j=0,1,2,3.
\end{equation*}
For $j=0,1,2$,
\begin{align*}
 \|D^j\mathfrak S_k(u,w;V)-D^j\mathfrak S_k(u',w';V')\|&\le C(1+\|V\|+\|V'\|+|\xi_k|
              +\|u\|+\|u'\|+\|w\|+\|w'\|)^C\\
 &\qquad\times(\|u-u'\|+\|w-w'\|+\|V-V'\|).
\end{align*}
\item The polynomial finite-horizon state-evolution bounds in \HIprop{prop:apr_est_se} and the averaged auxiliary-GFOM comparison in \HIprop{prop:se_aux_GFOM} hold. The auxiliary iterates satisfy the delocalization bound in \HIprop{prop:GFOM_deloc_asym} with a fixed polynomial growth exponent.
\item The gradient-descent bounds in \HIprop{prop:GD_l2_est}, the comparison between gradient descent and the auxiliary GFOM in \HIprop{prop:err_gd_GFOM}, the corrected-state bounds in \HIlem{lem:U_apr_err_est}, and the Onsager-coefficient comparison in \HIprop{prop:tau_rho_est_err} hold.
\end{enumerate}
The constants in the cited estimates have the same parameter dependence as in HI25, with additional dependence on the fixed width $q$ allowed.
\end{lemma}

\begin{proof}
The feed-forward and back-propagation maps are independent of the response link. Differentiating the squared loss gives the same gradient identity with $\boldsymbol\varphi$ in place of the scalar entrywise link. Moreover,
\begin{equation*}
  D\boldsymbol\varphi(w)
  =
  \operatorname{diag}\bigl(
    \varphi_1'(w_1),\ldots,\varphi_q'(w_q)
  \bigr).
\end{equation*}
For $j=1,2,3$ and $v_1,\ldots,v_j\in\R^q$,
\begin{equation*}
  [D^j\boldsymbol\varphi(w)[v_1,\ldots,v_j]]_b
  =\varphi_b^{(j)}(w_b)\prod_{h=1}^j(v_h)_b,
  \qquad b\in[q].
\end{equation*}
Thus, all mixed partial derivatives vanish. With operator norms on the multilinear derivatives, and with $D^0\boldsymbol\varphi=\boldsymbol\varphi$ measured in the Euclidean norm,
\begin{align*}
  \|D^j\boldsymbol\varphi(w)\|
  &\le C_q\Lambda(1+\|w\|)^{r_0},
  &&j=0,1,2,3,\\
  \|D^j\boldsymbol\varphi(w)-D^j\boldsymbol\varphi(w')\|
  &\le C_q\Lambda(1+\|w\|+\|w'\|)^{r_0}\|w-w'\|,
  &&j=0,1,2.
\end{align*}
The second bound follows by integrating the next derivative along the segment from $w$ to $w'$. These are the polynomial-growth and local-Lipschitz bounds used in the HI25 residual, pre-gradient, and derivative estimates.

To make the derivative bounds explicit, let $\Gsc_V(u)$ denote the network output as a function of its first-layer hidden state, with the deeper weights $V$ fixed. In column-vector notation its loss gradient is
\begin{align*}
 \mathfrak S_k(u,w;V)&=D\Gsc_V(u)^\top
       \{\Gsc_V(u)-\boldsymbol\varphi(w)-\xi_k e_1\},\\
 D_w \mathfrak S_k(u,w;V)&=-D\Gsc_V(u)^\top D\boldsymbol\varphi(w).
\end{align*}
The activation assumptions and the bounds above imply
\begin{align*}
 \|D^j \mathfrak S_k(u,w;V)\|
 &\le C(1+\|V\|+|\xi_k|+\|u\|+\|w\|)^C,
 &&j=0,1,2,3.
\end{align*}
For $j=0,1,2$, the corresponding difference bound is
\begin{align*}
 \|D^j \mathfrak S_k(u,w;V)-D^j \mathfrak S_k(u',w';V')\|
 &\le C(1+\|V\|+\|V'\|+|\xi_k|
              +\|u\|+\|u'\|+\|w\|+\|w'\|)^C\\
 &\qquad\times(\|u-u'\|+\|w-w'\|+\|V-V'\|).
\end{align*}
Here the derivatives $D^j$ are with respect to $(u,w)$, and the second bound follows by the product rule and integration along the segment joining the two arguments, including their deeper weights. In particular, a change in $D\boldsymbol\varphi(w)$ is controlled by $\|w-w'\|$.

Each auxiliary output row depends only on the corresponding input rows. The generic GFOM theorems allow these maps to vary across rows and coordinates. \Cref{lem:supp:input-localization} applies these theorems to smooth input cutoffs and removes the cutoffs using polynomial moment bounds. The deterministic gradient-descent comparison uses the loss gradient identity and the value bounds above. For the Onsager comparison, the finite polynomial expansion of the block inverse is controlled by the sum of the two state envelopes, as verified in \Cref{lem:gd-auxiliary-coefficient-comparison}. These arguments preserve every other hypothesis and give the stated componentwise-link conclusions.

\end{proof}

The derivative inequalities above also cover derivatives in the signal argument $w$. In particular, their difference bound controls changes in $D\boldsymbol\varphi(w)$ directly through $\|w-w'\|$. The state-evolution and auxiliary-GFOM bounds are established in \Cref{lem:supp:input-localization}; the gradient-descent and Onsager-coefficient estimates are established in \Cref{lem:gd-path-localization,lem:gd-auxiliary-coefficient-comparison}. The intermediate-state comparisons used later are stated explicitly in \Cref{lem:arm-wise-hi25-consequences}.

For arm $a$, take $\varphi_1=\varphi_a^\ast$ and $\varphi_j\equiv0$ for $j=2,\ldots,q$. Then, $\boldsymbol\varphi=\varphi_{a,[q]}^\ast$ and
\begin{equation*}
  (R_i,0,\ldots,0)^\top
  =
  \varphi_{a,[q]}^\ast
  \bigl((\beta_a^\top X_i,0,\ldots,0)^\top\bigr)
  +(\varepsilon_i,0,\ldots,0)^\top.
\end{equation*}
The construction applies to every value of $\varphi_a^\ast(0)$ allowed by the regularity assumption.

\begin{corollary}[Arbitrary fixed index directions under Gaussian design]
\label{cor:initialization-rotation}
Assume the working conditions above, except for the coordinate bounds on $\beta_a$ and $\gamma_j$ and the envelope condition \eqref{eq:arm-wise_admissibility}. Retain $\|\beta_a\|=1$, $\Upsilon^\top\Upsilon=I_\ell$, and the noise conditions \eqref{eq:noise-setup}--\eqref{eq:noise-rowwise-lln}. Suppose that the entire initialization vector $\mathcal W^{(0)}$ is independent of the logged sample and that
\begin{equation}
 \kappa_{0,d}\coloneqq
 1+\max_{a\in[K]}\left\{
 \|\xi_{\Is_a}\|_\infty
 +\|\sqrt d\,W_{a,1}^{(0)}\|_\infty
 +\max_{\alpha\in[2:L]}\|W_{a,\alpha}^{(0)}\|_{\op}\right\}
 =d^{o_p(1)}.
 \label{eq:rotation-initialization-envelope}
\end{equation}
Then, the conclusions of \Cref{thm:uniform-risk,thm:selection-main,thm:value_estimation} hold for arbitrary fixed orientations of the reward and policy-index directions.
\end{corollary}

\begin{proof}
Condition on $\mathcal W^{(0)}$ and let
\[
 S=\operatorname{span}\{\beta_a,\gamma_j,
           (W_{a,1}^{(0)})_{\cdot k}:a\in[K],j\in[\ell],k\in[q]\},
 \qquad r=\dim S\le K+\ell+Kq.
\]
Choose an orthonormal basis $B\in\R^{d\times r}$ of $S$ measurably in $\mathcal W^{(0)}$, for example by Gram--Schmidt applied in a fixed order and omitting dependent columns. Let $C\in\R^{d\times r}$ consist of the first $r$ columns of the real discrete cosine transform, so that $C^\top C=I_r$ and $\max_{i,j}|C_{ij}|\le\sqrt{2/d}$. Completing both frames to orthonormal bases by the same measurable procedure gives an orthogonal matrix $Q$ with $Q^\top B=C$. Thus, $Q$ depends only on the initialization and fixed directions. For $v\in S$,
\[
 \sqrt d\,\|Q^\top v\|_\infty
 =\sqrt d\,\|CB^\top v\|_\infty
 \le\sqrt2\,\|B^\top v\|_1
 \le\sqrt{2r}\,\|v\|.
\]
Apply this rotation jointly to the contexts and first-layer parameters:
\[
 X_i'=Q^\top X_i,\qquad \beta_a'=Q^\top\beta_a,
 \qquad \Upsilon'=Q^\top\Upsilon,
 \qquad W_{a,1}'{}^{(0)}=Q^\top W_{a,1}^{(0)}.
\]
Leave actions, rewards, noise, and deeper-layer initializations unchanged. Conditional on $\mathcal W^{(0)}$, the rotated contexts remain i.i.d.\ $N(0,I_d)$, independent of the actions and arm-specific noises. The signal and policy directions satisfy their required coordinate bounds with constant $\sqrt{2(K+\ell+Kq)}$. For each initial first-layer column,
\[
 \|(W_{a,1}^{(0)})_{\cdot k}\|
 \le\|\sqrt d\,W_{a,1}^{(0)}\|_\infty.
\]
Consequently the rotated envelope satisfies $\max_a\varkappa_{\ast,a}'\le C_0\kappa_{0,d}=d^{o_p(1)}$ for a constant $C_0$ depending only on $K,\ell,q$.

The conditional comparison arguments apply with all original initializations retained in the conditioning fields. In particular, condition on $\mathcal W^{(0)}$, the action labels, and the noises in an arm. The rotated directions and parameters are then fixed, and that arm's context rows are still independent standard Gaussians. The bounds in \Cref{lem:arm-wise-hi25-consequences,prop:supp:auxiliary_gfom_policy_index} are uniform in these fixed directions and initializations once their coordinate bounds and envelope are specified. Their error bounds are polynomial in $\varkappa_{\ast,a}'\log d$ times a negative power of $d$, so they vanish in probability under \eqref{eq:rotation-initialization-envelope}. Conditional probability bounds may be truncated at one before averaging over the initialization and noise. The empirical noise moments are unchanged by rotation. Conditioning additionally on all initializations also preserves the independence used when predictions are evaluated on another arm's observations. Thus, the risk and value consistency proofs apply to the rotated system, as does the finite-grid selection argument.

It remains to identify their conclusions with the original quantities. Induction in the simultaneous gradient updates gives
\[
 W_{a,1}'{}^{(t)}=Q^\top W_{a,1}^{(t)},\qquad
 W_{a,\alpha}'{}^{(t)}=W_{a,\alpha}^{(t)}\quad(\alpha\ge2).
\]
Indeed, $X_i'{}^\top W_{a,1}'{}^{(t)}=X_i^\top W_{a,1}^{(t)}$, so hidden states and deeper-layer gradients agree, and the first-layer gradient is multiplied by $Q^\top$. The signal and policy indices, hidden-state Hessians, estimated Onsager coefficients, and corrected predictions therefore agree. Hence, the criteria, selected iterations under the same tie rule, and value estimators agree pathwise.

Keep the original initialization in the rotated training field:
\[
 \sigma(\mathcal W^{(0)},(X_i',A_i,R_i)_{i\in[m]})
 =\sigma(\mathcal W^{(0)},\Ds_m)=\mathcal F_{\mathrm{tr}}.
\]
This equality holds because $Q$ is measurable in the retained initialization and the transformation of contexts is invertible given it; it does not require recovering the original initialization from its rotated weights. For a new Gaussian context $Z$ independent of $\mathcal F_{\mathrm{tr}}$, let $Z'=Q^\top Z$. Its conditional law is still $N(0,I_d)$, and predictions, policy weights, and reward means agree at the coupled contexts. Conditional test risks and fitted-model values are therefore equal. This proves the stated conclusions.
\end{proof}

\section{From Logged Data to Arm-wise State Evolution}

This section reduces the logged bandit data to the setting of \citet{han2025}. Conditional on the action labels and counts, each arm gives a Gaussian single-index regression problem with its own initialization and noise vector. We then append the target-policy index to the auxiliary GFOM iterates. This allows the state-evolution test functions to include the importance weights.

\subsection{Arm-wise conditioning and admissibility}

For $0<c_-<c_+<\infty$, let
\begin{equation*}
  E_{\mathrm{cnt}}
  \coloneqq\{c_-d\le N_a\le c_+d\text{ for every }a\in[K]\}.
\end{equation*}
For each arm $a$, set
\begin{equation*}
  \Ds_a\coloneqq\{(X_i,R_i):i\in\Is_a\},
  \qquad
  \widehat{\mathcal F}_a
  \coloneqq\sigma(\Is_a,X_{\Is_a},R_{\Is_a},W_a^{(0)}).
\end{equation*}
Define
\begin{equation*}
    \mathcal G_a
    \coloneqq
    \sigma\p{A_1,\ldots,A_m,\beta_a,\Upsilon,W_a^{(0)},\xi_{\Is_a}},
    \qquad
    \mathbb E_a^{(0)}[\cdot]
    \coloneqq
    \mathbb E\br{\cdot \mymid \mathcal G_a},
    \qquad
    \Prob_a^{(0)}(\cdot)
    \coloneqq
    \Prob(\cdot\mymid\mathcal G_a).
\end{equation*}

The envelope $\varkappa_{\ast,a}$ in \eqref{eq:arm-wise_kappa} is the HI25 envelope after the relabeling $n=d$ and $\mu_\ast=\beta_a$. Below we suppress the dimension subscript on the arm-wise noise law when there is no ambiguity.

\begin{suppproposition}[arm-wise conditioning and admissibility]
\label{prop:supp:arm-wise_conditioning}
Under the assumptions in \Cref{sec:formal-regime}, there exist constants $0<c_-<c_+<\infty$ such that $\Prob(E_{\mathrm{cnt}}^c)\to0$ and the following statements hold for each arm $a$.
\begin{enumerate}
\item Conditional on $(\Is_a,E_{\mathrm{cnt}})$, the observations in $\Ds_a$ are independent with common law
\begin{equation*}
  X\sim N(0,I_d),\qquad
  R=\varphi_a^\ast(\beta_a^\top X)+\xi_{a,d},\qquad
  \xi_{a,d}\sim P_{\xi,a,d},\quad \xi_{a,d}\perp X.
\end{equation*}
In particular, $\E\xi_{a,d}=0$ and $\E\xi_{a,d}^2=\sigma_{\xi,a,d}^2<\infty$.
\item On $E_{\mathrm{cnt}}$, the padded arm-wise system satisfies the hypotheses of \Cref{lem:hi25-component-link} under the relabeling
\begin{equation*}
  m=N_a,\qquad n=d,\qquad
  \mu_\ast=\beta_a,\qquad\phi_a=N_a/d\in[c_-,c_+].
\end{equation*}
\item For every fixed $t$, the predictor $f_{W_a^{(t)}}$, the parameters $W_a^{(t)}$, and the states $U_a^{(t)}$ and $\widehat U_a^{(t)}$ are $\widehat{\mathcal F}_a$-measurable.
\end{enumerate}
\end{suppproposition}

\begin{proof}
The law of large numbers and $m/d\to\phi$ give, for each $b\in[K]$,
\begin{equation*}
  \frac{N_b}{d}
  =
  \frac{N_b}{m}\frac{m}{d}
  \xrightarrow{p}
  \mu(b)\phi.
\end{equation*}
Because $K$ is fixed and $\min_b\mu(b)\ge c>0$, take $c_-=c\phi/2$ and $c_+=2\phi$. Then
\begin{equation*}
  \Prob(E_{\mathrm{cnt}}^c)
  \le\sum_{b=1}^K
  \Prob\{N_b/d\notin[c_-,c_+]\}
  \longrightarrow0.
\end{equation*}

Fix an arm $a$, a deterministic set $S\subseteq[m]$, and an action vector $a_{1:m}$ for which $\Is_a=S$. For bounded measurable functions $\{\vartheta_i\}_{i\in S}$, independence of the logged observations gives
\begin{equation*}
  \mathbb E\br{
    \prod_{i\in S}\vartheta_i(X_i,R_i)
    \mymid
    A_{1:m}=a_{1:m}
  }
  =
  \prod_{i\in S}
  \mathbb E\br{\vartheta_i(X,R)\mymid A=a}.
\end{equation*}
The right-hand side is the same for every action vector with $\Is_a=S$. It is therefore unchanged after averaging over such vectors, either conditionally on $\Is_a=S$ alone or jointly on $(\Is_a=S,E_{\mathrm{cnt}})$. This proves the asserted conditional i.i.d.\ law.

Context-independent logging gives $X\mid(A=a)\sim N(0,I_d)$. The reward assumptions give
\begin{equation*}
  R=\varphi_a^\ast(\beta_a^\top X)+\xi_{a,d},
  \qquad
  \mathbb E[\xi_{a,d}]=0,
  \qquad
  \mathbb E[\xi_{a,d}^2]=\sigma_{\xi,a,d}^2<\infty.
\end{equation*}
Moreover, $\xi_{a,d}$ is independent of $X$ under the arm-conditional law. Write $\xi_i=\varepsilon_i$ for $i\in S$ and $\xi_S=(\xi_i)_{i\in S}$. For bounded measurable $f_i:\R^d\to\R$ and $g_i:\R\to\R$,
\begin{align*}
  &\mathbb E\br{
    \prod_{i\in S}f_i(X_i)g_i(\xi_i)
    \mymid A_{1:m}=a_{1:m}
  } =
  \prod_{i\in S}
  \mathbb E[f_i(Z)]
  \mathbb E[g_i(\xi_{a,d})],
  \qquad Z\sim N(0,I_d).
\end{align*}
Hence, a regular conditional distribution satisfies
\begin{equation*}
  \mathbb E\br{
    \prod_{i\in S}f_i(X_i)
    \mymid A_{1:m},\xi_S
  }
  =
  \prod_{i\in S}\mathbb E[f_i(Z)]
  \qquad\text{a.s. on }\{\Is_a=S\}.
\end{equation*}
The initialization $W_a^{(0)}$ is independent of the logged sample, and $\beta_a$ and $\Upsilon$ are fixed directions. Therefore
\begin{equation*}
  \mathbb E\br{
    \prod_{i\in S}f_i(X_i)
    \mymid \mathcal G_a
  }
  =
  \prod_{i\in S}\mathbb E[f_i(Z)]
  \qquad\text{a.s. on }\{\Is_a=S\}.
\end{equation*}
Thus, conditioning on $\mathcal G_a$ leaves the arm-wise design rows i.i.d.\ $N(0,I_d)$. The event $E_{\mathrm{cnt}}$ is $\mathcal G_a$-measurable because it depends only on the action labels.

Under the relabeling $m=N_a$, $n=d$, $\mu_\ast=\beta_a$, and $\boldsymbol\varphi=\varphi_{a,[q]}^\ast$, HI25 Assumption (A1) follows on $E_{\mathrm{cnt}}$ from $N_a/d\in[c_-,c_+]$, and (A2) follows from the Gaussian design. Assumptions (A3)--(A4) are precisely the learning-rate and activation conditions in Assumption~\ref{ass:formal-nn-regularity}, and (A5) with coordinate-specific reward links follows from the same assumption and \Cref{lem:hi25-component-link}. The initialization is independent of the design by assumption. Comparing \eqref{eq:arm-wise_kappa} with HI25~\eqref{HI-def:kappa_ast} gives the stated identity for $\varkappa_{\ast,a}$. Thus, the relabeled arm-wise problem satisfies the hypotheses of the HI25 comparison results as extended in \Cref{lem:hi25-component-link}.

Finally, every finite GD iterate and algorithmic correction is a measurable function of $(\Is_a,X_{\Is_a},R_{\Is_a},W_a^{(0)})$. The theoretical correction $U_a^{(t)}$ also uses the state-evolution coefficients $\rho_{t,s}$, which are deterministic functions of the relabeled problem parameters, the realized arm-wise noise, the initialization, and the fixed hyperparameters. Since
\begin{equation*}
  \xi_{\Is_a}
  =
  \bigl(R_i-\varphi_a^\ast(\beta_a^\top X_i)\bigr)_{i\in\Is_a},
\end{equation*}
these coefficients are $\widehat{\mathcal F}_a$-measurable. Equivalently, for every fixed $r$ and $s\le r$, the recursion makes $(\rho_{r,s},D_{a,r},V_a^{(r)})$ measurable with respect to
\begin{equation*}
  \sigma(\Is_a,\beta_a,W_a^{(0)},\xi_{\Is_a})
  \subseteq\widehat{\mathcal F}_a.
\end{equation*}
Substitution into the definitions of $U_a^{(r)}$ and $\widehat U_a^{(r)}$ proves their measurability recursively. Thus, every object listed in the proposition is $\widehat{\mathcal F}_a$-measurable.
\end{proof}

\begin{lemma}[Fresh-draw conditioning]
\label{lem:fresh-draw-conditioning}
Let $\Theta_a$ be $\widehat{\mathcal F}_a$-measurable, and let $Q$ be independent of $\mathcal F_{\mathrm{tr}}$. If $\widehat{\mathcal F}_a\subseteq\mathcal H\subseteq\mathcal F_{\mathrm{tr}}$, then every nonnegative measurable function $b$ satisfies
\begin{equation*}
  \E[b(Q,\Theta_a)\mymid\mathcal H]
  =
  \int b(q,\Theta_a)\,dP_Q(q)
  \qquad\text{almost surely}.
\end{equation*}
The identity also holds for signed $b$ when $\int|b(q,\Theta_a)|\,dP_Q(q)<\infty$ almost surely, with conditional expectations interpreted by this finite fresh-draw integral.
\end{lemma}

\begin{proof}
The variable $\Theta_a$ is $\mathcal H$-measurable, and the conditional law of $Q$ given $\mathcal H$ is its marginal law. The identity follows first for bounded $b$ and then for nonnegative $b$ by monotone convergence. For signed $b$, apply the nonnegative identity to its positive and negative parts; the assumed absolute-integral bound makes both terms finite.
\end{proof}

In particular, the lemma applies with $\mathcal H$ equal to $\widehat{\mathcal F}_a$, $\sigma(\Ds_m,W_a^{(0)})$, or $\mathcal F_{\mathrm{tr}}$. These conditioning fields therefore give the same expectation for a fixed arm-wise predictor evaluated at a fresh draw.

\subsection{Auxiliary GFOM iterates and comparison bounds}

Fix an arm $a$, work on $E_{\mathrm{cnt}}$, and condition on the sigma-field $\mathcal G_a$ from \Cref{prop:supp:arm-wise_conditioning}. Expectations under this conditioning are denoted by $\E_a^{(0)}$. The arm-wise covariates remain random, and the HI25 setup applies with $m=N_a$, $n=d$, and $\phi_a=N_a/d$.

Write $\Is_a=\{i_{a,1}<\cdots<i_{a,N_a}\}$ and let
\begin{equation*}
  X_a
  \coloneqq
  \begin{pmatrix}
    X_{i_{a,1}}^\top\\
    \vdots\\
    X_{i_{a,N_a}}^\top
  \end{pmatrix}
  \in
  \mathbb R^{N_a\times d}
\end{equation*}
be the arm-wise design matrix.

Let $d_{L,q}\coloneqq d+(L-1)q$ and write
\begin{equation*}
  A_a\coloneqq \bigl(X_a/\sqrt d,\ 0\bigr)\in\mathbb R^{N_a\times d_{L,q}},
  \qquad
  \Upsilon_{\mathrm{pad},a}
  \coloneqq
  \begin{pmatrix}
    \sqrt d\,\Upsilon\\
    0_{(L-1)q\times\ell}
  \end{pmatrix}
  \in
  \mathbb R^{d_{L,q}\times\ell},
\end{equation*}
so that $A_a\Upsilon_{\mathrm{pad},a}=X_a\Upsilon$. Let $(u_a^{(t)},v_a^{(t)})$ denote the auxiliary GFOM iterates in \HISec{7.2} and \HIprop{prop:se_aux_GFOM}, using the padded reward link $\varphi_{a,[q]}^\ast$. The matrix $u_a^{(t)}$ has one row per observation, whereas $v_a^{(t)}$ has one row per input coordinate or padded parameter coordinate. We refer to the corresponding state-evolution descriptions as row-indexed and column-indexed, respectively. Define the iterates with appended target-policy index coordinates by
\begin{equation*}
  \widetilde u_a^{(t)}
  \coloneqq
  \bigl(u_a^{(t)},\,X_a\Upsilon\bigr)
  \in\mathbb R^{N_a\times(q+\ell)},
  \qquad
  \widetilde v_a^{(t)}
  \coloneqq
  \bigl(v_a^{(t)},\,0_{d_{L,q}\times\ell}\bigr)
  \in\mathbb R^{d_{L,q}\times(q+\ell)},
\end{equation*}
where the target-policy index coordinates are appended columnwise. Here $\mathrm{aux}$ denotes the original auxiliary GFOM block and $\mathrm{idx}$ denotes the appended index coordinate block. Also write
\begin{equation*}
  D_{a,-1}
  \coloneqq
  \bigl(\sqrt d\,\beta_a,\ 0_{d\times(q-1)}\bigr)\in\mathbb R^{d\times q}
\end{equation*}
for the arm-wise relabeling of the HI25 initial deterministic shift, and define its padded version
\begin{equation*}
  \bar\mu_{a,\ast,d}
  \coloneqq
  \begin{pmatrix}
    D_{a,-1}\\
    0_{(L-1)q\times q}
  \end{pmatrix}
  \in \mathbb R^{d_{L,q}\times q}.
\end{equation*}
Let $D_{a,0},\ldots,D_{a,t-1}$ denote the arm-wise deterministic shifts from \HIdef{def:gd_NN_se} under the relabeling $n=d$ and $\mu_\ast=\beta_a$.

We next establish the bounds needed for the state-evolution and Onsager comparisons. We first control the gradient-descent iterates, then prove an averaged comparison for the auxiliary GFOM, and finally compare the two recursions and their Onsager coefficients before returning to the appended policy index.

\begin{supplemma}[Bounds for gradient-descent iterates]
\label{lem:gd-path-localization}
Assume the working conditions of \Cref{sec:formal-regime}. Fix an arm on $E_{\mathrm{cnt}}$, condition on $\mathcal G_a$, and write $m=N_a$, $n=d$, $\phi=m/n$, and $\kappa=\varkappa_{\ast,a}$. Suppress the arm subscript, and fix a horizon $T\ge1$. For every $b_0>0$, there is $C_{b_0}<\infty$ such that
\begin{equation}
 \Prob\left\{
 \max_{0\le s\le T}\|XW_1^{(s)}\|_\infty
     >(C_{b_0}\kappa\log n)^{C_{b_0}}
 \mathrel{\big|}\mathcal G_a\right\}
 \le (C_{b_0}\kappa\log n)^{C_{b_0}}n^{-b_0}.
 \label{eq:gd-hidden-delocalization}
\end{equation}
For every fixed $r\ge1$, there is $C_r<\infty$ such that
\begin{equation}
 \E\left[\left\{
 1+\max_{0\le s\le T}\left(
    \|XW_1^{(s)}\|_\infty+\|W_1^{(s)}\|
       +\max_{2\le\alpha\le L}\|W_\alpha^{(s)}\|
 \right)\right\}^{r}\mathrel{\big|}\mathcal G_a\right]
 \le (C_r\kappa\log n)^{C_r}.
 \label{eq:gd-hidden-moments}
\end{equation}
The constants are uniform in $m,n,\kappa$ on $E_{\mathrm{cnt}}$.
\end{supplemma}

\begin{proof}
All probabilities and expectations in this proof are conditional on $\mathcal G_a$. Exponents denoted by $C_T$ depend only on the fixed horizon and regularity parameters, independently of the tail exponent $b$; multiplicative constants denoted by $C_b$ may depend on $b$. We take $n\ge3$; changing the constants covers smaller $n$. Set $A=X/\sqrt n$ and $w=X\beta_{[q]}$. The definition of $\kappa$ bounds $\sqrt n\|\beta\|_\infty$, $\sqrt n\|W_1^{(0)}\|_\infty$, the initial deeper weights, and $\|\xi\|_\infty$. In particular, $\|\beta\|+\|W_1^{(0)}\|\le C\kappa$. Gaussian spectral-norm and coordinate-tail bounds imply, for each $b>0$,
\begin{equation*}
 \Prob(\mathcal A_b^c)\le C_b n^{-b},\qquad
 \mathcal A_b=
 \{\|A\|_{\op}\le C_b,\ \|w\|_\infty\le C_b\kappa\sqrt{\log n}\}.
\end{equation*}
Also every fixed moment of $Q=1+\|A\|_{\op}+\kappa+\|w\|_\infty$ is bounded by $(C\kappa\log n)^C$.

We first control the weights without using a bound on individual hidden states. Let $\Gsc_V(u)$ denote the network output when the first-layer pre-activation is $u$ and the deeper weights are $V$. The bounded activation derivatives imply
\begin{align}
 \|\Gsc_V(u)\|&\le C(1+\|V\|)^C(1+\|u\|), \notag\\
 \|D_u \Gsc_V(u)\|&\le C(1+\|V\|)^C, \notag\\
 \|D_{V_\alpha}\Gsc_V(u)\|&\le C(1+\|V\|)^C(1+\|u\|).
 \label{eq:gd-network-value-bounds}
\end{align}
Here and below any fixed norm on the finite collection of deeper weights may be used. For $v=\sqrt n W_1$ and $u=Av$, the first-layer update is
\begin{equation*}
 v^+=v-\phi^{-1}\eta_1 A^\top\mathfrak S(u,w,V).
\end{equation*}
The first two bounds in \eqref{eq:gd-network-value-bounds} give
\begin{equation*}
 m^{-1/2}\|\mathfrak S(u,w,V)\|
 \le C(1+\|V\|)^C
       \{1+m^{-1/2}\|u\|+\|w\|_\infty^{r_0}+\kappa\}.
\end{equation*}
The last bound in \eqref{eq:gd-network-value-bounds} and Cauchy--Schwarz give, for each deeper-layer empirical gradient,
\begin{equation*}
 \left\|\frac1m\sum_{k=1}^m
 D_{V_\alpha}\Gsc_V(u_k)^\top
       \{\Gsc_V(u_k)-\boldsymbol\varphi(w_k)-\xi_k e_1\}\right\|
 \le C(1+\|V\|)^C
       \{1+m^{-1}\|u\|^2+\|w\|_\infty^{2r_0}+\kappa^2\}.
\end{equation*}
With $H_s=1+\|W_1^{(s)}\|+\max_{\alpha\ge2}\|W_\alpha^{(s)}\|$ and $m/n$ bounded above and below, these displays imply
\begin{equation}
 H_{s+1}\le(CQH_s)^C,\qquad
 \max_{0\le s\le T+1}H_s\le(CQ)^{C_T}.
 \label{eq:gd-global-norm-recursion}
\end{equation}
This establishes polynomial conditional moments for the weights. It also gives the crude bound $\max_{s\le T}\|XW_1^{(s)}\|_\infty\le\sqrt n(CQ)^{C_T}$.

We next obtain a bound uniform over a deterministic collection of deeper weight trajectories. For $B=(B^{(0)},\ldots,B^{(T-1)})$, with $B^{(s)}\in(\R^{q\times q})^{L-1}$, define
\begin{align}
 v_B^{(0)}&=\sqrt n W_1^{(0)}, \notag\\
 u_B^{(s+1)}&=Av_B^{(s)}, \notag\\
 v_B^{(s+1)}&=v_B^{(s)}-
       \phi^{-1}\eta_1^{(s)}A^\top
          \mathfrak S(u_B^{(s+1)},w,B^{(s)}),
       \qquad 0\le s<T.
 \label{eq:frozen-deeper-trajectory}
\end{align}
Include $u_B^{(T+1)}=Av_B^{(T)}$ when bounding the final hidden state. For each fixed $B$, this is a GFOM with rowwise maps. The initial signal state $w=A(\sqrt n\beta_{[q]})$ is produced by the same initial linear step as in the auxiliary GFOM construction; the starting vectors and all maps are deterministic under the conditional law. The maps have polynomial local Lipschitz bounds of fixed order, with constants at most $(C\kappa R)^C$ whenever $\max_s\|B^{(s)}\|\le R$, by the chain rule and the activation and reward-link assumptions. Applying \HIprop{prop:GFOM_deloc_asym}, with $x=C_b\log n$, and taking a union over the $O(n)$ coordinates and the fixed number of iterations, yields
\begin{equation}
 \sup_{\max_s\|B^{(s)}\|\le R}
 \Prob\left\{\max_{1\le s\le T+1}\|u_B^{(s)}\|_\infty
          >(C_b\kappa R\log n)^{C_T}\right\}
 \le C_b n^{-b}.
 \label{eq:frozen-path-tail}
\end{equation}
The supremum is outside the probability in this display. For all sufficiently large $n$, $x\le n$ as required by that proposition; the remaining finitely many $n$ are covered by increasing $C_b$.

To make the bound simultaneous over trajectories, we use that their parameter dimension $D=(L-1)q^2T$ is fixed. On $\mathcal A_b$, the same first-layer norm recursion as above, with $\|B\|\le R$, yields
\begin{equation*}
 \sup_{\|B\|\le R}\max_{s\le T}
       n^{-1/2}\|v_B^{(s)}\|
       \le(C_b\kappa R\log n)^{C_T}.
\end{equation*}
In particular every row norm is at most $\sqrt n(C_b\kappa R\log n)^{C_T}$. The polynomial local Lipschitz bound for $\mathfrak S$ therefore yields, for $\|B\|,\|B'\|\le R$ and $e_s=n^{-1/2}\|v_B^{(s)}-v_{B'}^{(s)}\|$,
\begin{equation*}
 e_0=0,\qquad
 e_{s+1}\le(n\kappa R\log n)^{C_T}
       \{e_s+\|B-B'\|\}.
\end{equation*}
Since $\|u_B^{(s+1)}-u_{B'}^{(s+1)}\|_\infty \le\sqrt n\|A\|_{\op}e_s$, finite iteration yields, after enlarging $C_T$,
\begin{equation}
 \max_{s\le T+1}\|u_B^{(s)}-u_{B'}^{(s)}\|_\infty
 \le(C_b n\kappa R\log n)^{C_T}\|B-B'\|.
 \label{eq:frozen-path-continuity}
\end{equation}
By \eqref{eq:gd-global-norm-recursion}, on $\mathcal A_b$ the actual deeper trajectory lies in the ball of radius $R=(C_b\kappa\log n)^{C_T}$. Choose a deterministic $\delta$-net of this ball, with $\delta=(C_b n\kappa R\log n)^{-C_T-1}$, so that every trajectory is within $\delta$ of a net point. The net can be chosen with cardinality at most
\begin{equation*}
 (1+C R/\delta)^D\le C_b n^{C_T}(\kappa\log n)^{C_T}.
\end{equation*}
Apply \eqref{eq:frozen-path-tail} at each net point with its tail exponent larger than $b_0+C_T+2$, and take a union bound over the net. By \eqref{eq:frozen-path-continuity}, replacing any trajectory by its nearest net point changes each hidden state by at most one. Thus, the bound extends to the whole ball, with failure probability at most $(C_{b_0}\kappa\log n)^{C_{b_0}}n^{-b_0}$. For the random choice $B^{(s)}=W_{[2:L]}^{(s)}$, induction in \eqref{eq:frozen-deeper-trajectory} gives $v_B^{(s)}=\sqrt n W_1^{(s)}$ exactly. No independence between this random path and $A$ is needed after the simultaneous bound has been established. This proves \eqref{eq:gd-hidden-delocalization}.

Finally, let $Z=\max_{s\le T}\|XW_1^{(s)}\|_\infty$ and let $\mathcal E_{b_0}$ be the event in \eqref{eq:gd-hidden-delocalization} on which $Z$ is bounded. The crude bound after \eqref{eq:gd-global-norm-recursion} yields $\E Z^{2r}\le n^r(C_r\kappa\log n)^{C_r}$. Consequently, Cauchy--Schwarz gives
\begin{align*}
 \E Z^r
 &\le(C_{b_0}\kappa\log n)^{rC_{b_0}}
       +(\E Z^{2r})^{1/2}\Prob(\mathcal E_{b_0}^c)^{1/2}\\
 &\le(C_{b_0,r}\kappa\log n)^{C_{b_0,r}}
       \{1+n^{(r-b_0)/2}\}.
\end{align*}
Taking $b_0>r$, and using the weight moment bounds already proved, establishes \eqref{eq:gd-hidden-moments}.
\end{proof}

The next lemma compares empirical averages of the auxiliary row and column histories with their state-evolution expectations. Fix an arm $a$, write $X=X_a$, $m=N_a$, $n=d$, and $\kappa=\varkappa_{\ast,a}$, and suppress the arm subscript below. Use the untruncated recursion in \HIdef{def:gd_NN_se} with the padded reward link. Its row history is $\mathfrak U^{([0:T])}$, with row maps $\Theta_k^{(s)}$. For each $i\in[n]$, let $\mathfrak V_i^{([1:T])}$ have the centered Gaussian law specified by $\operatorname{Cov}(\mathfrak V_i^{(s)},\mathfrak V_i^{(h)})=\Sigma_{s,h}$. The corresponding column states, written as rows, are
\begin{equation*}
 \Delta_i^{(s)}
 =D_{s;i\cdot}+\sum_{h=1}^s\mathfrak V_i^{(h)}\rho_{s,h}^{\top},
 \qquad i\in[n],\quad 0\le s\le T.
\end{equation*}
At $s=0$, the sum is empty and $\Delta_i^{(0)}=D_{0;i\cdot}$. All Gaussian laws here are conditional on $\mathcal G_a$.

\begin{supplemma}[Localization of the auxiliary GFOM]
\label{lem:supp:input-localization}
Assume the working conditions of \Cref{sec:formal-regime}. Fix an integer $T\ge1$, a pseudo-Lipschitz order $p\ge1$, a moment order $r\ge1$, and $\Lambda_\vartheta\ge2$. Let $\{\vartheta_k:\R^{q[0:T]}\to\R\}_{k=1}^m$ and $\{\psi_i:\R^{q[0:T]}\to\R\}_{i=1}^n$ be $\Lambda_\vartheta$-pseudo-Lipschitz of order $p$ and deterministic conditional on $\mathcal G_a$. There is a constant $C<\infty$ such that, on $E_{\mathrm{cnt}}$,
\begin{align*}
 &\E_a^{(0)}\left[\left|
   \frac1m\sum_{k=1}^m\vartheta_k(u_k^{([0:T])})
   -\frac1m\sum_{k=1}^m\E_a^{(0)}\vartheta_k\left(
     \{\Theta_k^{(s)}(\mathfrak U^{([0:s])})\}_{s=0}^T
   \right)\right|^r\right]\\
 &\quad+\E_a^{(0)}\left[\left|
   \frac1n\sum_{i=1}^n\psi_i(v_{1;i}^{([0:T])})
   -\frac1n\sum_{i=1}^n\E_a^{(0)}\psi_i\left(
     \{\Delta_i^{(s)}\}_{s=0}^T
   \right)\right|^r\right]
 \le (C\Lambda_\vartheta\kappa)^C n^{-1/C}.
\end{align*}
The same bound holds for test families on $\R^{(q+\ell)[0:T]}$ after replacing the empirical and Gaussian row histories, respectively, by
\begin{equation*}
 \{(u_k^{(s)},X_{k\cdot}\Upsilon)\}_{s=0}^T,
 \qquad
 \{(\Theta_k^{(s)}(\mathfrak U^{([0:s])}),
             \mathfrak U_\Upsilon)\}_{s=0}^T,
\end{equation*}
and the empirical and Gaussian column histories, respectively, by
\begin{equation*}
 \{(v_{1;i}^{(s)},0_{1\times\ell})\}_{s=0}^T,
 \qquad
 \{(\Delta_i^{(s)},0_{1\times\ell})\}_{s=0}^T.
\end{equation*}
Here $\mathfrak U_\Upsilon\sim N(0,I_\ell)$ is jointly Gaussian with $\mathfrak U^{([0:T])}$ and satisfies
\begin{equation*}
 \operatorname{Cov}(\mathfrak U^{(s)},\mathfrak U_\Upsilon)
 =n^{-1/2}D_{s-1}^{\top}\Upsilon,
 \qquad 0\le s\le T.
\end{equation*}
The constant $C$ depends only on $T,p,r,L,q,\ell,r_0,c_-,c_+$ and the fixed constants in \Cref{sec:formal-regime}, and is independent of $m,n,\kappa,\Lambda_\vartheta$ and the realized conditioning variables.
\end{supplemma}

\begin{proof}
All expectations and probabilities below are conditional on $\mathcal G_a$. Constants may depend on the fixed horizon, moment orders, depth, width, growth order, and the constants in \Cref{sec:formal-regime}. They are independent of $n$, $\kappa$, and the cutoff radius introduced below. We use the original deterministic sequences $V^{(s)}$ and $M^{(s)}$ throughout the construction.

Choose a smooth function $\chi:\R\to\R$ equal to the identity on $[-1,1]$ and constant outside $[-2,2]$, with $|\chi(x)|\le C|x|$. For $R\ge1$, set $\chi_R(x)=R\chi(x/R)$ and apply it coordinatewise. Replace the row loss gradient in $G_s^{\langle2\rangle}$ by
\begin{equation}
 \check{\mathfrak S}_{s,k,R}(u,w)
 =\mathfrak S_k\bigl(\chi_R(u),\chi_R(w),V^{(s-1)}\bigr),
 \qquad 1\le s\le T.
 \label{eq:input-cutoff-gradient}
\end{equation}
All other auxiliary maps, including $F_s^{\langle2\rangle}$ and its original $M^{(s-1)}$, are unchanged. The appended maps remain constant or zero. Check marks refer to this construction. In particular, $\check v_\alpha^{(s)}=v_\alpha^{(s)}=V_\alpha^{(s)}$ for $\alpha\ge2$: the corresponding columns of $A$ vanish, and the deterministic update is $V_\alpha^{(s-1)}-\eta_\alpha^{(s-1)}M_\alpha^{(s-1)}$.

We first record bounds that do not grow with $R$. Let $\Gsc_V(u)$ be the $q$-dimensional network output as a function of its first-layer hidden state. In the derivative calculations below, regard the row states and loss gradients as column vectors. The loss gradient is
\begin{equation*}
 \mathfrak S_k(u,w;V)=D \Gsc_V(u)^\top
 \{\Gsc_V(u)-\boldsymbol\varphi(w)-\xi_k e_1\}.
\end{equation*}
The activation and link assumptions, the product rule, and the chain rule give, for $j=0,1,2,3$,
\begin{equation}
 \|D^j \mathfrak S_k(u,w;V)\|
 \le C(1+\|V\|+|\xi_k|+\|u\|+\|w\|)^C.
 \label{eq:localization-polynomial-derivatives}
\end{equation}
Here derivatives are with respect to $(u,w)$, and any fixed norm on the finite collection of deeper-layer matrices may be used for $\|V\|$. For example, $D_w \mathfrak S_k(u,w;V)=-D \Gsc_V(u)^\top D\boldsymbol\varphi(w)$; this derivative does not contain the residual as a factor. The fourth activation derivative and the third link derivative suffice for \eqref{eq:localization-polynomial-derivatives}. Since $|\chi_R(x)|\le C|x|$ and $\|\chi_R^{(j)}\|_\infty\le C$ for $1\le j\le3$, the same polynomial bound holds for the cutoff map, uniformly in $R\ge1$. Also, bounded activation derivatives imply the sharper value bound
\begin{equation}
 \|\mathfrak S_k(u,w;V)\|+\|\check{\mathfrak S}_{s,k,R}(u,w)\|
 \le C\kappa^C(1+\|u\|+\|w\|^C),
 \label{eq:localization-linear-u}
\end{equation}
once the original $V^{(s-1)}$ is bounded by a polynomial in $\kappa$.

We establish this polynomial bound jointly with the state-evolution moment bounds, using the derivative estimates above so that the constants remain uniform in $R$. With the same column-vector convention, put
\begin{align*}
 P_{s,k}(x)&=\mathfrak S_k\bigl(\Theta_k^{(s)}(x),x^{(0)};V^{(s-1)}\bigr),\\
 J_{s,j,k}(x)&=D_{x^{(j)}}\Theta_k^{(s)}(x),
 \qquad K_{s,j,k}(x)=D_{x^{(j)}}P_{s,k}(x).
\end{align*}
The state evolution for the cutoff iteration follows from the generic asymmetric GFOM recursion in \HIdef{def:GFOM_se_asym}, with time shifted by one so that the deterministic initial states are $u^{(-1)}$ and $v^{(-1)}$. For each input coordinate $i\in[n]$, write the Gaussian innovation as a row $\check{\mathfrak V}_i^{(s)}\in\R^{1\times q}$. The corresponding column state satisfies
\begin{equation*}
 \check\Delta_i^{(s)\top}
 =\check{\mathfrak V}_i^{(s)\top}+\check\delta_s D_{-1;i\cdot}^{\top}
   +\sum_{h=1}^s(\check\tau_{s,h}+I_q\bm{1}_{h=s})
                       \check\Delta_i^{(h-1)\top}.
\end{equation*}
Expanding this affine recursion gives
\begin{equation*}
 \check\Delta_i^{(s)\top}
 =\check D_{s;i\cdot}^{\top}
  +\sum_{h=1}^s\check\rho_{s,h}\check{\mathfrak V}_i^{(h)\top}.
\end{equation*}
Its derivative with respect to $\check{\mathfrak V}_i^{(h)\top}$ is $\check\rho_{s,h}$, which is the row correction coefficient in the generic GFOM definition. On the column side, the generic correction sums the expected Jacobians of $-\phi_a^{-1}\eta_1^{(s-1)} \check{\mathfrak S}_{s,k,R}(\check\Theta_k^{(s)}(x),x^{(0)})$ with respect to the history coordinates, with weight $1/n$ over the $m$ rows. Since $\phi_a=m/n$, these coefficients are $\check\tau_{s,j}$ or $\check\delta_s$, as specified below. The Gaussian column innovations have covariance $\check\Sigma$, while averaging the second moments of the affine column states gives $\check\Omega+n^{-1}\check D^{\top}\check D$ on the row side. Thus, the coefficient recursions below follow directly from the generic GFOM definition. No symmetry of the Jacobian of the cutoff map is used. The padded column rows remain deterministic and equal the original $V^{(s)}$, as noted above. The untruncated recursion has the same algebra.

The response maps for the cutoff iteration use \eqref{eq:input-cutoff-gradient}, with the same original $V^{(s-1)}$. The exact row and Jacobian recursions are
\begin{align}
 \Theta_k^{(s)}(x)
 &=x^{(s)}-\phi_a^{-1}\sum_{h=1}^{s-1}
       \eta_1^{(h-1)}\rho_{s-1,h}P_{h,k}(x), \notag\\
 J_{s,j,k}(x)
 &=I_q\bm{1}_{j=s}-\phi_a^{-1}\sum_{h=1}^{s-1}
       \eta_1^{(h-1)}\rho_{s-1,h}K_{h,j,k}(x), \notag\\
 K_{s,j,k}(x)
 &=D_u \mathfrak S_k\bigl(\Theta_k^{(s)}(x),x^{(0)};V^{(s-1)}\bigr)
       J_{s,j,k}(x)
   +D_w \mathfrak S_k\bigl(\Theta_k^{(s)}(x),x^{(0)};V^{(s-1)}\bigr)\bm{1}_{j=0}.
 \label{eq:localization-response-jacobians}
\end{align}
A term with $j>h$ is zero. The same identities hold with check marks. Differentiating the last two identities once more gives polynomial bounds for $D^2\Theta_k^{(s)}$ and $D K_{s,j,k}$ from the bounds on $D\mathfrak S_k$ and $D^2\mathfrak S_k$.

More explicitly, suppose all coefficients through time $s-1$ and $V^{(s-1)}$ are bounded by $C\kappa^C$. The preceding recursions and \eqref{eq:localization-polynomial-derivatives} then give, uniformly in $k$ and $R$,
\begin{equation}
 \max_{1\le h\le s}\sum_{j=0}^2
   \left\{\|D^j\Theta_k^{(h)}(x)\|+\|D^jP_{h,k}(x)\|\right\}
 \le C\kappa^C(1+\|x\|)^C.
 \label{eq:localization-map-envelope}
\end{equation}
Here $D^0$ denotes the map itself and higher derivatives are with respect to the full history argument. The same bound holds for the checked maps. The covariance of the Gaussian history through time $s$ is $\Omega_{h-1,l-1}+n^{-1}D_{h-1}^\top D_{l-1}$, for $h,l\le s$. Its entries are bounded by $C\kappa^C$ under the induction hypothesis, so each fixed Gaussian moment is bounded by a polynomial in $\kappa$. The coefficient updates are exactly, for $1\le j,h\le s$,
\begin{align}
 \tau_{s,j}&=-\eta_1^{(s-1)}m^{-1}\sum_k\E K_{s,j,k}(\mathfrak U), \notag\\
 \delta_s&=-\eta_1^{(s-1)}m^{-1}\sum_k\E K_{s,0,k}(\mathfrak U), \notag\\
 \Sigma_{s,h}&=\phi_a^{-1}\eta_1^{(s-1)}\eta_1^{(h-1)}
       m^{-1}\sum_k\E P_{s,k}(\mathfrak U)P_{h,k}(\mathfrak U)^\top, \notag\\
 \rho_{s,j}&=I_q\bm{1}_{j=s}
      +\sum_{h=j+1}^s(\tau_{s,h}+I_q\bm{1}_{h=s})\rho_{h-1,j}, \notag\\
 D_s&=D_{-1}\delta_s^\top
      +\sum_{h=1}^sD_{h-1}(\tau_{s,h}^\top+I_q\bm{1}_{h=s}), \notag\\
 \Omega_{s,h}&=\sum_{i=1}^s\sum_{j=1}^h
                   \rho_{s,i}\Sigma_{i,j}\rho_{h,j}^\top.
 \label{eq:localization-coefficient-updates}
\end{align}
Thus, the new coefficients, and the entrywise maximum norm of $D_s$, are bounded by $C\kappa^C$. The original update for $V^{(s)}$ in step~(S5) of \HIdef{def:gd_NN_se} uses an expectation of a product of forward states and loss gradients; both have polynomial growth by the same network bounds. It therefore yields $\|V^{(s)}\|+\|M^{(s-1)}\|\le C\kappa^C$ and closes the joint induction. The initial bounds follow from $\|D_{-1}\|_\infty\vee\|D_0\|_\infty \vee\|V^{(0)}\|\le C\kappa$. This proves all the stated polynomial bounds for both recursions, uniformly over $R\ge1$.

We next compare their coefficients. Let $e_s$ be the maximum, over times $1,\ldots,s$, of the norm differences of $\tau,\delta,\rho,\Sigma,\Omega$, together with $\max_{h\le s}\|\check D_h-D_h\|_\infty$; set $e_0=0$. All matrix norms here are on fixed-dimensional matrices except the explicit entrywise norm on $D$. The Gaussian covariance matrices through time $s$ consequently satisfy
\begin{equation*}
 \|\operatorname{Cov}(\check{\mathfrak U}^{([0:s])})
       -\operatorname{Cov}(\mathfrak U^{([0:s])})\|_{\mathrm{op}}
 \le C\kappa^C e_{s-1}.
\end{equation*}
Indeed,
\begin{equation*}
 n^{-1}\|\check D_h^\top\check D_l-D_h^\top D_l\|_{\mathrm{op}}
 \le C\kappa^C\bigl(\|\check D_h-D_h\|_\infty
       +\|\check D_l-D_l\|_\infty\bigr).
\end{equation*}
Couple the two Gaussian histories by applying their positive semidefinite square roots to the same standard Gaussian vector. The square-root inequality $\|B^{1/2}-C^{1/2}\|_{\mathrm{op}}^2 \le C\|B-C\|_{\mathrm{op}}$ yields, for every fixed $b\ge1$,
\begin{equation}
 \E\|\check{\mathfrak U}^{([0:s])}
           -\mathfrak U^{([0:s])}\|^{2b}
 \le C\kappa^C e_{s-1}^{b}.
 \label{eq:localization-gaussian-difference}
\end{equation}
No nonsingularity of either covariance is required.

Fix an arbitrary $B>0$. At each step $s$, use the coupling of the histories through $s$ in \eqref{eq:localization-gaussian-difference}. The uniform covariance bounds and Gaussian tails give an event $H_{B,s}$ of probability at least $1-n^{-2B}$ on which both coupled histories have norm at most $C_B\kappa^C\sqrt{\log n}$. These stepwise couplings need not be consistent across $s$. By \eqref{eq:localization-map-envelope}, we can choose
\begin{equation}
 R=(C_B\kappa\log n)^C
 \label{eq:localization-radius}
\end{equation}
so large that all hidden-state and signal inputs in both recursions on each $H_{B,s}$, $s\le T$, lie in $[-R,R]$. The powers and constants in this choice do not depend on $R$. Hence, the cutoff and its derivatives agree with the identity at every such input.

We next compare the response maps and their Jacobians on $H_{B,s}$. For the coupled histories $x,y$, let $z=\|x-y\|+e_{s-1}$ and $L_n=C_B(\kappa\log n)^C$. Subtracting \eqref{eq:localization-response-jacobians}, using the mean value bound for $\mathfrak S_k$ and $D\mathfrak S_k$, and proceeding through $h=1,\ldots,s$, yields
\begin{align}
 \|\check\Theta_k^{(h)}(y)-\Theta_k^{(h)}(x)\|
 &\le L_n\left(z+\sum_{l<h}\|\check P_{l,k}(y)-P_{l,k}(x)\|\right), \notag\\
 \|\check J_{h,j,k}(y)-J_{h,j,k}(x)\|
 &\le L_n\left(e_{s-1}
          +\sum_{l<h}\|\check K_{l,j,k}(y)-K_{l,j,k}(x)\|\right), \notag\\
 \|\check P_{h,k}(y)-P_{h,k}(x)\|
 &\le L_n\left(\|x-y\|
          +\|\check\Theta_k^{(h)}(y)-\Theta_k^{(h)}(x)\|\right), \notag\\
 \|\check K_{h,j,k}(y)-K_{h,j,k}(x)\|
 &\le L_n\left(\|x-y\|
          +\|\check\Theta_k^{(h)}(y)-\Theta_k^{(h)}(x)\|
          +\|\check J_{h,j,k}(y)-J_{h,j,k}(x)\|\right).
 \label{eq:localization-response-differences}
\end{align}
All factors suppressed in these inequalities are bounded by \eqref{eq:localization-map-envelope}; $h\le s\le T$ is fixed. Iterating these finite inequalities therefore bounds each left-hand side by $C_B(\kappa\log n)^C(\|x-y\|+e_{s-1})$. On $H_{B,s}^c$, Cauchy--Schwarz and the uniform polynomial moment bounds give $C\kappa^C n^{-B}$ for the expectation of any one of these differences, or of its product with one further response factor. For covariance updates, use the exact product decomposition
\begin{equation*}
 \check P_s\check P_h^\top-P_sP_h^\top
   =(\check P_s-P_s)\check P_h^\top
       +P_s(\check P_h-P_h)^\top.
\end{equation*}
Equations \eqref{eq:localization-gaussian-difference} and \eqref{eq:localization-coefficient-updates} now yield
\begin{equation}
 e_s\le C_B(\kappa\log n)^C
          \{e_{s-1}^{1/2}+e_{s-1}+n^{-B}\}
 \le C_B(\kappa\log n)^C
          \{e_{s-1}^{1/2}+n^{-B}\}.
 \label{eq:localization-coefficient-error}
\end{equation}
For the last inequality use the already proved bound $e_{s-1}\le C\kappa^C$. The updates of $\rho,D,\Omega$ involve finite sums and products, so their differences follow by subtracting one factor at a time in \eqref{eq:localization-coefficient-updates}. Starting from $e_0=0$ and iterating \eqref{eq:localization-coefficient-error} gives $e_T\le C_B(\kappa\log n)^C n^{-B/2^{T-1}}$. Thus, for any prescribed $A>0$, choosing $B$ sufficiently large and then $R$ as in \eqref{eq:localization-radius} gives
\begin{equation}
 e_T\le C_A(\kappa\log n)^C n^{-A}.
 \label{eq:localization-arbitrary-rate}
\end{equation}
The bound holds for every $\kappa\ge1$.

It remains to compare the empirical iterations and apply the averaged GFOM theorem. The polynomial local-Lipschitz bounds from \eqref{eq:localization-polynomial-derivatives}, together with the affine $F$ maps, verify $(D^\ast2)''$ in \HIprop{prop:GFOM_deloc_asym} with a fixed polynomial exponent depending on the link growth order. The same verification holds uniformly in $R$ for the cutoff maps. Taking $x=C_B\log n$ and a union bound over the finitely many times and the $O(n)$ rows gives
\begin{equation*}
 \Prob\left(\max_{s\le T}\|u^{(s)}\|_\infty>R\right)\le n^{-B}
\end{equation*}
after increasing the constant and power in \eqref{eq:localization-radius}. On the event where this maximum is at most $R$, the two empirical recursions coincide exactly, by induction on time.

To control the contribution from the exceptional event, we derive moment bounds directly from the empirical recursion. The linear growth in $u$ in \eqref{eq:localization-linear-u}, together with $u^{(s)}=Av^{(s-1)}$ and the affine $v$ update, yields
\begin{equation*}
 m^{-1/2}\|u^{(s)}\|_F+n^{-1/2}\|v^{(s)}\|_F
 \le C\kappa^C(1+\|A\|_{\mathrm{op}})^C
       \left\{1+\left(m^{-1}\sum_k\|u_k^{(0)}\|^{2C}\right)^{1/2}\right\},
\end{equation*}
with the same bound for the cutoff iteration, uniformly in $R$. The Gaussian spectral-norm and signal moment bounds show that every fixed moment of the right-hand side is bounded by $C\kappa^C$, giving a bound of the form $C\kappa^C n^C$ for each individual row. Integrating the tail bound in \HIprop{prop:GFOM_deloc_asym} over $0\le x\le n$, and applying Cauchy--Schwarz with this crude bound beyond the threshold corresponding to $x=n$, improves this to
\begin{equation*}
 \max_k\E\|u_k^{([0:T])}\|^b
 +\max_i\E\|v_i^{([0:T])}\|^b
 +\max_k\E\|\check u_k^{([0:T])}\|^b
 +\max_i\E\|\check v_i^{([0:T])}\|^b
 \le C_b\kappa^{C_b}
\end{equation*}
for every fixed $b$. Gaussian moments give the same bound for the appended index. The pseudo-Lipschitz inequality, Jensen's inequality, and Cauchy--Schwarz on the exceptional event consequently show that, for arbitrarily large fixed $A$,
\begin{equation}
 \E\left|m^{-1}\sum_k\{
 \vartheta_k(\widetilde u_k^{([0:T])})
 -\vartheta_k(\widetilde{\check u}_k^{([0:T])})\}\right|^r
 \le (C_A\Lambda_\vartheta\kappa)^C n^{-A}.
 \label{eq:localization-empirical-comparison}
\end{equation}
The column-side version follows from the same calculation. When applying the generic averaged theorem, take its column tests to be zero on the $(L-1)q$ padded rows and multiply its column average by $\{n+(L-1)q\}/n$; this yields exactly the average over the first $n$ rows.

For the Gaussian comparison, we couple the two histories using the same policy index. The block covariance matrices $\boldsymbol\Omega_T=(\Omega_{s-1,h-1})_{s,h=0}^T$ and $\check{\boldsymbol\Omega}_T$ are positive semidefinite. Take independent standard Gaussian vectors $Z\in\R^n$ and $\zeta\in\R^{q(T+1)}$, and set
\begin{equation*}
 G=\boldsymbol\Omega_T^{1/2}\zeta,\quad
 \check G=\check{\boldsymbol\Omega}_T^{1/2}\zeta,\qquad
 \mathfrak U^{(s)}=G^{(s)}+n^{-1/2}D_{s-1}^\top Z,\quad
 \check{\mathfrak U}^{(s)}=\check G^{(s)}
                    +n^{-1/2}\check D_{s-1}^\top Z.
\end{equation*}
Use $\Upsilon^\top Z$ as the policy index in both systems. This has exactly the required joint covariances, and
\begin{equation*}
 \E\|\check{\mathfrak U}^{([0:T])}-\mathfrak U^{([0:T])}\|^{2b}
 \le C_b\left\{\sum_{s=0}^Tn^{-1}\|\check D_{s-1}-D_{s-1}\|_F^2
       +\|\check{\boldsymbol\Omega}_T-\boldsymbol\Omega_T\|_{\mathrm{op}}
       \right\}^b.
\end{equation*}
Combine \eqref{eq:localization-arbitrary-rate}, the response-map comparison \eqref{eq:localization-response-differences}, and the Gaussian tail bound used there. Every fixed moment of the row-state difference is at most $C_A(\kappa\log n)^C n^{-A}$ after increasing $B$ again. The common policy index has zero difference. The pseudo-Lipschitz inequality and Cauchy--Schwarz therefore give the Gaussian counterpart of \eqref{eq:localization-empirical-comparison}. For column states, couple $\mathfrak V_i$ and $\check{\mathfrak V}_i$ by the square roots of their $\Sigma$ block covariance matrices and use $\Delta_i^{(s)}=D_{s;i\cdot}+\sum_{h\le s}\mathfrak V_i^{(h)}\rho_{s,h}^\top$; the same conclusion follows from \eqref{eq:localization-arbitrary-rate}.

Finally the input-cutoff maps are globally Lipschitz, with Lipschitz constants and values at the origin bounded by $C(\kappa R)^C$. The appended constant index map has norm at most $C_\Upsilon$. The padded matrix satisfies $(D^\ast1)$ with column dimension $n+(L-1)q$, as verified in the augmented construction below. Apply \HIthm{thm:GFOM_se_asym_avg} to this cutoff system; its error is $(C\Lambda_\vartheta\kappa\log n)^C n^{-c}$ for fixed $c>0$. The two localization comparisons remove the cutoff. Absorbing the fixed power of $\log n$ into a smaller power of $n$ yields $(C\Lambda_\vartheta\kappa)^C n^{-1/C}$, as claimed.
\end{proof}

\begin{supplemma}[Comparison estimates for gradient descent and the Onsager coefficients]
\label{lem:gd-auxiliary-coefficient-comparison}
Under the conditions and notation of \Cref{lem:gd-path-localization}, let $(\hat u^{(s)},\hat v^{(s)})$ and $(u^{(s)},v^{(s)})$ denote the actual and auxiliary iterates, respectively. Thus, $\hat v_1^{(s)}=\sqrt n W_1^{(s)}$, the auxiliary deeper weights are $V^{(s)}$, and $\hat u^{(s+1)}=XW_1^{(s)}$. For every fixed $r\ge1$, there is $C_{T,r}<\infty$ such that
\begin{align}
 &\E\left[\max_{0\le s\le T}
 \left\{n^{-1/2}\|\hat v_1^{(s)}-v_1^{(s)}\|
      +\|W_{[2:L]}^{(s)}-V^{(s)}\|\right\}^{r}
       \mathrel{\big|}\mathcal G_a\right]
 \le(C_{T,r}\kappa\log n)^{C_{T,r}}n^{-1/C_{T,r}},
 \label{eq:gd-auxiliary-direct-comparison}\\
 &\E\left[
 \|\widehat{\bm\tau}^{[T]}-\bm\tau^{[T]}\|^r
 +\|\widehat{\bm\rho}^{[T]}-\bm\rho^{[T]}\|^r
       \mathrel{\big|}\mathcal G_a\right]
 \le(C_{T,r}\kappa\log n)^{C_{T,r}}n^{-1/C_{T,r}}.
 \label{eq:gd-coefficient-direct-comparison}
\end{align}
\end{supplemma}

\begin{proof}
We again work conditionally on $\mathcal G_a$ and use the Gaussian moment bounds and auxiliary comparison proved in \Cref{lem:supp:input-localization}.

For a deeper layer $\alpha$, write $\mathcal D_{\alpha,k}(z,w,V)$ for the gradient of the $k$th per-observation squared loss with respect to $V_\alpha$. It is the product of the forward activation and the back-propagated residual in the deeper-layer GD update. For fixed $V$, its first derivatives with respect to $(z,w)$ have polynomial growth by the chain rule. It is therefore pseudo-Lipschitz in $(z,w)$ of a fixed order, uniformly in $k$, with a constant polynomial in $1+\|V\|+\|\xi\|_\infty$. Define the auxiliary gradient error
\begin{equation*}
 \Delta_s=\max_{2\le\alpha\le L}
 \left\|\frac1m\sum_{k=1}^m
 \mathcal D_{\alpha,k}(u_k^{(s+1)},u_k^{(0)},V^{(s)})
                  -M_\alpha^{(s)}\right\|.
\end{equation*}
The definition of $M_\alpha^{(s)}$ is the corresponding Gaussian expectation. Apply \Cref{lem:supp:input-localization} entrywise to this fixed collection of pseudo-Lipschitz functions. For each fixed $r\ge1$,
\begin{equation}
 \E\Delta_s^r\le(C\kappa\log n)^C n^{-1/C}.
 \label{eq:auxiliary-gradient-error-direct}
\end{equation}
The constants may depend on $T,r$, and this convention applies throughout the rest of the proof.

Put
\begin{equation*}
 E_{1,s}=n^{-1/2}\|\hat v_1^{(s)}-v_1^{(s)}\|,
 \qquad E_{2,s}=\|W_{[2:L]}^{(s)}-V^{(s)}\|.
\end{equation*}
The two initializations agree, so $E_{1,0}=E_{2,0}=0$. By \Cref{lem:gd-path-localization}, the auxiliary delocalization bound, and the global weight bounds, for every $b>0$ there is an event $\mathcal E_b$ such that
\begin{equation*}
 \Prob(\mathcal E_b^c)\le(C_b\kappa\log n)^{C_b}n^{-b}
\end{equation*}
and, on $\mathcal E_b$, the actual and auxiliary row states, the common signal state, all deeper weights, and $\|A\|_{\op}$ are bounded by $P_b=(C_b\kappa\log n)^{C_b}$. The exact first-layer updates and the polynomial local Lipschitz bound for $\mathfrak S$ give, on this event,
\begin{align*}
 E_{1,s+1}
 &\le E_{1,s}+C\|A\|_{\op}m^{-1/2}
 \|\mathfrak S(\hat u^{(s+1)},u^{(0)},W_{[2:L]}^{(s)})
        -\mathfrak S(u^{(s+1)},u^{(0)},V^{(s)})\|\\
 &\le P_b^C(E_{1,s}+E_{2,s}).
\end{align*}
For the deeper update, add and subtract its empirical auxiliary gradient. The same local Lipschitz bound, followed by
\begin{equation*}
 m^{-1}\sum_k\|\hat u_k^{(s+1)}-u_k^{(s+1)}\|
 \le m^{-1/2}\|A\|_{\op}\|\hat v_1^{(s)}-v_1^{(s)}\|,
\end{equation*}
yields
\begin{equation*}
 E_{2,s+1}\le P_b^C(E_{1,s}+E_{2,s})+C\Delta_s.
\end{equation*}
Finite iteration therefore yields
\begin{equation*}
 \max_{s\le T}(E_{1,s}+E_{2,s})\,\bm1_{\mathcal E_b}
 \le P_b^C\sum_{s<T}\Delta_s.
\end{equation*}
Combine this inequality with \eqref{eq:auxiliary-gradient-error-direct}. The auxiliary moment bounds in \Cref{lem:supp:input-localization} hold without restriction to $\mathcal E_b$ and give polynomial conditional moments for the normalized auxiliary first-layer norm. On $\mathcal E_b^c$, Cauchy--Schwarz and these global norm moments give
\begin{equation*}
 \E\left[\max_{s\le T}(E_{1,s}+E_{2,s})^r\bm1_{\mathcal E_b^c}\right]
 \le\left\{\E\max_{s\le T}(E_{1,s}+E_{2,s})^{2r}\right\}^{1/2}
      \Prob(\mathcal E_b^c)^{1/2}
 \le(C_b\kappa\log n)^C n^{-b/2}.
\end{equation*}
Choosing $b$ sufficiently large proves \eqref{eq:gd-auxiliary-direct-comparison}.

We next compare the Onsager coefficients. For a fixed $t\le T$, let $L_k$ be the block diagonal matrix whose $s$th diagonal block is
\begin{equation*}
 \eta_1^{(s-1)}
 D_u\mathfrak S_k(\hat u_k^{(s)},u_k^{(0)},W_{[2:L]}^{(s-1)}),
 \qquad 1\le s\le t,
\end{equation*}
with the same column-Jacobian convention as the derivative cache. Let $L_k^{\mathrm{aux}}$ be the same expression with $(\hat u,W_{[2:L]})$ replaced by $(u,V)$, and, for a lower-triangular $(t-1)\times(t-1)$ block matrix $R$, set
\begin{equation}
 H_t(L,R)=-\sum_{j=0}^{t-1}
             \{-\phi^{-1}L\mathfrak O_t(R)\}^{j}L.
 \label{eq:coefficient-polynomial-map}
\end{equation}
Since $L\mathfrak O_t(R)$ is strictly lower block triangular,
\begin{equation*}
 H_t(L,R)=-\{I_{t,q}+\phi^{-1}L\mathfrak O_t(R)\}^{-1}L.
\end{equation*}
Thus, the inverse is a finite polynomial, and no lower bound on singular values is needed. In particular, telescoping the products in \eqref{eq:coefficient-polynomial-map} gives
\begin{align}
 \|H_t(L,R)-H_t(L',R')\|
 \le C(1+\|L\|+\|L'\|+\|R\|+\|R'\|)^C
             (\|L-L'\|+\|R-R'\|).
 \label{eq:coefficient-polynomial-stability}
\end{align}
The chain rule and the activation assumptions also give a polynomial local Lipschitz bound for $D_u\mathfrak S$ in its hidden state, signal state, and deeper weights. Since $L_k$ and $L_k^{\mathrm{aux}}$ have the same signal state, and all other arguments at both endpoints are bounded by $P_b$ on $\mathcal E_b$, this bound and \eqref{eq:coefficient-polynomial-stability} give
\begin{equation}
 \left\|\frac1m\sum_{k=1}^m
 \{H_t(L_k,\bm\rho^{[t-1]})
         -H_t(L_k^{\mathrm{aux}},\bm\rho^{[t-1]})\}\right\|
 \le P_b^C\max_{s<t}(E_{1,s}+E_{2,s}).
 \label{eq:actual-auxiliary-coefficient-average}
\end{equation}
Here the normalized row difference was bounded by Cauchy--Schwarz, as above. The actual row moments in \Cref{lem:gd-path-localization}, the auxiliary row moments, and the finite polynomial representation give polynomial conditional moments for both averages. Cauchy--Schwarz on $\mathcal E_b^c$, and \eqref{eq:gd-auxiliary-direct-comparison} on $\mathcal E_b$, therefore give a bound $(C\kappa\log n)^C n^{-1/C}$ for every fixed moment of the left-hand side of \eqref{eq:actual-auxiliary-coefficient-average}.

We next identify the Gaussian expectation of the auxiliary average. At a state-evolution row, let $L$ be the corresponding block diagonal matrix of loss-gradient Jacobians, including the factors $\eta_1^{(s-1)}$, and let $O=\mathfrak O_t(\bm\rho^{[t-1]})$. Differentiating the row-state recursion with respect to its Gaussian coordinates at times $1,\ldots,t$ gives the block response Jacobian
\begin{equation*}
 J=I_{t,q}-\phi^{-1}OLJ,
 \qquad J=(I_{t,q}+\phi^{-1}OL)^{-1}.
\end{equation*}
By definition, $\bm\tau^{[t]}$ is the negative row average of the expected derivative of the loss-gradient vector, with each time block multiplied by its learning rate. Thus
\begin{equation*}
 \bm\tau^{[t]}
 =-\E_{\pi_m}\E[LJ]
 =-\E_{\pi_m}\E[(I_{t,q}+\phi^{-1}LO)^{-1}L]
 =\E_{\pi_m}\E[H_t(L,\bm\rho^{[t-1]})].
\end{equation*}
The middle equality is the push-through identity, which also follows by expanding both nilpotent inverses. Here $\pi_m$ is an independent uniform row index, and the inner expectation is over the Gaussian state-evolution variables. The first derivatives of the composition of $H_t$ with the loss-gradient Jacobians have polynomial growth, so this composition is pseudo-Lipschitz of a fixed order. Applying \Cref{lem:supp:input-localization} to this auxiliary test function, and then using \eqref{eq:actual-auxiliary-coefficient-average}, shows that
\begin{equation*}
 \bar{\bm\tau}^{[t]}\coloneqq
    \frac1m\sum_k H_t(L_k,\bm\rho^{[t-1]})
 \quad\text{satisfies}\quad
 \E\|\bar{\bm\tau}^{[t]}-\bm\tau^{[t]}\|^r
 \le(C\kappa\log n)^C n^{-1/C}.
\end{equation*}

It remains to replace the state-evolution coefficients $\bm\rho^{[t-1]}$ in this average by their estimates. The algorithm gives the exact recursions
\begin{align*}
 \widehat{\bm\tau}^{[t]}
 &=\frac1m\sum_k H_t(L_k,\widehat{\bm\rho}^{[t-1]}),\\
 \widehat{\bm\rho}^{[t]}
 &=I_{t,q}+(\widehat{\bm\tau}^{[t]}+I_{t,q})
                    \mathfrak O_t(\widehat{\bm\rho}^{[t-1]}).
\end{align*}
Since these are finite polynomial recursions, induction and \Cref{lem:gd-path-localization} give polynomial conditional moments of every fixed order for both estimated coefficient matrices. The state-evolution coefficients have the same moment bounds by the induction in \Cref{lem:supp:input-localization}. By \eqref{eq:coefficient-polynomial-stability} and conditional Cauchy--Schwarz,
\begin{align*}
 \E\|\widehat{\bm\tau}^{[t]}-\bm\tau^{[t]}\|^r
 &\le(C\kappa\log n)^C
 \left\{n^{-1/C}
       +(\E\|\widehat{\bm\rho}^{[t-1]}
                    -\bm\rho^{[t-1]}\|^{2r})^{1/2}\right\},\\
 \E\|\widehat{\bm\rho}^{[t]}-\bm\rho^{[t]}\|^r
 &\le(C\kappa\log n)^C
 \left\{(\E\|\widehat{\bm\tau}^{[t]}
                    -\bm\tau^{[t]}\|^{2r})^{1/2}
       +(\E\|\widehat{\bm\rho}^{[t-1]}
                    -\bm\rho^{[t-1]}\|^{2r})^{1/2}\right\}.
\end{align*}
At $t=0$ the coefficient errors are zero. Inducting in $t$, simultaneously for every fixed moment order, proves \eqref{eq:gd-coefficient-direct-comparison}. Fixed powers of $\log n$ can be absorbed into a smaller positive power of $n$, so these estimates also have the form $(C\kappa)^C n^{-1/C}$.
\end{proof}

With $A=X/\sqrt n$, the corresponding row-state differences satisfy $\hat u^{(s+1)}-u^{(s+1)}=A(\hat v_1^{(s)}-v_1^{(s)})$, which relates their comparison to \eqref{eq:gd-auxiliary-direct-comparison}.

\begin{suppproposition}[auxiliary GFOM iterates with appended target-policy index coordinates]
\label{prop:supp:auxiliary_gfom_policy_index}
Under the assumptions in \Cref{sec:formal-regime}, fix an arm $a$ and $t\ge0$. On $E_{\mathrm{cnt}}$, conditional on $\mathcal G_a$, the following hold.
\begin{enumerate}
\item\label{item:supp:s2_decoupling} Appending the policy indices leaves the original GFOM iterates unchanged:
\begin{equation*}
  (\widetilde u_a^{(t)})^{\mathrm{aux}}=u_a^{(t)},
  \qquad
  (\widetilde v_a^{(t)})^{\mathrm{aux}}=v_a^{(t)},
  \qquad
  (\widetilde u_a^{(t)})^{\mathrm{idx}}=X_a\Upsilon,
  \qquad
  (\widetilde v_a^{(t)})^{\mathrm{idx}}=0.
\end{equation*}

\item\label{item:supp:s2_state_evolution} There is a centered jointly Gaussian history for the row state evolution,
\begin{equation*}
  \widetilde{\mathfrak U}_a^{(s)}
  =
  \bigl(\mathfrak U_a^{(s)},\mathfrak U_{\Upsilon,a}\bigr),
  \qquad
  s=0,\dots,t,
\end{equation*}
whose first $q$ coordinates coincide with the arm-wise Gaussian state evolution using the padded reward link in \HIprop{prop:se_aux_GFOM}. The final $\ell$ coordinates equal the same Gaussian coordinate vector for the policy index, $\mathfrak U_{\Upsilon,a}\sim N(0,I_\ell)$, at every time. The corresponding state-evolution maps for the rows satisfy
\begin{equation*}
  \widetilde\Theta_{a;k}^{(s)}
  \bigl(\widetilde{\mathfrak U}_a^{([0:s])}\bigr)
  =
  \Bigl(
    \Theta_{a;k}^{(s)}(\mathfrak U_a^{([0:s])}),
    \mathfrak U_{\Upsilon,a}
  \Bigr),
  \qquad
  s=0,\ldots,t,
\end{equation*}
where $\Theta_{a;k}^{(s)}$ is the original arm-wise row-indexed state evolution map. Moreover, for every $s=0,\ldots,t$,
\begin{equation*}
  \operatorname{Cov}\bigl(\mathfrak U_a^{(s)},\mathfrak U_{\Upsilon,a}\bigr)
  =
  d^{-1/2}D_{a,s-1}^{\top}\Upsilon,
  \qquad
  \operatorname{Var}(\mathfrak U_{\Upsilon,a})=I_\ell.
\end{equation*}

\item\label{item:supp:s2_pl} Fix a pseudo-Lipschitz order $\mathfrak p$, $\Lambda_\vartheta\ge2$, and $\mathfrak r\ge1$. Let $\{\vartheta_k\}_{k=1}^{N_a}$ be deterministic or $\mathcal G_a$-measurable test functions on $\mathbb R^{(q+\ell)[0:t]}$. After conditioning on $\mathcal G_a$, suppose that they form a deterministic family of $\Lambda_\vartheta$-pseudo-Lipschitz functions of order $\mathfrak p$. Then, there exist $c_t,C_t<\infty$ such that
\begin{align*}
  &\E_a^{(0)}\bigg[
    \bigg|
      \frac1{N_a}\sum_{k=1}^{N_a}
        \vartheta_k\bigl(\widetilde u_{a;k}^{([0:t])}\bigr)
      -
      \frac1{N_a}\sum_{k=1}^{N_a}
        \E_a^{(0)}
        \vartheta_k\Bigl(
          \bigl\{
            \widetilde\Theta_{a;k}^{(s)}
            (\widetilde{\mathfrak U}_a^{([0:s])})
          \bigr\}_{s=0}^t
        \Bigr)
    \bigg|^{\mathfrak r}
  \bigg]\\
  &\quad
  +
  \max_{\alpha\in[2:L]}\max_{s\in[1:t]}
    \E_a^{(0)}
    \bigl\|
      (\widetilde v_{a,\alpha}^{(s)})^{\mathrm{aux}}-V_{a,\alpha}^{(s)}
    \bigr\|_{\infty}^{\mathfrak r}
  \le
  (C_t\Lambda_\vartheta\varkappa_{\ast,a})^{c_t}\cdot d^{-1/c_t}.
\end{align*}

\item\label{item:supp:s2_delocalization} For each fixed $t$, after enlarging $C_t$,
\begin{equation*}
  \Prob_a^{(0)}
  \left(
    \|\widetilde u_a^{(t)}\|_{\infty}
    \vee
    \|\widetilde v_a^{(t)}\|_{\infty}
    >
    C_t(\varkappa_{\ast,a}\log d)^{C_t}
  \right)
  \le d^{-100}.
\end{equation*}
\end{enumerate}

\end{suppproposition}

\begin{proof}
Throughout, we fix an arm $a$ and suppress the subscript $a$, retaining $\phi_a$ for the arm-wise sample-to-dimension ratio. To match HI25 within this proof, use $m$ for the arm-wise sample size $N_a$, rather than the full logged sample size, and set $n=d$. Write $X\coloneqq X_a$, $\phi_a=m/n=N_a/d$, and $\beta\coloneqq\beta_a$. Retain the definitions above and set $n_{L,q}\coloneqq d_{L,q}$, $A\coloneqq A_a$, $\Upsilon_{\mathrm{pad}}\coloneqq\Upsilon_{\mathrm{pad},a}$, and $\bar\mu_{\ast,n}\coloneqq\bar\mu_{a,\ast,d}$. Thus, $A\Upsilon_{\mathrm{pad}}=X\Upsilon$. Under $\E_a^{(0)}[\cdot]$, before appending the target-policy index, the arm-wise padded problem is covered by \Cref{lem:hi25-component-link} under the relabeling $m=N_a$, $n=d$, $\mu_\ast=\beta_a$, and $\boldsymbol\varphi=\varphi_{a,[q]}^\ast$. Let $(u^{(t)},v^{(t)})$ denote the corresponding auxiliary GFOM iterates, and let $\Theta_k^{(s)}$ and $V_\alpha^{(s)}$ be the original arm-wise state evolution objects from \HIprop{prop:se_aux_GFOM}.

We append the target-policy indices by defining update maps with $q+\ell$ coordinates in each row:
\begin{alignat*}{2}
  \bigl[\widetilde F_t^{\langle1\rangle}(\tilde v^{([-1:t-1])})\bigr]^{\mathrm{aux}}
    &=
    F_t^{\langle1\rangle}\bigl((\tilde v^{([-1:t-1])})^{\mathrm{aux}}\bigr),
  &\qquad
  \bigl[\widetilde F_t^{\langle1\rangle}(\tilde v^{([-1:t-1])})\bigr]^{\mathrm{idx}}
    &=
    \Upsilon_{\mathrm{pad}},\\
  \bigl[\widetilde F_t^{\langle2\rangle}(\tilde v^{([-1:t-1])})\bigr]^{\mathrm{aux}}
    &=
    F_t^{\langle2\rangle}\bigl((\tilde v^{([-1:t-1])})^{\mathrm{aux}}\bigr),
  &\qquad
  \bigl[\widetilde F_t^{\langle2\rangle}(\tilde v^{([-1:t-1])})\bigr]^{\mathrm{idx}}
    &=
    0,\\
  \bigl[\widetilde G_t^{\langle1\rangle}(\tilde u^{([-1:t-1])})\bigr]^{\mathrm{aux}}
    &=
    G_t^{\langle1\rangle}\bigl((\tilde u^{([-1:t-1])})^{\mathrm{aux}}\bigr),
  &\qquad
  \bigl[\widetilde G_t^{\langle1\rangle}(\tilde u^{([-1:t-1])})\bigr]^{\mathrm{idx}}
    &=
    0,\\
  \bigl[\widetilde G_t^{\langle2\rangle}(\tilde u^{([-1:t])})\bigr]^{\mathrm{aux}}
    &=
    G_t^{\langle2\rangle}\bigl((\tilde u^{([-1:t])})^{\mathrm{aux}}\bigr),
  &\qquad
  \bigl[\widetilde G_t^{\langle2\rangle}(\tilde u^{([-1:t])})\bigr]^{\mathrm{idx}}
    &=
    0,
\end{alignat*}
and initialize
\begin{equation*}
  \widetilde u^{(-1)}=0_{m\times(q+\ell)},
  \qquad
  \widetilde v^{(-1)}=
  \bigl[\bar\mu_{\ast,n}\,\big|\,0_{n_{L,q}\times\ell}\bigr].
\end{equation*}
The resulting recursion has the same form as the original auxiliary GFOM, with $q+\ell$ state coordinates per row.

By induction on $t$, the first $q$ coordinates are the original auxiliary GFOM iterates, while the target-policy index coordinates satisfy
\begin{align*}
  (\widetilde u^{(t)})^{\mathrm{idx}}
  &=
  A\bigl[\widetilde F_t^{\langle1\rangle}\bigr]^{\mathrm{idx}}
  +\bigl[\widetilde G_t^{\langle1\rangle}\bigr]^{\mathrm{idx}}
  =A\Upsilon_{\mathrm{pad}}
  =X\Upsilon,\\
  (\widetilde v^{(t)})^{\mathrm{idx}}
  &=
  A^\top\bigl[\widetilde G_t^{\langle2\rangle}\bigr]^{\mathrm{idx}}
  +\bigl[\widetilde F_t^{\langle2\rangle}\bigr]^{\mathrm{idx}}
  =0.
\end{align*}
The first $q$ coordinates satisfy the original recursion because their update maps depend only on the histories of those coordinates. This proves item~\ref{item:supp:s2_decoupling} simultaneously for every $t\ge0$.

We first identify the formal state-evolution recursion. Its empirical comparison will follow from \Cref{lem:supp:input-localization}. Put
\begin{equation*}
 \bar A_0=\sqrt{\frac{n_{L,q}}n}\,(X,0_{m\times(L-1)q}),
 \qquad A=\frac{\bar A_0}{\sqrt{n_{L,q}}}.
\end{equation*}
Since $n_{L,q}/n=1+O(n^{-1})$ and $m/n_{L,q}$ is bounded above and away from zero on $E_{\mathrm{cnt}}$, $A$ satisfies $(D^\ast1)$ with the padded column dimension $n_{L,q}$. The zero columns are permitted by its variance-profile formulation. Every output row depends only on the corresponding input rows. The appended maps are constant or zero, and
\begin{equation*}
 \|\Upsilon_{\mathrm{pad}}\|_\infty
 =\max_{j\in[\ell]}\|\sqrt n\,\gamma_j\|_\infty\le C_\Upsilon.
\end{equation*}
The first $q$ coordinate maps have the polynomial local-Lipschitz bounds established in \Cref{lem:supp:input-localization}; their input-cutoff versions have global Lipschitz constants and values at zero bounded by a fixed polynomial in the cutoff radius and $\varkappa_{\ast,a}$. Thus, the hypotheses used by the averaged GFOM theorem and the polynomial-growth delocalization proposition are exactly those verified in that lemma.

For every history argument and every time index $j$, the relevant block Jacobians have the form
\begin{equation*}
  \nabla_j\widetilde F_t^{\langle h\rangle}
  =
  \begin{pmatrix}
    \nabla_jF_t^{\langle h\rangle}&0\\
    0&0
  \end{pmatrix},
  \qquad
  \nabla_j\widetilde G_t^{\langle h\rangle}
  =
  \begin{pmatrix}
    \nabla_jG_t^{\langle h\rangle}&0\\
    0&0
  \end{pmatrix}.
\end{equation*}
Indeed, the first $q$ output coordinates do not depend on the policy-index histories, and the policy-index outputs are constant or zero. The Onsager blocks involving the first $q$ coordinates, the coefficients $\rho_{t,s}$, and the deterministic shifts $D_t$ are therefore unchanged from the original HI25 recursion. All Jacobian blocks involving the appended coordinates are zero. Substitution into the generic asymmetric GFOM recursion in \HIdef{def:GFOM_se_asym} therefore gives the same first-$q$-coordinate state evolution, with the appended covariance computed next.

We now compute the augmented row covariance generated by that recursion. For $s\ge1$, let the column-side state entering the $s$th row covariance be
\begin{equation*}
  Q_j^{(s)}
  =
  (D_{s-1})_{j\cdot}
  +
  \sum_{r=1}^{s-1}
  \mathfrak V_j^{(r)}\rho_{s-1,r}^\top,
  \qquad j\in[n],
\end{equation*}
with the initialization $Q_j^{(0)}=(D_{-1})_{j\cdot}$. The centered column states satisfy $\mathbb E_a^{(0)}\mathfrak V_j^{(r)}=0$, so $\mathbb E_a^{(0)}Q_j^{(s)}=(D_{s-1})_{j\cdot}$. The column-side vector in the augmented recursion is $(Q_j^{(s)},\sqrt n\,\Upsilon_{j\cdot})$. Hence, the cross-time covariance block for augmented row states at times $s$ and $r$ is
\begin{equation*}
  \begin{pmatrix}
    \Omega_{s-1,r-1}+n^{-1}D_{s-1}^\top D_{r-1}
    &n^{-1/2}D_{s-1}^\top\Upsilon\\
    n^{-1/2}\Upsilon^\top D_{r-1}
    &I_\ell
  \end{pmatrix}.
\end{equation*}
The new blocks follow directly from
\begin{align*}
  \frac1n\sum_{j=1}^n
  \mathbb E_a^{(0)}[Q_j^{(s)}]^\top
  \sqrt n\,\Upsilon_{j\cdot}
  &=n^{-1/2}D_{s-1}^\top\Upsilon,\\
  \frac1n\sum_{j=1}^n
  n\,\Upsilon_{j\cdot}^\top\Upsilon_{j\cdot}
  &=\Upsilon^\top\Upsilon=I_\ell.
\end{align*}
Thus, the index--index covariance equals $I_\ell$ for every pair of times. If $J^{(s)}$ and $J^{(r)}$ are two such Gaussian index states, then
\begin{equation*}
  \mathbb E\|J^{(s)}-J^{(r)}\|^2
  =\operatorname{tr}(I_\ell+I_\ell-2I_\ell)=0.
\end{equation*}
Thus, the index states share the same Gaussian vector at every time.

For the appended index coordinates, the row-indexed iterate equals the rowwise linear feature $X\Upsilon$ at every time. Denote the common Gaussian index vector by $\mathfrak U_\Upsilon$. Since $\Upsilon^\top\Upsilon=I_\ell$ and $X_{k\cdot}^{\top}\sim N(0,I_n)$ under $\E_a^{(0)}$, this vector has law $N(0,I_\ell)$. Let
\begin{equation*}
  D_{-1}
  \coloneqq
  \bigl(\sqrt n\,\beta,\ 0_{n\times(q-1)}\bigr)\in\mathbb R^{n\times q},
\end{equation*}
and let $D_0,\ldots,D_{t-1}$ be the original arm-wise deterministic shifts from \HIdef{def:gd_NN_se}-(S4). We use the convention $\Omega_{-1,j}=\Omega_{j,-1}=0$ for $j=-1,0,\ldots,t-1$, consistent with $\mathfrak G^{(0)}=0$. To realize the original Gaussian state evolution explicitly, first suppose $t\ge1$ and let $(\mathfrak G^{(1)},\ldots,\mathfrak G^{(t)})\in(\R^q)^{[1:t]}$ be centered jointly Gaussian with covariance blocks
\begin{equation*}
  \operatorname{Cov}\bigl(\mathfrak G^{(s)},\mathfrak G^{(r)}\bigr)
  =
  \Omega_{s-1,r-1},
  \qquad
  s,r\in[1:t],
\end{equation*}
which exists because the block matrix
\begin{equation*}
  \bigl(\Omega_{s-1,r-1}\bigr)_{s,r=1}^t
  =
  0_{q\times q}\oplus\bm{\Omega}^{[t-1]}
  =
  0_{q\times q}\oplus
  \bigl(\bm{\rho}^{[t-1]}\bm{\Sigma}^{[t-1]}
  \bm{\rho}^{[t-1]\top}\bigr)
\end{equation*}
is positive semidefinite by \HI~\eqref{HI-def:rho_se_matrix}. When $t=0$, this Gaussian family is empty. In either case, let $Z\sim N(0,I_n)$ be independent of $(\mathfrak G^{(1)},\ldots,\mathfrak G^{(t)})$, and define
\begin{equation*}
  \mathfrak G^{(0)}\coloneqq 0,
  \qquad
  \mathfrak U^{(s)}
  \coloneqq
  \mathfrak G^{(s)} + n^{-1/2}D_{s-1}^{\top}Z,
  \qquad
  s=0,\ldots,t,
\end{equation*}
and
\begin{equation*}
  \mathfrak U_\Upsilon
  \coloneqq
  \Upsilon^\top Z.
\end{equation*}
Then, $(\mathfrak U^{(0)},\ldots,\mathfrak U^{(t)})$ is centered jointly Gaussian with
\begin{equation*}
  \operatorname{Cov}\bigl(\mathfrak U^{(s)},\mathfrak U^{(r)}\bigr)
  =
  \Omega_{s-1,r-1}+n^{-1}D_{s-1}^{\top}D_{r-1},
  \qquad
  s,r=0,\ldots,t,
\end{equation*}
which is exactly the covariance specification in \HIdef{def:gd_NN_se}-(S2) for the original arm-wise row-indexed state evolution. Moreover,
\begin{equation*}
  \operatorname{Cov}\bigl(\mathfrak U^{(s)},\mathfrak U_\Upsilon\bigr)
  =
  n^{-1/2}D_{s-1}^{\top}\Upsilon,
  \qquad
  \operatorname{Var}(\mathfrak U_\Upsilon)=I_\ell.
\end{equation*}
Thus, there is a centered jointly Gaussian extension
\begin{equation*}
  \widetilde{\mathfrak U}^{(s)}
  =
  \bigl(\mathfrak U^{(s)},\mathfrak U_\Upsilon\bigr),
  \qquad
  s=0,\ldots,t,
\end{equation*}
and the row-indexed state evolution map with appended target-policy index coordinates has the block form
\begin{equation*}
  \widetilde\Theta_k^{(s)}
  \bigl(\widetilde{\mathfrak U}^{([0:s])}\bigr)
  =
  \Bigl(
    \Theta_k^{(s)}(\mathfrak U^{([0:s])}),
    \mathfrak U_\Upsilon
  \Bigr),
  \qquad
  s=0,\ldots,t.
\end{equation*}
This proves item~\ref{item:supp:s2_state_evolution}.

For the empirical comparison, use the input-cutoff construction of \Cref{lem:supp:input-localization}, with the original $V^{(s)}$ and $M^{(s)}$ held fixed. Write its iterates as $(\check u^{(s)},\check v^{(s)})$ and append the same observed index. Then
\begin{equation*}
 \left\|\widetilde u_k^{([0:t])}
       -\widetilde{\check u}_k^{([0:t])}\right\|
 =\left\|u_k^{([0:t])}-\check u_k^{([0:t])}\right\|.
\end{equation*}
For an order-$\mathfrak p$ pseudo-Lipschitz test,
\begin{align*}
 &\left|\vartheta_k(\widetilde u_k^{([0:t])})
             -\vartheta_k(\widetilde{\check u}_k^{([0:t])})\right| \le\Lambda_\vartheta
  \left(1+\|u_k^{([0:t])}\|+\|\check u_k^{([0:t])}\|
       +2\sqrt{t+1}\|X_{k\cdot}\Upsilon\|\right)^{\mathfrak p-1}
  \|u_k^{([0:t])}-\check u_k^{([0:t])}\|.
\end{align*}
The moment bounds in the lemma control this empirical localization error, while its Gaussian coupling uses the same policy index $\Upsilon^\top Z$ for both histories. Since the test family is deterministic after conditioning on $\mathcal G_a$, the lemma's averaged comparison yields
\begin{align*}
 &\E_a^{(0)}\left[\left|
  \frac1{N_a}\sum_{k=1}^{N_a}\vartheta_k(\widetilde u_k^{([0:t])})
  -\frac1{N_a}\sum_{k=1}^{N_a}\E_a^{(0)}
   \vartheta_k\left(\left\{
    \widetilde\Theta_k^{(s)}(\widetilde{\mathfrak U}^{([0:s])})
                  \right\}_{s=0}^t\right)
                  \right|^{\mathfrak r}\right] \le (C_t\Lambda_\vartheta\varkappa_{\ast,a})^{c_t}d^{-1/c_t}.
\end{align*}
The deeper auxiliary blocks satisfy $(\widetilde v_\alpha^{(s)})^{\mathrm{aux}}=v_\alpha^{(s)}=V_\alpha^{(s)}$ exactly, because the padded columns of $A$ vanish. This proves item~\ref{item:supp:s2_pl}.

For delocalization, put $B_t=C_t(\varkappa_{\ast,a}\log d)^{C_t}$. The rowwise bound in \HIprop{prop:GFOM_deloc_asym}, with its tail parameter chosen as $C\log d$, gives after increasing $C_t$
\begin{align*}
  &\Prob_a^{(0)}\left(
    \|u^{(t)}\|_\infty\vee\|v^{(t)}\|_\infty>B_t
  \right)\\*
  &\quad\le
  (N_a+n_{L,q})
  \max_{k,j}
  \Prob_a^{(0)}\left(
    \|u_k^{(t)}\|_\infty\vee\|v_j^{(t)}\|_\infty>B_t
  \right)
  \le d^{-101},
\end{align*}
because $N_a+n_{L,q}=O(d)$ on $E_{\mathrm{cnt}}$ and $q$ is fixed. The appended coordinates are explicit:
\begin{equation*}
  (\widetilde u^{(t)})^{\mathrm{idx}}=X\Upsilon,
  \qquad
  (\widetilde v^{(t)})^{\mathrm{idx}}=0.
\end{equation*}
Each entry of $X\Upsilon$ is $N(0,1)$. Hence, for every $z>0$,
\begin{equation*}
  \Prob_a^{(0)}
  \bigl(\|X\Upsilon\|_\infty>z\bigr)
  \le2N_a\ell e^{-z^2/2}.
\end{equation*}
Taking $z=C\sqrt{\log d}$ and increasing $C$ makes this probability at most $d^{-101}$. Since $\varkappa_{\ast,a}\ge1$, this threshold is bounded by $B_t$ after increasing $C_t$. A union bound proves item~\ref{item:supp:s2_delocalization}.
\end{proof}

The appended coordinate vector is the observed index $X_{a;k\cdot}\Upsilon\in\R^\ell$, and the proposition identifies its joint state evolution with the auxiliary iterates, including their cross-covariances.

\section{Weighted Risk Estimation with the Empirical Noise Distribution}

We first compare the corrected empirical risk with a risk evaluated at an independent Gaussian context and a noise value sampled uniformly from the realized arm-wise noise values. The index $\pi$ in $\Es_{a,\pi}^{(t)}$ selects this noise value. This intermediate risk is used only in the proof; the procedure does not require the unobserved noise values. The next section replaces the sampled noise value by an independent draw from the arm-wise noise distribution.

\subsection{Auxiliary states and comparison estimates}

For the next comparisons, fix an arm $a$ and an iteration $t$. Let $\Is_a=\{i_{a,1}<\cdots<i_{a,N_a}\}$ and relabel the arm-wise sample by
\begin{equation*}
  X_k\coloneqq X_{i_{a,k}},\qquad
  R_k\coloneqq R_{i_{a,k}},\qquad
  \xi_k\coloneqq R_{i_{a,k}}-\varphi_a^\ast(\beta_a^\top X_{i_{a,k}}),
  \qquad
  k\in[N_a].
\end{equation*}
Let $X\in\mathbb R^{N_a\times d}$ denote the arm-wise design matrix with $k$-th row $X_k^\top$. Set $N\coloneqq N_a$, $\phi_a\coloneqq N/d$, $\beta\coloneqq \beta_a$, and $h\coloneqq h_a$. Define
\begin{equation*}
  \beta_{[q]}\coloneqq \bigl(\beta\,|\,0_{d\times(q-1)}\bigr)\in\mathbb R^{d\times q}.
\end{equation*}
Conditional on $\mathcal G_a$, which determines $N$, let $\pi_N$ be uniform on $[N]$ and independent of the arm-wise design, the fresh Gaussian vectors, and the state-evolution Gaussian variables. It remains uniform and independent of the fresh variables after conditioning on $\widehat{\mathcal F}_a$. Under this relabeling, the envelope $\varkappa_{\ast,a}$ in \eqref{eq:arm-wise_kappa} is exactly HI25's envelope in \eqref{HI-def:kappa_ast}. Hence, by \eqref{eq:envelope-polynomial-vanishing}, for every fixed $C,c>0$,
\begin{equation*}
  (\varkappa_{\ast,a}\log d)^C d^{-c}=o_p(1).
\end{equation*}
We use this repeatedly below to turn the polynomial error bounds at each fixed iteration into $o_p(1)$ remainders. Throughout, we work on $E_{\mathrm{cnt}}$ and condition on $\mathcal G_a$, using the expectation $\E_a^{(0)}$ and probability $\Prob_a^{(0)}$ from \Cref{prop:supp:arm-wise_conditioning}. The covariates $\{X_k\}_{k\in[N]}$ remain random and form the Gaussian HI25 design, while $\beta,\Upsilon,W_a^{(0)}$, and $\xi_{[N]}$ are fixed. Empirical quantities are measurable with respect to $\widehat{\mathcal F}_a$ from the same proposition, whereas the state-evolution comparisons are conditional on $\mathcal G_a$.

Let
\begin{equation*}
  \hat u^{(0)}\coloneqq X\beta_{[q]},\qquad
  \hat u^{(s)}\coloneqq XW_1^{(s-1)}\quad(s\ge1),
\end{equation*}
let $U^{(t)}$ be the theoretical corrected hidden-state matrix defined below, let $u^{(s)}$ be the original $q$-dimensional auxiliary GFOM iterate from Proposition~\ref{prop:supp:auxiliary_gfom_policy_index}, and let $V^{(s)}=(V_{\alpha}^{(s)})_{\alpha\in[2:L]}$ and $\rho_{t,s}$ be the arm-wise state-evolution objects. For each row $k$, write $R_{k;[q]}\coloneqq (R_k,0,\ldots,0)^\top$ and $\xi_{k;[q]}\coloneqq (\xi_k,0,\ldots,0)^\top$. Let
\begin{equation*}
  R_{[q]}
  \coloneqq
  \begin{pmatrix}
    R_{1;[q]}^\top\\
    \vdots\\
    R_{N;[q]}^\top
  \end{pmatrix},
  \qquad
  \xi_{[q]}
  \coloneqq
  \begin{pmatrix}
    \xi_{1;[q]}^\top\\
    \vdots\\
    \xi_{N;[q]}^\top
  \end{pmatrix}.
\end{equation*}
Recall the padded reward link in \eqref{eq:arm-component-link}. For a matrix $W\in\R^{N\times q}$, let
\begin{equation*}
  \varphi_{a,[q]}^\ast(W)
  \coloneqq
  \begin{pmatrix}
    \varphi_{a,[q]}^\ast(W_{1\cdot})^\top\\
    \vdots\\
    \varphi_{a,[q]}^\ast(W_{N\cdot})^\top
  \end{pmatrix}
  \in\R^{N\times q},
\end{equation*}
that is, the same padded link applied rowwise. Define the matrix-valued arm-wise residual map
\begin{equation*}
  \Rsc^{\mathrm{arm}}_{B}(U,W)
  \coloneqq
  \Gsc_{B}(U)-\varphi_{a,[q]}^\ast(W)-\xi_{[q]}
  \in\R^{N\times q},
\end{equation*}
and, for each row $k$, its rowwise version
\begin{equation*}
  \Rsc^{\mathrm{arm}}_{B;k}(u,w)
  \coloneqq
  \Gsc_{B;k}(u)-\varphi_{a,[q]}^\ast(w)-\xi_{k;[q]}
  \in\R^q.
\end{equation*}
When a row index is attached to $\Gsc$, it denotes the rowwise application of the same feed-forward map; the map $\Gsc_B$ itself has no additional dependence on $k$. Here $B=(B_2,\ldots,B_L)\in(\R^{q\times q})^{[2:L]}$ denotes the deeper-layer parameter tuple. Define the derivative of the row loss with respect to its hidden state by
\begin{equation*}
  \mathfrak S_k(u,w,B)
  \coloneqq
  \left\{\nabla_u\frac12
    \|\Rsc^{\mathrm{arm}}_{B;k}(u,w)\|^2\right\}^{\top}
  \in\R^{1\times q},
\end{equation*}
and let $\mathfrak S(U,W,B)$ collect these rows. This is the HI25 pre-gradient map with the padded reward link $\varphi_{a,[q]}^\ast$. Along the gradient-descent trajectory,
\begin{equation*}
  \mathfrak S_k(\hat u_k^{(s)},\hat u_k^{(0)},W_{[2:L]}^{(s-1)})
  =\Psi_{i_{a,k},a}^{(s-1)}.
\end{equation*}
The deeper-layer derivatives use the same padded residual. All HI25 comparisons invoked below therefore use the extension in \Cref{lem:hi25-component-link}. By the zero-column invariance established in \Cref{sec:formal-regime}, the padded residual is supported on its first coordinate along the gradient-descent, auxiliary, and state-evolution trajectories. Its squared norm is therefore exactly the scalar reward residual squared. Then, since $\hat u_k^{(0)}=(\beta^\top X_k,0,\ldots,0)$,
\begin{equation*}
  R_{k;[q]}=\varphi_{a,[q]}^\ast(\hat u_k^{(0)})+\xi_{k;[q]}.
\end{equation*}

For the comparisons below, write
\begin{equation*}
  \widehat U^{(t)}\coloneqq\widehat U_a^{(t)},\qquad
  W^{(t)}\coloneqq W_a^{(t)},\qquad
  \hat v_1^{(t)}\coloneqq\sqrt d\,W_1^{(t)},\qquad
  v_1^{(t)}\coloneqq v_{1,a}^{(t)}.
\end{equation*}
The auxiliary GFOM recursion makes $v_1^{(t)}$ a measurable function of the arm-wise design, response, initialization, and fixed problem parameters. Thus, $v_1^{(t)}$ is $\widehat{\mathcal F}_a$-measurable. Set
\begin{equation*}
  D_{-1}\coloneqq(\sqrt d\,\beta,0_{d\times(q-1)}).
\end{equation*}
For $s\ge0$, write $D_s\coloneqq D_{a,s}$ for the arm-wise deterministic state-evolution shifts. To compare $\widehat U^{(t)}$ with the auxiliary GFOM, we first replace the estimated Onsager coefficients by $\rho_{t,s}$, then replace the deeper weights in the loss gradients by $V^{(s-1)}$, and finally replace the GD hidden states by the auxiliary iterates. These three replacements define, respectively,
\begin{align*}
  U^{(t)}
  &\coloneqq
  \hat u^{(t+1)}
  +\phi_a^{-1}\sum_{s=1}^{t}\eta_1^{(s-1)}
  \mathfrak S(\hat u^{(s)},\hat u^{(0)},W_{[2:L]}^{(s-1)})
  \rho_{t,s}^\top,\\
  \bar U^{(t)}
  &\coloneqq
  \hat u^{(t+1)}
  +\phi_a^{-1}\sum_{s=1}^{t}\eta_1^{(s-1)}
  \mathfrak S(\hat u^{(s)},\hat u^{(0)},V^{(s-1)})\rho_{t,s}^\top,\\
  \bar U_{\mathrm{aux}}^{(t)}
  &\coloneqq
  u^{(t+1)}
  +\phi_a^{-1}\sum_{s=1}^{t}\eta_1^{(s-1)}
  \mathfrak S(u^{(s)},u^{(0)},V^{(s-1)})\rho_{t,s}^\top.
\end{align*}

Let $Z\sim N(0,I_d)$ be independent of $\mathcal F_{\mathrm{tr}}$, and define
\begin{equation*}
  G_t
  \coloneqq
  \left(
    d^{-1/2}v_1^{(t)\top}Z,\,
    d^{-1/2}D_{-1}^\top Z,\,
    \Upsilon^\top Z
  \right),
  \qquad
  G_{0,t}
  \coloneqq
  \left(
    \mathfrak U_a^{(t+1)},\,
    \mathfrak U_a^{(0)},\,
    \mathfrak U_{\Upsilon,a}
  \right).
\end{equation*}
Let $\widetilde\Sigma_t$ be the covariance of $G_t$ over the fresh draw $Z$, holding $(\mathcal G_a,v_1^{(t)})$ fixed, and let $\widetilde\Sigma_{0,t}$ be the conditional covariance of $G_{0,t}$ under $\E_a^{(0)}$.

\begin{lemma}[Comparison estimates for the arm-wise training iterates]
\label{lem:arm-wise-hi25-consequences}
Assume the conditions of \Cref{sec:formal-regime}. For every fixed $a\in[K]$, $t\in\mathbb Z_{\ge0}$, and $r\ge1$, there exists $C_{t,r}<\infty$ such that, on $E_{\mathrm{cnt}}$, with
\begin{equation*}
  \delta_{a,t,r}
  \coloneqq
  C_{t,r}(\varkappa_{\ast,a}\log d)^{C_{t,r}}d^{-1/C_{t,r}},
\end{equation*}
\begin{align*}
  &\E_a^{(0)}\left(N^{-1/2}\|\widehat U^{(t)}-U^{(t)}\|\right)^r
  +\E_a^{(0)}\left(N^{-1/2}\|U^{(t)}-\bar U^{(t)}\|\right)^r\\
  &\quad
  +\E_a^{(0)}\left(N^{-1/2}
    \|\bar U^{(t)}-\bar U_{\mathrm{aux}}^{(t)}\|\right)^r
  +\E_a^{(0)}\left(d^{-1/2}
    \|\hat v_1^{(t)}-v_1^{(t)}\|\right)^r\\
  &\quad
  +\max_{0\le s\le t}\max_{\alpha\in[2:L]}
    \E_a^{(0)}\|W_\alpha^{(s)}-V_\alpha^{(s)}\|_\infty^r
  \le\delta_{a,t,r}.
\end{align*}
For every fixed integer $p\ge1$, each of the following random variables has conditional $r$th moment at most $C_{t,p,r}(\varkappa_{\ast,a}\log d)^{C_{t,p,r}}$:
\begin{equation*}
  \max_{Q\in\{\widehat U^{(t)},U^{(t)},\bar U^{(t)},
    \bar U_{\mathrm{aux}}^{(t)}\}}N^{-1/2}\|Q\|,
  \quad
  d^{-1/2}\|\hat v_1^{(t)}\|,
  \quad
  d^{-1/2}\|v_1^{(t)}\|,
\end{equation*}
\begin{equation*}
  \max_{0\le s\le t}\max_{\alpha\in[2:L]}
  \bigl(\|W_\alpha^{(s)}\|_{\op}+\|V_\alpha^{(s)}\|_{\op}\bigr),
  \qquad
  \max_{1\le s\le t}\|\rho_{t,s}\|_{\op},
  \qquad
  \left\{\frac1N\sum_{k=1}^N
    (1+\|\hat u_k^{(0)}\|^{2p})\right\}^{1/2}.
\end{equation*}

Finally,
\begin{equation}
  \E_a^{(0)}
  \|\widetilde\Sigma_t-\widetilde\Sigma_{0,t}\|_\infty^r
  \le\delta_{a,t,r}.
  \label{eq:observable-covariance-comparison}
\end{equation}
\end{lemma}

\begin{proof}
After the relabeling $m=N_a$, $n=d$, and $\mu_\ast=\beta_a$, \Cref{prop:supp:arm-wise_conditioning} verifies the assumptions for the arm-wise problem with $\boldsymbol\varphi=\varphi_{a,[q]}^\ast$. The parameter and Onsager-coefficient comparisons follow from \Cref{lem:gd-auxiliary-coefficient-comparison}, applied at horizon $t+1$. We derive the corrected-state comparisons from those estimates and the deterministic loss-gradient bounds. For the first difference, the definitions give
\begin{align*}
  \widehat U^{(t)}-U^{(t)}
  ={}&
  \phi_a^{-1}\sum_{s=1}^t\eta_1^{(s-1)}
  \mathfrak S(\hat u^{(s)},\hat u^{(0)},W_{[2:L]}^{(s-1)})
  (\widehat\rho_{t,s}-\rho_{t,s})^\top.
\end{align*}
For $t\ge1$, the matrix norm inequality $\|AB\|_F\le\|A\|_F\|B\|_{\op}$ and conditional Cauchy--Schwarz give
\begin{align*}
  &\E_a^{(0)}(N^{-1/2}\|\widehat U^{(t)}-U^{(t)}\|)^r\\
  &\quad\le C_{t,r}\sum_{s=1}^t
  \left\{\E_a^{(0)}
    \left(N^{-1/2}\|
    \mathfrak S(\hat u^{(s)},\hat u^{(0)},W_{[2:L]}^{(s-1)})
    \|\right)^{2r}\right\}^{1/2}
  \left\{\E_a^{(0)}
    \|\widehat\rho_{t,s}-\rho_{t,s}\|_{\op}^{2r}
  \right\}^{1/2}\\
  &\quad\le\delta_{a,t,r}.
\end{align*}
The last step uses the coefficient comparison in \Cref{lem:gd-auxiliary-coefficient-comparison} at moment order $2r$, together with the loss-gradient moment bounds. At $t=0$, the difference is zero.

The next two differences are also exact:
\begin{align}
  U^{(t)}-\bar U^{(t)}
  ={}&
  \phi_a^{-1}\sum_{s=1}^t\eta_1^{(s-1)}
  \Bigl\{
    \mathfrak S(\hat u^{(s)},\hat u^{(0)},W_{[2:L]}^{(s-1)})
    -\mathfrak S(\hat u^{(s)},\hat u^{(0)},V^{(s-1)})
  \Bigr\}\rho_{t,s}^\top,
  \label{eq:population-deeper-replacement-difference}\\
  \bar U^{(t)}-\bar U_{\mathrm{aux}}^{(t)}
  ={}&
  \hat u^{(t+1)}-u^{(t+1)} \notag\\
  &+\phi_a^{-1}\sum_{s=1}^t\eta_1^{(s-1)}
  \Bigl\{
    \mathfrak S(\hat u^{(s)},\hat u^{(0)},V^{(s-1)})
    -\mathfrak S(u^{(s)},u^{(0)},V^{(s-1)})
  \Bigr\}\rho_{t,s}^\top.
  \label{eq:gd-auxiliary-state-difference}
\end{align}
For \eqref{eq:population-deeper-replacement-difference}, apply the loss-gradient difference bound in \HIprop{prop:apr_est_det_fcn}-(7), as extended by \Cref{lem:hi25-component-link}. Conditional H\"older's inequality separates the polynomial state and weight factors from the deeper-layer difference. Their moments are bounded by \Cref{lem:gd-path-localization}, the Gaussian signal moments, and \HIprop{prop:apr_est_se}, while \Cref{lem:gd-auxiliary-coefficient-comparison} bounds the deeper-layer difference. Hence
\begin{equation*}
  \E_a^{(0)}
  \left(N^{-1/2}\|U^{(t)}-\bar U^{(t)}\|\right)^r
  \le \delta_{a,t,r}.
\end{equation*}

To control the remaining state difference, we first bound the moments of $U^{(t)}$, $\bar U^{(t)}$, and $\bar U_{\mathrm{aux}}^{(t)}$. Their definitions, together with \HIprop{prop:apr_est_det_fcn}-(6), \HIprop{prop:GD_l2_est}, and \HIprop{prop:apr_est_se}, give, for every fixed $r\ge1$,
\begin{equation}
  \max_{Q\in\{U^{(t)},\bar U^{(t)},\bar U_{\mathrm{aux}}^{(t)}\}}
  \E_a^{(0)}
  \left(N^{-1/2}\|Q\|\right)^{2r}
  \le
  C_{t,r}(\varkappa_{\ast,a}\log d)^{C_{t,r}}.
  \label{eq:raw-intermediate-state-moments}
\end{equation}
Indeed, each state is the corresponding GD or auxiliary iterate plus a fixed number of pre-gradient terms. The rowwise deterministic bound for each such term has the form
\begin{equation*}
  C_t\kappa_B^{C_t}
  \left\{
    1+\|Z_k\|
    +\|\varphi_{a,[q]}^\ast(\hat u_k^{(0)})\|
    +|\xi_k|
  \right\},
\end{equation*}
where $Z$ is the corresponding GD or auxiliary state and $B$ is the corresponding deeper-layer tuple. Therefore
\begin{align*}
  N^{-1/2}\|\mathfrak S(Z,\hat u^{(0)},B)\|
  \le C_t\kappa_B^{C_t}\bigg[
    1+N^{-1/2}\|Z\|
    +\left\{\frac1N\sum_{k=1}^N
      \|\varphi_{a,[q]}^\ast(\hat u_k^{(0)})\|^2
    \right\}^{1/2}
    +\|\xi_{[N]}\|_\infty
  \bigg].
\end{align*}
Substitute this bound into each state definition. The finite sum inequality $(\sum_{j=1}^{t+1}x_j)^{2r}\le(t+1)^{2r-1}\sum_{j=1}^{t+1}x_j^{2r}$, followed by conditional H\"older's inequality and the imported parameter, coefficient, and iterate moments, proves \eqref{eq:raw-intermediate-state-moments}. The link term is controlled by the Gaussian moments of $\hat u_k^{(0)}$.

We now bound $\bar U^{(t)}-\bar U_{\mathrm{aux}}^{(t)}$ using \eqref{eq:gd-auxiliary-state-difference}. Since $u^{(0)}=\hat u^{(0)}$, \HIprop{prop:apr_est_det_fcn}-(7), with the minimum growth factor evaluated at the auxiliary point, yields
\begin{equation*}
  \begin{split}
  &\left\|
    \mathfrak S_k(\hat u_k^{(s)},u_k^{(0)},V^{(s-1)})
    -
    \mathfrak S_k(u_k^{(s)},u_k^{(0)},V^{(s-1)})
  \right\| \le
  C_t\kappa_{V^{(s-1)}}^{C_t}H_{k,s}
  \|\hat u_k^{(s)}-u_k^{(s)}\|,
  \end{split}
\end{equation*}
where
\begin{equation*}
  H_{k,s}
  \coloneqq
  1+\|u_k^{(s)}\|
  +\|\varphi_{a,[q]}^\ast(u_k^{(0)})\|
  +|\xi_k|.
\end{equation*}
Define the auxiliary delocalization event
\begin{equation*}
  E_{\mathrm{del},u}
  \coloneqq
  \left\{
    \max_{0\le s\le t+1}\|u^{(s)}\|_\infty
    \le C_t(\varkappa_{\ast,a}\log d)^{C_t}
  \right\}.
\end{equation*}
The polynomial local-Lipschitz bounds verified in \Cref{lem:supp:input-localization} allow us to apply \HIprop{prop:GFOM_deloc_asym}. Choosing its tail constant sufficiently large and taking a union bound over the $N_a=O(d)$ rows and $0\le s\le t+1$ yields
\begin{equation}
  \Prob_a^{(0)}(E_{\mathrm{del},u}^c)\le d^{-100}.
  \label{eq:intermediate-delocalization-event}
\end{equation}
On $E_{\mathrm{del},u}$, the polynomial growth of $\varphi_{a,[q]}^\ast$, the definition of $\varkappa_{\ast,a}$, and the imported state-evolution and coefficient bounds give
\begin{align*}
  &N^{-1/2}
  \|\bar U^{(t)}-\bar U_{\mathrm{aux}}^{(t)}\|
  \mathbf 1_{E_{\mathrm{del},u}} \le
  N^{-1/2}\|\hat u^{(t+1)}-u^{(t+1)}\| +
  C_t(\varkappa_{\ast,a}\log d)^{C_t}
  \sum_{s=1}^t
  N^{-1/2}\|\hat u^{(s)}-u^{(s)}\|.
\end{align*}
At $t=0$, the sum is empty. H\"older's inequality and \HIprop{prop:err_gd_GFOM}, as extended in \Cref{lem:hi25-component-link}, therefore give
\begin{equation*}
  \E_a^{(0)}\left[
    \left(
      N^{-1/2}
      \|\bar U^{(t)}-\bar U_{\mathrm{aux}}^{(t)}\|
    \right)^r
    \mathbf 1_{E_{\mathrm{del},u}}
  \right]
  \le
  C_{t,r}(\varkappa_{\ast,a}\log d)^{C_{t,r}}d^{-1/C_{t,r}}.
\end{equation*}

On $E_{\mathrm{del},u}^c$, the triangle inequality yields
\begin{equation*}
  N^{-1/2}\|\bar U^{(t)}-\bar U_{\mathrm{aux}}^{(t)}\|
  \le
  N^{-1/2}\|\bar U^{(t)}\|
  +N^{-1/2}\|\bar U_{\mathrm{aux}}^{(t)}\|.
\end{equation*}
Cauchy--Schwarz, \eqref{eq:raw-intermediate-state-moments}, and \eqref{eq:intermediate-delocalization-event} give
\begin{equation*}
  \E_a^{(0)}\left[
    \left(
      N^{-1/2}\|\bar U^{(t)}-\bar U_{\mathrm{aux}}^{(t)}\|
    \right)^r
    \mathbf 1_{E_{\mathrm{del},u}^c}
  \right]
  \le
  C_{t,r}(\varkappa_{\ast,a}\log d)^{C_{t,r}}d^{-50}.
\end{equation*}
Adding the last two bounds proves the $\bar U^{(t)}-\bar U_{\mathrm{aux}}^{(t)}$ comparison.

The remaining parameter and coefficient moment bounds follow from \Cref{lem:supp:input-localization,lem:gd-path-localization,lem:gd-auxiliary-coefficient-comparison}. The bound for $\widehat U^{(t)}$ follows from \eqref{eq:raw-intermediate-state-moments} and its comparison with $U^{(t)}$ proved above, while Gaussian moments give the bound involving $\hat u^{(0)}=X\beta_{[q]}$. At $t=0$, every state and parameter difference in the first display is zero by the initialization convention in \eqref{eq:state-evolution-initialization} and the empty-sum convention.

For \eqref{eq:observable-covariance-comparison}, the two covariance matrices are
\begin{equation*}
  \widetilde\Sigma_t
  =
  \begin{pmatrix}
    d^{-1}v_1^{(t)\top}v_1^{(t)}
    &d^{-1}v_1^{(t)\top}D_{-1}
    &d^{-1/2}v_1^{(t)\top}\Upsilon\\
    d^{-1}D_{-1}^\top v_1^{(t)}
    &d^{-1}D_{-1}^\top D_{-1}
    &d^{-1/2}D_{-1}^\top\Upsilon\\
    d^{-1/2}\Upsilon^\top v_1^{(t)}
    &d^{-1/2}\Upsilon^\top D_{-1}
    &I_\ell
  \end{pmatrix}
\end{equation*}
and
\begin{equation*}
  \widetilde\Sigma_{0,t}
  =
  \begin{pmatrix}
    \Omega_{t,t}+d^{-1}D_t^\top D_t
    &\Omega_{t,-1}+d^{-1}D_t^\top D_{-1}
    &d^{-1/2}D_t^\top\Upsilon\\
    \Omega_{-1,t}+d^{-1}D_{-1}^\top D_t
    &d^{-1}D_{-1}^\top D_{-1}
    &d^{-1/2}D_{-1}^\top\Upsilon\\
    d^{-1/2}\Upsilon^\top D_t
    &d^{-1/2}\Upsilon^\top D_{-1}
    &I_\ell
  \end{pmatrix},
\end{equation*}
where $\Omega_{t,-1}=\Omega_{-1,t}=0$. For the column-indexed comparisons, extend each test on the first $d$ rows by zero on the $(L-1)q$ padded rows and multiply it by $d_{L,q}/d$. The average normalized by $d_{L,q}$ then equals the required average normalized by $d$; this factor is bounded uniformly in $d$. For $b,c\in[q]$, apply the column-indexed auxiliary state evolution to the row tests
\begin{equation*}
  \vartheta_{i,b,c}(z^{([0:t])})
  =z_b^{(t)}z_c^{(t)},
  \qquad
  \vartheta'_{i,b,c}(z^{([0:t])})
  =z_b^{(t)}[D_{-1}]_{ic}.
\end{equation*}
These tests identify, respectively, the entries of $d^{-1}v_1^{(t)\top}v_1^{(t)}$ and $d^{-1}v_1^{(t)\top}D_{-1}$. Their state-evolution expectations are the corresponding entries of $\Omega_{t,t}+d^{-1}D_t^\top D_t$ and $\Omega_{t,-1}+d^{-1}D_t^\top D_{-1}$. The first test is pseudo-Lipschitz of order two. The second is rowwise Lipschitz with constant at most $\|D_{-1}\|_\infty\le\varkappa_{\ast,a}$. Applying the auxiliary state-evolution bound from \Cref{lem:supp:input-localization} to these tests and summing over the finitely many pairs $(b,c)$ controls the upper-left $2q\times2q$ block at the stated rate. The lower-right $(q+\ell)\times(q+\ell)$ blocks agree exactly.

For the remaining block, fix $b\in[q]$ and $j\in[\ell]$. Apply the column-indexed auxiliary state evolution at horizon $t+1$ to
\begin{equation*}
  \vartheta_{i,b,j}(z^{([0:t])})
  =\sqrt d\,(\gamma_j)_i z_b^{(t)}.
\end{equation*}
The delocalization assumption on $\gamma_j$ makes this a uniformly pseudo-Lipschitz family and yields
\begin{equation*}
  \E_a^{(0)}\left\|
    d^{-1/2}(v_1^{(t)})^\top\Upsilon
    -d^{-1/2}D_t^\top\Upsilon
  \right\|_\infty^r
  \le\delta_{a,t,r}.
\end{equation*}
The empirical coordinate in this comparison is exactly
\begin{equation*}
  \frac1d\sum_{i=1}^d
  \sqrt d\,(\gamma_j)_i[v_{1,i}^{(t)}]_b
  =
  \bigl[d^{-1/2}v_1^{(t)\top}\gamma_j\bigr]_b.
\end{equation*}
The state-evolution counterpart of $v_{1,i}^{(t)}$ is
\begin{equation*}
  \Delta_i^{(t)}=(D_t)_{i\cdot}
  +\sum_{s=1}^t\mathfrak V_i^{(s)}\rho_{t,s}^{\top},
\end{equation*}
with centered Gaussian $\mathfrak V_i^{(s)}$. Its conditional mean is therefore $(D_t)_{i\cdot}$, so the state-evolution average equals $[d^{-1/2}D_t^\top\gamma_j]_b$. This is also the corresponding entry of the cross-covariance
\begin{equation*}
  \operatorname{Cov}
  \bigl(\mathfrak U_a^{(t+1)},\mathfrak U_{\Upsilon,a}\bigr)
  =
  d^{-1/2}D_t^\top\Upsilon
\end{equation*}
by \Cref{prop:supp:auxiliary_gfom_policy_index}. Summing the moment bounds over the finitely many pairs $(b,j)$ completes the proof.
\end{proof}

\subsection{Comparison of weighted averages}

\begin{definition}[Test functions for weighted averages]
\label{def:weighted-test-functions}
Fix $a\in[K]$ and $t\in\mathbb Z_{\ge0}$. A family of functions $\Phi_{t,k}:\R\times\R^q\times\R^\ell\to\R$, $k\in[N]$, satisfies the comparison conditions if it is measurable with respect to $\sigma(\Is_a,\xi_{[N]},W_a^{(0)})$ and, conditional on $\mathcal G_a$, there are a fixed integer $p\ge1$ and $L_{a,t}\le C_t(\varkappa_{\ast,a}\log d)^{C_t}$ such that, for every $k\in[N]$, $z,z'\in\R$, $w\in\R^q$, and $s\in\R^\ell$,
\begin{align*}
  |\Phi_{t,k}(z,w,s)-\Phi_{t,k}(z',w,s)|
  &\le
  L_{a,t}(1+|z|+|z'|+\|w\|^p)|z-z'|,
  \\
  |\Phi_{t,k}(0,w,s)|
  &\le
  L_{a,t}(1+\|w\|^p).
\end{align*}
Moreover, with
\begin{equation*}
  \Theta_{\Phi,t,k}(u,w,s)
  \coloneqq
  \Phi_{t,k}\bigl(G_{V^{(t)}}(u),w,s\bigr),
\end{equation*}
the family $\{\Theta_{\Phi,t,k}\}_{k=1}^N$ satisfies the following pseudo-Lipschitz and origin bounds with the same $p$ and $L_{a,t}$:
\begin{align}
  |\Theta_{\Phi,t,k}(x)-\Theta_{\Phi,t,k}(x')|
  &\le
  L_{a,t}
  (1+\|x\|^{p-1}+\|x'\|^{p-1})\|x-x'\|,
  \label{eq:observable-joint-pl}\\
  |\Theta_{\Phi,t,k}(0)|
  &\le L_{a,t},
  \qquad x,x'\in\R^{2q+\ell}.
  \label{eq:observable-joint-origin}
\end{align}
\end{definition}

The first bound allows the Lipschitz factor to grow at most linearly in the scalar prediction. This lets us compare empirical averages using the normalized $\ell_2$ bounds for the corrected states. The composed state-evolution test function may have any fixed pseudo-Lipschitz order. For each family satisfying these conditions, the constants below depend only on that order and the fixed constants in the displayed bounds; they are uniform in $d$ and the row index $k$.

For such a family, we compare the corrected empirical average with the expectation of the same test at an independent context. With $Z\sim N(0,I_d)$ independent of $\mathcal F_{\mathrm{tr}}$ and of $\pi_N$, define these two quantities and the state-evolution expectation used to compare them by
\begin{align*}
  \widehat{\mathcal O}_{a}^{(t)}(\Phi)
  &\coloneqq
  \frac1N\sum_{k=1}^N
  \Phi_{t,k}\left(
    G_{W_{[2:L]}^{(t)}}(\widehat U_k^{(t)}),
    \hat u_k^{(0)},
    \Upsilon^\top X_k
  \right),\\
  \mathcal O_{a,\pi}^{(t)}(\Phi)
  &\coloneqq
  \E\left[
    \Phi_{t,\pi_N}\left(
      f_{W^{(t)}}(Z),
      d^{-1/2}D_{-1}^\top Z,
      \Upsilon^\top Z
    \right)
    \mymid
    \widehat{\mathcal F}_a
  \right],\\
  \Gamma_{a,\pi}^{(t)}(\Phi)
  &\coloneqq
  \E_a^{(0)}\left[
    \Phi_{t,\pi_N}\left(
      G_{V^{(t)}}(\mathfrak U_a^{(t+1)}),
      \mathfrak U_a^{(0)},
      \mathfrak U_{\Upsilon,a}
    \right)
  \right].
\end{align*}

\begin{suppproposition}[Comparison with expectations at independent contexts]
\label{prop:supp:weighted-observable-transfer}
Assume the conditions of \Cref{sec:formal-regime}. Fix $a\in[K]$, $t\in\mathbb Z_{\ge0}$, and a family satisfying \Cref{def:weighted-test-functions}. For every $r\ge1$, there is a constant $C_{t,r}<\infty$, depending only on the fixed pseudo-Lipschitz order and the constants in that definition, such that, on $E_{\mathrm{cnt}}$,
\begin{align}
  &\E_a^{(0)}
  |\widehat{\mathcal O}_{a}^{(t)}(\Phi)
    -\Gamma_{a,\pi}^{(t)}(\Phi)|^r
  +
  \E_a^{(0)}
  |\mathcal O_{a,\pi}^{(t)}(\Phi)
    -\Gamma_{a,\pi}^{(t)}(\Phi)|^r
  \le
  C_{t,r}(\varkappa_{\ast,a}\log d)^{C_{t,r}}
  d^{-1/C_{t,r}}.
  \label{eq:observable-transfer-bound}
\end{align}
Consequently,
\begin{equation*}
  \E_a^{(0)}
  |\widehat{\mathcal O}_{a}^{(t)}(\Phi)
    -\mathcal O_{a,\pi}^{(t)}(\Phi)|^r
  \le
  C_{t,r}(\varkappa_{\ast,a}\log d)^{C_{t,r}}
  d^{-1/C_{t,r}}.
\end{equation*}
\end{suppproposition}

\begin{proof}
\begin{samepage}
We first compare $\widehat{\mathcal O}_a^{(t)}(\Phi)$ with the state-evolution average $\Gamma_{a,\pi}^{(t)}(\Phi)$. We replace the deeper-layer weights and corrected states in stages, using \Cref{lem:arm-wise-hi25-consequences}. Define the resulting prediction vectors by
\begin{align*}
  z_k^{(0)}&\coloneqq
  G_{W_{[2:L]}^{(t)}}(\widehat U_k^{(t)}),\\
  z_k^{(1)}&\coloneqq
  G_{V^{(t)}}(\widehat U_k^{(t)}),\\
  z_k^{(2)}&\coloneqq
  G_{V^{(t)}}(U_k^{(t)}),\\
  z_k^{(3)}&\coloneqq
  G_{V^{(t)}}(\bar U_k^{(t)}),\\
  z_k^{(4)}&\coloneqq
  G_{V^{(t)}}(\bar U_{\mathrm{aux},k}^{(t)}).
\end{align*}
\end{samepage}
For $j=0,\ldots,4$, put
\begin{equation*}
  \mathcal O_j
  \coloneqq
  \frac1N\sum_{k=1}^N
  \Phi_{t,k}\left(
    z_k^{(j)},\hat u_k^{(0)},\Upsilon^\top X_k
  \right).
\end{equation*}
Thus, $\mathcal O_0=\widehat{\mathcal O}_a^{(t)}(\Phi)$.

The deterministic network bounds in \HIprop{prop:apr_est_det_fcn} give a random $\mathcal P_{a,t}$, with every fixed conditional moment under $\E_a^{(0)}$ bounded by a polynomial in $\varkappa_{\ast,a}\log d$, such that
\begin{align}
  N^{-1/2}\|z^{(0)}-z^{(1)}\|
  &\le
  \mathcal P_{a,t}
  \max_{\alpha\in[2:L]}
  \|W_\alpha^{(t)}-V_\alpha^{(t)}\|_\infty,
  \label{eq:observable-deeper-prediction-bound}\\
  N^{-1/2}\|z^{(1)}-z^{(2)}\|
  &\le
  \mathcal P_{a,t}N^{-1/2}
  \|\widehat U^{(t)}-U^{(t)}\|,
   \notag\\
  N^{-1/2}\|z^{(2)}-z^{(3)}\|
  &\le
  \mathcal P_{a,t}N^{-1/2}
  \|U^{(t)}-\bar U^{(t)}\|,
   \notag\\
  N^{-1/2}\|z^{(3)}-z^{(4)}\|
  &\le
  \mathcal P_{a,t}N^{-1/2}
  \|\bar U^{(t)}-\bar U_{\mathrm{aux}}^{(t)}\|.
  \label{eq:observable-third-state-bound}
\end{align}
For each $j=0,\ldots,3$, the first inequality in \Cref{def:weighted-test-functions}, followed by Cauchy--Schwarz over rows, yields
\begin{align*}
  |\mathcal O_j-\mathcal O_{j+1}|
  &\quad\le
  L_{a,t}
  \left\{\frac1N\sum_{k=1}^N
    (1+|z_k^{(j)}|+|z_k^{(j+1)}|
      +\|\hat u_k^{(0)}\|^p)^2
  \right\}^{1/2}\\
  &\qquad{}\times
  \left\{\frac1N\sum_{k=1}^N
    |z_k^{(j)}-z_k^{(j+1)}|^2
  \right\}^{1/2}.
\end{align*}
The first square-root factor has polynomial conditional moments by the deterministic forward-map bounds, \Cref{lem:arm-wise-hi25-consequences}, and the Gaussian moments of $\hat u^{(0)}$. More explicitly,
\begin{align*}
  \E_a^{(0)}|\mathcal O_j-\mathcal O_{j+1}|^r
  &\le L_{a,t}^r
  \left\{\E_a^{(0)}
    \left[\frac1N\sum_k
    (1+|z_k^{(j)}|+|z_k^{(j+1)}|+\|\hat u_k^{(0)}\|^p)^2
    \right]^r\right\}^{1/2}\\
  &\qquad\times
  \left\{\E_a^{(0)}
    (N^{-1/2}\|z^{(j)}-z^{(j+1)}\|)^{2r}
  \right\}^{1/2}\\
  &\le C_{t,r}(\varkappa_{\ast,a}\log d)^{C_{t,r}}
       \delta_{a,t,4r}^{1/4}.
\end{align*}
For the last inequality, apply conditional Cauchy--Schwarz once more to $\mathcal P_{a,t}$ and the corresponding error in \eqref{eq:observable-deeper-prediction-bound}-- \eqref{eq:observable-third-state-bound}, using the comparison lemma at moment order $4r$. Increasing $C_{t,r}$ and summing over $j$ gives
\begin{align}
  \sum_{j=0}^3\E_a^{(0)}|\mathcal O_j-\mathcal O_{j+1}|^r
  \le
  C_{t,r}(\varkappa_{\ast,a}\log d)^{C_{t,r}}
  d^{-1/C_{t,r}}.
  \label{eq:observable-algorithmic-transfer}
\end{align}

We next compare $\mathcal O_4$ with its state-evolution target. Put $M=\varkappa_{\ast,a}\log d$, and let $\mathcal T_M$ denote coordinatewise truncation to $[-M,M]$. Define
\begin{align*}
  \Psi_{t,k}^{(M)}(z^{([0:t+1])})
  &\coloneqq
  z^{(t+1)}
  +\phi_a^{-1}\sum_{s=1}^t\eta_1^{(s-1)}
  \mathfrak S_k\bigl(
    z^{(s)},\mathcal T_M(z^{(0)}),V^{(s-1)}
  \bigr)\rho_{t,s}^\top.
\end{align*}
Because $t$ is fixed, the local Lipschitz and polynomial-growth bounds for $\mathfrak S_k$ give fixed integers $p_0,C_0$ such that
\begin{align*}
  \|\Psi_{t,k}^{(M)}(z)-\Psi_{t,k}^{(M)}(z')\|
  &\le
  C_0(\varkappa_{\ast,a}\log d)^{C_0}
  (1+\|z\|^{p_0-1}+\|z'\|^{p_0-1})\|z-z'\|,\\
  \|\Psi_{t,k}^{(M)}(z)\|
  &\le
  C_0(\varkappa_{\ast,a}\log d)^{C_0}
  (1+\|z\|^{p_0}).
\end{align*}
Combining these inequalities with \eqref{eq:observable-joint-pl}--\eqref{eq:observable-joint-origin} shows that the functions
\begin{align*}
  \vartheta_{t,k,M}(\widetilde z^{([0:t+1])})
  \coloneqq
  \Phi_{t,k}\Bigl(
    &G_{V^{(t)}}\bigl[
      \Psi_{t,k}^{(M)}
      ((\widetilde z^{([0:t+1])})^{\mathrm{aux}})
    \bigr], \mathcal T_M((\widetilde z^{(0)})^{\mathrm{aux}}),
    (\widetilde z^{(0)})^{\mathrm{idx}}
  \Bigr)
\end{align*}
form a pseudo-Lipschitz family of fixed order with constant at most $C_0(\varkappa_{\ast,a}\log d)^{C_0}$. After conditioning on $\mathcal G_a$, this family is deterministic.

Define the truncated empirical and state-evolution averages
\begin{align*}
  \mathcal O_{4,M}
  &\coloneqq
  \frac1N\sum_{k=1}^N
  \vartheta_{t,k,M}
  \bigl(\widetilde u_k^{([0:t+1])}\bigr),\\
  \Gamma_{a,\pi,M}^{(t)}
  &\coloneqq
  \frac1N\sum_{k=1}^N
  \E_a^{(0)}
  \vartheta_{t,k,M}\left(
    \left\{
      \widetilde\Theta_k^{(s)}
      (\widetilde{\mathfrak U}^{([0:s])})
    \right\}_{s=0}^{t+1}
  \right).
\end{align*}
\Cref{prop:supp:auxiliary_gfom_policy_index}, applied at horizon $t+1$, gives
\begin{equation}
  \E_a^{(0)}
  |\mathcal O_{4,M}-\Gamma_{a,\pi,M}^{(t)}|^r
  \le
  C_{t,r}(\varkappa_{\ast,a}\log d)^{C_{t,r}}
  d^{-1/C_{t,r}}.
  \label{eq:observable-truncated-gfom}
\end{equation}

The truncated and untruncated empirical averages agree on the event
\begin{equation*}
  E_M^{\mathrm{emp}}\coloneqq
  \{\max_{k\in[N]}\|u_k^{(0)}\|_\infty\le M\}.
\end{equation*}
Indeed, on $E_M^{\mathrm{emp}}$,
\begin{equation*}
  \Psi_{t,k}^{(M)}(u_k^{([0:t+1])})
  =
  \bar U_{\mathrm{aux},k}^{(t)},
  \qquad
  \mathcal T_M(u_k^{(0)})=u_k^{(0)},
\end{equation*}
and therefore $\mathcal O_{4,M}=\mathcal O_4$. Since $u_k^{(0)}=(\beta^\top X_k,0,\ldots,0)$ and $\beta^\top X_k\sim N(0,1)$,
\begin{equation*}
  \Prob_a^{(0)}\bigl((E_M^{\mathrm{emp}})^c\bigr)
  \le2N e^{-M^2/2}
  \le d^{-100}
\end{equation*}
for all sufficiently large $d$, because $M=\varkappa_{\ast,a}\log d\ge\log d$. The polynomial-growth bounds above, the fixed-horizon moment bounds in \Cref{prop:supp:auxiliary_gfom_policy_index}, and \Cref{lem:arm-wise-hi25-consequences} imply
\begin{equation*}
  \E_a^{(0)}|\mathcal O_{4,M}-\mathcal O_4|^{2r}
  \le C_{t,r}(\varkappa_{\ast,a}\log d)^{C_{t,r}}.
\end{equation*}
Thus, Cauchy--Schwarz yields
\begin{align}
  \E_a^{(0)}|\mathcal O_{4,M}-\mathcal O_4|^r
  &=
  \E_a^{(0)}\left[
    |\mathcal O_{4,M}-\mathcal O_4|^r
    \mathbf 1_{(E_M^{\mathrm{emp}})^c}
  \right]\nonumber\\
  &\le
  \left\{\E_a^{(0)}
    |\mathcal O_{4,M}-\mathcal O_4|^{2r}
  \right\}^{1/2}
  \Prob_a^{(0)}((E_M^{\mathrm{emp}})^c)^{1/2}\nonumber\\
  &\le
  C_{t,r}(\varkappa_{\ast,a}\log d)^{C_{t,r}}d^{-50}.
  \label{eq:observable-empirical-truncation}
\end{align}

It remains to remove truncation from $\Gamma_{a,\pi,M}^{(t)}$. Define
\begin{align*}
  B_{t,k,M}
  &\coloneqq
  \phi_a^{-1}\sum_{s=1}^t\eta_1^{(s-1)}
  \Bigl[
    \mathfrak S_k\bigl(
      \Theta_k^{(s)}(\mathfrak U^{([0:s])}),
      \mathcal T_M(\mathfrak U^{(0)}),
      V^{(s-1)}
    \bigr) -
    \mathfrak S_k\bigl(
      \Theta_k^{(s)}(\mathfrak U^{([0:s])}),
      \mathfrak U^{(0)},
      V^{(s-1)}
    \bigr)
  \Bigr]\rho_{t,s}^\top.
\end{align*}
The state-evolution recursion yields the exact identity
\begin{align*}
  &\Psi_{t,k}^{(M)}\left(
    \left\{\Theta_k^{(s)}(\mathfrak U^{([0:s])})\right\}_{s=0}^{t+1}
  \right) =
  \mathfrak U^{(t+1)}+B_{t,k,M}.
\end{align*}
The difference inside each summand defining $B_{t,k,M}$ is bounded by
\begin{align*}
  &C_{t}(\varkappa_{\ast,a}\log d)^{C_t}
  \left(
    1+
    \|\Theta_k^{([0:t])}(\mathfrak U^{([0:t])})\|^{C_t}
    +\|\mathfrak U^{(0)}\|^{C_t}
  \right)\times
  \|\mathcal T_M(\mathfrak U^{(0)})
    -\mathfrak U^{(0)}\|.
\end{align*}
In particular, $B_{t,k,M}=0$ on $E_M^{\mathrm{se}}\coloneqq \{\|\mathfrak U^{(0)}\|_\infty\le M\}$. Gaussian moments, H\"older's inequality, and the tail bound $\Prob_a^{(0)}((E_M^{\mathrm{se}})^c)\le2q e^{-M^2/2}$ give, for every fixed $r\ge1$,
\begin{align}
  \E_a^{(0)}\|B_{t,\pi_N,M}\|^r
  &=\E_a^{(0)}\left[
    \|B_{t,\pi_N,M}\|^r\mathbf1_{(E_M^{\mathrm{se}})^c}
  \right]\nonumber\\
  &\le
  \left\{\E_a^{(0)}\|B_{t,\pi_N,M}\|^{2r}\right\}^{1/2}
  \Prob_a^{(0)}((E_M^{\mathrm{se}})^c)^{1/2}\nonumber\\
  &\le C_{t,r}(\varkappa_{\ast,a}\log d)^{C_{t,r}}
  e^{-c_{t,r}(\log d)^2}.
  \label{eq:observable-truncation-remainder}
\end{align}

Because $\pi_N$ is uniform and independent,
\begin{align*}
  \Gamma_{a,\pi,M}^{(t)}
  =
  \E_a^{(0)}\Bigl[
    \Phi_{t,\pi_N}\bigl(
      &G_{V^{(t)}}(
        \mathfrak U^{(t+1)}+B_{t,\pi_N,M}),
      \mathcal T_M(\mathfrak U^{(0)}),
      \mathfrak U_\Upsilon
    \bigr)
  \Bigr].
\end{align*}
Let
\begin{equation*}
  Y_M=
  (\mathfrak U^{(t+1)}+B_{t,\pi_N,M},
    \mathcal T_M(\mathfrak U^{(0)}),\mathfrak U_\Upsilon),
  \qquad
  Y=
  (\mathfrak U^{(t+1)},\mathfrak U^{(0)},\mathfrak U_\Upsilon).
\end{equation*}
The joint pseudo-Lipschitz condition yields
\begin{align*}
  &\left|
    \Theta_{\Phi,t,\pi_N}(Y_M)
    -\Theta_{\Phi,t,\pi_N}(Y)
  \right| \le
  L_{a,t}
  (1+\|Y_M\|^{p-1}+\|Y\|^{p-1})
  \left(
    \|B_{t,\pi_N,M}\|
    +\|\mathcal T_M(\mathfrak U^{(0)})
       -\mathfrak U^{(0)}\|
  \right).
\end{align*}
Conditional Jensen's inequality and Cauchy--Schwarz yield
\begin{align*}
  |\Gamma_{a,\pi,M}^{(t)}-\Gamma_{a,\pi}^{(t)}|^r
  &\le L_{a,t}^r
  \left\{\E_a^{(0)}
    (1+\|Y_M\|^{p-1}+\|Y\|^{p-1})^{2r}
  \right\}^{1/2}\\
  &\quad\times
  \left\{\E_a^{(0)}
    \left(\|B_{t,\pi_N,M}\|
      +\|\mathcal T_M(\mathfrak U^{(0)})-\mathfrak U^{(0)}\|
    \right)^{2r}
  \right\}^{1/2}.
\end{align*}
The first factor is bounded by a fixed polynomial in $\varkappa_{\ast,a}\log d$. Apply \eqref{eq:observable-truncation-remainder} at order $2r$ to the $B_{t,\pi_N,M}$ term in the second factor. Cauchy--Schwarz with the same Gaussian tail bound controls the remaining truncation error $\|\mathcal T_M(\mathfrak U^{(0)})-\mathfrak U^{(0)}\|$. Hence
\begin{equation}
  |\Gamma_{a,\pi,M}^{(t)}-\Gamma_{a,\pi}^{(t)}|^r
  \le
  C_{t,r}(\varkappa_{\ast,a}\log d)^{C_{t,r}}
  e^{-c_{t,r}(\log d)^2}.
  \label{eq:observable-state-truncation}
\end{equation}

The inequalities \eqref{eq:observable-truncated-gfom}, \eqref{eq:observable-empirical-truncation}, and \eqref{eq:observable-state-truncation}, followed by $|x+y+z|^r\le3^{r-1}(|x|^r+|y|^r+|z|^r)$, yield
\begin{equation*}
  \E_a^{(0)}
  |\mathcal O_4
    -\Gamma_{a,\pi}^{(t)}(\Phi)|^r
  \le
  C_{t,r}(\varkappa_{\ast,a}\log d)^{C_{t,r}}
  d^{-1/C_{t,r}}.
\end{equation*}
Finally,
\begin{equation*}
  |\mathcal O_0-\Gamma_{a,\pi}^{(t)}|
  \le
  \sum_{j=0}^3|\mathcal O_j-\mathcal O_{j+1}|
  +|\mathcal O_4-\Gamma_{a,\pi}^{(t)}|.
\end{equation*}
Together with \eqref{eq:observable-algorithmic-transfer}, this proves the first comparison in \eqref{eq:observable-transfer-bound}.

To compare the expectation at an independent context with the state-evolution expectation, define
\begin{align*}
  \mathcal J_0
  &\coloneqq\mathcal O_{a,\pi}^{(t)}(\Phi),\\
  \mathcal J_1
  &\coloneqq
  \E\left[
    \Phi_{t,\pi_N}\left(
      G_{V^{(t)}}(d^{-1/2}\hat v_1^{(t)\top}Z),
      d^{-1/2}D_{-1}^\top Z,
      \Upsilon^\top Z
    \right)
    \mymid\widehat{\mathcal F}_a
  \right],\\
  \mathcal J_2
  &\coloneqq
  \E\left[
    \Phi_{t,\pi_N}\left(
      G_{V^{(t)}}(d^{-1/2}v_1^{(t)\top}Z),
      d^{-1/2}D_{-1}^\top Z,
      \Upsilon^\top Z
    \right)
    \mymid\widehat{\mathcal F}_a
  \right].
\end{align*}
Holding the training sample fixed, the Gaussian identity
\begin{equation*}
  \E_Z\left\|
    d^{-1/2}(\hat v_1^{(t)}-v_1^{(t)})^\top Z
  \right\|^2
  =d^{-1}\|\hat v_1^{(t)}-v_1^{(t)}\|_F^2
\end{equation*}
and the first inequality in \Cref{def:weighted-test-functions} imply, after enlarging $\mathcal P_{a,t}$ if necessary,
\begin{align*}
  |\mathcal J_0-\mathcal J_1|
  &\le\mathcal P_{a,t}
    \max_{\alpha\in[2:L]}\|W_\alpha^{(t)}-V_\alpha^{(t)}\|_\infty,\\
  |\mathcal J_1-\mathcal J_2|
  &\le\mathcal P_{a,t}d^{-1/2}
    \|\hat v_1^{(t)}-v_1^{(t)}\|_F.
\end{align*}
Here $\mathcal P_{a,t}$ retains polynomial conditional moments: the network growth bounds and Gaussian integration express each multiplier in terms of the parameter norms in \Cref{lem:arm-wise-hi25-consequences}. Conditional Cauchy--Schwarz, followed by that lemma at order $2r$, yields
\begin{equation*}
  \E_a^{(0)}|\mathcal J_0-\mathcal J_1|^r
  +\E_a^{(0)}|\mathcal J_1-\mathcal J_2|^r
  \le
  C_{t,r}(\varkappa_{\ast,a}\log d)^{C_{t,r}}d^{-1/C_{t,r}}.
\end{equation*}
The two terms correspond, respectively, to replacing $W_{[2:L]}^{(t)}$ by $V^{(t)}$ and replacing $d^{-1/2}\hat v_1^{(t)\top}Z$ by $d^{-1/2}v_1^{(t)\top}Z$.

The quantity $\mathcal J_2$ is the expectation of $\Theta_{\Phi,t,\pi_N}$ under $G_t$. To compare it with $\Gamma_{a,\pi}^{(t)}$, let $\zeta\sim N(0,I_{2q+\ell})$ be independent of the training sample and $\pi_N$, and use the coupling
\begin{equation*}
  G_t'=\widetilde\Sigma_t^{1/2}\zeta,
  \qquad G_{0,t}'=\widetilde\Sigma_{0,t}^{1/2}\zeta.
\end{equation*}
The auxiliary iterate $v_1^{(t)}$ is $\widehat{\mathcal F}_a$-measurable, as noted before \Cref{lem:arm-wise-hi25-consequences}. The state-evolution parameters and covariance are $\mathcal G_a$-measurable and depend only on the arm-wise initialization, noise values, and fixed problem parameters; these quantities are also $\widehat{\mathcal F}_a$-measurable. Thus, conditional on $\widehat{\mathcal F}_a$, the coupled vectors have the respective laws used by $\mathcal J_2$ and $\Gamma_{a,\pi}^{(t)}$. Write $\E_\zeta$ for expectation over $\zeta$, keeping the training sample fixed. Since
\begin{equation*}
  \|A^{1/2}-B^{1/2}\|_{\op}
  \le C\|A-B\|_{\op}^{1/2},\qquad A,B\succeq0,
\end{equation*}
fixed dimensionality yields
\begin{align*}
  \E_\zeta\|G_t'-G_{0,t}'\|^{2r}
  &\le C_r\|\widetilde\Sigma_t-
    \widetilde\Sigma_{0,t}\|_\infty^r,\\
  \E_\zeta(1+\|G_t'\|^{p-1}+\|G_{0,t}'\|^{p-1})^{2r}
  &\le C_{p,r}
    (1+\|\widetilde\Sigma_t\|_{\op}
       +\|\widetilde\Sigma_{0,t}\|_{\op})^{r(p-1)}.
\end{align*}
Apply the joint pseudo-Lipschitz bound, Jensen's inequality over $(\zeta,\pi_N)$, and conditional Cauchy--Schwarz. Integrating over the training design then yields
\begin{align*}
  \E_a^{(0)}|\mathcal J_2-\Gamma_{a,\pi}^{(t)}(\Phi)|^r &\le C_{p,r}L_{a,t}^r
  \left\{\E_a^{(0)}
    (1+\|\widetilde\Sigma_t\|_{\op}
       +\|\widetilde\Sigma_{0,t}\|_{\op})^{r(p-1)}
  \right\}^{1/2} \times
  \left\{\E_a^{(0)}
    \|\widetilde\Sigma_t-\widetilde\Sigma_{0,t}\|_\infty^r
  \right\}^{1/2}\\
  &\le C_{t,r}(\varkappa_{\ast,a}\log d)^{C_{t,r}}
    d^{-1/C_{t,r}}.
\end{align*}
The first factor is controlled by the parameter moments and the covariance formulas in \Cref{lem:arm-wise-hi25-consequences}; the second is \eqref{eq:observable-covariance-comparison}. Finally,
\begin{equation*}
  \mathcal J_0-\Gamma_{a,\pi}^{(t)}
  =(\mathcal J_0-\mathcal J_1)+(\mathcal J_1-\mathcal J_2)
    +(\mathcal J_2-\Gamma_{a,\pi}^{(t)}).
\end{equation*}
The three bounds above and the finite sum inequality prove the second comparison in \eqref{eq:observable-transfer-bound}. Combining the two comparisons by the triangle inequality proves the final claim.
\end{proof}

\subsection{Application to squared residuals}

For each fixed arm $a\in[K]$, write
\begin{equation*}
  \Is_a=\{i_{a,1}<\cdots<i_{a,N_a}\},\qquad
  \xi_{k,a}
  =
  R_{i_{a,k}}-\varphi_a^\ast(\beta_a^\top X_{i_{a,k}}),
\end{equation*}
and let $\widehat{\mathcal F}_a$ be the empirical arm-wise sigma-field in \Cref{prop:supp:arm-wise_conditioning}. Define
\begin{equation*}
  \Es_{a,\pi}^{(t)}
  \coloneqq
  \E\left[
    w_a(Z_a)
    \left\{
      \varphi_a^\ast(\beta_a^\top Z_a)
      +\xi_{\pi_{N_a},a}
      -f_{W_a^{(t)}}(Z_a)
    \right\}^2
    \mymid
    \widehat{\mathcal F}_a
  \right],
\end{equation*}
where $Z_a\sim N(0,I_d)$ is independent of $\mathcal F_{\mathrm{tr}}$. On $\{N_a\ge1\}$, conditional on $\widehat{\mathcal F}_a$, the index $\pi_{N_a}$ is uniform on $[N_a]$ and independent of $Z_a$.

\begin{proposition}[Arm-wise risk comparison]
\label{prop:arm-wise-calibration}
Under the assumptions in \Cref{sec:formal-regime}, for every fixed arm $a$, $t\ge0$, and $r\ge1$, on $E_{\mathrm{cnt}}$, there exists $C_{t,r}<\infty$ such that
\begin{equation}
  \E_a^{(0)}
  |\widehat{\Es}_a^{(t)}-\Es_{a,\pi}^{(t)}|^r
  \le
  C_{t,r}(\varkappa_{\ast,a}\log d)^{C_{t,r}}d^{-1/C_{t,r}}.
  \label{eq:arm-wise-calibration-rate}
\end{equation}
Consequently,
\begin{equation*}
  \sup_{t\in\Ts}
  |\widehat{\Es}_a^{(t)}-\Es_{a,\pi}^{(t)}|
  \xrightarrow{p}0.
\end{equation*}
\end{proposition}

\begin{proof}
In the local notation above, take
\begin{equation*}
  \Phi_{t,k}^{\mathrm{risk}}(z,w,s)
  \coloneqq
  h(s)\{\varphi_a^\ast(w_1)+\xi_k-z\}^2.
\end{equation*}
The first two inequalities in \Cref{def:weighted-test-functions} follow from
\begin{align*}
  |\Phi_{t,k}^{\mathrm{risk}}(z,w,s)
    -\Phi_{t,k}^{\mathrm{risk}}(z',w,s)|
  &\le B_h\bigl(2|\varphi_a^\ast(w_1)|+2|\xi_k|
       +|z|+|z'|\bigr)|z-z'|,\\
  |\Phi_{t,k}^{\mathrm{risk}}(0,w,s)|
  &\le2B_h\bigl(|\varphi_a^\ast(w_1)|^2+|\xi_k|^2\bigr).
\end{align*}

For the joint condition, write $x=(u,w,s)$ and $r_k(x)=\varphi_a^\ast(w_1)+\xi_k-G_{V^{(t)}}(u)$. Then,
\begin{align*}
  |h(s)r_k(x)^2-h(s')r_k(x')^2|
  &\le B_h(|r_k(x)|+|r_k(x')|)|r_k(x)-r_k(x')| +L_h|r_k(x')|^2\|s-s'\|.
\end{align*}

The link bounds in \Cref{ass:formal-nn-regularity}, $|\xi_k|\le\varkappa_{\ast,a}$, and the network bounds in \HIprop{prop:apr_est_det_fcn} therefore imply
\begin{equation*}
  |\Theta_{\Phi^{\mathrm{risk}},t,k}(x)
    -\Theta_{\Phi^{\mathrm{risk}},t,k}(x')|
  \le C_t(\varkappa_{\ast,a}\log d)^{C_t}
    (1+\|x\|^{p-1}+\|x'\|^{p-1})\|x-x'\|,
\end{equation*}
where one may take the fixed integer $p=2\lceil1\vee r_0\rceil+1$. The same bounds control the value at the origin. Thus, \Cref{def:weighted-test-functions} holds with this $p$. The zero-padded output representation yields
\begin{equation*}
  \widehat{\mathcal O}_{a}^{(t)}(\Phi^{\mathrm{risk}})
  =
  \widehat{\Es}_a^{(t)},
  \qquad
  \mathcal O_{a,\pi}^{(t)}(\Phi^{\mathrm{risk}})
  =
  \Es_{a,\pi}^{(t)}.
\end{equation*}

\Cref{prop:supp:weighted-observable-transfer} gives \eqref{eq:arm-wise-calibration-rate}. For fixed $\epsilon>0$ and $r\ge1$, conditional Markov's inequality and a union bound yield
\begin{align*}
&\Prob\left(
  \sup_{t\in\Ts}
  |\widehat{\Es}_a^{(t)}-\Es_{a,\pi}^{(t)}|>\epsilon
\right) \le
\Prob(E_{\mathrm{cnt}}^c)
+\sum_{t\in\Ts}
\E\left[
  \mathbf 1_{E_{\mathrm{cnt}}}
  \min\left\{1,
    \epsilon^{-r}C_{t,r}
    (\varkappa_{\ast,a}\log d)^{C_{t,r}}d^{-1/C_{t,r}}
  \right\}
\right].
\end{align*}
Each integrand in the sum is bounded by one and converges in probability to zero by \eqref{eq:envelope-polynomial-vanishing}; hence, its expectation converges to zero. Since $\Prob(E_{\mathrm{cnt}}^c)\to0$, this proves the final display.
\end{proof}

\section{Independent Noise and Fitted-Value Comparisons}

The preceding risk comparison uses the empirical distribution of the training noises. We first bound prediction moments, then replace this distribution with the arm-specific noise law to obtain conditional test risk. We also derive the comparison of weighted fitted predictions used in the ADM and ADR proofs.

\subsection{Prediction moments}

\begin{proposition}[Fixed-grid prediction moment bound]
\label{prop:prediction-moment-bound}
Assume the conditions of Section~\ref{sec:formal-regime}, with $m/d\to\phi\in(0,\infty)$ and fixed $K,L,q,\Ts$. For every arm $a\in[K]$ and $Z\sim N(0,I_d)$ independent of $\mathcal F_{\mathrm{tr}}$, as $d\to\infty$,
\begin{equation}
  \sup_{t\in\Ts}
  \E\left[
    |f_{W_a^{(t)}}(Z)|^2
    \mymid
    \Ds_m,W_a^{(0)}
  \right]
  =
  d^{o_p(1)}.
  \label{eq:prediction-moment-bound}
\end{equation}
\end{proposition}

\begin{proof}
On $E_{\mathrm{cnt}}$, the arm-wise forms of \HIprop{prop:GD_l2_est} and \HIprop{prop:apr_est_se} give, uniformly on the fixed grid,
\begin{equation*}
  \|W_{a,1}^{(t)}\|_F=d^{o_p(1)},
  \qquad
  \max_{\alpha\in[2:L]}
  \|W_{a,\alpha}^{(t)}\|_{\op}=d^{o_p(1)}.
\end{equation*}
By \HIprop{prop:apr_est_det_fcn}-(1),
\begin{equation*}
  |f_{W_a^{(t)}}(z)|
  \le
  C_t\kappa_{W_{a,[2:L]}^{(t)}}^{C_t}
  \{1+\|z^\top W_{a,1}^{(t)}\|_\infty\}.
\end{equation*}
Conditional on $(\Ds_m,W_a^{(0)})$, the fitted matrices are fixed and $Z^\top W_{a,1}^{(t)}$ is Gaussian. Since $q$ is fixed,
\begin{align*}
  \E\left[
    \|Z^\top W_{a,1}^{(t)}\|_\infty^2
    \mymid \Ds_m,W_a^{(0)}
  \right]
  &\le
  \sum_{j=1}^q
  \E\left[
    |Z^\top(W_{a,1}^{(t)})_{\cdot j}|^2
    \mymid \Ds_m,W_a^{(0)}
  \right]\\
  &=
  \sum_{j=1}^q\|(W_{a,1}^{(t)})_{\cdot j}\|_2^2
  =
  \|W_{a,1}^{(t)}\|_F^2.
\end{align*}
Therefore, after changing $C_t$,
\begin{align*}
  \E\left[
    |f_{W_a^{(t)}}(Z)|^2
    \mymid \Ds_m,W_a^{(0)}
  \right] &\le
  C_t\kappa_{W_{a,[2:L]}^{(t)}}^{C_t}
  \{1+\|W_{a,1}^{(t)}\|_F^2\}
  =d^{o_p(1)}
\end{align*}
on $E_{\mathrm{cnt}}$, uniformly over the fixed grid $\Ts$. Since $\Prob(E_{\mathrm{cnt}}^c)\to0$, this proves \eqref{eq:prediction-moment-bound}.
\end{proof}

\subsection{From the empirical noise distribution to independent noise}

Retain the ordering, noise variables $\xi_{k,a}$, and empirical arm-wise sigma-field $\widehat{\mathcal F}_a$ from \Cref{prop:arm-wise-calibration}, and define
\begin{equation*}
  \Es_{a,\mathrm{ind}}^{(t)}
  \coloneqq
  \E\left[
    w_a(Z_a)
    \left\{
      \varphi_a^\ast(\beta_a^\top Z_a)
      +\xi_a^{\mathrm{new}}
      -f_{W_a^{(t)}}(Z_a)
    \right\}^2
    \mymid
    \widehat{\mathcal F}_a
  \right],
\end{equation*}
where $\xi_a^{\mathrm{new}}\sim P_{\xi,a,d}$ is independent of $(Z_a,\mathcal F_{\mathrm{tr}})$. The integrand depends on the training data only through the $\widehat{\mathcal F}_a$-measurable iterate $W_a^{(t)}$. Applying \Cref{lem:fresh-draw-conditioning} with $Q=(Z_a,\xi_a^{\mathrm{new}})$ shows that this conditional expectation is unchanged when $\widehat{\mathcal F}_a$ is replaced by $\mathcal F_{\mathrm{tr}}$. Thus, this definition is the risk in \eqref{eq:weighted-armwise-ind-risk}.

For a fixed arm $a\in[K]$ and iterate $t\in\Ts$, write
\begin{align*}
  m_t(z)
  &\coloneqq\varphi_a^\ast(\beta_a^\top z)-f_{W_a^{(t)}}(z),\\
  \bar\xi_a
  &\coloneqq\frac1{N_a}\sum_{k=1}^{N_a}\xi_{k,a},
  \qquad
  \overline{\xi_a^2}
  \coloneqq\frac1{N_a}\sum_{k=1}^{N_a}\xi_{k,a}^2,\\
  B_{a,t}
  &\coloneqq
  \E[w_a(Z_a)m_t(Z_a)\mymid\widehat{\mathcal F}_a],
  \qquad
  C_{w,a}\coloneqq\E[w_a(Z_a)].
\end{align*}
These quantities are finite under the conditions of Section~\ref{sec:formal-regime} by \Cref{prop:prediction-moment-bound}.

In the working regime of Section~\ref{sec:formal-regime}, positivity and proportional scaling give $\Prob(N_a\ge1)\to1$. The required bound on $B_{a,t}$ is verified below.

\begin{proposition}[Independent noise comparison]
\label{prop:independent-noise-comparison}
Fix $a\in[K]$ and a finite grid $\Ts$. Assume the known positive logging probabilities in Assumption~\ref{ass:no-empty-arms}, the noise law in \eqref{eq:noise-setup}, the empirical moment conditions in \eqref{eq:noise-rowwise-lln}, and $0\le w_a\le B_h<\infty$. Assume that $\Es_{a,\pi}^{(t)}$ and $\Es_{a,\mathrm{ind}}^{(t)}$ are finite almost surely for every $t\in\Ts$. For every $t\in\Ts$, on $\{N_a\ge1\}$,
\begin{equation}
  \Es_{a,\pi}^{(t)}-\Es_{a,\mathrm{ind}}^{(t)}
  =
  2\bar\xi_a B_{a,t}
  +
  \left(\overline{\xi_a^2}-\sigma_{\xi,a}^2\right)C_{w,a}.
  \label{eq:aux-independent-decomp}
\end{equation}
If $\Prob(N_a\ge1)\to1$ and $\sup_{t\in\Ts}|B_{a,t}|=d^{o_p(1)}$ as $d\to\infty$, then
\begin{equation*}
  \sup_{t\in\Ts}
  |\Es_{a,\pi}^{(t)}-\Es_{a,\mathrm{ind}}^{(t)}|
  \xrightarrow{p}0.
\end{equation*}
\end{proposition}

\begin{proof}
On $\{N_a\ge1\}$, conditional on $\widehat{\mathcal F}_a$, $\pi_{N_a}$ is uniform on $[N_a]$ and is independent of $Z_a$. Hence
\begin{equation*}
  \E[\xi_{\pi_{N_a},a}\mymid\widehat{\mathcal F}_a]
  =\bar\xi_a,
  \qquad
  \E[\xi_{\pi_{N_a},a}^2\mymid\widehat{\mathcal F}_a]
  =\overline{\xi_a^2}.
\end{equation*}
Expanding the two squared residuals therefore yields
\begin{align*}
  \Es_{a,\pi}^{(t)}
  &=
  \E\left[w_a(Z_a)m_t(Z_a)^2\mymid\widehat{\mathcal F}_a\right]
  +2\bar\xi_a B_{a,t}
  +\overline{\xi_a^2}C_{w,a},\\
  \Es_{a,\mathrm{ind}}^{(t)}
  &=
  \E\left[w_a(Z_a)m_t(Z_a)^2\mymid\widehat{\mathcal F}_a\right]
  +\sigma_{\xi,a}^2C_{w,a}.
\end{align*}
For the second identity, the cross term is zero because $\xi_a^{\mathrm{new}}$ is centered and independent of $(Z_a,\widehat{\mathcal F}_a)$. Subtracting the two identities proves \eqref{eq:aux-independent-decomp}.

We next verify the sufficient bound on $B_{a,t}$ under the conditions of Section~\ref{sec:formal-regime}. Weighted Cauchy--Schwarz and $0\le w_a\le B_h$ yield
\begin{align*}
  |B_{a,t}|^2
  &\le
  \E[w_a(Z_a)]
  \E\left[
    w_a(Z_a)m_t(Z_a)^2
    \mymid\widehat{\mathcal F}_a
  \right]\\
  &\le
  B_h^2
  \E\left[m_t(Z_a)^2\mymid\widehat{\mathcal F}_a\right].
\end{align*}
Moreover,
\begin{equation*}
  m_t(Z_a)^2
  \le
  2|\varphi_a^\ast(\beta_a^\top Z_a)|^2
  +2|f_{W_a^{(t)}}(Z_a)|^2.
\end{equation*}
The link term has finite deterministic expectation because $\beta_a^\top Z_a\sim N(0,1)$ and $\varphi_a^\ast$ has polynomial growth. For the prediction term, \Cref{lem:fresh-draw-conditioning} gives
\begin{equation*}
  \E\left[
    |f_{W_a^{(t)}}(Z_a)|^2
    \mymid\widehat{\mathcal F}_a
  \right]
  =
  \E\left[
    |f_{W_a^{(t)}}(Z_a)|^2
    \mymid\Ds_m,W_a^{(0)}
  \right].
\end{equation*}
The right-hand side is uniformly $d^{o_p(1)}$ on $\Ts$ by \Cref{prop:prediction-moment-bound}. Thus
\begin{equation}
  \sup_{t\in\Ts}|B_{a,t}|=d^{o_p(1)}.
  \label{eq:independent-noise-B-bound}
\end{equation}

Whether the bound on $B_{a,t}$ is assumed directly or obtained above, the empirical moment conditions \eqref{eq:noise-rowwise-lln} give
\begin{equation}
  \bar\xi_a=O_p(d^{-1/2}),
  \qquad
  \overline{\xi_a^2}-\sigma_{\xi,a}^2=o_p(1).
  \label{eq:independent-noise-lln}
\end{equation}
Choose $0<\eta<1/2$. Equations \eqref{eq:independent-noise-B-bound} and \eqref{eq:independent-noise-lln} give
\begin{equation*}
  |\bar\xi_a|\sup_{t\in\Ts}|B_{a,t}|
  =O_p(d^{-1/2})\,O_p(d^\eta)
  =O_p(d^{-1/2+\eta})
  =o_p(1).
\end{equation*}
Finally, $C_{w,a}\le B_h$, so on $\{N_a\ge1\}$, \eqref{eq:aux-independent-decomp} implies
\begin{align*}
  \sup_{t\in\Ts}
  |\Es_{a,\pi}^{(t)}-\Es_{a,\mathrm{ind}}^{(t)}|
  &\le
  2|\bar\xi_a|\sup_{t\in\Ts}|B_{a,t}| +
  |\overline{\xi_a^2}-\sigma_{\xi,a}^2|C_{w,a}.
\end{align*}
The right-hand side is $o_p(1)$. Since $\Prob(N_a=0)\to0$, the claimed convergence holds regardless of the definitions on empty arms.
\end{proof}

\subsection{Corrected predictions on logged actions}

For $a\in[K]$ and $t\in\Ts$, define
\begin{align*}
  \widehat M_{a,\diag}^{(t)}
  &\coloneqq
  \frac1{N_a}\sum_{i\in\Is_a}
  w_a(X_i)
  G_{W_{a,[2:L]}^{(t)}}(\widehat U_{i,a}^{(t)}),\\
  M_a^{(t)}
  &\coloneqq
  \E\left[
    w_a(Z)f_{W_a^{(t)}}(Z)
    \mymid\Ds_m,W_a^{(0)}
  \right],
\end{align*}
where $Z\sim N(0,I_d)$ is independent of $\mathcal F_{\mathrm{tr}}$. The first quantity averages corrected predictions on the training observations for arm $a$; the second averages the fitted network over an independent context.

\begin{proposition}[Weighted averages of corrected predictions]
\label{prop:diagonal-comparison}
Assume the conditions of Section~\ref{sec:formal-regime}, with fixed $K,L,q,\Ts$. For every $a\in[K]$, $t\in\Ts$, and $r\ge1$, on $E_{\mathrm{cnt}}$,
\begin{equation*}
  \E_a^{(0)}
  |\widehat M_{a,\diag}^{(t)}-M_a^{(t)}|^r
  \le
  C_{t,r}(\varkappa_{\ast,a}\log d)^{C_{t,r}}d^{-1/C_{t,r}}.
\end{equation*}
Consequently, as $d\to\infty$,
\begin{equation}
  \sup_{t\in\Ts}
  |\widehat M_{a,\diag}^{(t)}-M_a^{(t)}|
  \xrightarrow{p}0.
  \label{eq:diagonal-comparison}
\end{equation}
\end{proposition}

\begin{proof}
Take $\Phi_{t,k}^{\mathrm{fit}}(z,w,s)\coloneqq h(s)z$. Boundedness and Lipschitz continuity of $h$, together with \HIprop{prop:apr_est_det_fcn}-(1)(2), verify the conditions in \Cref{def:weighted-test-functions}. The two quantities in \Cref{prop:supp:weighted-observable-transfer} are
\begin{equation*}
  \widehat{\mathcal O}_{a}^{(t)}(\Phi^{\mathrm{fit}})
  =
  \widehat M_{a,\diag}^{(t)},
  \qquad
  \mathcal O_{a,\pi}^{(t)}(\Phi^{\mathrm{fit}})
  =
  \E\left[
    w_a(Z)f_{W_a^{(t)}}(Z)
    \mymid
    \widehat{\mathcal F}_a
  \right].
\end{equation*}
Because $W_a^{(t)}$ is $\widehat{\mathcal F}_a$-measurable and $Z$ is independent of both conditioning fields,
\begin{align*}
  \E\left[
    w_a(Z)f_{W_a^{(t)}}(Z)
    \mymid\widehat{\mathcal F}_a
  \right]&=
  \int w_a(z)f_{W_a^{(t)}}(z)\,dP_Z(z)\\
  &=
  \E\left[
    w_a(Z)f_{W_a^{(t)}}(Z)
    \mymid\Ds_m,W_a^{(0)}
  \right]
  =M_a^{(t)}.
\end{align*}
\Cref{prop:supp:weighted-observable-transfer} therefore gives the conditional moment bound. For fixed $\epsilon>0$ and $r\ge1$, conditional Markov's inequality yields
\begin{align*}
&\Prob\left(
  \sup_{t\in\Ts}
  |\widehat M_{a,\diag}^{(t)}-M_a^{(t)}|>\epsilon
\right) \le
\Prob(E_{\mathrm{cnt}}^c)
+\sum_{t\in\Ts}
\E\left[
  \mathbf 1_{E_{\mathrm{cnt}}}
  \min\left\{1,
    \epsilon^{-r}C_{t,r}
    (\varkappa_{\ast,a}\log d)^{C_{t,r}}d^{-1/C_{t,r}}
  \right\}
\right].
\end{align*}

Each integrand in the sum is bounded by one and converges in probability to zero by \eqref{eq:envelope-polynomial-vanishing}; hence, its expectation converges to zero. Since $\Prob(E_{\mathrm{cnt}}^c)\to0$, this proves \eqref{eq:diagonal-comparison}.
\end{proof}

\section{Uniform Consistency and Model Selection}
\label{sec:appendix-reduction}

We combine the arm-wise risk comparisons to prove uniform consistency of the risk criterion over the fixed iterate grid. We then derive the oracle inequality for model selection and the resulting bound on policy value error.

\subsection{Proof of Theorem~\ref{thm:uniform-risk}}
\label{app:thm-uniform-risk-proof}

\thmreduction*
\begin{proof}
Fix an arm $a$. By \Cref{prop:arm-wise-calibration},
\begin{equation*}
    \sup_{t \in \Ts}\abs{\widehat{\Es}_a^{(t)}-\Es_{a,\pi}^{(t)}}
    \xrightarrow{p}
    0.
\end{equation*}
By \Cref{prop:independent-noise-comparison},
\begin{equation*}
    \sup_{t \in \Ts}\abs{\Es_{a,\pi}^{(t)}-\Es_{a,\mathrm{ind}}^{(t)}}
    \xrightarrow{p}
    0.
\end{equation*}
Therefore
\begin{equation*}
    \sup_{t \in \Ts}\abs{\widehat{\Es}_a^{(t)} - \Es_{a,\mathrm{ind}}^{(t)}}
    \le
    \sup_{t \in \Ts}\abs{\widehat{\Es}_a^{(t)} - \Es_{a,\pi}^{(t)}}
    +
    \sup_{t \in \Ts}\abs{\Es_{a,\pi}^{(t)} - \Es_{a,\mathrm{ind}}^{(t)}},
\end{equation*}
and the right-hand side is $o_p(1)$. This proves the theorem.
\end{proof}

\subsection{Risk decomposition and a bound on policy value error}

Recall that, for deterministic measurable $g$,
\begin{equation*}
    \Rs_\nu^\circ(g)
    =\mathbb{E}\sum_{a=1}^K
    \nu(a\mymid X)\{g(X,a)-g^\ast(X,a)\}^2.
\end{equation*}
For an $\mathcal F_{\mathrm{tr}}$-measurable predictor, the expectation is conditional on $\mathcal F_{\mathrm{tr}}$ and uses an independent test observation, as in Section~\ref{sec:formal-regime}.

\begin{proposition}[Prediction risk bounds policy value error]
\label{prop:risk-controls-value}
Assume known context-independent logging with $\mu(a)>0$ for every arm, let $g^\ast(X,A)=\E[R\mymid X,A]$, and suppose $\E[R^2]<\infty$. For every deterministic measurable or $\mathcal F_{\mathrm{tr}}$-measurable predictor $g$ satisfying $\Rs_\nu^\circ(g)<\infty$ almost surely,
\begin{align*}
    |\Vs_g(\nu)-\Vs(\nu)|
    &\le\sqrt{\Rs_\nu^\circ(g)},\\
    \Rs_\nu(g)
    &=\Rs_\nu(g^\ast)+\Rs_\nu^\circ(g).
\end{align*}
\end{proposition}

\begin{proof}[Proof of \Cref{prop:risk-controls-value}]
Let $\mathcal H$ denote either the trivial sigma-field or $\mathcal F_{\mathrm{tr}}$, let $\tilde g$ be $\mathcal H$-measurable, and let $(X,A,R)$ be an independent test draw. Since the logging policy is context independent,
\begin{align*}
    \Vs_{\tilde g}(\nu) - \Vs(\nu)
    &=
    \mathbb{E}\br{
        \sum_{a=1}^K
        \nu(a\mymid X)
        \p{\tilde g(X,a)-g^\ast(X,a)}
        \mymid
        \mathcal H
    }\\
    &=
    \mathbb{E}\br{
        w_A(X)\p{\tilde g(X,A)-g^\ast(X,A)}
        \mymid
        \mathcal H
    }.
\end{align*}
Conditional Cauchy--Schwarz yields
\begin{align*}
    \abs{\Vs_{\tilde g}(\nu) - \Vs(\nu)}
    &\le
    \mathbb{E}\br{w_A(X)\mymid\mathcal H}^{1/2}
    \mathbb{E}\br{
        w_A(X)\p{\tilde g(X,A)-g^\ast(X,A)}^2
        \mymid
        \mathcal H
    }^{1/2}.
\end{align*}
The first factor equals one because
\begin{align*}
    \mathbb E[w_A(X)\mymid X,\mathcal H]
    &=
    \sum_{a=1}^K
    \mu(a)\frac{\nu(a\mymid X)}{\mu(a)}
    =1,
\end{align*}
and the square of the second factor is
\begin{align*}
    \mathbb{E}\br{
        w_A(X)\p{\tilde g(X,A)-g^\ast(X,A)}^2
        \mymid
        \mathcal H
    } =
    \mathbb{E}\br{
        \sum_{a=1}^K\nu(a\mymid X)
        \p{\tilde g(X,a)-g^\ast(X,a)}^2
        \mymid
        \mathcal H
    }
    =\Rs_\nu^\circ(\tilde g).
\end{align*}
This proves the value bound.

For the risk decomposition, write $\varepsilon\coloneqq R-g^\ast(X,A)$ and $\delta_{\tilde g}(X,A)\coloneqq g^\ast(X,A)-\tilde g(X,A)$. Then, $R-\tilde g(X,A)=\varepsilon+\delta_{\tilde g}(X,A)$, and expanding the square yields
\begin{align*}
    \Rs_\nu(\tilde g)
    &=
    \mathbb E\br{w_A(X)\varepsilon^2\mymid\mathcal H}
    +\mathbb E\br{
        w_A(X)\delta_{\tilde g}(X,A)^2
        \mymid\mathcal H
    } +
    2\mathbb E\br{
        w_A(X)\varepsilon\delta_{\tilde g}(X,A)
        \mymid
        \mathcal H
    }.
\end{align*}
The cross term is zero, since iterated conditioning yields
\begin{align*}
    \mathbb E\br{
        w_A(X)\varepsilon\delta_{\tilde g}(X,A)
        \mymid\mathcal H
    } =
    \mathbb E\br{
        w_A(X)\delta_{\tilde g}(X,A)
        \mathbb E[\varepsilon\mymid X,A,\mathcal H]
        \mymid\mathcal H
    }
    =0.
\end{align*}

The first remaining term is $\Rs_\nu(g^\ast)$ and the second is $\Rs_\nu^\circ(\tilde g)$, proving the decomposition. Taking $\mathcal H$ to be trivial gives the deterministic statement, while taking $\mathcal H=\mathcal F_{\mathrm{tr}}$ gives its conditional out-of-sample version.
\end{proof}

These finite-sample conclusions use only the logging probabilities and the conditional mean of the reward; they do not require the neural-network assumptions.

\subsection{Proof of Theorem~\ref{thm:selection-main}}
\label{app:thm-selection-main-proof}

\thmselectionmain*
\begin{proof}
Set
\begin{equation*}
  \Delta_a
  \coloneqq
  \sup_{t\in\Ts}
  |\widehat\Es_a^{(t)}-\Es_{a,\mathrm{ind}}^{(t)}|.
\end{equation*}
Let $t_a^\star$ be the first minimizer of $\Es_{a,\mathrm{ind}}^{(t)}$ on the fixed ordered grid $\Ts$. Since $\hat t_a$ minimizes $\widehat\Es_a^{(t)}$,
\begin{align*}
  \Es_{a,\mathrm{ind}}^{(\hat t_a)}
  &\le
  \widehat\Es_a^{(\hat t_a)}+\Delta_a\\
  &\le
  \widehat\Es_a^{(t_a^\star)}+\Delta_a\\
  &\le
  \Es_{a,\mathrm{ind}}^{(t_a^\star)}+2\Delta_a.
\end{align*}
The definition of $t_a^\star$ also gives the lower bound, so
\begin{equation}
  0
  \le
  \Es_{a,\mathrm{ind}}^{(\hat t_a)}
  -\min_{t\in\Ts}\Es_{a,\mathrm{ind}}^{(t)}
  \le2\Delta_a.
  \label{eq:finite-grid-selection}
\end{equation}
For every $\epsilon>0$, the fixed number of arms gives
\begin{equation*}
  \Prob\left(\max_{a\in[K]}\Delta_a>\epsilon\right)
  \le
  \sum_{a=1}^K\Prob(\Delta_a>\epsilon)
  \longrightarrow0
\end{equation*}
by \Cref{thm:uniform-risk}. Thus, $\max_a\Delta_a=o_p(1)$.

We next identify the part of the arm-wise test risk that depends on the predictor. Write $\delta_{a,t}(Z_a)\coloneqq g^\ast(Z_a,a)-g_{a,t}(Z_a)$. Since $g_{a,t}$ is $\widehat{\mathcal F}_a$-measurable, \Cref{lem:fresh-draw-conditioning} gives
\begin{equation*}
  \E\left[
    \nu(a\mymid Z_a)\delta_{a,t}(Z_a)^2
    \mymid\widehat{\mathcal F}_a
  \right]
  =
  \E\left[
    \nu(a\mymid Z_a)\delta_{a,t}(Z_a)^2
    \mymid\mathcal F_{\mathrm{tr}}
  \right]
  =
  \Rs_{\nu,a}^\circ(g_{a,t}).
\end{equation*}
Since $\mu(a)w_a(Z_a)=\nu(a\mymid Z_a)$,
\begin{align*}
  \mu(a)\Es_{a,\mathrm{ind}}^{(t)}
  &=
  \E\left[
    \nu(a\mymid Z_a)
    \{\xi_a^{\mathrm{new}}+\delta_{a,t}(Z_a)\}^2
    \mymid\widehat{\mathcal F}_a
  \right]\\
  &=
  C_a+\Rs_{\nu,a}^\circ(g_{a,t})
  +2\E\left[
    \nu(a\mymid Z_a)\xi_a^{\mathrm{new}}\delta_{a,t}(Z_a)
    \mymid\widehat{\mathcal F}_a
  \right],
\end{align*}
where
\begin{equation*}
  C_a
  \coloneqq
  \mu(a)\E\left[
    w_a(Z_a)(\xi_a^{\mathrm{new}})^2
    \mymid\widehat{\mathcal F}_a
  \right]
  =\sigma_{\xi,a}^2\E[\nu(a\mymid Z_a)].
\end{equation*}
Thus, $C_a$ does not depend on $t$. The cross term is zero because $\E[\xi_a^{\mathrm{new}}\mymid Z_a,\widehat{\mathcal F}_a]=0$. Therefore
\begin{equation}
  \mu(a)\Es_{a,\mathrm{ind}}^{(t)}
  =C_a+\Rs_{\nu,a}^\circ(g_{a,t}),
  \qquad t\in\Ts.
  \label{eq:arm-risk-decomposition}
\end{equation}
Substituting \eqref{eq:arm-risk-decomposition} into \eqref{eq:finite-grid-selection} and canceling $C_a$ gives
\begin{equation*}
  0
  \le
  \Rs_{\nu,a}^\circ(g_{a,\hat t_a})
  -\min_{t\in\Ts}\Rs_{\nu,a}^\circ(g_{a,t})
  \le
  2\mu(a)\Delta_a.
\end{equation*}

By \Cref{lem:fresh-draw-conditioning}, each arm-wise test risk has the same value when conditioned on $\widehat{\mathcal F}_a$ or on $\mathcal F_{\mathrm{tr}}$. Therefore, for every $\tb=(t_1,\ldots,t_K)\in\Ts^K$, conditioning on the fresh logged action in the definition of $\Rs_\nu$ gives
\begin{equation*}
    \Rs_\nu(g_{\tb})
    =
    \sum_{a=1}^K \mu(a)\Es_{a,\mathrm{ind}}^{(t_a)}
    \qquad
    \text{for every } \tb \in \Ts^K
\end{equation*}
so minimization over $\Ts^K$ separates across arms:
\begin{equation*}
  \min_{\tb\in\Ts^K}\Rs_\nu(g_{\tb})
  =
  \sum_{a=1}^K\mu(a)
  \min_{t\in\Ts}\Es_{a,\mathrm{ind}}^{(t)}.
\end{equation*}
Consequently, \eqref{eq:finite-grid-selection} implies
\begin{align*}
  0
  &\le
  \Rs_\nu(g_{\thatb})
  -\min_{\tb\in\Ts^K}\Rs_\nu(g_{\tb})\\
  &=
  \sum_{a=1}^K\mu(a)
  \left\{
    \Es_{a,\mathrm{ind}}^{(\hat t_a)}
    -\min_{t\in\Ts}\Es_{a,\mathrm{ind}}^{(t)}
  \right\}\\
  &\le
  2\sum_{a=1}^K\mu(a)\Delta_a
  =o_p(1).
\end{align*}
By \Cref{prop:risk-controls-value}, $\Rs_\nu(g)=\Rs_\nu(g^\ast)+\Rs_\nu^\circ(g)$, where $\Rs_\nu(g^\ast)$ is constant in $g$. Subtracting this common term from the preceding oracle inequality gives
\begin{equation*}
  \Rs_\nu^\circ(g_{\thatb})
  \le
  \min_{\tb\in\Ts^K}\Rs_\nu^\circ(g_{\tb})+o_p(1).
\end{equation*}
Applying the value bound from the same proposition now yields
\begin{align*}
  |\Vs_{g_{\thatb}}(\nu)-\Vs(\nu)|^2
  &\le
  \Rs_\nu^\circ(g_{\thatb})\\
  &\le
  \min_{\tb\in\Ts^K}\Rs_\nu^\circ(g_{\tb})+o_p(1),
\end{align*}
which proves all claims.
\end{proof}

\subsection{Suboptimal stopping after full-data refitting}

With a fixed validation fraction, cross-validation estimates the test risk of models trained at a different limiting sample-to-dimension ratio. The following example shows that ten-fold cross-validation can correctly rank these models yet select a suboptimal iterate after full-data refitting.

Let $F_1,\ldots,F_{10}$ partition $[m]$ independently of the observations, with $|F_j|=m/10+o(m)$. Write $g_{a,t}^{(-j)}$ for the arm-$a$ model trained on $\Is_a\setminus F_j$, using the same initialization and learning rates as the full-data model $g_{a,t}$. On $\Ts=\{0,1\}$, define
\begin{equation*}
 \widehat{\mathcal E}^{\rm CV}_a(t)
 \coloneqq
 \frac1{N_a}\sum_{j=1}^{10}\sum_{i\in\Is_a\cap F_j}
 \{R_i-g_{a,t}^{(-j)}(X_i)\}^2,
 \qquad
 \hat t_a^{\rm CV}\in\arg\min_{t\in\Ts}
 \widehat{\mathcal E}^{\rm CV}_a(t).
\end{equation*}
The importance weights equal one in the example below. Let $t_a^\star\in\arg\min_{t\in\Ts}\Rs_{\nu,a}^\circ(g_{a,t})$ be the full-data risk minimizer, and set $\hat{\mathbf t}^{\rm CV}=(\hat t_1^{\rm CV},\ldots,\hat t_K^{\rm CV})$. Break ties toward $t=0$ and use any measurable convention when a required arm count is zero; the latter event has probability tending to zero.

\begin{samepage}
\begin{proposition}[Suboptimal stopping after cross-validation]
\label{prop:cv-stopping-reversal}
Let $X_i\overset{\rm iid}{\sim}N(0,I_d)$, let $K\ge2$ be fixed, and assume context-independent logging with $\mu(a)=\nu(a\mid x)=1/K$ and $m/d\to K$. Take noiseless rewards $R_i=g^\ast(X_i,A_i)=\beta^\top X_i$, where $\beta=d^{-1/2}(1,\ldots,1)^\top$. For every full-data and fold-complement fit, use the depth-two, width-one network with identity activation, initialization $W_{a,1}^{(0)}=0$, $W_{a,2}^{(0)}=1$, and first-layer step size $\eta=39/40$. The second-layer step size may be any fixed positive number satisfying Assumption~\ref{ass:formal-nn-regularity}. Then, as $d\to\infty$,
\begin{equation*}
 \Prob\bigl(\hat t_a^{\rm CV}=0,\ t_a^\star=1
             \text{ for every }a\in[K]\bigr)\longrightarrow1,
\end{equation*}
and the full-data model selected by CV has a nonvanishing excess prediction risk over the best candidate iterates:
\begin{equation*}
 \Rs_\nu^\circ(g_{\hat{\mathbf t}^{\rm CV}})
 -\min_{\mathbf t\in\{0,1\}^K}\Rs_\nu^\circ(g_{\mathbf t})
 \xrightarrow{p}\frac{39}{800}.
\end{equation*}
\end{proposition}
\end{samepage}

\begin{proof}
For a training set of $n$ observations in an arm, let $S_n=n^{-1}\sum_{i=1}^nX_iX_i^\top$. The deeper-layer gradient is zero at initialization, and one simultaneous gradient-descent update yields $g_1(x)=\eta x^\top S_n\beta$. For an independent $Z\sim N(0,I_d)$, the initialization has prediction risk $\|\beta\|^2=1$, while the first iterate has conditional prediction risk
\begin{equation}
 \begin{aligned}
 r_n&\coloneqq\E[(g_1(Z)-\beta^\top Z)^2\mid S_n]
 =\|\eta S_n\beta-\beta\|^2
 \\
 &\xrightarrow{p}r(\alpha)\coloneqq(1-\eta)^2+\frac{\eta^2}{\alpha},
 \qquad n/d\to\alpha\in(0,\infty).
 \end{aligned}
 \label{eq:cv-one-step-risk}
\end{equation}
To verify the limit, Gaussian rotational invariance allows us to take $\beta=e_1$. Write $z_i=X_{i1}$ and $s_n=n^{-1}\sum_i z_i^2\to_p1$. Conditional on $(z_i)_{i=1}^n$, the other coordinates of $S_n\beta$ are independent $N(0,s_n/n)$ variables. Thus
\begin{equation*}
 r_n\ \overset{\rm d}{=}\ (\eta s_n-1)^2
       +\eta^2\frac{s_n}{n}U_{d-1},
 \qquad U_{d-1}\sim\chi^2_{d-1}\text{ independent of }s_n.
\end{equation*}
Since $U_{d-1}/d\to_p1$, this proves \eqref{eq:cv-one-step-risk}. The calculation also applies conditional on the action labels and fold partition when the training count is random and $n/d\to_p\alpha$.

Let $v_{a,j}=|\Is_a\cap F_j|$ and $n_{a,-j}=N_a-v_{a,j}$. Uniform logging and $m/d\to K$ give, jointly over the fixed numbers of arms and folds,
\begin{equation*}
 \frac{N_a}{d}\xrightarrow{p}1,\qquad
 \frac{v_{a,j}}{d}\xrightarrow{p}\frac1{10},\qquad
 \frac{n_{a,-j}}{d}\xrightarrow{p}\frac9{10}.
\end{equation*}
Write $r_{a,-j}$ for the conditional prediction risk of $g_{a,1}^{(-j)}$. By \eqref{eq:cv-one-step-risk}, $r_{a,-j}\to_p r(9/10)$. Conditional on the action labels, fold partition, and training data outside fold $j$, the validation residuals are independent $N(0,r_{a,-j})$ variables. Their mean square therefore has conditional mean $r_{a,-j}$ and variance $2r_{a,-j}^2/v_{a,j}=o_p(1)$. Conditional Chebyshev's inequality and a finite union over arms and folds yield
\begin{equation*}
 \frac1{v_{a,j}}\sum_{i\in\Is_a\cap F_j}
 \{R_i-g_{a,1}^{(-j)}(X_i)\}^2
 =r_{a,-j}+o_p(1).
\end{equation*}
Since the CV criterion averages these fold errors with weights $v_{a,j}/N_a$, we obtain, for every arm,
\begin{align*}
 \widehat{\mathcal E}^{\rm CV}_a(0)&\xrightarrow{p}1,
 &\widehat{\mathcal E}^{\rm CV}_a(1)&\xrightarrow{p}
 r(9/10)=\frac{1691}{1600}>1,\\
 K\Rs_{\nu,a}^\circ(g_{a,0})&=1,
 &K\Rs_{\nu,a}^\circ(g_{a,1})&\xrightarrow{p}
 r(1)=\frac{761}{800}<1.
\end{align*}
The strict inequalities imply the stated selection event. On that event, the excess prediction risk equals
\begin{equation*}
 \frac1K\sum_{a=1}^K
 \{1-K\Rs_{\nu,a}^\circ(g_{a,1})\}
 \xrightarrow{p}1-\frac{761}{800}=\frac{39}{800}.
\end{equation*}
This proves both conclusions.
\end{proof}

The example satisfies the working assumptions: the reward link and activation are identity maps, the noise is zero, the importance weights are constant, and $\sqrt d\,\|\beta\|_\infty=1$ with bounded initializations. It concerns a fixed validation fraction; Appendix~\ref{app:exp-vanishing-holdouts} examines two vanishing-fraction alternatives empirically.

\section{Value Estimation Without Sample Splitting and Consequences}
\label{app:proof-value-estimation}

We now prove that ADM consistently estimates $\Vs_{g_{\tb}}(\nu)$ and ADR consistently estimates $\Vs(\nu)$ when both are evaluated on the reward-model training sample. For each arm, the proof treats predictions on that arm's training observations using the Onsager correction. Predictions on observations assigned to other arms satisfy conditional laws of large numbers. In the ADR estimator, the resulting conditional expectations of the fitted predictions cancel.

For $a,b\in[K]$ and $t\in\Ts$, define
\begin{align*}
  M_b^{(t)}
  &\coloneqq
  \E\left[
    w_b(Z)f_{W_b^{(t)}}(Z)
    \mymid
    \mathcal F_{\mathrm{tr}}
  \right],\\
  \overline M_{a,b}^{(t)}
  &\coloneqq
  \frac1{N_a}\sum_{i\in\Is_a}
  w_b(X_i)f_{W_b^{(t)}}(X_i),
  \qquad a\ne b,\\
  \overline R_a
  &\coloneqq
  \frac1{N_a}\sum_{i\in\Is_a}w_a(X_i)R_i,
  \qquad
  R_a^\star
  \coloneqq
  \E[w_a(X)g^\ast(X,a)],
\end{align*}
where $Z\sim N(0,I_d)$ is independent of $\mathcal F_{\mathrm{tr}}$. By \Cref{lem:fresh-draw-conditioning}, the displayed definition of $M_b^{(t)}$ equals the version conditioned on $\sigma(\Ds_m,W_b^{(0)})$ in \Cref{prop:diagonal-comparison}; the integrand depends on the training data only through $W_b^{(t)}$.

\begin{lemma}[Convergence of the averages used in value estimation]
\label{lem:value-sample-averages}
Assume the conditions of Section~\ref{sec:formal-regime}, with $m/d\to\phi\in(0,\infty)$ and fixed $K,L,q,\Ts$. Then, as $d\to\infty$,
\begin{align*}
  \max_{a\in[K]}\sup_{t\in\Ts}
  |\widehat M_{a,\diag}^{(t)}-M_a^{(t)}|
  &\xrightarrow{p}0,\\
  \max_{a\ne b}\sup_{t\in\Ts}
  |\overline M_{a,b}^{(t)}-M_b^{(t)}|
  &\xrightarrow{p}0,\\
  \max_{a\in[K]}\left|\frac{N_a}{m}-\mu(a)\right|
  &=O_p(d^{-1/2}),\\
  \max_{a\in[K]}|\overline R_a-R_a^\star|
  &=o_p(1),\\
  \max_{a\in[K]}\sup_{t\in\Ts}|M_a^{(t)}|
  &=d^{o_p(1)}.
\end{align*}
Under the sub-Gaussian noise condition in \Cref{ass:single-index-logging}, the reward-average bound strengthens to $O_p(d^{-1/2})$.
\end{lemma}

\begin{proof}
The diagonal fitted-value bound follows from \Cref{prop:diagonal-comparison} and a finite union over the arms. For $a\ne b$, let
\begin{equation*}
  \widehat{\mathcal F}_b
  \coloneqq
  \sigma(\Is_b,X_{\Is_b},R_{\Is_b},W_b^{(0)}),
\end{equation*}
and let
\begin{equation*}
  \widehat{\mathcal F}_{a,b}
  \coloneqq
  \sigma(A_1,\ldots,A_m,X_{\Is_b},R_{\Is_b},W_b^{(0)}).
\end{equation*}
The count event $E_{\mathrm{cnt}}$ is measurable with respect to $\widehat{\mathcal F}_{a,b}$. For the off-diagonal comparison, work on this event, so $N_a\ge1$. Conditional on $\widehat{\mathcal F}_{a,b}$, the arm-$b$ predictor is fixed, while $\{X_i:i\in\Is_a\}$ are i.i.d.\ $N(0,I_d)$ and independent of that predictor. Because $Z$ is independent of $\mathcal F_{\mathrm{tr}}$, the definition of $M_b^{(t)}$ is unchanged if $\mathcal F_{\mathrm{tr}}$ is replaced by $\widehat{\mathcal F}_b$; hence, $M_b^{(t)}$ is $\widehat{\mathcal F}_{a,b}$-measurable. It follows that the variables $Y_{i,b}^{(t)}\coloneqq w_b(X_i)f_{W_b^{(t)}}(X_i)$, $i\in\Is_a$, are conditionally i.i.d.\ with mean $M_b^{(t)}$. The same conditioning argument shows that the second moment of $w_b(Z)f_{W_b^{(t)}}(Z)$ agrees under $\widehat{\mathcal F}_{a,b}$ and $\mathcal F_{\mathrm{tr}}$. Boundedness of $w_b$ and \Cref{prop:prediction-moment-bound} therefore give
\begin{equation*}
  \sup_{t\in\Ts}
  \E[|Y_{i,b}^{(t)}|^2\mymid\widehat{\mathcal F}_{a,b}]
  =d^{o_p(1)}.
\end{equation*}
Therefore,
\begin{align}
  \E\left[
    |\overline M_{a,b}^{(t)}-M_b^{(t)}|^2
    \mymid\widehat{\mathcal F}_{a,b}
  \right]
  &=
  \frac1{N_a}
  \operatorname{Var}\left(
    Y_{i,b}^{(t)}\mymid\widehat{\mathcal F}_{a,b}
  \right)
  \nonumber\\
  &\le
  \frac1{N_a}
  \E\left[
    |Y_{i,b}^{(t)}|^2
    \mymid\widehat{\mathcal F}_{a,b}
  \right]
  \le d^{-1+o_p(1)},
  \label{eq:offdiagonal-second-moment}
\end{align}
uniformly over the fixed grid. For every $\epsilon>0$, conditional Chebyshev's inequality and measurability of $E_{\mathrm{cnt}}$ give
\begin{align*}
 \Prob\left(|\overline M_{a,b}^{(t)}-M_b^{(t)}|>\epsilon\right) \le\Prob(E_{\mathrm{cnt}}^c)
 +\E\left[\mathbf 1_{E_{\mathrm{cnt}}}
   \min\left\{1,
     \frac{\E[|Y_{i,b}^{(t)}|^2\mymid\widehat{\mathcal F}_{a,b}]}
          {N_a\epsilon^2}
   \right\}\right].
\end{align*}
The integrand is bounded by one and, by \eqref{eq:offdiagonal-second-moment}, converges to zero in probability. Its expectation therefore converges to zero. Since $\Prob(E_{\mathrm{cnt}}^c)\to0$, a finite union over $(a,b,t)$ proves the second claim.

For the count bound, $N_a\sim\operatorname{Binomial}(m,\mu(a))$, so
\begin{equation*}
  \E\left[\frac{N_a}{m}\right]=\mu(a),
  \qquad
  \operatorname{Var}\left(\frac{N_a}{m}\right)
  =\frac{\mu(a)\{1-\mu(a)\}}{m}.
\end{equation*}
For every $M>0$, Chebyshev's inequality and a union bound give
\begin{align*}
 \Prob\left(\sqrt d\max_a|N_a/m-\mu(a)|>M\right)
 &\le\sum_{a=1}^K
 \frac{d\,\mu(a)\{1-\mu(a)\}}{mM^2}
 \le\frac{C}{M^2},
\end{align*}
where $C$ is uniform in $d$ because $m\asymp d$ and $K$ is fixed. Letting $M\to\infty$ proves the third claim.

For the reward-average comparison, continue to work on $E_{\mathrm{cnt}}$. Decompose
\begin{align*}
  \overline R_a-R_a^\star
  &=\frac1{N_a}\sum_{i\in\Is_a}
    \{w_a(X_i)g^\ast(X_i,a)-R_a^\star\} +\E[w_a(X)]\frac1{N_a}\sum_{i\in\Is_a}\varepsilon_i +\frac1{N_a}\sum_{i\in\Is_a}
    \{w_a(X_i)-\E[w_a(X)]\}\varepsilon_i.
\end{align*}
For the first term, conditional on the action labels, independence and the link moment bound give
\begin{align*}
 \E\left[
 \left|\frac1{N_a}\sum_{i\in\Is_a}
 \{w_a(X_i)g^\ast(X_i,a)-R_a^\star\}\right|^2
 \mymid A_{1:m}\right] =
 \frac{\operatorname{Var}\{w_a(X)g^\ast(X,a)\}}{N_a}
 \le\frac{C}{N_a}
 \le\frac{C}{c_-d}
\end{align*}
on $E_{\mathrm{cnt}}$. Hence, this term is $O_p(d^{-1/2})$. The second term is also $O_p(d^{-1/2})$ by \eqref{eq:noise-rowwise-lln} and $\E[w_a(X)]\le B_h$. For the last term, condition additionally on the realized noises and write
\begin{equation*}
 S_a\coloneqq\frac1{N_a}\sum_{i\in\Is_a}
 \{w_a(X_i)-\E[w_a(X)]\}\varepsilon_i.
\end{equation*}
The contexts are conditionally independent, so
\begin{align*}
 \E[S_a\mymid A_{1:m},\varepsilon_{1:m}]&=0,\\
 \E[S_a^2\mymid A_{1:m},\varepsilon_{1:m}]
 &=\frac{\operatorname{Var}(w_a(X))}{N_a^2}
   \sum_{i\in\Is_a}\varepsilon_i^2\\
 &\le\frac{B_h^2}{N_a}\|\xi_{\Is_a}\|_\infty^2
 \le\frac{B_h^2}{c_-d}\varkappa_{\ast,a}^2
 =d^{-1+o_p(1)}
\end{align*}
on $E_{\mathrm{cnt}}$. Conditional Chebyshev's inequality yields, for $\epsilon>0$,
\begin{equation*}
 \Prob(|S_a|>\epsilon)
 \le\Prob(E_{\mathrm{cnt}}^c)
 +\E\left[\mathbf1_{E_{\mathrm{cnt}}}
   \min\left\{1,
     \frac{B_h^2\varkappa_{\ast,a}^2}{c_-d\epsilon^2}
   \right\}\right]
 \longrightarrow0.
\end{equation*}
As above, the integrand is bounded by one and converges to zero in probability, so its expectation converges to zero. Combining the three terms and taking a finite union over the arms proves the fourth claim.

Under the sub-Gaussian noise condition in \Cref{ass:single-index-logging}, $\operatorname{Var}(w_a(X)R\mymid A=a)$ is uniformly bounded. Conditional on $\Is_a$, the weighted rewards are i.i.d.\ with mean $R_a^\star$ and variance bounded by a constant. Thus, on $E_{\mathrm{cnt}}$,
\begin{equation*}
 \E[|\overline R_a-R_a^\star|^2\mymid A_{1:m}]
 =\frac{\operatorname{Var}(w_a(X)R\mymid A=a)}{N_a}
 \le\frac{C}{c_-d}.
\end{equation*}
For $M>0$, conditional Chebyshev's inequality and a union bound yield
\begin{equation*}
 \Prob\left(\sqrt d\max_a|\overline R_a-R_a^\star|>M\right)
 \le\Prob(E_{\mathrm{cnt}}^c)+\frac{KC}{c_-M^2}.
\end{equation*}
Taking $d\to\infty$ and then $M\to\infty$ proves the sharper bound.

Finally, weighted Cauchy--Schwarz yields
\begin{align*}
  |M_a^{(t)}|^2
  &\le
  \E[w_a(Z)]
  \E\left[
    w_a(Z)|f_{W_a^{(t)}}(Z)|^2
    \mymid\mathcal F_{\mathrm{tr}}
  \right]\\
  &\le
  B_h^2
  \E\left[
    |f_{W_a^{(t)}}(Z)|^2
    \mymid\mathcal F_{\mathrm{tr}}
  \right].
\end{align*}
By \Cref{lem:fresh-draw-conditioning}, the same conditional prediction-moment bound holds with conditioning on $\mathcal F_{\mathrm{tr}}$. The fixed-grid bound in \Cref{prop:prediction-moment-bound}, followed by a finite union over arms, proves the last claim.
\end{proof}

\begin{proof}[Proof of \Cref{thm:value_estimation}]
For $i\in\Is_a$, abbreviate the corrected observed-action prediction by
\begin{equation*}
  c_{i,a}^{(t)}
  \coloneqq
  G_{W_{a,[2:L]}^{(t)}}(\widehat U_{i,a}^{(t)}).
\end{equation*}
Define the comparison errors
\begin{align*}
  \delta_{\diag}
  &\coloneqq
  \max_{a\in[K]}\sup_{t\in\Ts}
  |\widehat M_{a,\diag}^{(t)}-M_a^{(t)}|,\\
  \delta_{\mathrm{off}}
  &\coloneqq
  \max_{a\ne b}\sup_{t\in\Ts}
  |\overline M_{a,b}^{(t)}-M_b^{(t)}|,\\
  \delta_N
  &\coloneqq
  \max_{a\in[K]}|N_a/m-\mu(a)|,\\
  \delta_R
  &\coloneqq
  \max_{a\in[K]}|\overline R_a-R_a^\star|,\\
  B_M
  &\coloneqq
  \max_{a\in[K]}\sup_{t\in\Ts}|M_a^{(t)}|.
\end{align*}
By \Cref{lem:value-sample-averages},
\begin{equation}
  \delta_{\diag}=o_p(1),
  \quad
  \delta_{\mathrm{off}}=o_p(1),
  \quad
  \delta_N=O_p(d^{-1/2}),
  \quad
  \delta_R=o_p(1),
  \quad
  B_M=d^{o_p(1)}.
  \label{eq:value-comparison-errors}
\end{equation}

\paragraph{ADM.}
For $i\in\Is_a$, \eqref{eq:adm-model} and $\nu(b\mymid x)=\mu(b)w_b(x)$ give the rowwise identity
\begin{align*}
  \sum_{b=1}^K\nu(b\mymid X_i)\widehat g_{\tb}(X_i,b)
  &=
  \sum_{b\ne a}
  \mu(b)w_b(X_i)f_{W_b^{(t_b)}}(X_i)
  +\mu(a)w_a(X_i)c_{i,a}^{(t_a)}.
\end{align*}
Summing first within each logged-action set and then over the arms yields
\begin{equation}
    \widehat{\Vs}_{\ADM}(\tb)
    =
    \sum_{a=1}^K \frac{N_a}{m}
    \br{
        \sum_{b \neq a}\mu(b)\,\overline M_{a,b}^{(t_b)}
        +
        \mu(a)\,\widehat M_{a,\diag}^{(t_a)}
    }.
    \label{eq:adm-decomposition}
\end{equation}

Let
\begin{equation*}
  T_{\ADM}(\tb)
  \coloneqq
  \sum_{a=1}^K \mu(a)
  \left\{
    \sum_{b\ne a}\mu(b)M_b^{(t_b)}
    +\mu(a)M_a^{(t_a)}
  \right\}.
\end{equation*}
Subtracting $T_{\ADM}(\tb)$ from \eqref{eq:adm-decomposition} yields
\begin{align*}
 \widehat\Vs_{\ADM}(\tb)-T_{\ADM}(\tb) &=\sum_{a=1}^K\frac{N_a}{m}
 \Bigl\{\sum_{b\ne a}\mu(b)
   (\overline M_{a,b}^{(t_b)}-M_b^{(t_b)}) + \mu(a)(\widehat M_{a,\diag}^{(t_a)}-M_a^{(t_a)})\Bigr\}\\
 &\qquad+\sum_{a=1}^K\left\{\frac{N_a}{m}-\mu(a)\right\}
   \sum_{b=1}^K\mu(b)M_b^{(t_b)}.
\end{align*}
The second sum is zero because $\sum_aN_a/m=\sum_a\mu(a)=1$. Since the weights in the first sum are nonnegative,
\begin{align*}
 \sup_{\tb\in\Ts^K}
 |\widehat\Vs_{\ADM}(\tb)-T_{\ADM}(\tb)|
 &\le\sum_{a=1}^K\frac{N_a}{m}
   \{(1-\mu(a))\delta_{\mathrm{off}}+\mu(a)\delta_{\diag}\}\\*
 &\le\delta_{\mathrm{off}}+\delta_{\diag}
 =o_p(1).
\end{align*}
Equivalently,
\begin{equation}
    \widehat{\Vs}_{\ADM}(\tb)
    =
    \sum_{a=1}^K \mu(a)
    \br{
        \sum_{b \neq a}\mu(b)\,M_b^{(t_b)}
        +
        \mu(a)\,M_a^{(t_a)}
    }
    +
    o_p(1),
    \label{eq:adm-almost-there}
\end{equation}
uniformly over $\tb \in \Ts^K$.

Now
\begin{align*}
    \sum_{a=1}^K \mu(a)\sum_{b \neq a}\mu(b)\,M_b^{(t_b)}
    &=
    \sum_{b=1}^K \mu(b)\p{1-\mu(b)}M_b^{(t_b)}.
\end{align*}
Adding the logged-action term gives
\begin{equation*}
    \sum_{a=1}^K \mu(a)
    \br{
        \sum_{b \neq a}\mu(b)\,M_b^{(t_b)}
        +
        \mu(a)\,M_a^{(t_a)}
    }
    =
    \sum_{a=1}^K \mu(a)\,M_a^{(t_a)}.
\end{equation*}
Finally,
\begin{align}
    \sum_{a=1}^K \mu(a)\,M_a^{(t_a)}
    &=
    \mathbb{E}\br{
        \sum_{a=1}^K \mu(a)\,w_a(Z)\,f_{W_a^{(t_a)}}(Z)
        \mymid
        \mathcal F_{\mathrm{tr}}
    }
    \nonumber\\
    &=
    \mathbb{E}\br{
        \sum_{a=1}^K \nu(a \mymid Z)\,g_{\tb}(Z,a)
        \mymid
        \mathcal F_{\mathrm{tr}}
    }
    \nonumber\\
    &=
    \Vs_{g_{\tb}}(\nu).
    \label{eq:adm-target}
\end{align}
The last equality is the definition of the policy value under the fitted reward model, conditional on the training data and initialization. Substituting \eqref{eq:adm-target} into \eqref{eq:adm-almost-there} proves the uniform ADM claim.

\paragraph{ADR.}
For $i\in\Is_a$, the ADR summand satisfies
\begin{align*}
  &\sum_{b=1}^K\nu(b\mymid X_i)\widehat g_{\tb}(X_i,b)
  +w_a(X_i)\{R_i-c_{i,a}^{(t_a)}\}\\
  &\quad=
  \sum_{b\ne a}\mu(b)w_b(X_i)f_{W_b^{(t_b)}}(X_i)
  +w_a(X_i)R_i
  +\{\mu(a)-1\}w_a(X_i)c_{i,a}^{(t_a)}\\
  &\quad=
  \sum_{b\ne a}\mu(b)w_b(X_i)f_{W_b^{(t_b)}}(X_i)
  +w_a(X_i)R_i
  -\{1-\mu(a)\}w_a(X_i)c_{i,a}^{(t_a)}.
\end{align*}
Here we used $\nu(b\mymid x)=\mu(b)w_b(x)$ and $\nu(a\mymid x)-w_a(x)=-\p{1-\mu(a)}w_a(x)$ for the observed arm. Summing within each $\Is_a$ and then over $a$ gives
\begin{equation}
    \widehat{\Vs}_{\ADR}(\tb)
    =
    \sum_{a=1}^K \frac{N_a}{m}
    \br{
        \sum_{b \neq a}\mu(b)\,\overline M_{a,b}^{(t_b)}
        +
        \overline R_a
        -
        \p{1-\mu(a)}\widehat M_{a,\diag}^{(t_a)}
    }.
    \label{eq:adr-decomposition}
\end{equation}

Define
\begin{equation*}
  T_{\ADR}(\tb)
  \coloneqq
  \sum_{a=1}^K\mu(a)
  \left\{
    \sum_{b\ne a}\mu(b)M_b^{(t_b)}
    +R_a^\star
    -\{1-\mu(a)\}M_a^{(t_a)}
  \right\}.
\end{equation*}
Subtracting $T_{\ADR}(\tb)$ from \eqref{eq:adr-decomposition} yields
\begin{align*}
 \widehat\Vs_{\ADR}(\tb)-T_{\ADR}(\tb) &=\sum_{a=1}^K\frac{N_a}{m}
 \Bigl\{\sum_{b\ne a}\mu(b)
    (\overline M_{a,b}^{(t_b)}-M_b^{(t_b)}) +(\overline R_a-R_a^\star)
    -(1-\mu(a))(\widehat M_{a,\diag}^{(t_a)}-M_a^{(t_a)})\Bigr\}\\
 &\qquad+\sum_{a=1}^K\left\{\frac{N_a}{m}-\mu(a)\right\}
    \{R_a^\star-M_a^{(t_a)}\}.
\end{align*}
Here the common term $\sum_b\mu(b)M_b^{(t_b)}$ again cancels from the second sum. For $0<\eta<1/2$, the count bound and $B_M=d^{o_p(1)}$ give
\begin{equation*}
 \delta_N B_M
 =O_p(d^{-1/2})\,O_p(d^\eta)
 =O_p(d^{-1/2+\eta})
 =o_p(1).
\end{equation*}
Since $\max_a|R_a^\star|\le C$ by the link and weight assumptions, the triangle inequality yields
\begin{align*}
 \sup_{\tb\in\Ts^K}
 |\widehat\Vs_{\ADR}(\tb)-T_{\ADR}(\tb)|
 &\le\delta_{\mathrm{off}}+\delta_R+\delta_{\diag}
     +K\delta_N(C+B_M)\\
 &=o_p(1)
\end{align*}
by \eqref{eq:value-comparison-errors} and $\delta_NB_M=o_p(1)$. Therefore
\begin{equation}
    \widehat{\Vs}_{\ADR}(\tb)
    =
    \sum_{a=1}^K \mu(a)
    \br{
        \sum_{b \neq a}\mu(b)\,M_b^{(t_b)}
        +
        R_a^\star
        -
        \p{1-\mu(a)}M_a^{(t_a)}
    }
    +
    o_p(1),
    \label{eq:adr-almost-there}
\end{equation}
uniformly over $\tb \in \Ts^K$.

To identify the leading term, exchange the order of summation:
\begin{align*}
 T_{\ADR}(\tb)
 &=\sum_{a=1}^K\mu(a)R_a^\star
   +\sum_{b=1}^K\mu(b)M_b^{(t_b)}\sum_{a\ne b}\mu(a) -\sum_{a=1}^K\mu(a)(1-\mu(a))M_a^{(t_a)}\\
 &=\sum_{a=1}^K\mu(a)R_a^\star
   +\sum_{b=1}^K\mu(b)(1-\mu(b))M_b^{(t_b)} -\sum_{a=1}^K\mu(a)(1-\mu(a))M_a^{(t_a)}\\
 &=\sum_{a=1}^K\mu(a)R_a^\star\\
 &=\E\left[\sum_{a=1}^K\nu(a\mymid X)g^\ast(X,a)\right]
 =\Vs(\nu).
\end{align*}
Substitution into \eqref{eq:adr-almost-there} proves the uniform ADR claim. Finally, uniformity over $\Ts^K$ gives the selected-model statements because $\thatb\in\Ts^K$:
\begin{align*}
  |\widehat\Vs_{\ADM}(\thatb)-\Vs_{g_{\thatb}}(\nu)|
  &\le
  \sup_{\tb\in\Ts^K}
  |\widehat\Vs_{\ADM}(\tb)-\Vs_{g_{\tb}}(\nu)|
  \xrightarrow{p}0,\\
  |\widehat\Vs_{\ADR}(\thatb)-\Vs(\nu)|
  &\le
  \sup_{\tb\in\Ts^K}
  |\widehat\Vs_{\ADR}(\tb)-\Vs(\nu)|
  \xrightarrow{p}0.
\end{align*}
\end{proof}

\begin{corollary}[Uniform ADM--ADR comparison]
\label{cor:adr-selected}
Under the assumptions of \Cref{thm:value_estimation}, the estimators in \eqref{eq:adm} and \eqref{eq:adr} satisfy, as $d\to\infty$,
\begin{equation*}
    \sup_{\mathbf{t} \in \mathcal{T}^K}
    \Bigl|
        \widehat{\mathcal{V}}_{\ADR}(\mathbf{t})
        -
        \widehat{\mathcal{V}}_{\ADM}(\mathbf{t})
        +
        \mathcal{V}_{g_{\mathbf{t}}}(\nu)
        -
        \mathcal{V}(\nu)
    \Bigr|
    \xrightarrow{p}
    0.
\end{equation*}
In particular,
\begin{equation*}
    \widehat{\Vs}_{\ADR}(\thatb)
    -
    \widehat{\Vs}_{\ADM}(\thatb)
    =
    -\p{\Vs_{g_{\thatb}}(\nu)-\Vs(\nu)}
    +
    o_p(1).
\end{equation*}
\end{corollary}
\begin{proof}
For each $\tb \in \Ts^K$,
\begin{equation*}
    \widehat{\Vs}_{\ADR}(\tb) - \widehat{\Vs}_{\ADM}(\tb)
    =
    \br{\widehat{\Vs}_{\ADR}(\tb) - \Vs(\nu)}
    -
    \br{\widehat{\Vs}_{\ADM}(\tb) - \Vs_{g_{\tb}}(\nu)}
    -
    \br{\Vs_{g_{\tb}}(\nu) - \Vs(\nu)}.
\end{equation*}
The first two brackets are uniformly $o_p(1)$ by \Cref{thm:value_estimation}. Rearranging yields the uniform claim. The selected-model statement follows by setting $\tb=\thatb$, since $\thatb\in\Ts^K$.
\end{proof}

\subsection{Error of uncorrected same-sample DR}
\label{app:same-sample-empirical-process}

We first recall the empirical-process condition used in classical analyses of DR estimators. We then examine whether this condition holds over the neural-network class considered here and derive the estimation error along the fitted gradient-descent trajectory. Let \(P_{m,d}\) be the empirical measure of the training sample. For a function that depends on the fitted model, \(P_d\) denotes expectation over an independent observation \(Q=(X,A,R)\), conditional on \(\mathcal F_{\mathrm{tr}}\). Write \(\mathbb G_{m,d}=\sqrt m(P_{m,d}-P_d)\). Consider the DR estimator obtained by substituting the fitted networks \(g_\tb\) into \(\psi_g\) and averaging over their training sample, without the Onsager correction:
\begin{equation}
  \widehat{\Vs}_{\DR}^{\mathrm{nv}}(\tb)
  \coloneqq
  \frac1m\sum_{i=1}^m\psi_{g_\tb}(Q_i).
  \label{eq:naive-dr-definition}
\end{equation}
Define \(H_{g,d}(Q)\coloneqq\psi_g(Q)-\psi_{g^\ast}(Q)\). Known context-independent logging gives \(P_dH_{g,d}=0\) for every realized integrable \(\mathcal F_{\mathrm{tr}}\)-measurable \(g\). Consequently, the exact decomposition
\begin{equation*}
  \sqrt m\{
    \widehat{\Vs}_{\DR}^{\mathrm{nv}}(\tb)-\Vs(\nu)
  \}
  =
  \mathbb G_{m,d}\{\psi_{g^\ast}-\Vs(\nu)\}
  +\mathbb G_{m,d}H_{g_\tb,d}
\end{equation*}
holds for the training-measurable fitted model \(g_\tb\).

\begin{remark}[Donsker conditions and cross-fitting]
When the data distribution \(P\) is fixed, asymptotic stochastic equicontinuity yields \(\mathbb G_mH_{\widehat g}=o_p(1)\) if, with probability tending to one, \(H_{\widehat g}\) belongs to a fixed \(P\)-Donsker class and \(\|H_{\widehat g}\|_{P,2}=o_p(1)\). Cross-fitting evaluates each fitted reward model on observations excluded from its training sample. For each fold, conditional on its fitted model, the evaluation observations are independent with law \(P\). With a fixed number of folds, conditional Chebyshev's inequality therefore makes the cross-fitted empirical-process term \(o_p(1)\) when each fitted score error converges to zero in \(L^2(P)\), without a Donsker condition \citep{chernozhukov2018double}. A fixed iterate grid gives finitely many models for each realized training sample. It does not put those models in a fixed Donsker class: as the data vary, each iterate ranges over a class whose input dimension grows with \(d\).
\end{remark}

The next proposition yields a sequence of classes whose functions converge uniformly to zero in \(L^2(P_d)\), but whose empirical-process supremum does not converge to zero. Thus, local stochastic equicontinuity fails over the full class of permitted predictors. This does not determine the error of the particular predictors produced by gradient descent, which we analyze separately below.

Consider the admissible submodel $g^\ast=0$ and $\nu=\mu$. For a fixed arm $a\in[K]$, define
\begin{align*}
 g_{\theta,d}(x,b)
 &\coloneqq\mathbf 1\{b=a\}\theta^\top x,\qquad
 \|\theta\|_2\le d^{-1/2},\\
 H_{\theta,d}(Q)
 &\coloneqq\psi_{g_{\theta,d}}(Q)-\psi_{g^\ast}(Q)
 =\{\mu(a)-\mathbf 1(A=a)\}\theta^\top X,
\end{align*}
and let $\mathcal H_d\coloneqq\{H_{\theta,d}:\|\theta\|_2\le d^{-1/2}\}$.

\begin{samepage}
\begin{proposition}[Failure of local stochastic equicontinuity]
\label{prop:local-equicontinuity-failure}
Under the Gaussian design and known logging conditions in Assumptions~\ref{ass:single-index-logging} and~\ref{ass:no-empty-arms}, with fixed $K\ge2$ and $m/d\to\phi\in(0,\infty)$, the class $\mathcal H_d$ satisfies
\begin{align*}
 \sup_{H\in\mathcal H_d}\|H\|_{P_d,2}
 &=\sqrt{\frac{\mu(a)\{1-\mu(a)\}}{d}}
 \longrightarrow0,\\
 \sup_{H\in\mathcal H_d}|\mathbb G_{m,d}H|
 &\xrightarrow{p}\sqrt{\mu(a)\{1-\mu(a)\}}.
\end{align*}
\end{proposition}
\end{samepage}

\begin{proof}
Independence of \(A\) and \(X\) gives
\[
  \|H_{\theta,d}\|_{P_d,2}^2
  =
  \mu(a)\{1-\mu(a)\}\|\theta\|_2^2,
\]
which proves the first claim. Since \(P_dH_{\theta,d}=0\), Euclidean duality gives
\[
  \sup_{H\in\mathcal H_d}|\mathbb G_{m,d}H|
  =
  \frac1{\sqrt d}
  \left\|
    \frac1{\sqrt m}\sum_{i=1}^m
    \{\mu(a)-\mathbf 1(A_i=a)\}X_i
  \right\|_2.
\]
Conditional on \(A_{1:m}\), the vector inside the norm has distribution \(N(0,\tau_m^2I_d)\), where
\[
  \tau_m^2
  =
  \frac1m\sum_{i=1}^m
  \{\mu(a)-\mathbf 1(A_i=a)\}^2
  \xrightarrow{p}
  \mu(a)\{1-\mu(a)\}.
\]
The result follows from \(\|Z_d\|_2/\sqrt d\xrightarrow{p}1\) for \(Z_d\sim N(0,I_d)\).
\end{proof}

\begin{remark}
The predictors in \Cref{prop:local-equicontinuity-failure} are represented by the permitted network with depth two, width one, and identity activation: take \(W_1=\theta\) and \(W_2=1\). The architecture and scaling assumptions therefore do not imply local stochastic equicontinuity over the entire predictor class.
\end{remark}

We now analyze the DR estimator formed from the fitted gradient-descent iterates. Order the arm-\(a\) observations as \(\Is_a=\{i_{a,1},\ldots,i_{a,N_a}\}\), and define the weighted average of the uncorrected predictions on that arm's training observations by
\begin{equation*}
  \widetilde M_{a,\diag}^{(t)}
  \coloneqq
  \frac1{N_a}\sum_{i\in\Is_a}
  w_a(X_i)
  G_{W_{a,[2:L]}^{(t)}}(X_i^\top W_{a,1}^{(t)}).
\end{equation*}
Write $V_a^{(t)}\coloneqq(V_{a,\alpha}^{(t)})_{\alpha\in[2:L]}$ for the deeper-layer quantities in the state evolution, and define the state-evolution approximation to this training-sample average by
\begin{align}
  M_{a,d}^{\mathrm{tr},(t)}
  \coloneqq
  \frac1{N_a}\sum_{k=1}^{N_a}
  \E_a^{(0)}\Big[
    &h_a(\mathfrak U_{\Upsilon,a})
    G_{V_a^{(t)}}\bigl(
      \Theta_{a;k}^{(t+1)}
      (\mathfrak U_a^{([0:t+1])})
    \bigr)
  \Big],
  \label{eq:raw-diagonal-state-evolution-target}
\end{align}
and let
\begin{equation}
  D_{a,d}(t)
  \coloneqq
  M_{a,d}^{\mathrm{tr},(t)}-M_a^{(t)},
  \qquad
  \mathfrak b_d(\tb)
  \coloneqq
  -\sum_{a=1}^K
  \mu(a)\{1-\mu(a)\}D_{a,d}(t_a).
  \label{eq:naive-dr-state-evolution-gap}
\end{equation}
Here
\[
  M_a^{(t)}
  =
  \E\left[
    w_a(Z)f_{W_a^{(t)}}(Z)
    \mymid\Ds_m,W_a^{(0)}
  \right]
\]
is the expected weighted prediction at an independent context, conditional on the fitted model. Thus, \(D_{a,d}(t)\) is the difference between the state-evolution approximation to the training-sample average and the corresponding conditional expectation at an independent context. Both quantities can depend on \(d\) and the training data; neither is assumed to have a limit. The weighted sum \(\mathfrak b_d(\tb)\) will describe the leading estimation error of DR on the training sample.

\begin{lemma}[Uncorrected predictions on training observations]
\label{lem:raw-diagonal-comparison}
Under the conditions of Section~\ref{sec:formal-regime}, with fixed $K,L,q,\Ts$, for every arm $a\in[K]$, as $d\to\infty$,
\[
  \sup_{t\in\Ts}
  \left|
    \widetilde M_{a,\diag}^{(t)}
    -M_{a,d}^{\mathrm{tr},(t)}
  \right|
  \xrightarrow{p}0.
\]
\end{lemma}

\begin{proof}
Fix \(a\) and condition as in \Cref{prop:supp:arm-wise_conditioning}. In the local arm-wise notation used in the comparison proof, the uncorrected first-layer state is $\hat u_k^{(t+1)}=X_{i_{a,k}}^\top W_{a,1}^{(t)}$. The gradient-descent comparison in \Cref{lem:gd-auxiliary-coefficient-comparison}, at horizon \(t+1\), gives
\[
  N_a^{-1/2}
  \|\hat u^{(t+1)}-u^{(t+1)}\|_F=o_p(1),
  \qquad
  \max_{\alpha\in[2:L]}
  \|W_{a,\alpha}^{(t)}-V_{a,\alpha}^{(t)}\|_\infty=o_p(1),
\]
with the stated conditional polynomial-moment rates. By \HIprop{prop:apr_est_det_fcn}-(1)(2), there is a random \(\mathcal P_{a,t}\), whose fixed conditional moments are bounded by a polynomial in \(\varkappa_{\ast,a}\log d\), such that
\begin{align*}
 N_a^{-1/2}
 \left\|
 G_{W_{a,[2:L]}^{(t)}}(\hat u^{(t+1)})
 -G_{V_a^{(t)}}(u^{(t+1)})
 \right\|_2 \le
 \mathcal P_{a,t}
 \left\{
 N_a^{-1/2}\|\hat u^{(t+1)}-u^{(t+1)}\|_F
 +\max_{\alpha\in[2:L]}
 \|W_{a,\alpha}^{(t)}-V_{a,\alpha}^{(t)}\|_\infty
 \right\}.
\end{align*}
Since $|h_a|\le B_h$, Cauchy--Schwarz yields
\begin{align*}
 &\left|\frac1{N_a}\sum_{k=1}^{N_a}
 h_a(\Upsilon^\top X_{i_{a,k}})
 \left\{G_{W_{a,[2:L]}^{(t)}}(\hat u_k^{(t+1)})
       -G_{V_a^{(t)}}(u_k^{(t+1)})\right\}\right|\\
 &\quad\le\frac{B_h}{\sqrt{N_a}}
 \left\|G_{W_{a,[2:L]}^{(t)}}(\hat u^{(t+1)})
       -G_{V_a^{(t)}}(u^{(t+1)})\right\|_2
 =o_p(1).
\end{align*}

By the bounded Lipschitz property of \(h_a\) and the forward-map derivative bounds, the test function
\[
  \vartheta_{a,t}(z^{([0:t+1])})
  =
  h_a\bigl((z^{(0)})^{\mathrm{idx}}\bigr)
  G_{V_a^{(t)}}\bigl((z^{(t+1)})^{\mathrm{aux}}\bigr)
\]
is, conditionally on \(\mathcal G_a\), pseudo-Lipschitz of order two, with pseudo-Lipschitz constant bounded by
\[
  C_t\left\{1+\max_{\alpha\in[2:L]}
  \|V_{a,\alpha}^{(t)}\|_{\mathrm{op}}\right\}^{C_t}.
\]
After conditioning on $\mathcal G_a$, this constant is deterministic. Apply item~\ref{item:supp:s2_pl} of \Cref{prop:supp:auxiliary_gfom_policy_index} at horizon \(t+1\) with this constant. The resulting bound is polynomial in the displayed constant. The conditional moment bounds in \Cref{lem:arm-wise-hi25-consequences}, together with \eqref{eq:envelope-polynomial-vanishing}, therefore give the comparison with the target in \eqref{eq:raw-diagonal-state-evolution-target}:
\[
  \frac1{N_a}\sum_{k=1}^{N_a}
  h_a(\Upsilon^\top X_{i_{a,k}})
  G_{V_a^{(t)}}(u_k^{(t+1)})
  -M_{a,d}^{\mathrm{tr},(t)}
  =o_p(1).
\]
Combining this with the preceding comparison between the fitted and auxiliary predictions, and taking a finite union over \(t\in\Ts\), proves the claim.
\end{proof}

\begin{proposition}[Error of the DR estimator without sample splitting]
\label{prop:naive-dr-state-evolution-bias}
Assume the conditions of Section~\ref{sec:formal-regime}, with fixed $K,L,q,\Ts$. Let $\mathfrak b_d$ be defined by \eqref{eq:naive-dr-state-evolution-gap}. For every $\tb\in\Ts^K$, the estimators in \eqref{eq:naive-dr-definition} and \eqref{eq:adr} satisfy the exact identity
\begin{align}
  \widehat{\Vs}_{\DR}^{\mathrm{nv}}(\tb)
  -\widehat{\Vs}_{\ADR}(\tb)
  =
  -\sum_{a=1}^K\frac{N_a}{m}\{1-\mu(a)\}
  \left[
    \widetilde M_{a,\diag}^{(t_a)}
    -\widehat M_{a,\diag}^{(t_a)}
  \right].
  \label{eq:naive-minus-adr-exact}
\end{align}
Moreover,
\begin{equation}
  \sup_{\tb\in\Ts^K}
  \left|
    \widehat{\Vs}_{\DR}^{\mathrm{nv}}(\tb)
    -\Vs(\nu)-\mathfrak b_d(\tb)
  \right|
  \xrightarrow{p}0.
  \label{eq:naive-dr-bias-characterization}
\end{equation}
Consequently, for any \(\mathcal F_{\mathrm{tr}}\)-measurable choice \(\widehat\tb_d\in\Ts^K\), the estimator \(\widehat{\Vs}_{\DR}^{\mathrm{nv}}(\widehat\tb_d)\) is consistent for \(\Vs(\nu)\) if and only if $\mathfrak b_d(\widehat\tb_d)\xrightarrow{p}0$. In particular, it is inconsistent whenever there is an \(\epsilon>0\) such that
\[
  \limsup_{d\to\infty}
  \Prob\left(
    |\mathfrak b_d(\widehat\tb_d)|>\epsilon
  \right)>0.
\]
\end{proposition}

\begin{proof}
For each observation \(i\in\Is_a\), \(\widehat{\Vs}_{\DR}^{\mathrm{nv}}\) and \(\widehat{\Vs}_{\ADR}\) use the same reward and the same predictions for arms \(b\ne a\). Their summands differ by \(-\{1-\mu(a)\}w_a(X_i)\) times the difference between the uncorrected and corrected predictions for arm \(a\). Summing these differences gives \eqref{eq:naive-minus-adr-exact}.

By \Cref{lem:raw-diagonal-comparison,prop:diagonal-comparison}, uniformly on the fixed grid,
\[
  \widetilde M_{a,\diag}^{(t)}
  -\widehat M_{a,\diag}^{(t)}
  =
  D_{a,d}(t)+o_p(1).
\]
For the average of the uncorrected predictions, boundedness of $h_a$ and Cauchy--Schwarz give
\[
  |\widetilde M_{a,\diag}^{(t)}|
  \le
  B_h\left\{
    \frac1{N_a}\sum_{i\in\Is_a}
    |f_{W_a^{(t)}}(X_i)|^2
  \right\}^{1/2}.
\]
The deterministic forward-map bounds and the normalized GD-state moment bounds make the right-hand side $d^{o_p(1)}$, uniformly on the fixed grid. The same argument with $\widehat U_a^{(t)}$ and \Cref{lem:arm-wise-hi25-consequences} gives $\sup_t|\widehat M_{a,\diag}^{(t)}|=d^{o_p(1)}$. Thus, the error from replacing $N_a/m$ by $\mu(a)$ in \eqref{eq:naive-minus-adr-exact} is bounded by
\begin{align*}
 &\sup_{\tb\in\Ts^K}\left|
   \sum_{a=1}^K\left\{\frac{N_a}{m}-\mu(a)\right\}(1-\mu(a))
    \{\widetilde M_{a,\diag}^{(t_a)}-\widehat M_{a,\diag}^{(t_a)}\}
 \right|\\
 &\quad\le K\max_a|N_a/m-\mu(a)|\,
   \max_{a,t}\{|\widetilde M_{a,\diag}^{(t)}|
                +|\widehat M_{a,\diag}^{(t)}|\}\\
 &\quad=O_p(d^{-1/2})\,d^{o_p(1)}=o_p(1).
\end{align*}
Combining this bound with the preceding uniform comparison gives
\[
  \sup_{\tb\in\Ts^K}
  \left|
    \widehat{\Vs}_{\DR}^{\mathrm{nv}}(\tb)
    -\widehat{\Vs}_{\ADR}(\tb)
    -\mathfrak b_d(\tb)
  \right|
  \xrightarrow{p}0.
\]
The ADR part of \Cref{thm:value_estimation} proves \eqref{eq:naive-dr-bias-characterization}. Evaluating the uniform statement at \(\widehat\tb_d\) proves the final claim.
\end{proof}

The condition concerns the sum over arms, so it allows cancellation between the arm-wise contributions.

\begin{corollary}[Non-tightness of the empirical-process term]
\label{cor:naive-score-nontight}
Assume the conditions of \Cref{prop:naive-dr-state-evolution-bias}. Let \(\widehat\tb_d\in\Ts^K\) be \(\mathcal F_{\mathrm{tr}}\)-measurable and set \(H_{\widehat\tb_d,d}=\psi_{g_{\widehat\tb_d}}-\psi_{g^\ast}\). Then
\begin{equation}
  \frac1{\sqrt m}
  \mathbb G_{m,d}H_{\widehat\tb_d,d}
  =
  \mathfrak b_d(\widehat\tb_d)+o_p(1).
  \label{eq:fitted-score-bias-scale}
\end{equation}
If the separation condition in \Cref{prop:naive-dr-state-evolution-bias} holds, then \(\mathbb G_{m,d}H_{\widehat\tb_d,d}\) is not tight. In particular, no classes \(\mathcal H_d\) can both contain \(H_{\widehat\tb_d,d}\) with probability tending to one and satisfy
\[
  \sup_{H\in\mathcal H_d}
  |\mathbb G_{m,d}H|=O_p(1)
\]
uniformly in \(d\).
\end{corollary}

\begin{proof}
Pathwise, \(P_dH_{\widehat\tb_d,d}=0\), so
\begin{align*}
  \frac1{\sqrt m}\mathbb G_{m,d}H_{\widehat\tb_d,d}
  &=P_{m,d}H_{\widehat\tb_d,d}\\*
  &=\widehat{\Vs}_{\DR}^{\mathrm{nv}}(\widehat\tb_d)
    -P_{m,d}\psi_{g^\ast}.
\end{align*}
By \Cref{prop:naive-dr-state-evolution-bias}, \(\widehat{\Vs}_{\DR}^{\mathrm{nv}}(\widehat\tb_d)-\Vs(\nu)\) is \(\mathfrak b_d(\widehat\tb_d)+o_p(1)\). To control the empirical average that uses $g^\ast$, work on $E_{\mathrm{cnt}}$ and write
\begin{align*}
 P_{m,d}\psi_{g^\ast}-\Vs(\nu)
 &=\frac1m\sum_{i=1}^m
   \left\{\sum_{a=1}^K\nu(a\mymid X_i)g^\ast(X_i,a)-\Vs(\nu)\right\} +\sum_{a=1}^K\frac{N_a}{m}
   \frac1{N_a}\sum_{i\in\Is_a}w_a(X_i)\varepsilon_i.
\end{align*}

The first term is $O_p(d^{-1/2})$ by the uniformly bounded signal second moment. Each weighted noise average in the second term is $o_p(1)$ by the reward-average argument in \Cref{lem:value-sample-averages}. Since $\Prob(E_{\mathrm{cnt}}^c)\to0$, it follows that $P_{m,d}\psi_{g^\ast}-\Vs(\nu)=o_p(1)$, proving \eqref{eq:fitted-score-bias-scale}. Under the separation condition, there exist $\epsilon,\delta>0$ and a subsequence $d_j$ such that
\begin{equation*}
 \Prob(|\mathfrak b_{d_j}(\widehat\tb_{d_j})|>\epsilon)>\delta.
\end{equation*}
Let
\begin{equation*}
 r_d\coloneqq m^{-1/2}\mathbb G_{m,d}H_{\widehat\tb_d,d}
                -\mathfrak b_d(\widehat\tb_d)=o_p(1).
\end{equation*}
For every fixed $M>0$, eventually $M<\epsilon\sqrt m/2$ along this subsequence, and
\begin{align*}
 \Prob\left(|\mathbb G_{m,d_j}H_{\widehat\tb_{d_j},d_j}|>M\right)
 &\ge\Prob\left(
   |\mathfrak b_{d_j}(\widehat\tb_{d_j})|>\epsilon,
   |r_{d_j}|\le\epsilon/2\right)\\
 &\ge\delta-\Prob(|r_{d_j}|>\epsilon/2)
 =\delta-o(1).
\end{align*}
This contradicts tightness. If a class $\mathcal H_d$ contains the fitted score error with probability tending to one, then
\begin{align*}
 \Prob\left(|\mathbb G_{m,d}H_{\widehat\tb_d,d}|>M\right)
 &\le\Prob(H_{\widehat\tb_d,d}\notin\mathcal H_d) +\Prob\left(\sup_{H\in\mathcal H_d}|\mathbb G_{m,d}H|>M\right).
\end{align*}
Thus, the stated uniform class bound would imply tightness and cannot hold.
\end{proof}

The preceding result applies when the DR estimation error does not converge to zero. The following example satisfies this condition after a single gradient-descent update.

\proponestepnaivebias*

\begin{proof}
Work on $E_{\mathrm{cnt}}$; its complement has probability tending to zero. Since $f_{W_a^{(0)}}=0$, $W_{a,1}^{(0)}=0$, and $W_{a,2}^{(0)}=1$, the two gradients are
\begin{align*}
 \nabla_{W_{a,1}}\mathcal L_a(W_a^{(0)})
 &=\frac1{N_a}\sum_{i\in\Is_a}(0-c)X_iW_{a,2}^{(0)}
 =-c\overline X_a,\\
 \nabla_{W_{a,2}}\mathcal L_a(W_a^{(0)})
 &=\frac1{N_a}\sum_{i\in\Is_a}(0-c)X_i^\top W_{a,1}^{(0)}
 =0.
\end{align*}
The simultaneous update therefore gives
\begin{equation*}
 W_{a,1}^{(1)}=\eta c\overline X_a,
 \qquad W_{a,2}^{(1)}=1,
 \qquad g_{a,1}(x)=\eta c\,x^\top\overline X_a.
\end{equation*}
Since \(w_a=1\) and \(\Vs(\nu)=c\), direct substitution into \eqref{eq:naive-dr-definition} gives
\[
  \widehat{\Vs}_{\DR}^{\mathrm{nv}}-c
  =
  \eta c
  \left[
    \sum_{a=1}^K
    \mu(a)\overline X^\top\overline X_a
    -
    \sum_{a=1}^K
    \frac{N_a}{m}\|\overline X_a\|_2^2
  \right],
  \qquad
  \overline X=\frac1m\sum_{i=1}^mX_i.
\]
Conditional on the action labels,
\[
  \frac{N_a}{m}\|\overline X_a\|_2^2
  \overset{d}{=}
  \frac{\chi_{d,a}^2}{m}
  \xrightarrow{p}\frac1\phi.
\]
Also,
\[
  \overline X^\top\overline X_a
  =
  \frac{N_a}{m}\|\overline X_a\|_2^2
  +
  \sum_{b\ne a}
  \frac{N_b}{m}\overline X_b^\top\overline X_a
  \xrightarrow{p}\frac1\phi,
\]
because, for $a\ne b$, conditional independence gives
\[
  \E[\overline X_b^\top\overline X_a\mid A_{1:m}]=0,
  \qquad
  \E[(\overline X_b^\top\overline X_a)^2\mid A_{1:m}]
  =\frac{d}{N_aN_b}=O(d^{-1})
\]
on $E_{\mathrm{cnt}}$. Thus, every cross-arm inner product is $o_p(1)$. The two sums in the expression for $\widehat{\Vs}_{\DR}^{\mathrm{nv}}-c$ therefore converge to \(1/\phi\) and \(K/\phi\), respectively. Since $\Prob(E_{\mathrm{cnt}}^c)\to0$, the claimed limit follows. The example uses a constant smooth reward link, zero noise, constant importance weights, a smooth identity activation, and deterministic sample-independent initialization, so it satisfies the working assumptions. By \Cref{prop:naive-dr-state-evolution-bias}, this limit also implies
\[
  \mathfrak b_d(1,\ldots,1)
  \xrightarrow{p}
  -\frac{\eta c(K-1)}{\phi},
\]
so the separation condition holds.
\end{proof}

\begin{remark}[Asymptotic linearity and inference]
Define the DR estimator that uses the true reward function by
\[
  \widehat{\Vs}_{\mathrm{or}}
  =\frac1m\sum_{i=1}^m\psi_{g^\ast}(Q_i).
\]
Let $\Delta_{i,a}=\widehat g_\tb(X_i,a)-g^\ast(X_i,a)$ be the adjusted prediction error for observation $i$ and arm $a$. Then
\[
  \sqrt m\{
    \widehat{\Vs}_{\ADR}(\tb)-\widehat{\Vs}_{\mathrm{or}}
  \}
  =\frac1{\sqrt m}\sum_{i=1}^m
  \left[
    \sum_{a=1}^K\nu(a\mid X_i)\Delta_{i,a}
    -w_{A_i}(X_i)\Delta_{i,A_i}
  \right].
\]
To obtain the asymptotic linear expansion
\[
  \sqrt m\{\widehat{\Vs}_{\ADR}(\widehat\tb)-\Vs(\nu)\}
  =\frac1{\sqrt m}\sum_{i=1}^m
    \{\psi_{g^\ast}(Q_i)-\Vs(\nu)\}+o_p(1),
\]
the preceding prediction-error sum, evaluated at $\tb=\widehat\tb$, must be $o_p(1)$. The consistency results prove only that it is $o_p(\sqrt m)$.

Sufficient conditions for the expansion are
\[
  \sqrt m\{\widehat{\Vs}_{\ADR}(\widehat\tb)
             -\widehat{\Vs}_{\mathrm{or}}\}
  -\frac1{\sqrt m}\sum_{i=1}^m H_{g_{\widehat\tb},d}(Q_i')
  =o_p(1),
\]
together with $\mathcal J_{\mathrm{var}}(g_{\widehat\tb})=o_p(1)$. Here $\mathcal J_{\mathrm{var}}$ is the conditional variance of the score error in \Cref{prop:adr-proxy-main}, and $Q_1',\ldots,Q_m'$ are i.i.d. observations from the logged-data law, independent of $\mathcal F_{\mathrm{tr}}$. Write the independent sum as
\begin{equation*}
 S_m'\coloneqq\frac1{\sqrt m}\sum_{i=1}^m
 H_{g_{\widehat\tb},d}(Q_i').
\end{equation*}
Conditional independence and \Cref{prop:adr-proxy-main} give
\begin{align*}
 \E[S_m'\mymid\mathcal F_{\mathrm{tr}}]&=0,\\
 \operatorname{Var}(S_m'\mymid\mathcal F_{\mathrm{tr}})
 &=\frac1m\sum_{i=1}^m
   \operatorname{Var}(H_{g_{\widehat\tb},d}(Q_i')
                      \mymid\mathcal F_{\mathrm{tr}})\\
 &=\mathcal J_{\mathrm{var}}(g_{\widehat\tb}).
\end{align*}
Consequently, for every $\epsilon>0$,
\begin{equation*}
 \Prob(|S_m'|>\epsilon)
 \le\E\left[\min\left\{1,
   \epsilon^{-2}\mathcal J_{\mathrm{var}}(g_{\widehat\tb})
 \right\}\right]\longrightarrow0
\end{equation*}
when $\mathcal J_{\mathrm{var}}(g_{\widehat\tb})=o_p(1)$. The assumed comparison then gives $\sqrt m(\widehat\Vs_{\ADR}(\widehat\tb)-\widehat\Vs_{\mathrm{or}}) =S_m'+o_p(1)=o_p(1)$. By the selection guarantee and \Cref{prop:adr-proxy-main}, the variance condition follows if
\[
  \min_{\tb\in\Ts^K}\Rs_\nu^\circ(g_\tb)=o_p(1).
\]
Our assumptions permit prediction error that does not converge to zero, and the comparisons proved above do not establish the required $o_p(1)$ remainder at the $\sqrt m$ scale.

In the nonparametric model with known logging probabilities and unrestricted context and reward distributions, $\psi_{g^\ast}-\Vs(\nu)$ is the efficient influence function \citep{kennedy2024semiparametric}. The Gaussian single-index model considered here imposes additional restrictions, so its efficiency bound requires a separate calculation. The notation $\sigma_{\mathrm{eff}}^2$ in \Cref{cor:independent-score-variance} denotes the variance of $\psi_{g^\ast}$.

The state-evolution results in this paper assume that the arm-wise aspect ratios stay bounded above and away from zero, with fixed network width and training horizon. Extending these results to $m/d\to\infty$, growing width, or a growing training horizon requires new bounds that are uniform in the corresponding parameters.
\end{remark}

\subsection{Variance of the DR score on an independent observation}
\label{app:DR-var-control}

We evaluate $\psi_g(X,A,R)$ on an observation independent of the sample used to fit $g$. Conditional on $\mathcal F_{\mathrm{tr}}$, this score has mean $\mathcal V(\nu)$ under the known logging policy. The next proposition decomposes its conditional variance and bounds the part that depends on $g$ by the target-policy-weighted prediction risk. To obtain the sampling variance of ADR, one must also account for covariances between the adjusted terms evaluated on different training observations.

For an independent observation $(X,A,R)$ and a reward model $g$, define
\begin{align*}
 v_{\mathrm{intrinsic}}(X)
 &\coloneqq\sum_{a=1}^K
   \frac{\nu(a\mymid X)^2}{\mu(a)}\sigma_{\xi,a}^2,\\
 V_{\mathrm{model}}(X,g)
 &\coloneqq\sum_{a=1}^K\frac{\nu(a\mymid X)^2}{\mu(a)}
   \{g(X,a)-g^\ast(X,a)\}^2 -\left\{\sum_{a=1}^K\nu(a\mymid X)
   \{g(X,a)-g^\ast(X,a)\}\right\}^2,\\
 V_{\mathrm{intrinsic}}
 &\coloneqq\operatorname{Var}\{\psi_{g^\ast}(X,A,R)\},\\
 \mathcal J_{\mathrm{var}}(g)
 &\coloneqq\E[V_{\mathrm{model}}(X,g)\mymid\mathcal F_{\mathrm{tr}}],\\
 \mathcal J_{\DR}(g)
 &\coloneqq\E\left[
   \sum_{a=1}^K\nu(a\mymid X)\frac{1-\mu(a)}{\mu(a)}
   \{g(X,a)-g^\ast(X,a)\}^2
   \mymid\mathcal F_{\mathrm{tr}}\right].
\end{align*}
For deterministic $g$, the conditioning in these definitions can be omitted. Also set
\begin{equation*}
 c_-(\mu)\coloneqq\min_{a\in[K]}\frac{1-\mu(a)}{\mu(a)},
 \qquad
 c_+(\mu)\coloneqq\max_{a\in[K]}\frac{1-\mu(a)}{\mu(a)}.
\end{equation*}

\begin{proposition}[Excess variance of the DR score]
\label{prop:adr-proxy-main}
\label{prop:dr-proxy}
Assume $K\ge2$, known context-independent logging with $\mu(a)>0$ for every arm, and rewards $R=g^\ast(X,A)+\varepsilon$ whose noise law satisfies \eqref{eq:noise-setup}. Suppose $\E[R^2]<\infty$. Let $g$ be a deterministic measurable reward model or an $\mathcal F_{\mathrm{tr}}$-measurable fitted reward model, with $\Rs_\nu^\circ(g)<\infty$ almost surely. Let $(X,A,R)$ follow the logged-data law independently of $\mathcal F_{\mathrm{tr}}$. Then, almost surely,
\begin{align*}
 \operatorname{Var}(\psi_g\mymid X,\mathcal F_{\mathrm{tr}})
 &=v_{\mathrm{intrinsic}}(X)+V_{\mathrm{model}}(X,g),\\
 \operatorname{Var}(\psi_g\mymid\mathcal F_{\mathrm{tr}})
 &=V_{\mathrm{intrinsic}}+\mathcal J_{\mathrm{var}}(g),\\
 \operatorname{Var}(\psi_g-\psi_{g^\ast}\mymid\mathcal F_{\mathrm{tr}})
 &=\mathcal J_{\mathrm{var}}(g).
\end{align*}
Moreover,
\begin{align*}
 0\le\mathcal J_{\mathrm{var}}(g)&\le\mathcal J_{\DR}(g),\\
 c_-(\mu)\Rs_\nu^\circ(g)
 &\le\mathcal J_{\DR}(g)
 \le c_+(\mu)\Rs_\nu^\circ(g).
\end{align*}
Under the additional assumptions of \Cref{thm:selection-main}, the selector in \eqref{def:that} satisfies
\begin{equation}
 0\le\mathcal J_{\DR}(g_{\thatb})
       -\min_{\tb\in\Ts^K}\mathcal J_{\DR}(g_\tb)
 \le2\sum_{a=1}^K(1-\mu(a))\Delta_a=o_p(1),
 \label{eq:separable-dr-proxy-oracle}
\end{equation}
where $\Delta_a=\sup_{t\in\Ts} |\widehat\Es_a^{(t)}-\Es_{a,\mathrm{ind}}^{(t)}|$.
\end{proposition}

\begin{proof}[Proof of \Cref{prop:dr-proxy}]
It suffices to prove the conditional statement. The deterministic case is the same argument with a trivial $\mathcal F_{\mathrm{tr}}$. Let $(X,A,R)$ be an independent test draw and write $\Delta_a(X)\coloneqq g^\ast(X,a)-g(X,a)$, $\nu_a\coloneqq\nu(a\mymid X)$, and $\mu_a\coloneqq\mu(a)$. Because $R=g^\ast(X,A)+\varepsilon$ and $\mathbb E[\varepsilon\mymid X,A,\mathcal F_{\mathrm{tr}}]=0$,
\begin{align*}
    \mathbb E\br{
        \psi_g(X,A,R)
        \mymid
        X,\mathcal F_{\mathrm{tr}}
    }
    &=
    \sum_{a=1}^K\nu_a g(X,a)
    +\sum_{a=1}^K
    \mu_a\frac{\nu_a}{\mu_a}
    \{g^\ast(X,a)-g(X,a)\}\\*
    &=
    \sum_{a=1}^K \nu_a g^\ast(X,a),
\end{align*}
and the centered score is
\begin{align*}
    \psi_g(X,A,R)
    -
    \mathbb E\br{
        \psi_g(X,A,R)
        \mymid
        X,\mathcal F_{\mathrm{tr}}
    }
    &=
    \frac{\nu(A \mymid X)}{\mu(A)}\,\varepsilon
    \\
    &\quad+
    \br{
        \frac{\nu(A \mymid X)}{\mu(A)}\,\Delta_A(X)
        -
        \sum_{a=1}^K \nu_a\Delta_a(X)
    }.
\end{align*}
Denote the second term on the right by $Q_A(X)$. The two terms are conditionally orthogonal because
\begin{align*}
  \E\left[
    \frac{\nu(A\mymid X)}{\mu(A)}\varepsilon Q_A(X)
    \mymid X,\mathcal F_{\mathrm{tr}}
  \right] =
  \sum_{a=1}^K
  \mu_a\frac{\nu_a}{\mu_a}Q_a(X)
  \E[\varepsilon\mymid X,A=a,\mathcal F_{\mathrm{tr}}]
  =0.
\end{align*}
Moreover,
\begin{align*}
  \E\left[
    \left\{\frac{\nu(A\mymid X)}{\mu(A)}\varepsilon\right\}^2
    \mymid X,\mathcal F_{\mathrm{tr}}
  \right]
  &=
  \sum_{a=1}^K\frac{\nu_a^2}{\mu_a}\sigma_{\xi,a}^2.
\end{align*}
Therefore,
\begin{align*}
    \operatorname{Var}\p{
        \psi_g(X,A,R)
        \mymid
        X,\mathcal F_{\mathrm{tr}}
    }
    &=
    \sum_{a=1}^K \frac{\nu_a^2}{\mu_a}\sigma_{\xi,a}^2
    +
    \operatorname{Var}_{A\sim\mu(\cdot)}\!\left(
        \frac{\nu(A \mymid X)}{\mu(A)}\,\Delta_A(X)
        \mymid
        X,\mathcal F_{\mathrm{tr}}
    \right).
\end{align*}
The model-dependent term is
\begin{align*}
 \operatorname{Var}_{A\sim\mu(\cdot)}\left(
    \frac{\nu(A\mymid X)}{\mu(A)}\Delta_A(X)
    \mymid X,\mathcal F_{\mathrm{tr}}\right)
 &=\sum_{a=1}^K\mu_a
    \left\{\frac{\nu_a}{\mu_a}\Delta_a(X)\right\}^2
       -\left\{\sum_{a=1}^K\nu_a\Delta_a(X)\right\}^2\\*
 &=\sum_{a=1}^K\frac{\nu_a^2}{\mu_a}\Delta_a(X)^2
       -\left\{\sum_{a=1}^K\nu_a\Delta_a(X)\right\}^2\\*
 &=V_{\mathrm{model}}(X,g).
\end{align*}
Applying the law of total variance conditionally on $\mathcal F_{\mathrm{tr}}$ yields
\begin{align*}
    \operatorname{Var}\p{
        \psi_g(X,A,R)
        \mymid
        \mathcal F_{\mathrm{tr}}
    }
    &=
    \mathbb E\br{
        \sum_{a=1}^K \frac{\nu_a^2}{\mu_a}\sigma_{\xi,a}^2
        \mymid
        \mathcal F_{\mathrm{tr}}
    }
    +
    \operatorname{Var}\p{
        \sum_{a=1}^K \nu_a g^\ast(X,a)
        \mymid
        \mathcal F_{\mathrm{tr}}
    }
    +
    \mathcal J_{\mathrm{var}}(g).
\end{align*}
The first two terms are constant in $g$. Setting $g=g^\ast$ identifies their sum with $\operatorname{Var}(\psi_{g^\ast}\mymid\mathcal F_{\mathrm{tr}})$, which equals $V_{\mathrm{intrinsic}}$ because the test draw is independent of $\mathcal F_{\mathrm{tr}}$. In addition,
\begin{align*}
 \psi_g(X,A,R)-\psi_{g^\ast}(X,A,R)&=Q_A(X),\\
 \E[Q_A(X)\mymid X,\mathcal F_{\mathrm{tr}}]&=0,\\
 \operatorname{Var}(\psi_g-\psi_{g^\ast}\mymid\mathcal F_{\mathrm{tr}})
 &=\E[\operatorname{Var}(Q_A(X)\mymid X,\mathcal F_{\mathrm{tr}})
       \mymid\mathcal F_{\mathrm{tr}}]\\
 &=\mathcal J_{\mathrm{var}}(g).
\end{align*}
This proves the third variance identity.

It remains to bound the model-dependent term. Conditionally on $(X,\mathcal F_{\mathrm{tr}})$, let $B$ have law $\nu(\cdot\mymid X)$ independently of $A$, and set
\begin{equation*}
  Y\coloneqq
  \left\{\frac{\mathbf 1_{\{A=B\}}}{\mu(B)}-1\right\}\Delta_B(X).
\end{equation*}
Then
\begin{align*}
    \mathbb E\br{
        Y
        \mymid
        A,X,\mathcal F_{\mathrm{tr}}
    }
    &=
    \sum_{b=1}^K\nu_b
    \left\{\frac{\mathbf 1_{\{A=b\}}}{\mu_b}-1\right\}
    \Delta_b(X)\\
    &=
    \frac{\nu(A \mymid X)}{\mu(A)}\,\Delta_A(X)
    -
    \sum_{a=1}^K \nu_a\Delta_a(X),
\end{align*}
so
\begin{equation*}
  \operatorname{Var}\left(
    \E[Y\mymid A,X,\mathcal F_{\mathrm{tr}}]
    \mymid X,\mathcal F_{\mathrm{tr}}
  \right)
  =V_{\mathrm{model}}(X,g).
\end{equation*}
On the other hand, conditional on $B=b$,
\begin{align*}
  \E[Y\mymid B=b,X,\mathcal F_{\mathrm{tr}}]
  &=
  \left\{
    \mu_b\left(\frac1{\mu_b}-1\right)
    -(1-\mu_b)
  \right\}\Delta_b(X)
  =0,\\
  \E[Y^2\mymid B=b,X,\mathcal F_{\mathrm{tr}}]
  &=
  \left\{
    \mu_b\left(\frac1{\mu_b}-1\right)^2
    +(1-\mu_b)
  \right\}\Delta_b(X)^2\\
  &=
  \frac{1-\mu_b}{\mu_b}\Delta_b(X)^2.
\end{align*}
Therefore
\begin{align*}
    \operatorname{Var}\p{
      Y\mymid X,\mathcal F_{\mathrm{tr}}
    }
    &=
    \sum_{a=1}^K
    \nu_a
    \frac{1-\mu(a)}{\mu(a)}\,
    \Delta_a(X)^2.
\end{align*}
Finally, applying the law of total variance conditionally on $(X,\mathcal F_{\mathrm{tr}})$ gives
\begin{align*}
  \operatorname{Var}(Y\mymid X,\mathcal F_{\mathrm{tr}})
  &=
  \E\left[
    \operatorname{Var}(Y\mymid A,X,\mathcal F_{\mathrm{tr}})
    \mymid X,\mathcal F_{\mathrm{tr}}
  \right]\\
  &\quad+
  \operatorname{Var}\left(
    \E[Y\mymid A,X,\mathcal F_{\mathrm{tr}}]
    \mymid X,\mathcal F_{\mathrm{tr}}
  \right)\\
  &\ge V_{\mathrm{model}}(X,g).
\end{align*}
Averaging over $X$ proves $0\le\mathcal J_{\mathrm{var}}(g)\le\mathcal J_{\mathrm{DR}}(g)$.

Because $K\ge2$ and $\mu(a)>0$ for every arm, each $\mu(a)<1$. Hence, $c_-(\mu)>0$. For each fixed arm $a$,
\begin{equation*}
    c_-(\mu)\,\nu(a \mymid X)
    \le
    \nu(a \mymid X)\frac{1-\mu(a)}{\mu(a)}
    \le
    c_+(\mu)\,\nu(a \mymid X).
\end{equation*}
Multiplying by $\{g(X,a)-g^\ast(X,a)\}^2$, summing over $a$, and taking conditional expectation proves the sandwich bound. Finally, \eqref{eq:finite-grid-selection} bounds each arm's excess prediction error relative to its best candidate by $2\mu(a)\Delta_a$, because its noise contribution is independent of the iterate. Since
\[
 \mathcal J_{\DR}(g_\tb)
 =\sum_{a=1}^K\frac{1-\mu(a)}{\mu(a)}
       \Rs_{\nu,a}^\circ(g_{a,t_a}),
\]
its minimum over $\Ts^K$ is the sum of the armwise minima. Multiplying the armwise error bounds by $(1-\mu(a))/\mu(a)$ and summing proves \eqref{eq:separable-dr-proxy-oracle}.
\end{proof}

Thus, the selection guarantee controls an upper bound on the excess variance of a DR score evaluated on an independent observation.

\corindscorevariance*
\begin{proof}
By \Cref{prop:adr-proxy-main},
\begin{align*}
 \operatorname{Var}(\psi_{g_{\thatb}}\mymid\mathcal F_{\mathrm{tr}})
 &=\sigma_{\mathrm{eff}}^2+\mathcal J_{\mathrm{var}}(g_{\thatb})\\
 &\le\sigma_{\mathrm{eff}}^2+\mathcal J_{\DR}(g_{\thatb})\\
 &\le\sigma_{\mathrm{eff}}^2
   +\min_{\tb\in\Ts^K}\mathcal J_{\DR}(g_\tb)+o_p(1).
\end{align*}
If $\min_{\tb\in\Ts^K}\Rs_\nu^\circ(g_\tb)=o_p(1)$, then \Cref{thm:selection-main} and the same proposition give
\begin{align*}
 0\le\mathcal J_{\mathrm{var}}(g_{\thatb})
 &\le c_+(\mu)\Rs_\nu^\circ(g_{\thatb})\\
 &\le c_+(\mu)\min_{\tb\in\Ts^K}\Rs_\nu^\circ(g_\tb)+o_p(1)
 =o_p(1).
\end{align*}
Substitution into the first variance identity proves the second claim.
\end{proof}

\section{Sharper Convergence Rates}
\label{sec:appendix-quantitative-refinements}

We now give polynomial convergence rates under the strengthened moment condition
\begin{equation}
    \max_{a \in [K]}\mathbb{E}\br{\varkappa_{\ast,a}^s}
    \le
    C_s(\log d)^{C_s}
    \qquad
    \text{for each fixed } s \ge 1.
    \label{eq:poly-envelope}
\end{equation}
Because $\varkappa_{\ast,a}$ contains $\|\xi_{\Is_a}\|_\infty$, this condition is stronger than the envelope assumption used above. It provides the polylogarithmic second- and fourth-moment bounds for the triangular-array noise laws used below.

\begin{proposition}[Polynomial rates under the strengthened envelope]
\label{prop:poly-rates}
Assume the conditions of Section~\ref{sec:formal-regime} and \eqref{eq:poly-envelope}, with fixed $K,L,q,\Ts$. There exists $\beta_{\HI}>0$, independent of $d$, such that, for every $0<\beta<\beta_0\coloneqq\min\{\beta_{\HI},1/2\}$, as $d\to\infty$, the corrected risk criterion satisfies
\begin{align*}
    \max_{a \in [K]}
    \sup_{t \in \Ts}
    \abs{\widehat{\Es}_a^{(t)} - \Es_{a,\mathrm{ind}}^{(t)}}
    &=
    O_p\p{d^{-\beta}}
\end{align*}
and the value estimators satisfy
\begin{align*}
    \sup_{\tb \in \Ts^K}
    \abs{\widehat{\Vs}_{\ADM}(\tb) - \Vs_{g_{\tb}}(\nu)}
    &=
    O_p\p{d^{-\beta}},\\
    \sup_{\tb \in \Ts^K}
    \abs{\widehat{\Vs}_{\ADR}(\tb) - \Vs(\nu)}
    &=
    O_p\p{d^{-\beta}}.
\end{align*}
\end{proposition}

\begin{proof}[Proof of \Cref{prop:poly-rates}]
Apply \Cref{prop:supp:weighted-observable-transfer} to the squared-residual and fitted-value observables. Because $K$ and $\Ts$ are fixed, these conditional first-moment bounds have a common exponent $\beta_{\HI}>0$. For either $D_{a,t}=\widehat\Es_a^{(t)}-\Es_{a,\pi}^{(t)}$ or $D_{a,t}=\widehat M_{a,\diag}^{(t)}-M_a^{(t)}$, \eqref{eq:poly-envelope} therefore gives
\begin{equation*}
 \E[\mathbf1_{E_{\mathrm{cnt}}}|D_{a,t}|]
 \le C(\log d)^C d^{-\beta_{\HI}}.
\end{equation*}
For $0<\beta<\beta_{\HI}$, Markov's inequality and a finite union yield
\begin{align*}
 \Prob\left(\max_{a,t}|D_{a,t}|>d^{-\beta}\right)
 &\le\Prob(E_{\mathrm{cnt}}^c)
   +d^\beta\sum_{a,t}\E[\mathbf1_{E_{\mathrm{cnt}}}|D_{a,t}|]\\*
 &\le\Prob(E_{\mathrm{cnt}}^c)
   +C(\log d)^C d^{-(\beta_{\HI}-\beta)}
 \longrightarrow0.
\end{align*}
Applying this bound to the two errors yields
\begin{align}
  \max_{a\in[K]}\sup_{t\in\Ts}
  |\widehat\Es_a^{(t)}-\Es_{a,\pi}^{(t)}|
  &=O_p(d^{-\beta}),
  \label{eq:poly-training-risk}\\
  \max_{a\in[K]}\sup_{t\in\Ts}
  |\widehat M_{a,\diag}^{(t)}-M_a^{(t)}|
  &=O_p(d^{-\beta}).
  \label{eq:poly-diagonal}
\end{align}

We next use the exact noise identity \eqref{eq:aux-independent-decomp}. Since $\varkappa_{\ast,a}$ contains the maximum realized arm-wise noise, \eqref{eq:poly-envelope} with $s=2,4$ implies
\begin{equation*}
  \mu(a)\E[|\varepsilon_i|^s\mymid A_i=a]
  =\E[\mathbf 1_{\{A_i=a\}}|\varepsilon_i|^s]
  \le
  \E[\varkappa_{\ast,a}^s],
\end{equation*}
and hence
\begin{equation*}
  \max_{a\in[K]}\E|\xi_{a,d}|^s
  \le(\log d)^{C_s},
  \qquad s\in\{2,4\},
\end{equation*}
after enlarging $C_s$. Conditional on $A_{1:m}$, the arm-wise noises are i.i.d., and hence, on $E_{\mathrm{cnt}}$,
\begin{align*}
 \operatorname{Var}(\bar\xi_a\mymid A_{1:m})
 &=\frac{\sigma_{\xi,a}^2}{N_a}
 \le\frac{(\log d)^C}{c_-d},\\
 \operatorname{Var}(\overline{\xi_a^2}\mymid A_{1:m})
 &=\frac{\operatorname{Var}(\xi_{a,d}^2)}{N_a}
 \le\frac{\E|\xi_{a,d}|^4}{N_a}
 \le\frac{(\log d)^C}{c_-d}.
\end{align*}
Their conditional means are zero and $\sigma_{\xi,a}^2$, respectively. Conditional Chebyshev's inequality and a finite union therefore give
\begin{equation*}
  \max_a|\bar\xi_a|
  +\max_a|\overline{\xi_a^2}-\sigma_{\xi,a}^2|
  =O_p\{d^{-1/2}(\log d)^C\}.
\end{equation*}
The deterministic prediction bound used in \Cref{prop:prediction-moment-bound}, together with \eqref{eq:poly-envelope}, also gives
\begin{equation*}
  \max_a\sup_{t\in\Ts}|B_{a,t}|=O_p\{(\log d)^C\}.
\end{equation*}
Substitution in \eqref{eq:aux-independent-decomp} gives
\begin{align*}
 \max_a\sup_{t\in\Ts}
 |\Es_{a,\pi}^{(t)}-\Es_{a,\mathrm{ind}}^{(t)}|
 &\le2\max_a|\bar\xi_a|\max_{a,t}|B_{a,t}|
   +B_h\max_a|\overline{\xi_a^2}-\sigma_{\xi,a}^2|\\
 &=O_p\{d^{-1/2}(\log d)^C\}
 =O_p(d^{-\beta})
\end{align*}
for every $0<\beta<1/2$, since $d^{\beta-1/2}(\log d)^C\to0$. The triangle inequality and \eqref{eq:poly-training-risk} now prove the risk-criterion claim for $0<\beta<\beta_0$.

Finally, the proof of \Cref{lem:value-sample-averages} becomes quantitative under \eqref{eq:poly-envelope}. Equation \eqref{eq:poly-diagonal} controls the diagonal terms. The conditional second moment in \eqref{eq:offdiagonal-second-moment} is $O_p\{d^{-1}(\log d)^C\}$. Conditional Chebyshev's inequality gives the corresponding bound for the off-diagonal averages, while the centered reward averages and arm-count deviations are $O_p\{d^{-1/2}(\log d)^C\}$. Moreover, $B_M=\sup_{a,t}|M_a^{(t)}|=O_p\{(\log d)^C\}$. In the notation of \eqref{eq:value-comparison-errors}, these bounds give
\begin{align*}
 \delta_{\diag}&=O_p(d^{-\beta}),\\
 \delta_{\mathrm{off}}+\delta_R+\delta_N(1+B_M)
 &=O_p\{d^{-1/2}(\log d)^C\}
 =O_p(d^{-\beta})
\end{align*}
for $0<\beta<\beta_0$. The bounds in the proof of \Cref{thm:value_estimation} consequently yield
\begin{align*}
 \sup_{\tb\in\Ts^K}
 |\widehat\Vs_{\ADM}(\tb)-\Vs_{g_\tb}(\nu)|
 &\le\delta_{\mathrm{off}}+\delta_{\diag}
 =O_p(d^{-\beta}),\\
 \sup_{\tb\in\Ts^K}
 |\widehat\Vs_{\ADR}(\tb)-\Vs(\nu)|
 &\le\delta_{\mathrm{off}}+\delta_R+\delta_{\diag}
      +K\delta_N(C+B_M)
 =O_p(d^{-\beta}).
\end{align*}
\end{proof}

\section{Additional Experiments and Implementation Details}
\label{app:exp-details}

\subsection{Synthetic data and network training}

Unless stated otherwise, the Gaussian experiments use independent observations with $X_i\sim\mathcal N(0,I_d)$, $A_i\sim\operatorname{Unif}([K])$, and $\varepsilon_i\sim\mathcal N(0,0.8^2)$, where the contexts, actions, and noises are mutually independent. The rewards are
\[
R_i=\varphi_{A_i}^\ast(\beta_{A_i}^\top X_i)+\varepsilon_i.
\]
The two-arm baseline uses $\varphi_1^\ast(y)=\tanh(y)$, $\varphi_2^\ast(y)=\tanh(2y)$, and the logistic target policy $\nu(1\mid x)=\sigma(2\beta_1^\top x)$, where $\sigma(u)=(1+e^{-u})^{-1}$. The signal directions are fixed, dense, and orthonormal. Other reward links and policies are specified below.

For each arm, we train the network in Appendix~\ref{sec:formal-regime} with sigmoid activation, depth $L=2$, and width $q=5$. First-layer entries are initialized independently from $\mathcal N(0,1/d)$ and deeper-layer entries from $\mathcal N(0,1/q)$, with columns $2,\ldots,q$ of the output matrix set to zero. Initial weights are fixed across datasets and shared by all procedures and folds. Every layer is updated by full-batch gradient descent on the unweighted loss $\mathcal L_a$, normalized by its arm-specific training count except in the leave-one-out diagnostic below. Candidate times are $\mathcal T=\{0,\ldots,30\}$. We use $\eta=2$ unless stated otherwise; the increasing-ratio experiment uses $\eta=1$ for every method.

\subsection{Stopping rules and comparison estimators}
\label{app:exp-stopping}

For a full-data model, the proposed stopping rule minimizes \eqref{eq:weighted-criterion} separately for each arm. We compare it with the rule that minimizes the criterion obtained by replacing the corrected predictions with $f_{W_a^{(t)}}(X_i)$, and with ten-fold cross-validation. In the experiment with several target policies, we also minimize the corrected criterion with $w_a\equiv1$ to isolate the effect of weighting the stopping rule for the target policy. All rules choose the smallest minimizing iteration. For cross-validation, a random partition of the complete logged sample defines the folds independently of the observations. We train each arm's network on the complement of each fold and pool its weighted squared residuals over the held-out observations from that arm. The selected number of steps is then used to fit the model on the full arm sample from the common initialization. Consequently, every stopping rule is evaluated on the same full-data training trajectory. Its selection regret is
\[
\mathcal R_\nu^\circ(g_{\hat{\mathbf t}})
-\min_{\mathbf t\in\mathcal T^K}\mathcal R_\nu^\circ(g_{\mathbf t}).
\]
The minimum separates over arms. The reward-noise contribution does not depend on the stopping times, so the same regret is obtained using $\mathcal R_\nu$.

For risk estimation in Figures~\ref{fig:calibration} and~\ref{fig:non-gaussian-growth}, we compare the criterion with $\mathcal E_{a,\mathrm{ind}}^{(t)}$ in \eqref{eq:weighted-armwise-ind-risk}, including the reward-noise contribution. The reported RMSE is the square root of the mean squared discrepancy, where the mean is taken over arms, candidate iterations, and datasets. Selection regret is reported on the total target-risk scale.

For ten-fold cross-fitting (CF10), we randomly partition the complete logged sample into ten folds independently of the observations. Corrected stopping minimizes \eqref{eq:weighted-criterion} on each arm's complete training-complement path. For inner-validation stopping, we split the complement within each arm into $70\%$ for training and $30\%$ for selection. These partitions are independent of the contexts and rewards. We select by held-out weighted squared residuals, restart from the common initialization, and refit on the complete arm-specific complement. Both procedures use the same folds and candidate iterations, exclude the evaluation fold from fitting and selection, and average ordinary held-out DR scores with augmentation coefficient one.

The inverse propensity scoring (IPS) and self-normalized inverse propensity scoring (SNIPS) estimators use the same logged sample:
\[
\widehat\Vs_{\mathrm{IPS}}
=\frac1m\sum_{i=1}^m w_{A_i}(X_i)R_i,
\qquad
\widehat\Vs_{\mathrm{SNIPS}}
=\frac{\sum_{i=1}^m w_{A_i}(X_i)R_i}
       {\sum_{i=1}^m w_{A_i}(X_i)}.
\]
For the direct comparison of the Onsager correction, we evaluate the DR estimator in \eqref{eq:naive-dr-definition} and ADR at identical full-data models and stopping times. We likewise compare ADM with the unadjusted empirical average of the same fitted predictions. The target in this latter comparison is $\Vs_{g_{\hat{\mathbf t}}}(\nu)$, evaluated separately for each fitted model.

We also evaluate each arm's selected predictor on contexts logged under other arms, without changing its training sample or stopping time:
\begin{equation}
\widehat\Vs_{\mathrm{cross}}=
\sum_{a=1}^K\frac1{m-N_a}
\sum_{i\notin\mathcal I_a}
\nu(a\mid X_i)f_{W_a^{(\hat t_a)}}(X_i).
\label{eq:cross-arm-dm}
\end{equation}
Here $\hat t_a$ is selected by the corrected criterion using only arm-$a$ data. Under context-independent logging, conditional on the action labels, the other arms' contexts are independent of this fitted predictor; this is the off-diagonal independence argument in Lemma~\ref{lem:value-sample-averages}. The denominator is the actual number $m-N_a$ of evaluation contexts. Like DM and ADM in this comparison, cross-arm DM targets the fitted-model value $\Vs_{g_{\hat{\mathbf t}}}(\nu)$, rather than the true policy value.

Figure~\ref{fig:value-correction} compares these estimators on the same 200 datasets and selected models per dimension. ADM reduces MSE relative to cross-arm DM by $50.7\%$--$57.4\%$; all five paired pointwise $95\%$ intervals favor ADM. Both estimators' bias intervals contain zero in every setting. The gain is therefore in precision for the fitted-model value.

\subsection{Vanishing-fraction holdouts}
\label{app:exp-vanishing-holdouts}

We compare two holdout sizes fixed before running the experiment, $h=\lceil\sqrt m\rceil$ and $h=\lceil m^{2/3}\rceil$, on the same five Gaussian, uniform-logging settings and 200 datasets per dimension as Figure~\ref{fig:calibration}. Both satisfy $h\to\infty$ and $h/m\to0$. For each dataset, one permutation independent of the observations defines nested holdouts of these sizes. Each arm's network is trained on its complement from the common initialization. We average weighted squared residuals over that arm's held-out observations and select the smallest minimizing iteration. The selected iteration is then evaluated on the same full-data trajectory as the other stopping rules, corresponding to full-data refitting.

Table~\ref{tab:vanishing-holdout-accuracy} reports the risk-estimation RMSE and selection regret defined in Appendix~\ref{app:exp-stopping}. Both holdouts have larger errors and regret than the corrected criterion and ten-fold CV at every dimension. For both metrics, all paired pointwise $95\%$ intervals favor the corrected criterion and CV over the holdouts; risk-estimation comparisons use MSE. Intervals use within-dataset differences, as described in Appendix~\ref{app:exp-reference}. The two schedules thus show a finite-sample accuracy--cost trade-off: correction is more accurate, while the holdouts are cheaper (Appendix~\ref{app:exp-runtime}); their asymptotic comparison remains open.

\begin{table}[!htbp]
\centering
\caption{Risk estimation and full-data stopping selection with $m=2d$.
Each entry yields risk-estimation RMSE / mean selection regret over the same 200 datasets per dimension.}
\label{tab:vanishing-holdout-accuracy}
\small \setlength{\tabcolsep}{3pt}
\begin{tabular*}{\textwidth}{@{\extracolsep{\fill}}rcccc@{}}
\toprule
$d$ & Corrected & Ten-fold CV & $h=\lceil\sqrt m\rceil$ & $h=\lceil m^{2/3}\rceil$\\
\midrule
$128$ & $0.3466$ / $0.0274$ & $0.4100$ / $0.0393$ & $1.2601$ / $0.3710$ & $0.8611$ / $0.2505$\\
$256$ & $0.2481$ / $0.0141$ & $0.2898$ / $0.0186$ & $1.0045$ / $0.2128$ & $0.6390$ / $0.1207$\\
$512$ & $0.1695$ / $0.0095$ & $0.2155$ / $0.0130$ & $0.7814$ / $0.1738$ & $0.4726$ / $0.0512$\\
$1024$ & $0.1161$ / $0.0043$ & $0.1726$ / $0.0093$ & $0.6772$ / $0.0991$ & $0.3782$ / $0.0296$\\
$2048$ & $0.0811$ / $0.0025$ & $0.1537$ / $0.0082$ & $0.5527$ / $0.0621$ & $0.3017$ / $0.0191$\\
\bottomrule
\end{tabular*}
\end{table}

\subsection{Target-policy weighting}
\label{app:exp-policy-weighting}

Figure~\ref{fig:policy-weighting} uses three arms with reward links $\varphi_1^\ast(y)=\tfrac12+\tanh(y)+\tfrac12\tanh^2(y)$, $\varphi_2^\ast(y)=-\tfrac12+\tanh(2y)-\tfrac12\tanh^2(y)$, and $\varphi_3^\ast(y)=\tfrac14+\cos(y)$. Writing $z=(\beta_1^\top x,\beta_2^\top x,\beta_3^\top x)$, the four policies are uniform, $\operatorname{softmax}(2z)$, $\operatorname{softmax}(-2z)$, and $\operatorname{softmax}(2(z_2,z_3,z_1))$. We use $m=3d$ at $d\in\{128,256,512\}$, with 200 datasets per dimension and the network and training schedule above. Every policy uses the same logged datasets, training trajectories, and Onsager correction, so the comparison changes only the weights used for selection.

Averaging within each dataset over the three nonuniform policies, weighting reduces prediction risk relative to unweighted corrected selection by $0.0315$, $0.0311$, and $0.0327$ across the three dimensions. The corresponding reductions relative to weighted ten-fold CV are $0.0281$, $0.0267$, and $0.0293$. Weighting reduces risk for $\operatorname{softmax}(2z)$ and $\operatorname{softmax}(-2z)$ at every dimension, with the larger reduction for the former. For the uniform policy, weighted and unweighted selection coincide. For the cyclically permuted policy, paired intervals for weighted versus unweighted selection contain zero at all three dimensions. The corrected weighted rule has lower mean regret than weighted CV in every policy--dimension setting; its paired interval excludes zero except for $\operatorname{softmax}(2z)$ at $d=128$.

\subsection{Value estimation at a fixed sample-to-dimension ratio}
\label{app:exp-fixed-ratio}

We use the five $m=2d$ settings of Figure~\ref{fig:fixed-ratio-value}, with the fitting and selection protocols in Appendix~\ref{app:exp-stopping}. To compare the selected complement-trained models, we compute their target-policy prediction errors on independent test contexts, conditional on each training complement, and average across folds using held-out fold sizes, using the numerical references in Appendix~\ref{app:exp-reference}.

\begin{table}[!htbp]
\centering
\caption{Prediction error of the selected complement-trained models with
$m=2d$. The last column is corrected stopping minus inner validation, with a paired pointwise $95\%$ Monte Carlo interval over 200 datasets.}
\label{tab:cf-selected-risk}
\small \setlength{\tabcolsep}{4pt}
\begin{tabular*}{\textwidth}{@{\extracolsep{\fill}}rccc@{}}
\toprule
$d$ & Inner validation & Corrected stopping & Difference [$95\%$ interval]\\
\midrule
$128$ & $0.8242$ & $0.4754$ & $-0.3488\;[-0.3735,-0.3241]$\\
$256$ & $0.7276$ & $0.4703$ & $-0.2574\;[-0.2814,-0.2334]$\\
$512$ & $0.6874$ & $0.4777$ & $-0.2097\;[-0.2273,-0.1922]$\\
$1024$ & $0.6173$ & $0.4678$ & $-0.1495\;[-0.1640,-0.1350]$\\
$2048$ & $0.5828$ & $0.4705$ & $-0.1123\;[-0.1258,-0.0988]$\\
\bottomrule
\end{tabular*}
\end{table}

For the bias--variance decomposition, let $e_r$ be a procedure's policy-value error on dataset $r$, with $B=200$ datasets, and let $\bar e=B^{-1}\sum_r e_r$. We report empirical squared bias $\bar e^2$, variance $B^{-1}\sum_r(e_r-\bar e)^2$, and MSE $B^{-1}\sum_r e_r^2$. The first two sum exactly to the third before rounding. The variance uses denominator $B$ and measures variation across complete value estimates with the initializations fixed. Table~\ref{tab:fixed-ratio-decomposition} reports these components and Student $t$ intervals for the mean squared error; the intervals quantify Monte Carlo uncertainty across datasets.

\begin{table}[!htbp]
\centering
\caption{Empirical bias--variance decomposition for policy-value estimation
with $m=2d$ and 200 datasets per dimension. CF10 uses corrected stopping. Squared bias, variance, and MSE are multiplied by $m$. The final column gives pointwise $95\%$ Monte Carlo intervals for MSE; squared bias and variance are descriptive empirical components, defined in the text.}
\label{tab:fixed-ratio-decomposition}
\small \setlength{\tabcolsep}{4pt}
\begin{tabular*}{\textwidth}{@{\extracolsep{\fill}}rlrrrr@{}}
\toprule
$d$ & Method & Squared bias & Variance & MSE & $95\%$ interval for MSE\\
\midrule
$128$ & ADR & $0.00007$ & $1.352$ & $1.352$ & $[1.072,\,1.632]$ \\
 & CF10 & $0.00134$ & $1.398$ & $1.399$ & $[1.117,\,1.681]$ \\
 & IPS & $0.00106$ & $1.364$ & $1.365$ & $[1.088,\,1.643]$ \\
 & SNIPS & $0.00163$ & $1.334$ & $1.335$ & $[1.062,\,1.609]$ \\
\addlinespace[3pt]
$256$ & ADR & $0.00058$ & $1.491$ & $1.491$ & $[1.203,\,1.779]$ \\
 & CF10 & $0.00031$ & $1.625$ & $1.625$ & $[1.310,\,1.939]$ \\
 & IPS & $0.00085$ & $1.667$ & $1.667$ & $[1.340,\,1.995]$ \\
 & SNIPS & $0.00056$ & $1.601$ & $1.602$ & $[1.278,\,1.925]$ \\
\addlinespace[3pt]
$512$ & ADR & $0.00049$ & $1.473$ & $1.473$ & $[1.189,\,1.757]$ \\
 & CF10 & $0.00005$ & $1.508$ & $1.508$ & $[1.225,\,1.791]$ \\
 & IPS & $0.00938$ & $1.665$ & $1.674$ & $[1.364,\,1.985]$ \\
 & SNIPS & $0.00831$ & $1.572$ & $1.581$ & $[1.291,\,1.870]$ \\
\addlinespace[3pt]
$1024$ & ADR & $0.00160$ & $1.646$ & $1.647$ & $[1.283,\,2.011]$ \\
 & CF10 & $0.00356$ & $1.654$ & $1.658$ & $[1.310,\,2.005]$ \\
 & IPS & $0.01433$ & $1.723$ & $1.737$ & $[1.413,\,2.061]$ \\
 & SNIPS & $0.01408$ & $1.637$ & $1.651$ & $[1.338,\,1.964]$ \\
\addlinespace[3pt]
$2048$ & ADR & $0.00130$ & $1.372$ & $1.373$ & $[1.112,\,1.634]$ \\
 & CF10 & $0.00045$ & $1.407$ & $1.408$ & $[1.133,\,1.683]$ \\
 & IPS & $0.00188$ & $1.590$ & $1.592$ & $[1.294,\,1.890]$ \\
 & SNIPS & $0.00135$ & $1.540$ & $1.541$ & $[1.251,\,1.832]$ \\
\bottomrule
\end{tabular*}
\end{table}

The bias intervals for ADR and both CF10 procedures contain zero at every dimension. Paired $95\%$ MSE intervals favor ADR over corrected-stopping CF10 at $d=256$ and over inner-validation CF10 at $d=256,2048$; the remaining intervals contain zero. Corrected stopping lowers CF10's MSE point estimate at every dimension, with a paired interval excluding zero only at $d=2048$. Prediction-risk improvements need not translate proportionally into value-MSE gains: DR variance combines errors across arms, and cross-fitted estimates also include covariance between folds.

ADR has lower MSE than IPS at $d=256,512,2048$ and than SNIPS at $d=2048$, with paired $95\%$ intervals below zero. The remaining intervals contain zero. These comparisons measure finite-sample MSE: with known logging, IPS is unbiased and consistent under the stated moment conditions.

\subsection{Risk curves and leave-one-out refitting}
\label{app:exp-risk-paths}

\begin{figure}[t]
\centering
\includegraphics[width=\textwidth]{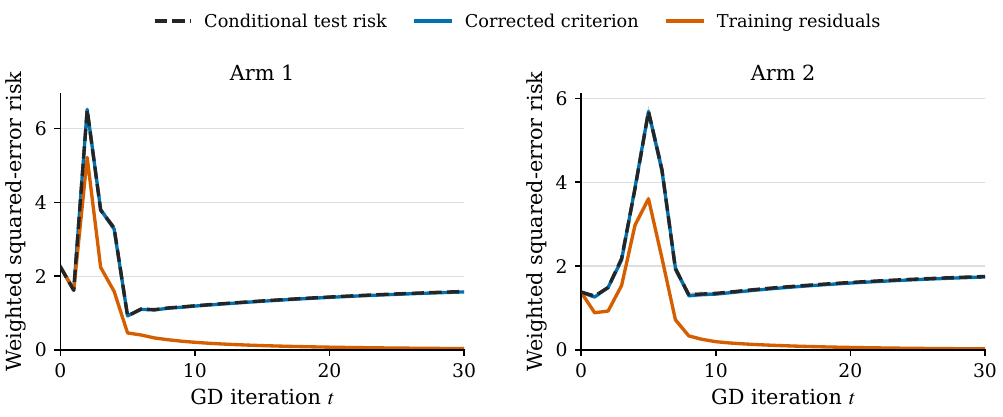}
\caption{Importance-weighted arm-wise risk along training at $d=512$
and $m=1024$. Lines are averages over 200 datasets; the test-risk reference includes reward noise and is conditional on each fitted network. Shading gives pointwise $95\%$ Monte Carlo intervals across datasets.}
\label{fig:risk-paths}
\end{figure}

Figure~\ref{fig:risk-paths} shows the two-arm risk curves at $d=512$ and $m=1024$. In later iterations, the training-residual criterion decreases while the test risk rises; the corrected criterion follows the test risk.

We also compare corrected and unadjusted predictions with networks retrained after omitting the evaluated observation. All layers are retrained from the same initialization and with the same learning rates, retaining the original arm denominator $N_a$ in the loss. We use $d\in\{128,256,512\}$ and $m=2d$, with 100 independent datasets per dimension and 32 rows sampled uniformly without replacement per dataset. The discrepancy is the root mean squared prediction difference over these rows and datasets.

\begin{figure}[t]
\centering
\includegraphics[width=\textwidth]{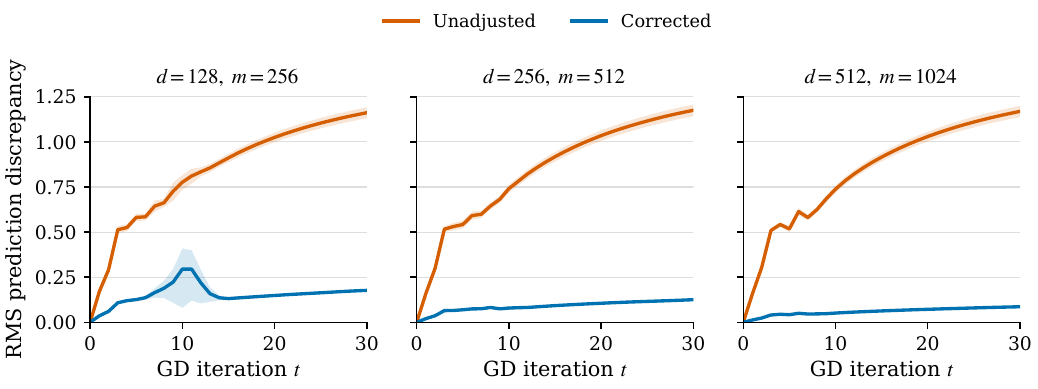}
\caption{Prediction discrepancy from a network retrained after omitting
the evaluated observation. Every layer is retrained, with the original arm denominator $N_a$ retained. At each dimension, we use 100 datasets and sample 32 rows per dataset. Lines show root mean squared discrepancy over rows and datasets; shading gives pointwise $95\%$ Monte Carlo intervals.}
\label{fig:deletion-paths}
\end{figure}

Figure~\ref{fig:deletion-paths} shows this discrepancy along training. At $t=30$, the corrected prediction's root mean squared discrepancy decreases from $0.177$ to $0.086$ as $d$ increases from $128$ to $512$, while the unadjusted prediction's discrepancy remains near $1.17$.

\subsection{Nonuniform logging}
\label{app:exp-nonuniform}

We repeat the two-arm Gaussian experiment with context-independent logging probabilities $\mu(1)\in\{0.5,0.25,0.1\}$ and $\mu(2)=1-\mu(1)$. We use $d\in\{512,1024,2048\}$ and $m=2d$, with 200 independent datasets at each setting. The reward links, target policy, initial weights, network architecture, and training schedule remain as specified above, with $\eta=2$. The uniform-logging reference uses new datasets generated for this comparison.

Each arm's criterion uses $w_a(X)=\nu(a\mid X)/\mu(a)$. We measure error in estimating total target-policy test risk by comparing $\sum_{a=1}^2\mu(a)\widehat{\mathcal E}_a^{(t_a)}$ with $\mathcal R_\nu(g_{\mathbf t})$, and form the corresponding comparisons for training residuals and cross-validation. The reported RMSE is the square root of the mean squared error over all $31\times31$ pairs of stopping times and datasets. This differs from the arm-wise RMSE in Figure~\ref{fig:calibration}. Selection regret uses the total prediction risk $\mathcal R_\nu^\circ$, with every rule evaluated on the same full-data training trajectories.

\begin{figure}[t]
\centering
\includegraphics[width=\textwidth]{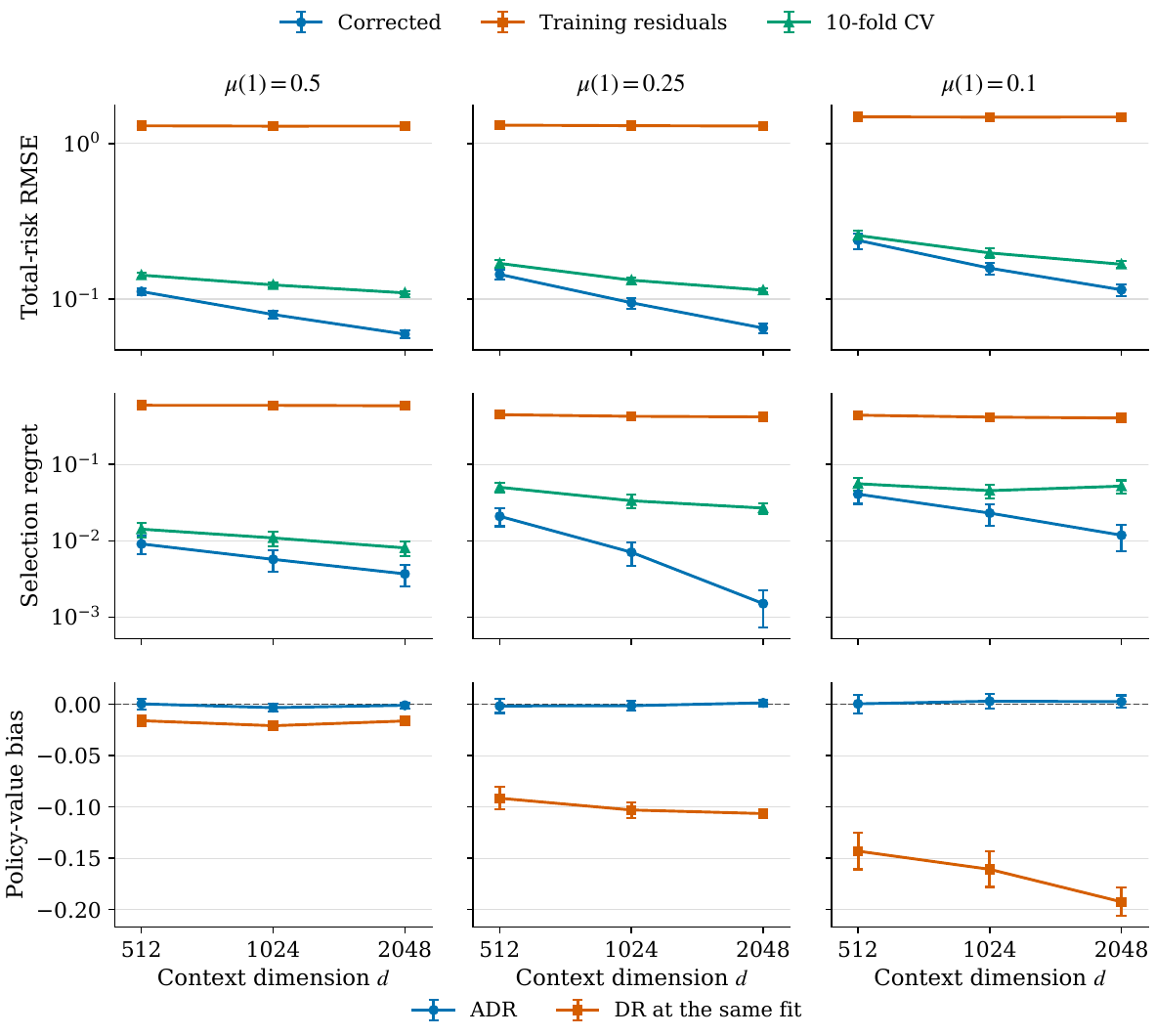}
\caption{Risk estimation, stopping-time selection, and value bias under
context-independent logging, with $\mu(2)=1-\mu(1)$, $m=2d$, and 200 datasets per setting. Top: root mean squared error in estimating the full-data models' total target-policy test risk, with squared errors averaged over all pairs of stopping times and datasets. Middle: mean total prediction-risk regret of the selected full-data models. These two rows use logarithmic vertical axes. Bottom: empirical bias of DR and ADR at identical models and stopping times selected by the corrected criterion. Bars give pointwise $95\%$ Monte Carlo intervals across datasets.}
\label{fig:nonuniform-logging}
\end{figure}

Figure~\ref{fig:nonuniform-logging} shows that the corrected criterion's total-risk RMSE and mean selection regret decrease with dimension under each logging distribution. Its empirical mean squared risk-estimation error is smaller than cross-validation's in all nine settings; the paired $95\%$ Monte Carlo interval for the difference excludes zero except at $\mu(1)=0.1,d=512$. The paired selection-regret intervals favor the corrected rule in every setting. For the rare arm at $\mu(1)=0.1,d=512$, however, its arm-wise risk RMSE is slightly higher than cross-validation's, despite the lower total-risk error.

The bottom row of Figure~\ref{fig:nonuniform-logging} compares DR and ADR at identical full-data models and stopping times selected by the corrected criterion. The correction reduces absolute empirical bias in every setting. For $\mu(1)=0.1$, the DR bias changes from $-0.143$ to $-0.193$ as dimension increases from $512$ to $2048$, while the corresponding ADR biases are $0.0004$ and $0.0025$. All nine ADR bias intervals include zero. This comparison isolates the effect of correcting the predictions used for value estimation.

\subsection{Non-Gaussian contexts}
\label{app:exp-nongaussian}

We repeat the two-arm experiment at $d\in\{256,1024,2048\}$ and $m=2d$ with independent Rademacher coordinates and with independent coordinates distributed as $\sqrt{3/5}\,T_5$, where $T_5$ has a Student $t$ distribution with five degrees of freedom. Both distributions have mean zero and variance one. At each dimension, we retain the Gaussian experiment's reward links, target policy, initial weights, and training schedule, with independent uniform logging. Each setting uses 200 independent logged datasets.

\begin{figure}[t]
\centering
\includegraphics[width=\textwidth]{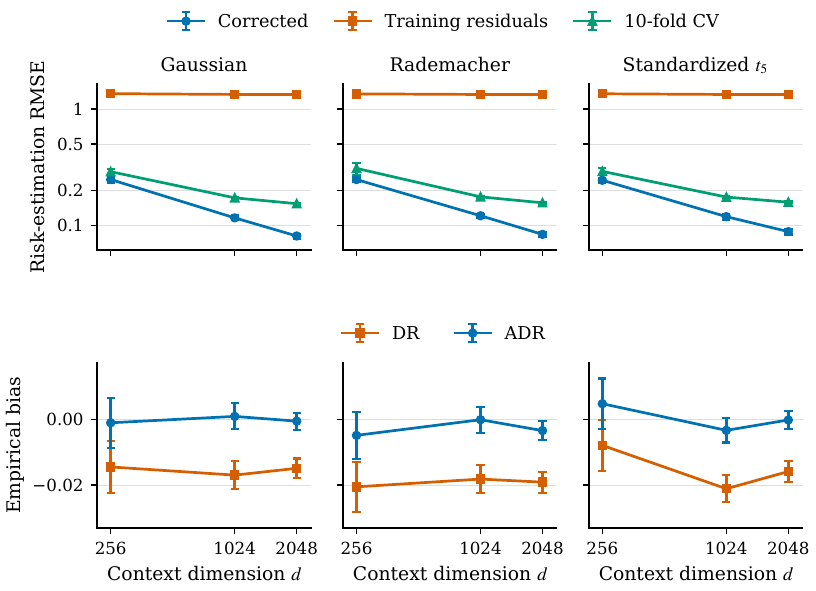}
\caption{Risk estimation and value estimation under three context
distributions, with $m=2d$ and 200 datasets per setting. Top: root mean squared error in estimating the full-data models' conditional arm-wise test risks, with squared errors averaged over both arms, all candidate iterations, and datasets. These risks include reward noise and use $w_a(X)=2\nu(a\mid X)$; the vertical axis is logarithmic. Bottom: empirical bias of DR and ADR relative to the true policy value, at identical models and stopping times selected by the corrected criterion. Bars give pointwise $95\%$ Monte Carlo intervals across datasets.}
\label{fig:non-gaussian-growth}
\end{figure}

The upper panels of Figure~\ref{fig:non-gaussian-growth} compare risk estimation under the three context laws. As dimension increases, the corrected criterion's RMSE decreases from $0.248$ to $0.084$ for Rademacher contexts and from $0.244$ to $0.088$ for standardized $t_5$ contexts. At $d=2048$, the corresponding RMSEs are $0.156$ and $0.158$ for cross-validation, and $1.335$ and $1.334$ for training residuals. Table~\ref{tab:non-gaussian-selection} compares the resulting stopping decisions on the full-data training trajectories. For both non-Gaussian laws, the corrected rule has smaller mean selection regret than ten-fold cross-validation at every dimension, with the paired $95\%$ intervals below zero.

The lower panels compare DR and ADR at the same full-data models and stopping times selected by the corrected criterion, with the true policy value computed for each context law as in Appendix~\ref{app:exp-reference}. The correction reduces absolute empirical bias in all six non-Gaussian settings. At $d=2048$, the DR and ADR biases are $-0.0191$ and $-0.0034$ for Rademacher contexts, and $-0.0159$ and $-0.0002$ for standardized $t_5$ contexts.

\begin{table}[!htbp]
\centering
\caption{Stopping-time selection under three context distributions,
with $m=2d$ and 200 datasets per setting. Regret is measured by the total target-policy prediction risk $\mathcal R_\nu^\circ$ relative to the best pair of stopping times on the full-data trajectories. Parentheses give Monte Carlo standard errors. The last two columns report the paired regret difference between the corrected rule and ten-fold cross-validation, with pointwise $95\%$ Monte Carlo intervals.}
\label{tab:non-gaussian-selection}
\small \setlength{\tabcolsep}{4pt}
\begin{tabular*}{\textwidth}{@{\extracolsep{\fill}}lrccc@{}}
\toprule
Context law & $d$ & Corrected regret & Corrected minus CV & $95\%$ interval\\
\midrule
Gaussian & $256$ & $0.0141$ $(0.0020)$ & $-0.0045$ & $[-0.0082,\,-0.0008]$ \\
 & $1024$ & $0.0043$ $(0.0007)$ & $-0.0050$ & $[-0.0070,\,-0.0031]$ \\
 & $2048$ & $0.0025$ $(0.0005)$ & $-0.0057$ & $[-0.0076,\,-0.0038]$ \\
\addlinespace[3pt]
Rademacher & $256$ & $0.0138$ $(0.0018)$ & $-0.0066$ & $[-0.0106,\,-0.0025]$ \\
 & $1024$ & $0.0054$ $(0.0007)$ & $-0.0063$ & $[-0.0086,\,-0.0041]$ \\
 & $2048$ & $0.0042$ $(0.0006)$ & $-0.0059$ & $[-0.0077,\,-0.0041]$ \\
\addlinespace[3pt]
Standardized $t_5$ & $256$ & $0.0154$ $(0.0018)$ & $-0.0089$ & $[-0.0127,\,-0.0050]$ \\
 & $1024$ & $0.0066$ $(0.0009)$ & $-0.0041$ & $[-0.0062,\,-0.0020]$ \\
 & $2048$ & $0.0027$ $(0.0005)$ & $-0.0059$ & $[-0.0077,\,-0.0040]$ \\
\bottomrule
\end{tabular*}
\end{table}

\subsection{Numerical reference values and uncertainty}
\label{app:exp-reference}

Reference test risks are conditional on the fitted models. For Gaussian contexts, we integrate the joint Gaussian vector of signal and policy indices and fitted first-layer projections using two independent scrambled Sobol rules with antithetic points, each containing $2^{14}$ points. For non-Gaussian contexts, we average two independent Monte Carlo estimates, each using 16,384 draws and their negatives from the specified context law. The reward-noise contribution is integrated analytically. All reference draws are independent of the logged data and are used only to assess the procedures; they do not enter fitting, selection, or the value estimators.

The true policy value is computed by Gaussian integration, a finite sum over binomial sign counts for Rademacher contexts, or numerical inversion of the signal sums' characteristic functions for standardized $t_5$ contexts. Errors of DM, ADM, and cross-arm DM are measured against the average of the two independent numerical references for each fitted model's value.

For normalized value MSE, we compute $\sigma_{\mathrm{eff}}^2=\operatorname{Var}\{\psi_{g^\ast}(X,A,R)\}$ by Gaussian integration and use the true-reward score average $\widehat\Vs_{\mathrm{or}}=m^{-1}\sum_i\psi_{g^\ast}(Q_i)$ as a control variate. Specifically, we average
\[
1+\frac{m}{\sigma_{\mathrm{eff}}^2}
\left\{(\widehat\Vs-\Vs(\nu))^2
      -(\widehat\Vs_{\mathrm{or}}-\Vs(\nu))^2\right\}
\]
over datasets; the control squared error has mean $\sigma_{\mathrm{eff}}^2/m$.

All pointwise $95\%$ intervals use Student $t$ quantiles across independent logged datasets. They describe Monte Carlo uncertainty in performance, rather than uncertainty in policy value from a single dataset. RMSE intervals take square roots of the MSE limits after truncating the lower limit at zero. Comparisons use within-dataset differences; in the target-policy experiment, averages over nonuniform policies are also formed within each dataset before computing intervals.

\subsection{Value estimation as sample size outgrows dimension}
\label{app:exp-increasing-ratio}
We compare ADR, DR with ten-fold cross-fitting (CF10), inverse propensity scoring (IPS), and self-normalized IPS (SNIPS) as the number of observations increases relative to the context dimension. We use the two-arm model above and train every reward network with $\eta=1$ over candidate iterations $\{0,\ldots,30\}$. Both corrected-stopping and inner-validation CF10 use the protocols in Appendix~\ref{app:exp-stopping}; each evaluation fold is excluded from fitting and selection. We~first~fix $d=256$ and increase $m\in\{512,2048,8192,32768\}$. We also increase both dimension and sample-to-dimension ratio along $(d,m)\in\{(256,2048),(512,8192),(1024,32768)\}$. Each setting uses 200 independent datasets, except those with $m=32768$, which use 100.

We report normalized mean squared error, $m\E[(\widehat\Vs-\Vs(\nu))^2]/\sigma_{\mathrm{eff}}^2$, where $\sigma_{\mathrm{eff}}^2=\operatorname{Var}\{\psi_{g^\ast}(X,A,R)\}$. The normalization $\sigma_{\mathrm{eff}}^2/m$ is the exact MSE of DR using the true reward function, whose normalized MSE is therefore one for every $(d,m)$. At a fixed data law, $\sigma_{\mathrm{eff}}^2$ is also the classical semiparametric variance bound for known logging and otherwise unrestricted context and reward distributions. We use it here as an oracle reference. We estimate normalized MSE using the control-variate calculation in Appendix~\ref{app:exp-reference}.

Figure~\ref{fig:ratio-precision} shows ADR and both CF10 procedures approaching normalized MSE one as sample size outgrows dimension. At $(d,m)=(1024,32768)$, ADR and corrected-stopping CF10 give $1.000$ and $1.013$, respectively. At both endpoints with $m=32768$, paired ADR-minus-corrected-CF10 intervals contain zero, while ADR's MSE is lower than IPS and SNIPS with paired intervals below zero.

\subsection{Computational cost}
\label{app:exp-runtime}

ADR requires one full-data training trajectory per arm. Its response recursion adds $O(N_aq^3T^3)$ operations and $O(N_aT^2q^2)$ storage (Appendix~\ref{sec:onsager}). Corrected-stopping CF10 repeats this calculation on ten training complements. Table~\ref{tab:runtime} shows that ADR is $5.5$--$8.9$ times faster in the three benchmark settings. CV and inner-validation CF10 are faster in the smallest setting, while ADR is faster in the larger two.

Both tables report median wall time over five shared datasets per setting, using the two-arm Gaussian model and training settings above, with $\eta=2$ when $m=2d$ and $\eta=1$ at $m=32768$. Each measurement includes training, selection, refitting where required, and final value estimation; common data generation and reference integration are excluded. Runs use one numerical thread, with a separate warm-up excluded. The studies were conducted on different CPUs, so comparisons use the controls within each table.

\begin{table}[!htbp]
\centering
\caption{Median runtime in seconds for complete selection and value
estimation on an Intel Xeon E5-2680 v3. Both CF10 procedures exclude each evaluation fold from fitting and selection. CV includes full-data refitting and raw DR scoring.}
\label{tab:runtime}
\small \setlength{\tabcolsep}{4pt}
\begin{tabular*}{\textwidth}{@{\extracolsep{\fill}}rrcccc@{}}
\toprule
$d$ & $m$ & ADR & \shortstack{Ten-fold CV\\+ refit} & \shortstack{CF10\\(inner validation)} & \shortstack{CF10\\(corrected stopping)}\\
\midrule
$512$ & $1024$ & $2.55$ & $1.44$ & $1.34$ & $13.97$\\
$2048$ & $4096$ & $8.15$ & $16.68$ & $13.26$ & $66.74$\\
$1024$ & $32768$ & $51.84$ & $86.02$ & $133.29$ & $462.31$\\
\bottomrule
\end{tabular*}
\end{table}

The single holdouts in Table~\ref{tab:holdout-runtime} are faster than ADR. Together with Appendix~\ref{app:exp-vanishing-holdouts}, this identifies the trade-off: the correction improves risk estimation and stopping selection, while a single holdout costs less. The CV and holdout timings include full-data refitting and raw DR scoring; these procedures compare selection costs, since refitting alone does not correct same-sample DR bias.

\begin{table}[!htbp]
\centering
\caption{Median runtime in seconds in a separate matched holdout study on
an Intel Xeon Gold 6230R. All procedures include selection and final scoring; CV and holdouts also include full-data refitting.}
\label{tab:holdout-runtime}
\small \setlength{\tabcolsep}{4pt}
\begin{tabular*}{\textwidth}{@{\extracolsep{\fill}}rrcccc@{}}
\toprule
$d$ & $m$ & ADR & \shortstack{Ten-fold CV\\+ refit} & \shortstack{Holdout + refit\\$h=\lceil\sqrt m\rceil$} & \shortstack{Holdout + refit\\$h=\lceil m^{2/3}\rceil$}\\
\midrule
$512$ & $1024$ & $1.66$ & $0.71$ & $0.13$ & $0.14$\\
$2048$ & $4096$ & $7.99$ & $12.97$ & $1.82$ & $1.69$\\
$1024$ & $32768$ & $54.68$ & $71.61$ & $13.61$ & $14.02$\\
\bottomrule
\end{tabular*}
\end{table}

\end{document}